\documentclass[PHD]{macro/neu_msthesis}

\title{Macro-Action-Based Multi-Agent/Robot \\Deep Reinforcement Learning under \\ Partial Observability}

\author{Yuchen Xiao}

\dept{Khoury College of Computer Sciences}

\degreename{Computer Sciences}

\field{Computer Sciences}

\submitdate{Aug 2022}

\numberofmembers{3}

\principaladviser{Christopher Amato}
\firstreader{Leslie Kaelbling}
\secondreader{Robert Platt}

\thirdreader{Lawson Wong}

\usepackage{amsfonts}			
\usepackage{times}		

\newcommand{\ifno}[1]{}

\usepackage{multirow}
\makeindex
\usepackage{makeidx}
\usepackage{acronym}

\usepackage{url}
\ifnum\pdfoutput>0
\usepackage[pdftex]{graphicx}
\else
\usepackage{graphicx}
\fi

\ifnum\pdfoutput>0
\usepackage[hidelinks,pdftex,
bookmarks=true,
bookmarksnumbered=true,
hypertexnames=false,
breaklinks=true           
]{hyperref}
\else
\usepackage[hidelinks,hypertex,
bookmarks=true,
bookmarksnumbered=true,
hypertexnames=false,
breaklinks=true           
]{hyperref}
\fi

\hypersetup{				
pdfauthor = {\authorRef},
pdftitle = {\titleRef},
pdfsubject = {\expandafter{\degreeRef} thesis submitted to Northeastern University},
pdfkeywords = {add keywords here}
pdfcreator = {LaTeX with hyperref package},
pdfproducer = {dvips + ps2pdf}}

\usepackage{amsmath,amsfonts,bm}

\def\eqref#1{equation~\ref{#1}}

\def\1{\bm{1}}

\DeclareMathAlphabet{\mathsfit}{\encodingdefault}{\sfdefault}{m}{sl}
\SetMathAlphabet{\mathsfit}{bold}{\encodingdefault}{\sfdefault}{bx}{n}

\newcommand{\E}{\mathbb{E}}

\DeclareMathOperator*{\argmax}{arg\,max}

\usepackage{epsfig}
\usepackage{wrapfig}
\usepackage{subcaption}
\usepackage[font=small]{caption}
\usepackage{algorithm}
\usepackage{algorithmicx}
\usepackage[noend]{algpseudocode}
\usepackage{multirow}
\usepackage{multicol}
\usepackage{booktabs}
\usepackage{tcolorbox}

\begin{document}

\pdfbookmark[1]{Cover}{cover}

\titlepage

\begin{frontmatter}

\begin{dedication}
To my family.
\end{dedication}


\begin{acknowledgements}

First and foremost, I would like to thank my advisor Christopher Amato for his incredible support, insightful feedback, and invaluable guidance throughout my Ph.D., and for providing me with the freedom to pursue the research topics that I was truly passionate about. 
He trained me in the leadership of conducting research work and the ability to present research work well.
He was also very supportive to me in my career and kept giving me confidence.
His generosity made my entire Ph.D. journey much more enjoyable.
His principles of being a great advisor and scholar will continue to guide me throughout my life. 

I also would like to extend my sincere appreciation to the remaining members of my thesis committee, Leslie Pack Kaelbling, Robert Platt, and Lawson L.S. Wong, for their time, interesting questions, and constructive feedback on this dissertation. 

I am also grateful to my collaborators. A special mention must go to Sammie Katt, who had been incredibly helpful from beginning to end, such as helping me (with a fully mechanical engineering background) speed up on learning fundamentals of computer sciences and providing effective suggestions on improving my programming skills. 
I really enjoyed working with him on my first project and all the discussions on RL-related topics, as well as the memorable experience of attending AAMAS 2019 together.  
I would especially like to thank the students including Joshua Hoffman, Weihao Tan, and Tian Xia for their great efforts and contributions to my thesis project. 
I still remember the moments when we worked together on experiments and paper writing until very late before the conference deadlines. 
I would also thank Andrea Baisero for meaningful discussions on the theory part of many papers. 
I am thankful to Andreas Ten Pas and Marcus Gualtieri who provided technical support for making the Fetch Robot perform manipulation in my projects.
I also would like to thank Xueguang Lyu, Trong Nghia Hoang, Brett Daley, Kavinayan Sivakumar, Xingyu Lu, and Chengguang Xu for being brilliant and fantastic collaborators.

I have also had the pleasure of working with and interacting with a number of other students and friends including Zhen Zeng, Shayegan Omidshafiei, Yinxiao Li, Dian Wang, David Slayback, Hai Nguyen, Enrico Marchesini, Kevin Esslinger, Pushyami Kaveti, Haojie Huang, Ondrej Biza, Tarik Kelestemur, Shuo Jiang, Xupeng Zhu, Molly Ohman, Kaiyu Zheng, Xiaobei Guo, Wendi Cui, Yuhong Du, and many others. 

Finally, I would love to extend my sincere gratitude to my family. 
I am incredibly thankful to my parents, Ying Guan and Kecheng Xiao, for spending so much time with me growing up, shaping me as a person, encouraging me in tough times, and always supporting me in my life. 
I also would love to thank my wife, Mang Zhang, for all the times she has supported me and been with me.
I thank my daughter, MingXue (Ella) Xiao, for letting me experience the miracle of life and birth, and for making me understand more about the love and support my parents have given me. 

\end{acknowledgements}


\begin{abstract}

    \noindent The state-of-the-art multi-agent reinforcement learning (MARL) methods have provided promising solutions to a variety of complex problems. 
    Yet, these methods all assume that agents perform synchronized primitive-action executions so that they are not genuinely scalable to long-horizon real-world multi-agent/robot tasks that inherently require agents/robots to asynchronously reason about high-level action selection at varying time durations. 
    The Macro-Action Decentralized Partially Observable Markov Decision Process (MacDec-POMDP) is a general formalization for asynchronous decision-making under uncertainty in fully cooperative multi-agent tasks. 
    In this thesis, we first propose a group of value-based RL approaches for MacDec-POMDPs, where agents are allowed to perform asynchronous learning and decision-making with macro-action-value functions in three paradigms: decentralized learning and control, centralized learning and control, and centralized training for decentralized execution (CTDE). 
    Building on the above work, we formulate a set of macro-action-based policy gradient algorithms under the three training paradigms, where agents are allowed to directly optimize their parameterized policies in an asynchronous manner. 
    We evaluate our methods both in simulation and on real robots over a variety of realistic domains. Empirical results demonstrate the superiority of our approaches in large multi-agent problems and validate the effectiveness of our algorithms for learning high-quality and asynchronous solutions with macro-actions.

\end{abstract}

\pdfbookmark[1]{Table of Contents}{contents}
\tableofcontents
\listoffigures
\newpage\ssp
\listoftables


\chapter*{List of Acronyms}
\addcontentsline{toc}{chapter}{List of Acronyms}

\begin{acronym}

\acro{MDPs}{Markov Decision Processes}.

\acro{POMDPs}{Partially Observable Markov Decision Processes}.

\acro{Dec-POMDPs}{Decentralized Partially Observable Markov Decision Processes}.

\acro{MacDec-POMDPs}{Macro-Action Decentralized Partially Observable Markov Decision Processes}.

\acro{RL}{Reinforcement Learning}.

\acro{DQN}{Deep Q-network}.

\acro{DDQN}{Double Deep Q-network}.

\acro{RNN}{Recurrent Neural Network}.

\acro{DRQN}{Deep Recurrent Q-network}.

\acro{MARL}{Multi-Agent Reinforcement Learning}.

\acro{IQL}{Independent Q-Learning}.

\acro{Dec-HDRQN}{Decentralized Hysteresis DRQN}.

\acro{IAC}{Independent Actor-Critic}.

\acro{CTDE}{Centralized Training for Decentralized Execution}.

\acro{IACC}{Independent Actor with Centralized Critic}.

\acro{Dec-HDDRQN}{Decentralized Hysteresis Double DRQN with Macro-Actions}.

\acro{Cen-DDRQN}{Centralized Double DRQN with Macro-Actions}.

\acro{MacDec-DDRQN}{Macro-Action-Based Decentralized Double Deep Recurrent Q-Net}.

\acro{Parallel-MacDec-DDRQN}{Parallel Macro-Action-Based Decentralized Double Deep Recurrent Q-Net}.

\acro{Mac-IAC}{Macro-Action-Based Independent Actor-Critic}.

\acro{Mac-CAC}{Macro-Action-Based Centralized Actor-Critic}.

\acro{Naive Mac-IACC}{Naive Macro-Action-Based Independent Actor with Centralized Critic}.

\acro{Mac-IAICC}{Naive Macro-Action-Based Independent Actor with Individual Centralized Critic}.

\acro{ROS}{Robot Operating System}.

\end{acronym}


\end{frontmatter}


\pagestyle{headings}
\setlength\parindent{1.4em}


\chapter{Introduction}

\section{Overview}
\label{chap:intro:overview}

\noindent More and more autonomous agents/robots are (going to be) deployed in a variety of real-world applications.
Examples include office service~\cite{X:ServiceRobot}, package delivery~\cite{MURRAY2020368}, and agriculture inspection~\cite{agricultural}, search and rescue~\cite{SR}, autonomous vehicles~\cite{waymo17,TangICCV2019}, sports~\cite{JOLLY2007589, DeepMindSoccer} and others.
In all of these scenarios, it is likely necessary for each agent to take into account the presence and the effect of other agents in order to behave rationally, which naturally results in multi-agent decision-making systems. 
Notably, real-world environments are often unstructured and fast-changing, and agents' reasoning has to rely on partial knowledge of the environment, noisy sensing, stochastic outcomes, and many other surrounding uncertainties. 
Furthermore, fast and perfect communication is difficult and expensive to achieve in general. 
Agents may have to perform sequential decisions based on only local information and cope with uncertainty about each other. Collaboration over agents thus becomes more difficult.  
All of these facts make multi-agent decision-making under partial observability an important challenge.

Multi-agent reinforcement learning (MARL) is a promising framework to generate solutions for these kinds of multi-agent problems. 
Recently, by leveraging deep neural networks to deal with large state and observation input, deep MARL has attracted great attention and achieved many successes in solving challenging multi-agent problems.  
Unfortunately, the state-of-the-art deep MARL methods~\cite{COMA,MAAC,QPLEX,MADDPG,DecHDRQN,WQMIX,QMIX,QTRAN,VDAC,DOP,ROMA} are not truly applicable to solve large and long-horizon multi-agent tasks in the real-world, because they are originally developed for cases where agents synchronously execute primitive-actions at every time step.  

Synchronizing decisions across multiple agents in realistic settings is problematic, since it requires agents to wait for each other's termination in order to make their next decisions together, which indicates that agents have to communicate about termination reliably during online execution.  
Such a synchronization over agents potentially restricts the efficiency, flexibility, and robustness of the entire system.
Many real-world multi-agent tasks are complex in terms of requiring a set of subtasks to be finished and perhaps involving heterogeneous agents. 
Agents may split into a number of groups or act in a number of roles to focus on different subtasks but still collaborate as a whole. 
Synchronizing all agents is not a practical manner, and instead, agents should operate in an asynchronous way.  
Moreover, real-world tasks are often long-horizon so it is intractable to solve them using the methods that only allow agents to perform synchronized learning and execution at the primitive level. 
To have scalable solution methods for solving these problems, it is also promising to incorporate hierarchical structure into agents' learning and execution.
Problem decomposition and sub-task allocation can be attained through hierarchies to reduce complexity, and meanwhile, asynchronization over agents can be directly supported in the manner that agents select high-level durative choices at different time steps.


\begin{figure}[t!]
    \centering
    \subcaptionbox{\label{intro_a}Robots serve two humans with tool delivery.}
        [0.51\linewidth]{\includegraphics[width=8cm, height=4.4cm]{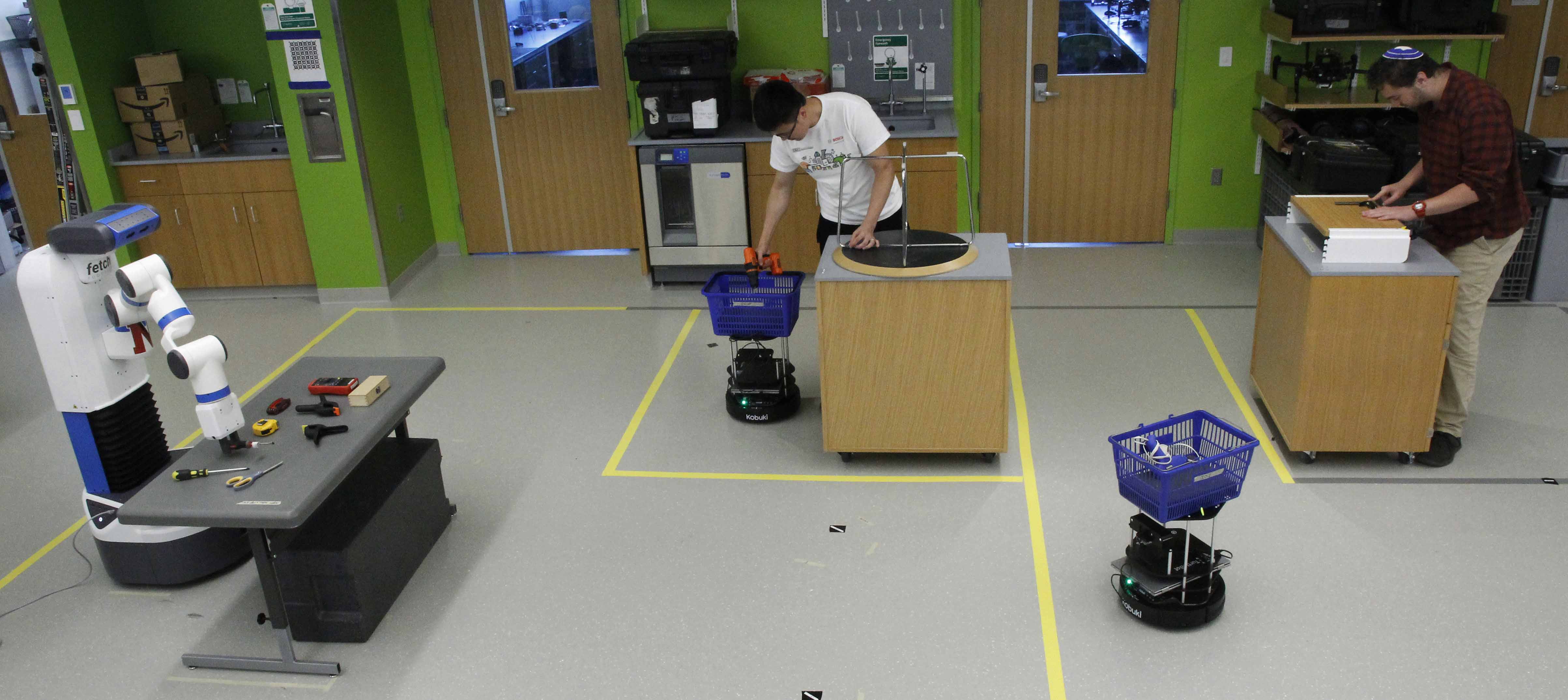}}
    ~
    \centering
    \subcaptionbox{\label{intro_b}Tool Passing.}
        [0.20\linewidth]{\includegraphics[height=4.4cm]{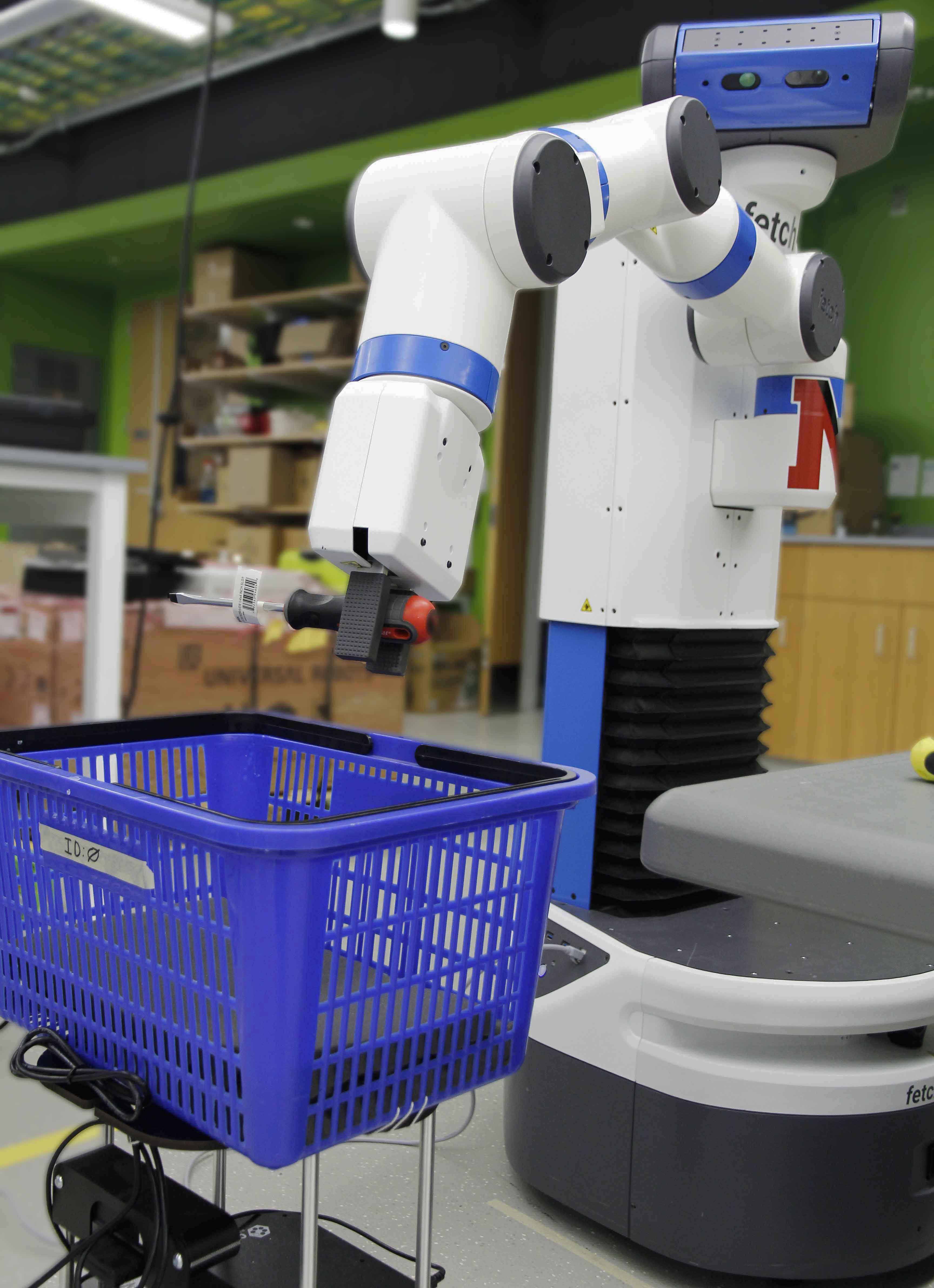}}
    ~
    \centering
    \subcaptionbox{\label{intro_c}Tool Delivery.}
        [0.20\linewidth]{\includegraphics[height=4.4cm]{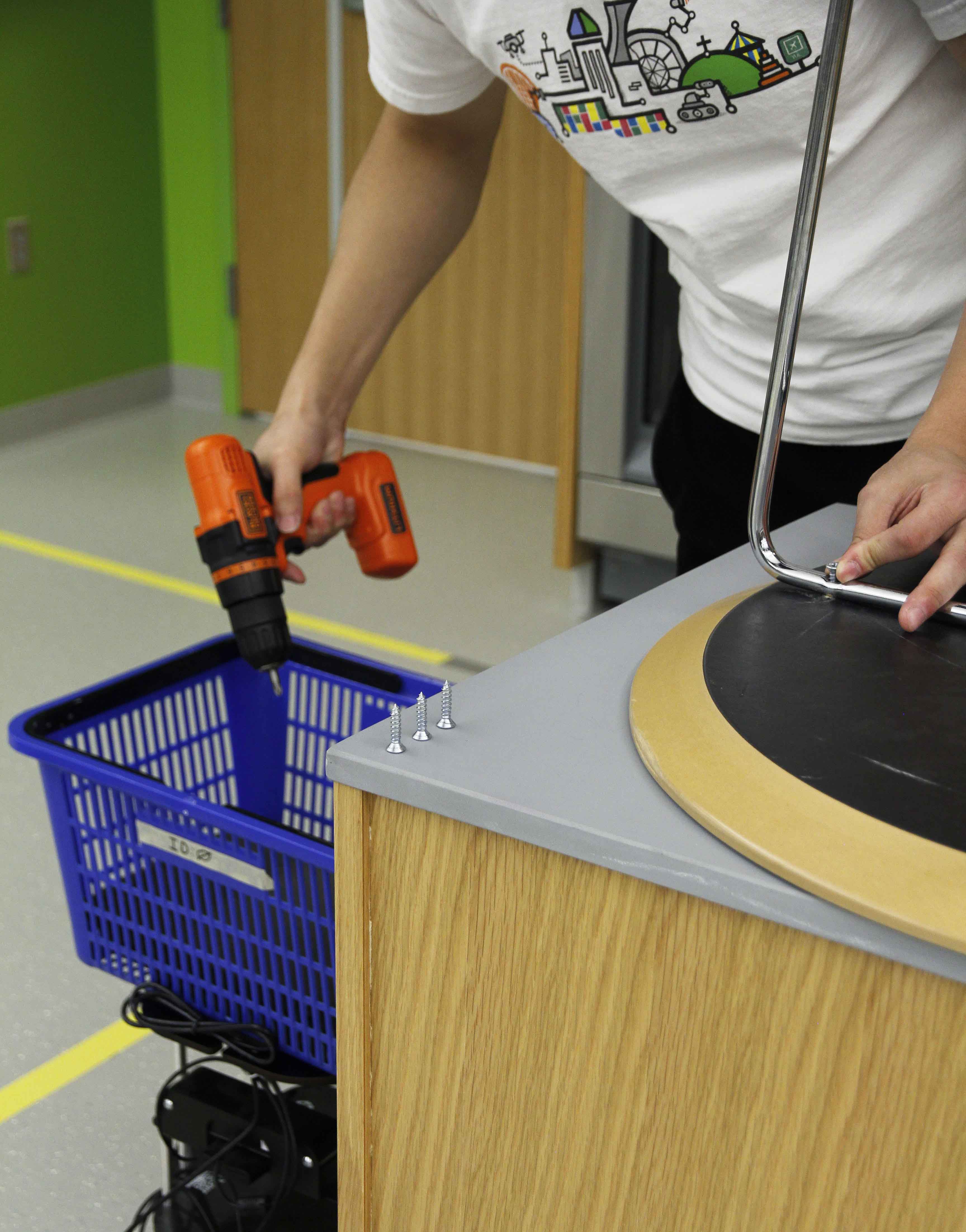}}
    \caption{Example of a real-world multi-robot tool delivery task.} 
    \label{chap:intro:example}
\end{figure}

Consider in an assembly room (Fig.~\ref{intro_a}), a team of autonomous robots is assisting two humans with tool delivery service. 
In order to support humans more efficiently, robots have to be able to predict potential tools each human will need and collaborate with each other to perform tool searching on a table (Fig.~\ref{intro_a}), tool passing (Fig.~\ref{intro_b}), and tool delivery (Fig.~\ref{intro_c}). 
Learning high-quality cooperative behavior for this task in such a large, stochastic, and uncertain environment is challenging for robots, especially when humans work in separate rooms and communication is limited, where each robot must behave in a decentralized way according to only local information. 
When robots use current state-of-the-art methods to solve this task, they have to synchronously learn and reason about collaboration at the motor-control level, which actually makes the problem even more challenging to be solved.
An essential fact is that robots have often been equipped with controllers for navigation, manipulation, and detection, and they do not need to learn these skills again in multi-robot settings.
Instead, a more promising way is to allow agents to reason about collaboration over high-level choices (e.g., grasp a tool on the table, pass a tool to a mobile robot, or deliver a tool to a human) that can be modeled by using existing controllers. 
As these high-level actions usually take different amounts of time to complete, agents' decision-making at the high-level appears asynchronous such that they can start and end their own high-level actions at different time steps. 
Thus, there is a great need for another set of learning frameworks considering such asynchronicity over agents.

Typically, fully cooperative multi-agent problems under a partially observable environment with actions taking various durations are modeled as \emph{Macro-Action Decentralized Partially Observable Markov Decision Processes} (MacDec-POMDPs).
Macro-Actions, as an instance of temporally-extended actions, can represent high-level robot controllers (e.g., navigation to a waypoint or grasping an object) and are naturally able to be selected and terminated by agents in an asynchronous way.
Planning methods have been developed for MacDec-POMDPs and their usefulness has been demonstrated in realistic robotics problems (e.g., warehouse box pushing~\cite{ICRA15MacDec}, bartender and waiters~\cite{RSS15}, package delivery~\cite{IJRR17DecPOSMDP,GDICE}, and health-aware disaster relief~\cite{MacroObsICRA17}).
However, so far, learning methods for MacDec-POMDPs are limited to the case where each agent's decentralized policy is represented as a finite-state-controller (FSC), and the objective is to learn the parameters of distributions over actions and transitions for each node via expectation-maximization algorithms~\cite{MiaoAAAI16,MiaoIROS17}.
More importantly, unlike the general online model-free MARL settings, these learning methods rely on a batch of pre-collected data without letting agents continue interacting with the world during training.  

Therefore, the objective of this thesis is to develop principled and scalable model-free MARL algorithms for MacDec-POMDPs that are feasible for solving large, long-horizon, and realistic multi-agent/robot domains and robust to software and hardware uncertainties. Although the work is based on MacDec-POMDPs, the proposed approaches would be applicable for other models (e.g., Dec-POSMDP~\cite{IJRR17DecPOSMDP}) with temporally-extended actions. 

\section{Literature Review}
\label{chap:intro:related_work}

\noindent \textbf{Cooperative Multi-Agent Reinforcement Learning.} 
The \emph{Decentralized Partially Observable Markov Process} (Dec-POMDP)~\cite{Oliehoek}  offers a general framework to model multi-agent cooperative domains, where agents individually make local decisions while considering uncertainty in action outcomes, noisy sensing, and information about other agents. 
Many MARL methods have been developed for Dec-POMDPs.
\emph{Independent learning} (IL) framework is the simplest solution allowing each agent to learn a decentralized policy in Dec-POMDPs~\cite{COMA,tan1993multi}.
Although IL may sometimes work well, it encounters a crucial theoretical issue: the environment becomes non-stationary from each agent's perspective as other agents explore and update policies.
This so-called environmental non-stationarity is known to generate a high variance on value and gradient estimations and impedes agents from collaborating well.
An extreme way to address this is to learn a centralized policy. However, centralized control is not preferable due to the strict requirement on fast and perfect online communications over agents, which is often difficult to be maintained in real-world settings. 
Recently, \emph{centralized training for decentralized execution} (CTDE)~\cite{OliehoekSV08,KraemerB16} paradigm attracts the most attention in MARL community, as it handles offline training by utilizing global information while allowing online execution in a decentralized way based on only local information. 
Current MARL methods mainly implement CTDE in two patterns: one is based on the actor-critic framework that learns a centralized critic to calculate each actor's policy gradients~\cite{COMA, MAAC, MADDPG,VDAC, DOP,  SQDDPG, LIIR, LICA, CM3}; the other relies on factorizing centralized Q-value to each decentralized Q-value via networks with variant types of constraints~\cite{QPLEX,WQMIX,QMIX,ROMA,vdn18,MAVEN}. 
Despite these methods having generated remarkable solutions on common benchmarks~\cite{MADDPG,smac}, they are still not feasible for long-horizon large-scale real-world multi-agents/robots tasks. 
As discussed above, the vital reason is that these methods require agents to synchronously execute primitive-actions, which actually contradicts the asynchronous nature of multi-robot systems in practice. 
Also, when collaborations consist of long-horizon sequential actions over agents, it would be super intractable to reason about these at such a low-level at every time step.          

This thesis aims at developing scalable MARL frameworks by allowing agents to perform asynchronous learning and execution in a hierarchical manner. In the following paragraphs, we focus on highlighting the differences between the contributions of this thesis and other existing multi-agent hierarchical reinforcement learning frameworks. 

\textbf{Multi-Agent Hierarchical Reinforcement Learning}. 
To scale up learning in MARL problems, 
hierarchy has been introduced into multi-agent scenarios.
One line of hierarchical MARL is still focusing on learning primitive-action-based policy for each agent while leveraging a hierarchical structure to achieve knowledge transfer~\cite{tianpei:neurips21}, credit assignment~\cite{AhilanFedual}, and low-level policy factorization over agent~\cite{OptionResq}.   
In these works, as the decision-making over agents is still limited at the low-level, none of them has been evaluated in large-scale realistic domains. 
Instead, by having macro-actions, our methods equip agents with the potential capability of exploiting abstracted skills, sub-task allocation, and problem decomposition via hierarchical decision-making, which is critical for scaling up to real-world multi-robot tasks. 

Another line of the research allows agents to learn both a high-level policy and a low-level policy, but the  methods either force agents to perform a high-level choice at every time step~\cite{schroeder:nips19,MAIntro-OptionQ} or require all agents' high-level decisions
have the same time duration~\cite{NachumAPGK19,WangK0LZITF20,wang:iclr2021,HAVEN,JiachenMAHRL}, where agents are actually synchronized at both levels. In contrast, our frameworks are more general and applicable to real-world multi-robot systems because they allow agents to asynchronously execute at a high-level without synchronization or waiting for all agents to terminate. 

Recently, some asynchronous hierarchical approaches have been developed. \cite{wu2021spatial} extends Deep Q-Networks~\cite{DQN} to learn a high-level pixel-wise spatial-action-value map for each agent in a fully decentralized learning way. 
Our work, however, accepts any representations of high-level actions.
\cite{MendaCGBTKW19} frame multi-agent asynchronous decision-making problems as event-driven processes with one assumption on the acceptable of losing the ability to capture low-level interaction between agents within an event duration and the other on homogeneous agents, but our frameworks rely on the time-driven simulator used for general multi-agent and single-agent RL problems and do not have the above assumptions. 
\cite{DOC} adapt a single-agent~\emph{option-critic} framework~\cite{OptionAC} to multi-agent domains to learn all components (e.g., low-level policy, high-level abstraction, high-level policy) from scratch, but learning at both levels is difficult and the proposed method does not perform well even in small TeamGrid~\cite{TeamGrid} scenarios. 
More important to note is that 
none of the existing works provides a general asynchronous and hierarchical multi-agent reinforcement learning framework to solve multi-agent problems with macro-actions under partial observability.

\section{Thesis Contributions}
\label{chap:intro:con}

\noindent In this thesis, we formulate the first set of value-based learning frameworks and the first set of policy-gradient-based learning frameworks for MacDec-POMDPs.
The resulting algorithms allow agents to asynchronously learn and execute in three different manners: decentralized learning and control, centralized learning and control, and centralized training for decentralized execution (CTDE). 
Similar to how primitive-actions are defined for an agent in conventional RL problems, this thesis assumes a set of macro-actions has been given for the agents, which is consistent with the fact that many controllers have been developed for real robots. 
Our contributions then aim to allow agents to asynchronously learn and execute high-level policies over macro-actions. 

\textbf{Macro-Action-Based Decentralized Q-Learning}. We first propose a decentralized learning framework that allows agents to asynchronously learn decentralized macro-action-value functions using Deep Q-Learning (DQN)~\cite{DQN} while relying on a new replay buffer called Macro-Action Concurrent Experiences Relay Trajectories (Mac-CERTs). 
Mac-CERTs maintain high-level transition information for agents and involve an independent reward accumulation mechanism depending on each agent's own macro-action execution status.  
Asynchronous updates over agents are achieved in the way that each agent accesses its own sequential experiences and performs \emph{temporal-difference} (TD) updates when its own macro-action terminates.
This contribution offers a principled approach for learning each agent's macro-action values and processing asynchronous information over agents from a decentralized perspective.  

\textbf{Macro-Action-Based Centralized Q-Learning}. We also propose a centralized learning framework that is based on DQN and allows agents to learn a centralized macro-action-value function while generating Macro-Action Joint Experiences Relay Trajectories (Mac-JERTs) in a new relay buffer. 
MacJERTs accumulate rewards for each joint macro-action depending on its termination defined as the time step when $any$ agent's macro-action ends.     
Updates for the centralized macro-action value function only occur at the termination step of each joint macro-action, but more importantly, a new TD-loss with a conditional target prediction is proposed to allow the learner to properly consider agents' asynchronous macro-action execution. 
This contribution offers a principled approach for learning joint macro-action values and processing asynchronous information over agents from a centralized perspective.  

\textbf{Macro-Action-Based CTDE Q-Learning}. To improve the quality of macro-action-based decentralized policies for large and complex multi-robot problems, we develop a Macro-Action-Based Decentralized Double Deep Recurrent Q-Net (MacDec-DDRQN), the underlying idea is to train each decentralized Q-net using a centralized Q-net for target macro-action selection in the TD updating. 
The potential benefits of the way of using the centralized Q-net include removing maximum bias, ameliorating the effect of environmental non-stationarity, and facilitating better collaboration. 
A variant called Parallel-MacDec-DDRQN with two separate training environments is also designed to further improve the learning performance in certain classes of problems.
This is the first work that successfully incorporates macro-actions into a CTDE paradigm and deploys the learned decentralized policies on a real-world multi-robot system.

Policy gradient methods have different theoretical properties and can fit better with different types of tasks than value-based methods, such as being more scalable in the action space. Motivated by this evidence, this thesis also contributes to formulating the first set of macro-action-based multi-agent actor-critic algorithms as follows.   

\textbf{Macro-Action-Based Independent Actor-Critic}. We first incorporate Mac-CERTs into an independent actor-critic framework, referred to Mac-IAC, where each agent can learn an on-policy macro-action-value function as the critic and independently update its parameterized policy when the corresponding macro-action terminates.  
The proposed framework offers a general way to adapt any primitive-action-based independent actor-critic learning algorithm (e.g., IPPO~\cite{IPPO}) to train decentralized policies over macro-actions.
Mac-IAC naturally allows fully online learning and may also be able to generate high-quality solutions for certain domains.  

\textbf{Macro-Action-Based Centralized Actor-Critic}. We also incorporate Mac-JERTs into an actor-critic framework, referred to as Mac-CAC, where agents learn an on-policy joint macro-action-value function as the centralized critic and optimize a parameterized centralized policy when each joint macro-action terminates. 
Mac-CAC may be still preferred when online full communication is available and can certainly act as an important baseline for performance analysis in comparisons.

\textbf{Macro-Action-Based Independent Actor with Individual Centralized Critic}. Although in the case with primitive-actions, independent actor with a centralized critic (IACC) has become the most popular implementation format of CTDE with policy gradients, it is particularly challenging in macro-action-based multi-agent settings. 
It is hard to determine what type of centralized critic to use and how to learn it for decentralized policy optimization, because there is a significant difference in macro-action execution status and the corresponding cumulative reward between the decentralized perspective and the centralized perspective. 
To address this challenge, we first show that naively incorporating macro-actions into IACC leads to unstable learning and performs worse when the asynchronicity over agents' macro-action execution gets more complex; 
and then we propose a new algorithm called Mac-IAICC that allows agents to leverage centralized information by learning an individual centralized critic but trained with respect to each own macro-action execution.
Mac-IAICC achieves the best performance in solving a variety of large and realistic multi-agent domains, and the learned decentralized policies successfully solve a large and long-horizon real-world multi-robot tool delivery task.     

The contributions of this thesis build up a foundation for future macro-action-based MARL algorithm development, including other work on asynchronous and hierarchical multi-agent decision-making, to solve even larger and more complex problems with various sources of uncertainty.   

\section{Overall Structure}
\label{chap:intro:str}

The thesis is organized in the following structure:

\begin{itemize}
    \item{Chapter~\ref{chap:BG} provides an overview of sequential decision-making models for both single-agent and multi-agent problems as well as the corresponding important reinforcement learning frameworks.}
    \item{Chapter~\ref{chap:paper1} presents the first principled formulations for learning macro-action-value functions in fully decentralized and fully centralized manners based on deep Q-networks, and shows the details of two new replay buffers designed for each case accordingly. Experimental evaluations demonstrate the effectiveness of our methods in dealing with agents' asynchronous macro-action execution and the advantage of learning with macro-actions over primitive-actions and the scalability of our methods. The technical content of this chapter was first published in the paper: "Macro-Action-Based Deep Multi-Agent Reinforcement Learning", Yuchen Xiao, Joshua Hoffman, Christopher Amato, in the Conference on Robot Learning (CoRL), 2019~\cite{xiao_corl_2019}.}
    \item{Chapter~\ref{chap:paper2} introduces the first macro-action-based CTDE algorithm that uses a centralized Q-net in the optimization of decentralized Q-nets. Experimental evaluations validate the efficiency and the practical nature of the proposed method by achieving near-centralized results in simulation and having real robots accomplish a warehouse tool delivery task in an efficient way. The technical content of this chapter was first presented in the paper: "Learning Multi-Robot Decentralized Macro-Action-Based Policies via a Centralized Q-Net", Yuchen Xiao, Joshua Hoffman, Tian Xia, Christopher Amato, in the International Conference on Robotics and Automation (ICRA), 2020~\cite{xiao_icra_2020}.}
    \item{Chapter~\ref{chap:paper3} describes a set of asynchronous multi-agent actor-critic methods that allow agents to directly optimize asynchronous (macro-action-based) policies in three training paradigms: decentralized learning, centralized learning, and centralized training for decentralized execution. Experimental evaluations (in simulation and on hardware) in a variety of realistic domains confirm the superiority of the proposed methods for learning high-quality and asynchronous solutions in large multi-agent problems. The technical content of this chapter was first discussed in the paper: "Asynchronous Actor-Critic for Multi-Agent Reinforcement Learning", Yuchen Xiao, Weihao Tan, Christopher Amato, in the Conference on Neural Information Processing Systems (NeurIPS), 2022~\cite{xiao_neurips_2022}.}
    \item{Chapter~\ref{chap:conclude} summarizes the contributions of this thesis and also discusses potential research topics for future work.}
    \item{The Appendix displays the visualization of learned behaviors under the considered domains, and lists the hyper-parameters used for generating the presented results.}

\end{itemize}



\chapter{Background}
\label{chap:BG}

\noindent In this chapter, we first present formal definitions of models for single-agent and multi-agent decision-making problems. We also discuss the representative single-agent deep reinforcement learning algorithms and provide an overview of popular multi-agent deep reinforcement learning paradigms in the end.  

\section{Markov Decision Processes}
\label{chap:BG:MDPs}

\noindent Single-agent sequential decision making in fully observable environments can be modeled as a Markov decision process (MDP), defined as a tuple $\langle S, A, T, O, R \rangle$~\cite{MDPs}, where:

\begin{itemize}
        \vspace{-1mm}
        \item{$S$ is a finite set of environment states;} 
            \vspace{-1mm}
        \item{$A$ is a set of primitive-actions;}
            \vspace{-1mm}
        \item{$T$ is a state transition probability function, $T: S\times A \times S \rightarrow [0,1]$, that indicates the probability of transitioning from a state $s$ to a next state $s'$ if a primitive-action $a$ is taken, represented as $T(s, a, s') = P(s'\mid s, a)$;}
            \vspace{-1mm}
        \item{$R$ is a reward function, $R: S\times A \rightarrow \mathbb{R}$, that assigns a immediate reward $r(s,a)$ for taking a primitive-action $a$ in a state $s$.} 
\end{itemize}

The objective of MDP solution methods is to find a policy $\pi(s\mid a)$: $S \times A \rightarrow [0,1]$ that maximizes the expected cumulative discounted reward from a start state $s_0$, denoted as:

\begin{equation}
    V^{\pi}(s_{(0)})=\mathbb{E}\Biggr[\sum_{t=0}^{\infty}\gamma^tr(s_{(t)},a_{(t)})| s_{(0)}, \pi\Biggr]
\end{equation}
where, $\gamma\in[0,1)$ is a discount factor to maintain finite sums, and $s_0$ is associated with a designed initial state distribution $b_0$.

\section{Partially Observable Markov Decision Processes}
\label{chap:BG:POMDPs}

\noindent The POMDP~\cite{Kaelbling:L}, as an extension of the MDP, provides a framework for sequential decision-making problems with uncertainty in observations as well as action outcomes. 
Formally, a POMDP can be defined as a tuple $\left\langle S, A, \Omega, T, O, R\right\rangle$, where: 

\begin{itemize}
        \vspace{-1mm}
    \item{$S$ is a set of environment states;} 
        \vspace{-1mm}
    \item{$A$ is a set of primitive-actions;} 
        \vspace{-1mm}
    \item{$\Omega$ is a set of primitive-observations;}
        \vspace{-1mm}
    \item{$T$ is a state transition probability function, $T: S\times A \times S \rightarrow [0,1]$, that indicates the probability of transitioning from a state $s$ to a next state $s'$ if a primitive-action $a$ is taken, represented as $T(s, a, s') = P(s'\mid s, a)$;}
        \vspace{-1mm}
    \item{$O$ is an observation function, $O: S \times A \times \Omega \rightarrow [0,1]$, that is the probability of observing $o$ in a resulting state $s'$ after taking an action $a$, represented as $O(s', a, o) = P(o\mid s', a)$;}
        \vspace{-1mm}
    \item{$R$ is a reward function, $R: S\times A \rightarrow \mathbb{R}$, that assigns a immediate reward $r(s,a)$ for taking a primitive-action $a$ in a state $s$.} 
\end{itemize}

Because the world is partially observable, the agent lacks access to the underlying state $s$. 
Instead, the agent keeps track of an observation-action history $h$ to make a decision at every time step.
We consider the case where the agent's policy is defined as $\pi(a\mid h): H \times A \rightarrow [0,1]$, a mapping from history space to action space.
Accordingly, the objective of POMDP solution methods is to find a policy that optimizes the expected sum of discounted rewards from an initial state: 
\begin{equation}
    V^{\pi}(s_{(0)}) = \mathbb{E}\Biggl[\sum_{t=0}^{\infty}\gamma^t r(s_{(t)}, a_{(t)}) \mid s_{(0)}, \pi \Biggr],
\end{equation}
where the discount factor $\gamma \in [0, 1)$ determines the impact of future rewards on current decision-making, and the initial state is determined by a distribution $b_0$.

\section{Decentralized POMDPs}
\label{chap:BG:Dec-POMDPs}

\noindent In fully cooperative decentralized multi-agent domains with both state and outcome uncertainties, each agent must choose actions individually based on local observations. This setting is described as a decentralized partially observable Markov decision process (Dec-POMDP)~\cite{Oliehoek}. 

More formally, a Dec-POMDP is represented as a tuple $\langle I, S, A, \Omega, T, O, R \rangle$, where: 
\begin{itemize}
            \vspace{-1mm}
        \item{$I$ is a finite set of agents;} 
            \vspace{-1mm}
        \item{$S$ is a finite set of environment states;} 
            \vspace{-1mm}
        \item{$A= \times_{i\in I} A_i$ is the joint primitive-action space over each agent's primitive-action set $A_i$;} 
            \vspace{-1mm}
        \item{$\Omega = \times_{i\in I} \Omega_i$ is the joint primitive-observation space over each agent's primitive-observation set $\Omega_i$;} 
            \vspace{-1mm}
        \item{$T$ is a state transition probability function, $T: S\times A \times S \rightarrow [0,1]$, that indicates the probability of transitioning from a state $s$ to a next state $s'$ after agents take a joint primitive-action $\vec{a}$, represented as $T(s, \vec{a}, s') = P(s'\mid s,\vec{a})$;}
            \vspace{-1mm}
        \item{$O$ is an observation probability function, $O: \Omega\times A \times S \rightarrow [0,1]$, that denotes the probability of receiving a joint primitive-observation $\vec{o}$ when a joint primitive-action $\vec{a}$ were taken and arriving in state $s'$, represented as $O(\vec{o},\vec{a},s')=P(\vec{o}\mid \vec{a}, s')$;} 
            \vspace{-1mm}
        \item{$R$ is a reward function, $R: S\times A \rightarrow \mathbb{R}$, that assigns a shared immediate reward $r(s,\vec{a})$ to agents for taking a joint primitive-action $\vec{a}$ in a state $s$.} 
\end{itemize}

At every time step, agents synchronously execute a joint primitive-action $\vec{a}=\times_{i\in I} a_i$, each individually selected by an agent using a policy $\pi_i: H^A_i\times A_i\rightarrow [0,1]$, a mapping from local primitive-observation-action history $H^A_i$ to primitive-actions. 
In finite-horizon Dec-POMDPs, the objective of solution methods is to find a joint policy $\vec{\pi}=\times_i\pi_i$ that maximizes the expected sum of discounted rewards starting from an initial state $s_0$ as:
\begin{equation}
    V^{\pi}(s_{(0)})=\mathbb{E}\big[\sum_{t=0}^{\mathbb{H}-1}\gamma^tr(s_{(t)},\vec{a}_{(t)})\mid s_{(0)}, \vec{\pi}\big]
\end{equation}
where, $\gamma\in[0,1]$ is a discount factor, and $\mathbb{H}$ is the number of (primitive) time steps until the problem terminates (the horizon).

\section{Macro-Action Decentralized POMDPs}
\label{chap:BG:MecDec-POMDPs}

\noindent To allow asynchronous execution among agents with temporally-extended actions that can begin and end at different time steps, the macro-action decentralized partially observable Markov decision process (MacDec-POMDP)~\cite{AAMAS14AKK,AmatoJAIR19} incorporates the \emph{option} framework~\cite{Sutton:1999} into the Dec-POMDP by defining each agent $i$'s macro-action as a tuple:
\begin{equation}
m_i=\langle I_{m_i}, \pi_{m_i}, \beta_{m_i}\rangle 
\end{equation}
consisting of an initiation set $I_{m_i}\subset H^M_i$ that defines how to initiate a macro-action based on macro-observation-action history $H^M_i$ at the high-level; 
a low-level policy $\pi_{m_i}:H^A_i\times A_i\rightarrow [0,1]$ for the execution of a macro-action; 
and a stochastic termination function $\beta_{m_i}:H^A_i\rightarrow[0,1]$ that determines how to terminate a macro-action based on primitive-observation-action history $H^A_i$ at the low-level.  
Formally, a MacDec-POMDP is defined as $\langle I, S, A, M, \Omega, \zeta, T, O, Z, R\rangle$, where 
\begin{itemize}
        \vspace{-1mm}
    \item{$I$, $S$, $A$, $\Omega$, $T$, $O$, $R$ are the same as defined in the Dec-POMDP;} 
        \vspace{-1mm}
    \item{$M=\times_{i\in I}M_i$ is the set of joint macro-actions with $M_i$ being a finite macro-action space for each agent $i$;}
        \vspace{-1mm}
    \item{$\zeta=\times_{i\in I}\zeta_i$ is the set of joint macro-observations with $\zeta_i$ being a finite macro-observation space for each agent $i$;}
        \vspace{-1mm}
    \item{$Z=\times_{i\in I} Z_i$ is a joint macro-observation likelihood model with $Z_i: \zeta_i\times M_i \times S \rightarrow[0,1]$ as the probability of agent $i$ receiving a macro-observation $z_i\in\zeta_i$ when it completes a macro-action $m_i\in M_i$ and arrives in a state $s'$. Hence, $Z(z_i,m_i,s')=P(z_i\mid m_i,s')$.}
\end{itemize}

During execution, each agent independently selects a macro-action $m_i$ using a high-level policy $\Psi_i:H^M_i\times M_i\rightarrow[0,1]$, a mapping from macro-observation-action history to macro-actions, and captures a macro-observation $z_i \sim Z_i(z_i,m_i,s')$ when the macro-action terminates in a state $s'$. 
The objective of solving MacDec-POMDPs with the finite horizon is to find a joint high-level policy $\vec{\Psi}=\times_{i\in I} \Psi_i$ that maximizes the value: 
\begin{equation}
V^{\vec{\Psi}}(s_{(0)})=\mathbb{E}\Big[\sum_{t=0}^{\mathbb{H}-1}\gamma^tr\big(s_{(t)},\vec{a}_{(t)}\big)\mid s_{(0)}, \vec{\pi}, \vec{\Psi}\Big] 
\end{equation}
with a discount factor $\gamma\in[0,1]$ and a problem horizon $\mathbb{H}$.

\section{Single-Agent Reinforcement Learning}
\label{chap:BG:RL}

\noindent In this thesis, we focus on model-free reinforcement learning (RL), where the agent aims to learn an optimal policy by interacting with the environment without explicit world models (e.g., $T$, $O$ and $R$) as prior knowledge. 
Model-free RL methods can be categorized into two classes: (a) value-based approaches that learn the values of actions and then select actions based on the learned values; and (b) policy gradient approaches that directly learn a parameterized policy to select actions.  
In this section, we provide a brief introduction to the representative single-agent deep RL algorithms over the two classes, where deep neural networks are used as function approximators and policies.

\subsection{DQN, Double-DQN, and DRQN}

\noindent Q-learning~\cite{Watkins1992} is a classical model-free RL algorithm to learn the optimal action-value function defined as:
\begin{align}
    Q^*(s,a) &= \max_{\pi}Q^{\pi}(s,a)\\
    &= \max_{\pi}\mathbb{E}_{\pi}\Biggr[\sum_{t'=t}^{\infty}\gamma^{t'-t} r_t \mid s_t=s, a_t=a \Biggr] 
\end{align}
The Bellman optimality equation for the optimal action-value function is written as: 
\begin{equation}
    Q^*(s,a) = \mathbb{E}_{s'}\biggr[r + \gamma\max_{a'}Q^*(s',a') \mid s,a\biggr]
    \label{boe}
\end{equation}
According to Eq.~\ref{boe}, Q-learning iteratively updates an action-value function $Q(s,a)$ to estimate the optimal one via a \emph{temporal-difference} (TD) control:
\begin{equation}
    Q^{k+1}(s,a) = Q^{k}(s,a) + \alpha \big[r + \gamma\max_{a'}Q^k(s',a') - Q^k(s,a)\big] 
\end{equation}
with a learning rate $\alpha\in(0,1)$ and a TD-error computed in the above brackets. Q-learning has a convergence guarantee to $Q^*(s,a)$ as all state-action pairs are visited and updated sufficiently. As the learned action-value function $Q(s,a)$ is independent of the policy being followed, Q-learning is an off-policy learning approach. Typically, during learning, the agent uses an $\epsilon$-greedy policy for choosing action randomly with a probability of $\epsilon$, which encourages the agent to continue exploring the environment. 

The Deep Q-Network (DQN) \cite{DQN} extends Q-learning to include a deep neural network as a function approximator. 
DQN learns $Q_\theta(s,a)$, parameterized with $\theta$, by minimizing the loss:
\begin{equation}
    \mathcal{L}(\theta)=\E_{<s, a, s' r>\sim\mathcal{D}}\Big[\big(y - Q_{\theta}(s, a)\big)^2 \Big]
\end{equation}
where,
\begin{equation}
y=r + \gamma\max_{a'}Q_{\theta^-}(s',a') 
\end{equation}
A target action-value function $Q_{\theta^-}$ ($\theta^-$ is an outdated copy of $\theta$) and an experience replay buffer $\mathcal{D}$~\cite{Lin1992} are implemented for stable learning. 

In order to deal with the maximization bias introduced by the maximization operation over estimated values, the idea behind Double Q-learning~\cite{DoubleQ} is generalized to DQN, called Double-DQN (DDQN)~\cite{DDQN}, by rewriting the target value calculation as:
\begin{equation}
y = r + \gamma Q_{\theta^-}(s', \argmax_{a'}Q_{\theta}(s',a')) 
\end{equation}
where, two Q-networks are used to reduce overestimations on action values.

Both DQN and DDQN assume full access to environment states, however, we consider environments to be partially observable. 
To address the partial observability, Deep Recurrent Q-Networks (DRQN)~\cite{DRQN} apply Long Short-Term Memory (LSTM) cells~\cite{LSTM} to maintain an internal hidden state from the agent's observation-action history $h$. The corresponding action-value function $Q_{\theta}(h,a)$ is then updated by minimizing the following loss: 
\begin{equation}
    \mathcal{L}(\theta)=\E_{<o, a, o' r>\sim\mathcal{D}}\Big[\big(y - Q_{\theta}(h, a)\big)^2 \Big]
\end{equation}
where,
\begin{equation}
y=r + \gamma\max_{a'}Q_{\theta^-}(hao',a') 
\end{equation}

In our work, we extend DDQN with a recurrent layer, called DDRQN, to learn macro-action-value functions that allow agents to perform asynchronous and hierarchical decision-making. 
This is done for both decentralized learning and centralized learning in Chapter~\ref{chap:paper1}. 

\subsection{Actor-Critic Policy Gradient}

\noindent Policy gradient is another popular reinforcement learning technique with the aim of directly optimizing a parameterized policy $\pi_{\theta}$ by performing gradient ascent on the policy's performance defined as $J(\theta) = \mathbb{E}_{\pi_{\theta}}[\sum_{t=0}^{\infty}\gamma^tr_t]$.
Based on the single-agent \emph{policy gradient theorem}~\cite{Sutton:1999} in MDPs, we can simply adapt it to POMDPs by having the following on-policy gradient with respect to a policy's parameters:
\begin{equation}
    \nabla_\theta J (\theta)=\mathbb{E}_{\pi_{\theta}}[\nabla_\theta\log\pi_\theta(a\mid h)Q^{\pi_\theta}(h,a)]
    \label{pg}
\end{equation}

In \emph{actor-critic} framework~\cite{konda2000actor}, the on-policy action-value $Q^{\pi_\theta}$ is approximated by learning an action-value function $Q^{\pi_\theta}_\phi$ (critic) via \emph{temporal-difference} (TD) learning, and the policy $\pi_\theta$ (actor) is optimized by following the gradient in Eq~\ref{pg}.   
Policy gradient methods often involve high variance in gradient estimation. 
To reduce the variance, people often train a state-value function in MDPs but a history-value function $V^{\pi_\theta}_{\mathbf{w}}(h)$ in POMDPs as the critic, and
this critic is used to provide a one-step bootstrap estimation and as a baseline~\cite{Sutton:1999}, which ends up with an \emph{advantage actor-critic} (A2C) policy gradient that can be written as:  

\begin{equation}
    \nabla_{\theta}J(\theta) = \mathbb{E}_{\pi_{\theta}}\Big[\nabla_\theta\log\pi_\theta(a\mid h)A(a,h)\Big]
    \label{a2c}
\end{equation}
where, 
\begin{equation}
    A(h,a) = r(h,a) + V^{\pi_\theta}_\mathbf{w}(h') - V^{\pi_\theta}_\mathbf{w}(h)
    \label{a2c:adv}
\end{equation}
In this thesis, critics in all policy gradient methods are trained by minimizing the following $n$-step TD loss:

\begin{equation}
    \mathcal{L}(\mathbf{w})=\E_{\pi_\theta}\Big[\big(y_t - V^{\pi_\theta}_{\mathbf{w}}(h_t)\big)^2 \Big]
\end{equation}
where
\begin{equation}
    y_t=\sum_{t'=t}^{t+n-1}\gamma^{t'-t}r_t + \gamma^nV^{\pi_\theta}_{\mathbf{w}^-}(h_{t+n})
\end{equation}
and a target network with parameters $\mathbf{w}^-$ periodically copied from $\mathbf{w}$. This $n$-step bootstrap return is also implemented in the advantage value estimator (Eq.~\ref{a2c:adv}).

\section{Multi-Agent Reinforcement Learning}

\noindent In multi-agent reinforcement learning (MARL), there are multiple agents interacting with the same environment by perceiving input and selecting actions as well as considering the effect of each other in order to optimize each own decisions. Across this thesis, we consider full cooperative settings under partial observability, where the objective of agents is to maximize the global return. In this section, we introduce three standard MARL training paradigms: centralized learning, decentralized learning, and centralized training for decentralized execution (CTDE), and we also discuss the corresponding algorithms under each paradigm.   

\subsection{Centralized Learning and Control}

\noindent Perhaps the most straightforward way to solve cooperative MARL problems is fully centralized learning and control. 
In particular, we treat all agents as a single big agent to learn a centralized policy $\bm{\pi}(\vec{a}\mid\vec{h})$, a mapping from the joint observation-action history space to the joint action space, and then all the single-agent RL algorithms can be directly applied here.  

In theory, with access to the joint information over agents, the centralized policy learned in this training paradigm is guaranteed to converge to the global optimal behavior.
However, in practice, this framework suffers from two fundamental challenges: (a) the joint action space exponentially increases with respect to the number of agents, which potentially causes learning to be very slow and likely traps into a local optimum due to approximation error on action-values; (b) in order to perform centralized control, it is necessary to guarantee fast and perfect online communication over agents, which is often impossible to be achieved in many real-world settings. 

\subsection{Decentralized Learning and Control}
\label{chap:BG:MARL:Dec}

\noindent Because of the aforementioned issues in the fully centralized case, having a decentralized policy for each agent is preferable, where each agent independently makes decisions based on only local information.   

{\bf Independent Q-Learning (IQL)}~\cite{tan1993multi} is the simplest approach to learn decentralized policies for agents. It extends Q-learning to multi-agent scenarios by allowing each agent to independently learn its own action-value function. In the version with DQN~\cite{IDQL}, each agent $i$'s action-value function is annotated as $Q_{\theta_i}(s,a_i)$. 
In partially observable environments, we can use the DRQN to represent the action-value function as $Q_{\theta_i}(h_i,a_i)$ for each agent $i$, where $h_i$ indicates each agent's local observation-action history.
While decentralized policies can be directly learned in this simple decomposition manner, each independent learner suffers from several innate limitations: 
the difficulty in achieving efficient credit assignment as each agent maintains Q-values only for individual actions but receives global rewards depending on joint actions; 
the dilemma of the environmental non-stationarity from a local perspective caused by the existence of other learning agents; 
and the tendency to settle at a local optimum (a shadowed equilibrium~\cite{FuldaV07}) due to rare information sharing over agents, resulting in agents' local best choices to be suboptimal system behavior.

{\bf Decentralized Hysteresis DRQN (Dec-HDRQN)}~\cite{DecHDRQN} is one representative decentralized learning method to improve solution quality in Dec-POMDPs.  
It combines Hysteretic Q-learning~\cite{HQL} with DRQN, where each agent uses two learning rates $\alpha$ and $\beta$ to update an individual action-value function such as: 

\begin{equation}
    Q_{\theta_i}(h_i,a_i) = 
    \begin{cases}
        Q_{\theta_i}(h_i,a_i) + \beta\delta & \text{if } \delta\le 0\\
        Q_{\theta_i}(h_i,a_i) + \alpha\delta & \text{otherwise}
    \end{cases}
\end{equation}
when the TD error $\delta=r+\gamma\max_{a_i'} Q_{\theta_{i^-}}(h_i',a_i') - Q_{\theta_i}(h_i,a_i)$ is negative, a smaller learning rate $\beta$ is used, and $\alpha$ is a normal learning rate used otherwise. 
This facilitates multi-agent learning by making each agent robust against negative updating due to teammates' mistakes. 
Meanwhile, a new replay buffer called \emph{Concurrent Experience Replay Trajectories} (CERTs) 
is introduced to assist with the non-stationarity issue, by sampling concurrent experiences for training, which encourages each agent's policy to be optimized in the same direction. 

{\bf Independent Actor-Critic (IAC)}~\cite{COMA} is a straightforward extension of the single-agent A2C to multi-agent cases. Similar to IQL, in this framework, each agent independently optimizes its own actor $\pi_{\theta_i}$ and critic $V^{\pi_{\theta_i}}_{\mathbf{w}_i}$ purely based on local experiences.
Accordingly, the independent policy gradient is formulated as:

\begin{equation}
    \nabla_{\theta_i}J(\theta_i) = \mathbb{E}_{\vec{\pi}_{\theta}}\Big[\nabla_{\theta_i}\log\pi_{\theta_i}(a_i|h_i)A(a_i,h_i)\Big]
    \label{ia2c}
\end{equation}
where,
\begin{equation}
A(a_i,h_i) = r + V^{\pi_{\theta_i}}_{\mathbf{w}_i}(h_i') - V^{\pi_{\theta_i}}_{\mathbf{w}_i}(h_i)
\end{equation}
and a shared reward $r$ over agents is assigned by the global reward function $R$.  
Although IAC may sometimes work in practice, it still suffers from the same inherent issues mentioned above in IQL.
Nevertheless, an essential attribute of independent learning is that agents are able to conduct fully online learning. 

\subsection{Centralized Training for Decentralized Execution}
\label{chap:BG:CTDE}

\noindent In recent years, centralized training for decentralized execution (CTDE) has shown considerable promise in learning high-quality decentralized policies in Dec-POMDPs.
To address the main difficulties encountered in independent learning, CTDE provides agents with access to global information during offline training while maintaining decentralized online execution based on local information. 
This paradigm is potentially more feasible to solve real-world multi-agent tasks, where the policies are first trained in a simulator and then deployed on the real system.   
   
{\bf Value Function Factorization} has become a standard implementation of CTDE that decouples a joint Q-value function into individual Q-value functions as each agent's policy~\cite{QPLEX,WQMIX,QMIX,ROMA,MAVEN} in order to avoid the exponential size of the joint action space. This kind of architecture, as a result, is more scalable to large multi-agent problems in terms of the number of agents. 
More concretely, each agent optimizes its own Q-net by minimizing the following TD loss of a joint but factored Q-net:  
\begin{equation}
    \mathcal{L}(\vec{\theta},\psi)=\mathbb{E}_D\Big[\big(y^{tot} - Q^{tot}_{\vec{\theta},\psi}(\mathbf{x}, \vec{h}, \vec{a})\big)^2 \Big]
\end{equation}
where,
\begin{equation}
    Q^{tot}_{\vec{\theta},\psi}(\mathbf{x}, \vec{h}, \vec{a}) = f_{\psi}\big(\mathbf{x}, \{Q_{\theta_i}(h_i,a_i)\}_{i=1}^N\big)
\end{equation}
\begin{equation}
    y^{tot} = r + \gamma\max_{\vec{a}\,'} Q^{tot}_{\vec{\theta}^-, \psi^-}(\mathbf{x}', \vec{h}', \vec{a}\,') 
\end{equation}
and $\mathbf{x}$ represents extra accessible global signals (e.g., the environment state ). 
However, it is important to note that, in these methods, the function $f_\psi$ enforces a particular constraint on the relationship between the centralized Q-values and decentralized Q-values (e.g., a linear summation constraint, a non-linear monotonic constraint, or other weighted constraints). 
These constraints actually place different representational limitations on joint Q-value functions, such as the true joint Q-value function of a given domain cannot be represented with these constraints.

{\bf Independent Actor with Centralized Critic (IACC)} is another widespread exploitation of the CTDE paradigm.
The state-of-the-art MARL policy gradient approaches~\cite{COMA,MAAC,MADDPG,VDAC,DOP,SQDDPG,LIIR,LICA,CM3} utilize IACC in variant ways and have achieved significant successes in solving many challenging multi-agent tasks. 
The vital idea of IACC framework is to train a \emph{centralized critic} that is allowed to capture global information, and then use it to direct the optimization of each decentralized actor that conditions on only local information. The resulting policy gradient can be formulated as:
\begin{equation}
    \nabla_{\theta_i} J(\theta_i) = \mathbb{E}_{\vec{\pi}_\theta}\biggr[\nabla_{\theta_i}\log\pi_{\theta_i}(a_i|h_i)Q_\phi^{\vec{\pi}_\theta}(\mathbf{x},\vec{a})\biggr]
\end{equation}
where, the centralized critic, $Q_\phi^{\vec{\pi}_\theta}(\mathbf{x},\vec{a})$, is updated in an on-policy learning way by minimizing the following loss:

\begin{equation}
    \mathcal{L}(\phi) = \mathbb{E}_{\vec{\pi}_\theta}\biggr[\big(Q_\phi^{\vec{\pi}_\theta}(\mathbf{x}, \vec{a})-y\big)^2\biggr]
\end{equation}
where,
\begin{equation}
    y = r + \gamma Q_{\phi^-}^{\vec{\pi}_{\theta^-}}(\mathbf{x'}, a_1',...a_i'...,a_n') \mid _{a_i'\sim\pi_{\theta_i^-}(h_i')}
\end{equation}
and each agent's target policy $\pi_{\theta_i^-}$ is used to sample the next action to calculate the target prediction in order to further stabilize the learning. 
By accessing all agents' actions, this centralized critic is favored for its stationary learning targets and overcomes the major environmental non-stationary issue in independent learning. 
Additionally, many variants of global information can be included in $\mathbf{x}$, such as a true environmental state, a joint observation, a joint action-observation history, or even certain mixed combinations, which are helpful in facilitating the update of decentralized policies to optimize global cooperative performance. 
Apart from these positive effects of using a centralized critic, it certainly introduces extra variance in each agent's decentralized policy gradient estimation depending on other agents' actions~\cite{Lyu_aamas_2021,DOP}. Therefore, we consider the version of IACC with a joint history-value function as the critic to reduce the variance, and the decentralized policy gradient can be rewritten as:  
\begin{equation}
    \nabla_{\theta_i} J (\theta_i)=\mathbb{E}_{\vec{\pi}_{\vec{\theta}}}\Big[\nabla_{\theta_i}\log\pi_{\theta_i}(a_i\mid h_i)A(\mathbf{x},\vec{a})\Big]
\end{equation}
\begin{equation}
    A(\mathbf{x}, \vec{a}) = r + \gamma V^{\vec{\pi}_{\vec{\theta}}}_{\mathbf{w}}(\mathbf{x'})- V^{\vec{\pi}_{\vec{\theta}}}_{\mathbf{w}}(\mathbf{x})
    \label{IACC-V}
\end{equation}



\chapter{Macro-Action-Based Decentralized and Centralized Q-Learning}
\label{chap:paper1}

\section{Introduction}
\label{chap:paper1:intro}

\noindent As more robots are deployed in various settings, these robots must be able to act and learn in environments with other agents in them. A number of methods have been developed for solving the resulting multi-robot (or more generally multi-agent) learning problem. 
In particular, significant progress has been made on multi-agent deep reinforcement learning to solve challenging tasks in cooperative as well as competitive scenarios (e.g.,~\cite{COMA,MADDPG,DecHDRQN,QMIX}).
However, current methods assume that actions are modeled as primitive operations and synchronized action execution over agents.

In real-world multi-robot cooperative tasks, however, robots often select and complete actions at different times. 
Such asynchronous collaboration requires a different set of methods that consider these different completion times. 
Macro-action-based frameworks allow asynchronous action selection and termination while also naturally representing high-level robot controllers (e.g., navigation to a waypoint or grasping an object). 
In the multi-agent case, the 
\textit{Macro-Action Decentralized Partially Observable Markov Decision Process} (MacDec-POMDP)~\cite{AAMAS14AKK,AmatoJAIR19} extends the \textit{options framework}~\cite{Sutton:1999} to partially observable multi-agent domains. 
Planning methods have been developed for MacDec-POMDPs which have been demonstrated in realistic robotics problems~\cite{ICRA15MacDec,RSS15,IJRR17DecPOSMDP,MiaoAAAI16}, but only limited learning settings have been considered~\cite{MiaoIROS17}. 

Nevertheless, a principled way is still missing to generalize the above multi-agent deep reinforcement learning methods to macro-action-based robotics problems. In this chapter, we bridge this gap by: (a) proposing a \emph{decentralized} macro-action-based learning method that is based on DQN
~\cite{DQN} and generates Macro-Action Concurrent Experience Replay Trajectories (Mac-CERTs) to properly maintain macro-action trajectories for each agent; (b) introducing a \emph{centralized} macro-action-based learning method that is also based on DQN and generates  Macro-Action Joint Experience Replay Trajectories (Mac-JERTs) to maintain time information in macro-action trajectories along with a conditional target prediction method for learning a centralized joint macro-action-value function. 
Decentralized learning of decentralized policies is needed for online learning by the agents, but is difficult due to the noisy and limited learning signals of each agent and the apparent non-stationarity of the domain. Centralized learning of centralized policies is important when full communication is available during execution or as an intermediate step in generating decentralized policies in a centralized manner. 

We test our methods in simulation against state-of-the-art (primitive-action-based) methods. 
The results demonstrate that our methods are able to achieve much higher performance than learning with primitive actions and are scalable to large environment spaces. 
We believe the proposed methods are promising for learning in realistic multi-robot settings.

\section{Approach}
\label{chap:paper1:app}

\noindent In multi-robot deep reinforcement learning with macro-actions, the highly asynchronous execution of macro-actions motivates a need for a principled way for updating values and maintaining replay buffers. 
In this section, we introduce two approaches for solving these problems 
for learning decentralized (Section \ref{chap:paper1:app:MacDecQ}) and centralized (Section \ref{chap:paper1:app:MacCenQ}) policies. 
In each case, we assume the agent(s) can observe the current macro-action, macro-observation, and reward at each time step. That is, we do not have access to the primitive-level actions and observations, but we could indirectly calculate the duration of a macro-action by counting time steps. 

\subsection{Macro-Action-Based Decentralized Q-Learning}
\label{chap:paper1:app:MacDecQ}

\begin{figure}[t!]
    \centering 
    \includegraphics[height=4.1cm]{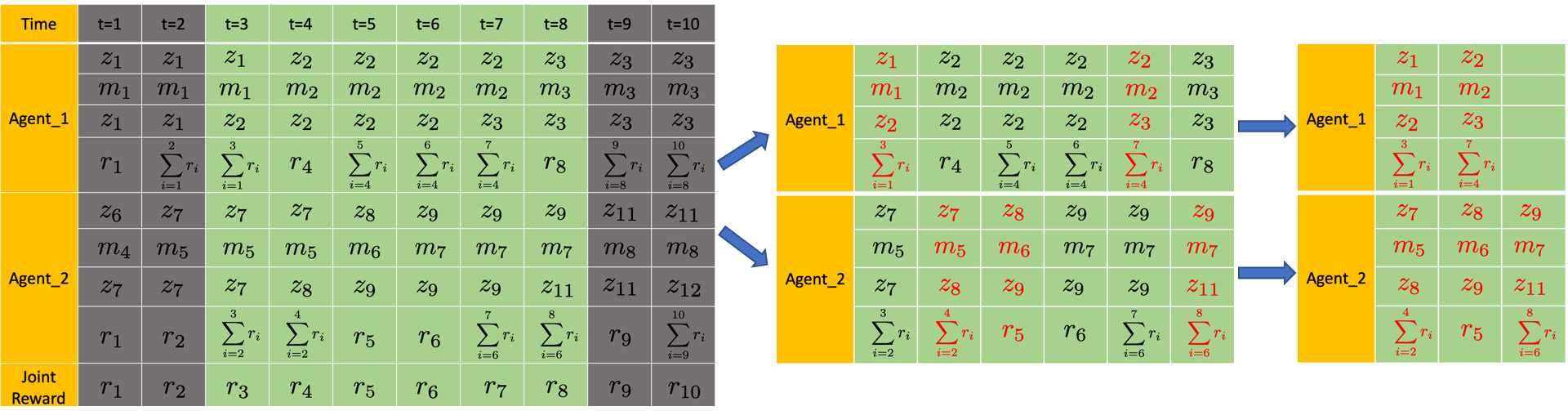}
    \caption{An example of Mac-CERTs. Two agents first sample concurrent sequential experiences (green area) from the replay buffer; The valid experience (when the macro-action terminates, marked as red), is then selected to compose a squeezed sequential experience for each agent. Note that the subscript in the buffer is for distinguishing different macro-actions and macro-observations rather than agents.}
    \label{decbuffer}
\end{figure}

\noindent In the decentralized case, each agent only has access to its own macro-actions and macro-observations as well as the joint reward at each time step. 
As a result, there are several choices for how information is maintained. 
For example, each agent could maintain exact the information mentioned above (as seen on the left side of Fig.~\ref{decbuffer}), the time-step information can be removed (losing the duration information), or some other representation could be used that explicitly calculates time. 
We choose the middle approach. 
As a result, updates only need to take place for each agent after the completion of its own macro-action, and we introduce a replay buffer based on \emph{Macro-Action Concurrent Experience Replay Trajectories} (Mac-CERTs) visualized in Fig.~\ref{decbuffer}.

More concretely, under a macro-action-observation history $h_i$, 
each agent independently selects a macro-action $m_i$ via a macro-action-based decentralized policy $\Psi(m_i|h_i)$ and maintains an accumulating reward, $r^c(h_i,m_i,\tau_i) = \sum_{t=t_{m_i}}^{t_{m_i}+\tau_i-1}\gamma^{t-t_{m_i}} r_t$, 
for the macro-action from its first time-step $t_{m_i}$ to a termination time-step $t_{m_i}+\tau_i-1$. 
The agent then obtains a new macro-observation $z_i'$ with the probability $P(z_i'\mid h_i, m_i, \tau_i)$ and results in a new history $h_i'=\langle h_i,m_i,z_i'\rangle$ under the transition model $P(h_i',\tau_i \mid h_i,m_i)$. 
Correspondingly, the experience tuple collected by each agent $i$ is represented as $\langle z, m, z', r^c  \rangle_i$, where $z_i$ is the macro-observation used for choosing the macro-action $m_i$. 
We can write down the Bellman equation for each agent $i$ under a given high-level policy $\Psi_i$ as :
\begin{align}
    Q^{\Psi_i}(h_i,m_i) &= \E_{h_i',\tau_i \mid h_i, m_i}\Big[r^c(h_i,m_i,\tau_i) + \gamma^{\tau_i} V^{\Psi_i}(h_i')\Big]\\
        & =\E_{\tau_i \mid h_i, m_i} \biggr[\sum_{h_i'}P(h_i'\mid h_i, m_i, \tau_i)\Big[r^c(h_i,m_i,\tau_i) + \gamma^{\tau_i} V^{\Psi_i}(h_i')\Big]\biggr] \\
        & = \E_{\tau_i \mid h_i, m_i} \biggr[r^c(h_i,m_i,\tau_i) + \gamma^{\tau_i}\sum_{h_i'}P(h_i'\mid h_i, m_i, \tau_i)V^{\Psi_i}(h_i')\biggr]\\
        & = \E_{\tau_i \mid h_i, m_i} \biggr[r^c(h_i,m_i,\tau_i) + \gamma^{\tau_i}\sum_{z_i'\in\zeta_i}P(z_i' \mid h_i, m_i, \tau_i)V^{\Psi_i}(h_i')\biggr]\\ 
        & =  \E_{\tau_i \mid h_i, m_i} \biggr[r^c(h_i,m_i,\tau_i) + \gamma^{\tau_i}\sum_{z_i'\in\zeta_i}P(z_i' \mid h_i, m_i, \tau_i)V^{\Psi_i}(h_i,m_i,z_i')\biggr] \\
        & = r^c(h_i,m_i) + \sum_{z_i'\in\zeta_i}P(z_i' \mid h_i, m_i)V^{\Psi_i}(h_i,m_i,z_i')
\label{bellman1}
\end{align}
where, $P(z_i'\mid h_i,m_i,\tau_i) = \mathbb{E}_{\vec{h}_{/i}\mid h_i,\tau_i}\Biggr[\mathbb{E}_{s\mid \vec{h}}\biggr[\mathbb{E}_{\vec{m}_{/i}|\vec{h}_{/i}}\Bigr[\mathbb{E}_{\vec{\tau}_{/i}\mid\vec{h}_{/i},\vec{m}_{/i}}\bigr[\mathbb{E}_{s' \mid s,\vec{m},\vec{\tau}}[P(z_i'\mid m_i,s')]\bigr]\Bigr]\biggr]\Biggr]$; 
$P(\tau_i\mid h_i, m_i)$ is the stochastic execution time-step cost of a macro-action associated with its termination condition $\beta_{m_i}$.

In each training iteration, agents first sample a concurrent mini-batch of sequential experiences (either random traces with the same length or entire episodes) from the replay buffer $\mathcal D$. Each sampled sequential experience is further cleaned up by filtering out the experiences when the corresponding macro-action is still executing. This disposal procedure finally results in a mini-batch of `squeezed' sequential experiences for each agent's training. A specific example is shown in Fig.~\ref{decbuffer} (assuming $\gamma = 1$).       

In this work, we implement Dec-HDRQN with Double Q-learning (Dec-HDDRQN) to train the decentralized macro-action-value function $Q_{\theta_i}(h_i, m_i)$ (in Eq.~\ref{bellman1}) for each agent $i$. 
Each agent updates its own macro-action-value function by minimizing the loss:
\begin{equation}
    \mathcal{L}(\theta_i)=\mathbb{E}_{<z, m, z', r^c>_i\sim\mathcal{D}}\Big[\bigl(y_i - Q_{\theta _i}(h_i, m_i)\bigr)^2 \Big] \end{equation} 
where,
\begin{equation}
    y_i = r^c_i + \gamma Q_{\theta_i^-}\bigl(h_i', \argmax_{m_i'} Q_{\theta_i}(h_i',m_i')\bigr)
\end{equation} 

\subsection{Macro-Action-Based Centralized Q-Learning}
\label{chap:paper1:app:MacCenQ}

\noindent Achieving centralized control in the macro-action setting needs to learn a joint macro-action-value function $Q(\vec h, \vec m)$. This requires a way to correctly accumulate the rewards for each joint macro-action.  This is actually more complicated than the decentralized case because there is no obvious update step (i.e., there may never be a time when all agents have terminated their macro-actions together). As a result, we use the idea of updating when \emph{any} agent terminates a macro-action \cite{AmatoJAIR19,AAMAS14AKK}. But this makes updating and maintaining a buffer more complicated than in Section\ref{chap:paper1:app:MacDecQ}. 

In this case, we introduce a centralized replay buffer that we call \emph{Macro-Action Joint Experience Replay Trajectories} (Mac-JERTs). Instead of independently maintaining a cumulative reward for each macro-action, agents share a joint cumulative reward $\vec{r}^{\,c}(\vec{h}, \vec{m}, \vec{\tau}) = \sum_{t=t_{\vec{m}}}^{t_{\vec{m}}+\vec{\tau}-1}\gamma^{t-t_{\vec{m}}} r_t$ for each joint macro-action $\vec{m}$, where $t_{\vec{m}}$ is the time-step when a joint macro-action $\vec{m}$ starts, and $t_{\vec{m}}+\vec{\tau}-1$ is the ending time-step of $\vec{m}$ when \emph{any} agent finishes its macro-action. Here, we can write down the Bellman equation under a centralized macro-action-based policy $\bm{\Psi}$ as:
\begin{align}
    Q^{\Psi}(\vec{h},\vec{m}) &= \E_{\vec{h}',\vec{\tau} \mid \vec{h}, \vec{m}}\Bigr[\vec{r}^{\,c}(\vec{h},\vec{m}, \vec{\tau})+ \gamma^{\vec{\tau}} V^{\Psi}(\vec{h}')\Bigr]\\
        & = \E_{\vec{\tau}\mid \vec{h},\vec{m}}\biggr[\sum_{\vec{h}'}P(\vec{h}'\mid \vec{h}, \vec{m}, \vec{\tau})\Bigr[\vec{r}^{\,c}(\vec{h},\vec{m},\vec{\tau}) + \gamma^{\vec{\tau}} V^{\Psi}(\vec{h}')\Bigr]\biggr] \\
        & = \E_{\vec{\tau}\mid \vec{h},\vec{m}}\biggr[\vec{r}^{\,c}(\vec{h},\vec{m},\vec{\tau}) + \gamma^{\vec{\tau}}\sum_{\vec{h}'}P(\vec{h}'\mid \vec{h}, \vec{m}, \vec{\tau})V^{\Psi}(\vec{h}')\biggr] \\
        & = \E_{\vec{\tau}\mid \vec{h},\vec{m}}\biggr[\vec{r}^{\,c}(\vec{h},\vec{m},\vec{\tau}) + \gamma^{\vec{\tau}}\sum_{\vec{z}\,'\in\zeta}P(\vec{z}\,' \mid \vec{h}, \vec{m}, \vec{\tau})V^{\Psi}(\vec{h}')\biggr] \\
        & = \E_{\vec{\tau}\mid \vec{h},\vec{m}}\biggr[\vec{r}^{\,c}(\vec{h},\vec{m},\vec{\tau}) + \gamma^{\vec{\tau}}\sum_{\vec{z}\,'\in\zeta}P(\vec{z}\,' \mid \vec{h}, \vec{m}, \vec{\tau})V^{\Psi}(\vec{h},\vec{m},\vec{z}\,')\biggr] \\
        & = \vec{r}^{\,c}(\vec{h}, \vec{m}) + \sum_{\vec{z}\,'\in\zeta}P(\vec{z}\,' \mid \vec{h}, \vec{m})V^{\Psi}(\vec{h},\vec{m},\vec{z}\,')
        \label{bellman2}
\end{align}
where, $P(\vec{z}\,'\mid \vec{h},\vec{m},\vec{\tau}) = \mathbb{E}_{s|\vec{h}}\big[\mathbb{E}_{s' \mid s,\vec{m},\vec{\tau}}[P(\vec{z}\,'|\vec{m},s')]\big]$;  $P(\vec{\tau}\mid \vec{h}, \vec{m})$ is the stochastic execution time-step cost of a joint macro-action associated with the termination condition of each agent's macro-action.

\begin{figure}[t!]
    \centering
    \includegraphics[height=4cm]{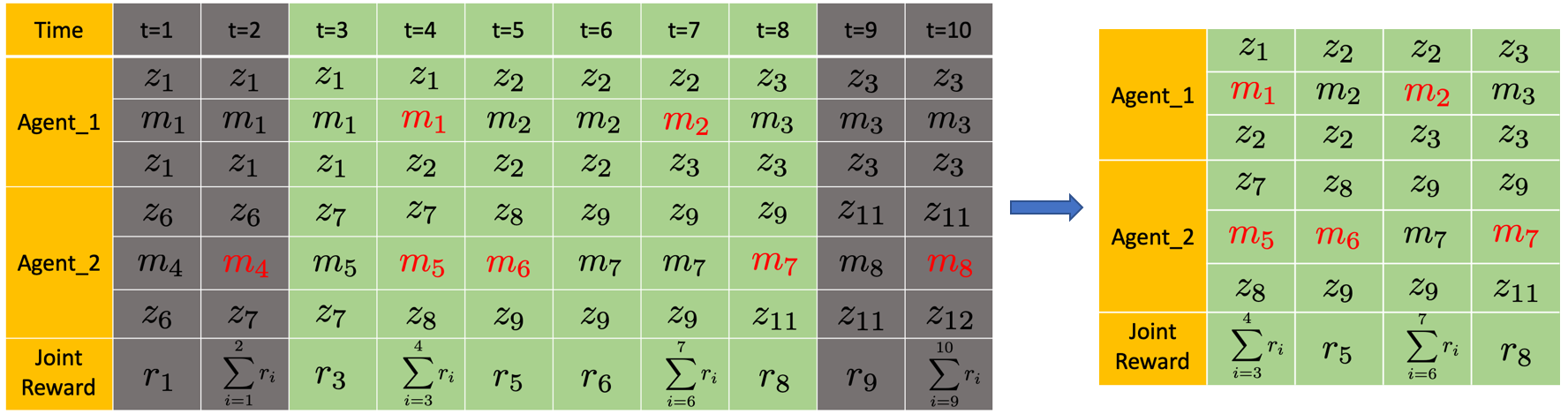}
    \caption{An example of Mac-JERTs. A joint sequential experience (green area) is first sampled from the memory buffer, and then, depending on the termination of each joint macro-action, a squeezed sequential experience is generated for the centralized training. Each agent's macro-action, which is responsible for the termination of the joint one, is marked in red.}
    \label{cenbuffer}
\end{figure}

In our work, we use Double-DRQN (DDRQN) to train the centralized macro-action-value function. In each training iteration, a mini-batch of sequential joint experiences is first sampled from Mac-JERTs, and then a similar filtering operation, as presented in Section~\ref{chap:paper1:app:MacDecQ}, is used to obtain the `squeezed' joint experiences (shown in Fig.~\ref{cenbuffer}). But, in this case, only one joint reward is maintained that accumulates from the selection of any agent's macro-action to the completion of any (possibly other) agent's macro-action. 

Using the squeezed joint sequential experiences, the centralized macro-action-value function (in Eq.~\ref{bellman2}) at time-step $t$, $Q_{\phi}(\vec{h}_{(t)}, \vec{m}_{(t)})$, is trained end-to-end to minimize the following loss:

\begin{equation}
    \mathcal{L}(\phi)=\mathbb{E}_{<\vec{z}_{(t)}, \vec{m}_{(t)}, \vec{z}_{(t+1)}, \vec{r}^c_{(t)} >\sim\mathcal{D}}\biggr[\Bigr(y_{(t)} - Q_{\phi}\bigr(\vec{h}_{(t)}, \vec{m}_{(t)}\bigr)\Bigr)^2 \biggr]
\end{equation}
where,
\begin{equation}
    \label{uncondi}
    y_{(t)} = \vec{r}^{\,c}_{(t)} + \gamma Q_{\phi^-}\Bigr(\langle\vec{h}_{(t)}, \vec{m}_{(t)}, \vec{z}_{(t+1)}\rangle,\argmax_{\vec{m}_{(t+1)}}Q_{\phi}\bigr(\langle\vec{h}_{(t)}, \vec{m}_{(t)},\vec{z}_{(t+1)}\rangle,\vec{m}_{(t+1)}\bigr)\Bigr)
\end{equation}

The next joint macro-action selection part in Eq.~\ref{uncondi} implies that at the next step all agents will switch to a new macro-action. However, this is often not true. For example, in Fig.~\ref{cenbuffer}, the last three squeezed sequential experiences show that only one of the agents changes its macro-action per step. Therefore, the more agents that are not switching macro-actions, the less accurate the prediction that Eq.~\ref{uncondi} will make. In order to have a more correct value estimation for a joint macro-action, here, we propose a \emph{conditional target prediction} as:

\begin{equation}
    \label{condi}
    y_{(t)} = \vec{r}^{\,c}_{(t)} + \gamma Q_{\phi^-}\Bigr(\langle\vec{h}_{(t)},\vec{m}_{(t)}, \vec{z}_{(t+1)}\rangle, \argmax_{\vec{m}_{(t+1)}}Q_{\phi}\bigr(\langle\vec{h}_{(t)},\vec{m}_{(t)},\vec{z}_{(t+1)}\rangle, \vec{m}_{(t+1)} \mid \vec{m}^{\text{undone}}_{(t)}\bigr)\Bigr)
\end{equation}

\noindent where, $\vec{m}^{\text{undone}}_{(t)}$ is the joint-macro-action over the agents who have not terminated the macro-actions at time-step $t$ and will continue running it at next step. 
The comparison of the training results using the two different predictions is discussed in Section~\ref{chap:paper1:exp}.  

\section{Simulated Experiments}
\label{chap:paper1:exp}

\subsection{Experimental Setup}
\label{chap:paper1:exp:domain}

\begin{figure}[h!]
    \centering
    \begin{subfigure}{.3\textwidth}
        \centering
        \includegraphics[height=3.5cm]{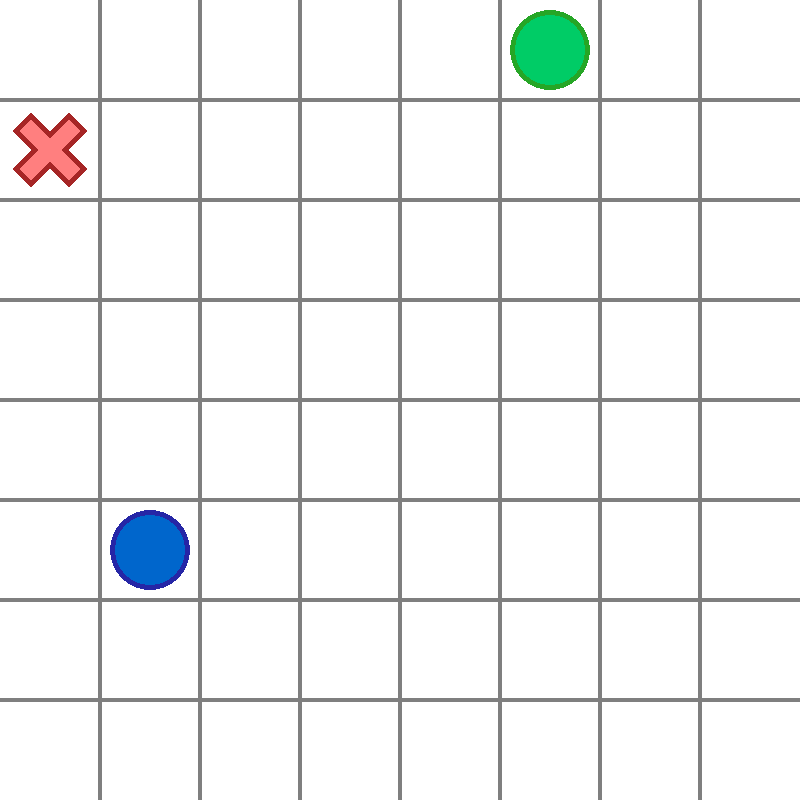}
        \caption{Capture Target}
        \label{ctma}
    \end{subfigure}
    \begin{subfigure}{.3\textwidth}
        \centering
        \includegraphics[height=3.5cm]{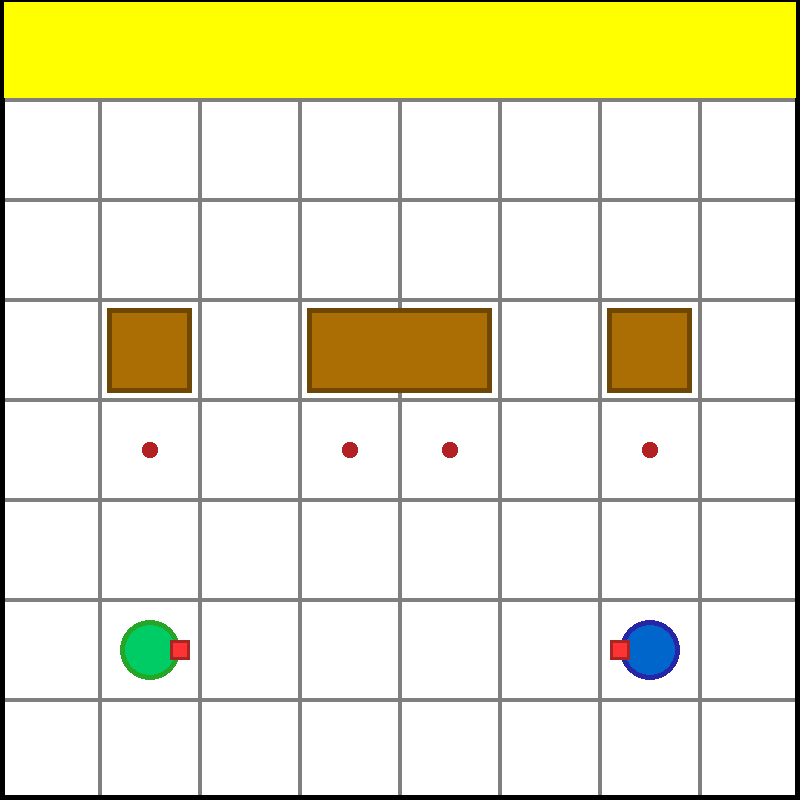}
        \caption{Box Pushing}
        \label{bpma}
    \end{subfigure}
    \begin{subfigure}{.35\textwidth}
        \centering
        \includegraphics[height=3.5cm]{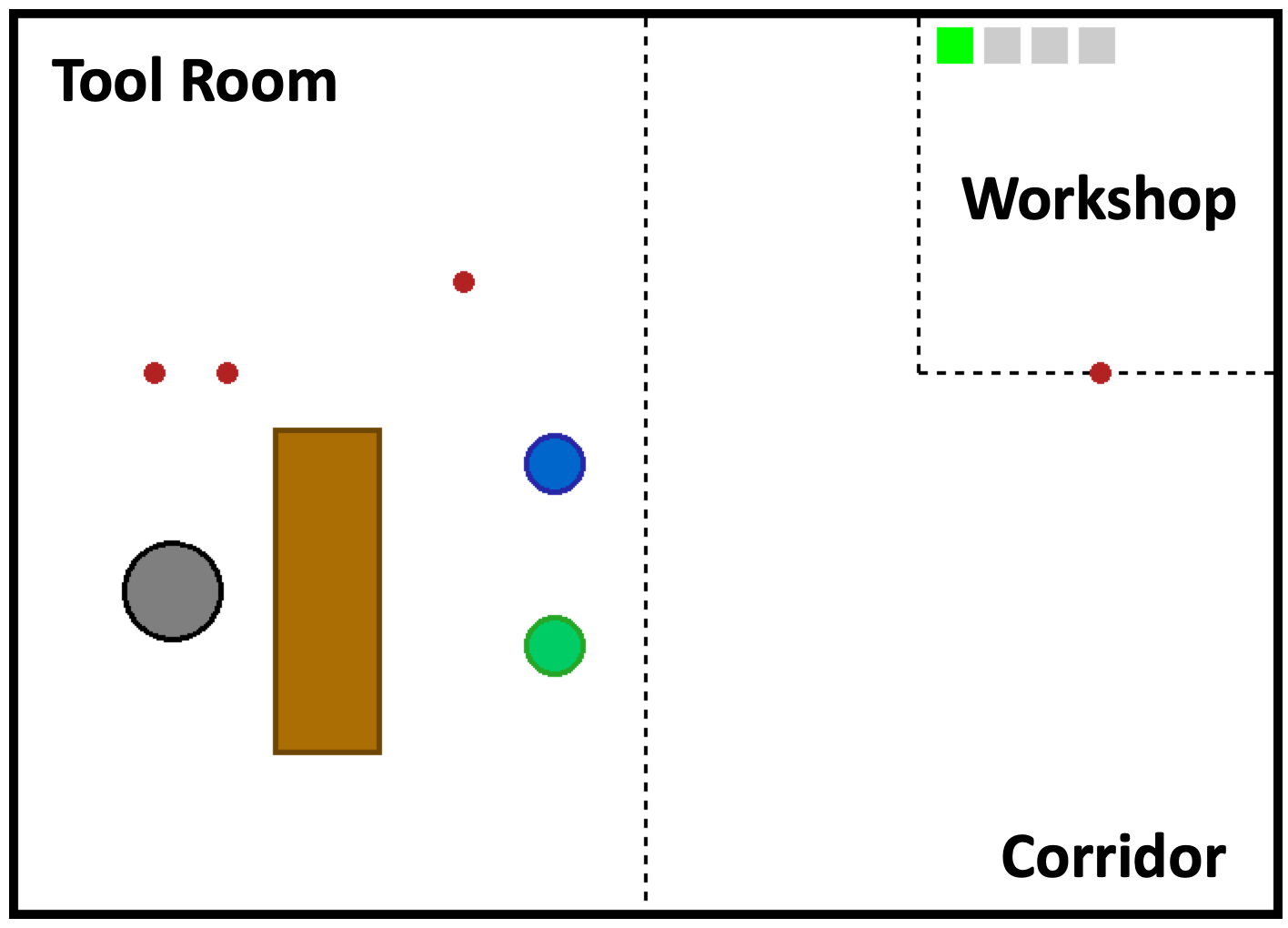}
        \caption{Warehouse Tool Delivery}
        \label{wtdma}
    \end{subfigure}
    \caption{Experimental environments}
    \label{domains}
\end{figure}

\noindent We evaluate our approaches on three different domains (Fig.~\ref{domains}): (a) Capture Target, a variant of an existing multi-agent-single-target (MAST) domain~\cite{DecHDRQN}; (b) Box Pushing, a benchmark Dec-POMDP domain~\cite{SZuai07}; 
(c) Warehouse Tool Delivery Domain inspired by human-robot interaction. 
Note that the macro-actions, defined in domains that we consider in this paper, are quite simple. It will not always be so straightforward, but we leave macro-action design and selection for future work. 
Typically, we also include primitive-actions into the macro-action set (as one-step macro-actions), which gives agents the chance to learn more complex policies that use both when it is necessary.

{\bf Capture Target}. In Fig.~\ref{ctma}, two robots (green and blue circles) are tasked with capturing a randomly moving target (red cross). A terminal reward +1  can only be obtained when the two robots capture the target simultaneously. The macro-observations here are the same as the primitive (low-level) ones: each agent's own location (fully observable) and the target's location (partially observable with a flickering probability of 0.3). In the primitive action version~\cite{DecHDRQN}, each agent has four moving actions (\emph{up, down, left, right}) and a \emph{stay} action. In the macro-action case, there are only two macro-actions for each agent: \emph{\textbf{Move\_to\_Target}}, navigates the robot towards the target and keeps updating the target's location according to the low-level observation; It terminates when the robot reaches the observed target's position. Note that if the target is flicked, the agent will continue moving towards the previously observed one; \emph{\textbf{Stay}}, is the same as the primitive one and lasts only 1 time-step.     

\textbf{Box Pushing}. This is a well-known cooperative robotics problem originally presented in~\cite{SZuai07}. Fig.~\ref{bpma} displays one example of this problem in a grid world. There are two small boxes and one big box in the environment. The objective of the two robots is to cooperate on pushing the big box (middle brown square), which is only movable when two robots push it together, to the goal area (yellow bar at the top). 
The difficulties come from the partial observability (each robot is only allowed to observe one cell in front) and two small boxes which attract the robots to learn the sub-optimal policy, such as pushing one small box on each own.

In the primitive action version, each agent has four actions: \textit{move forward}, \textit{turn left}, \textit{turn right} and \textit{stay}. The small box moves forward one grid cell when any robot faces it and executes the \textit{move forward} action. The big box is only movable when the two robots face it in two parallel cells and move forward together. The robot can only observe one of five states in the cell in front of it: empty, teammate, boundary, small box, or big box. During execution, the agents get $-0.1$ reward per step. Successfully pushing the big box to the goal area results in a $+100$ reward or a $+10$ reward for each small box. Either hitting the boundary or pushing the big box alone generates a $-5$ penalty.

In the macro-action version, besides the one-step macro-actions \textbf{\textit{Turn-Left}}, \textbf{\textit{Turn-Right}}, and \textbf{\textit{Stay}}, we include three long-term macro-actions: \emph{\textbf{Move-to-Small-Box(i)}}, navigates the robot to the red waypoint below one of the small boxes and ends with facing the box; 
\textbf{\textit{Move-to-Big-Box}}, navigates the robot to one of the waypoints below the big box and facing it; \textbf{\textit{Push}}, lets the robot keep moving forward until touching the environment's boundary, hitting the big box on its own, or pushing a box to the goal area.  
Note that, the boxes are only allowed to be pushed toward the north, and each episode terminates either one of the boxes pushed to the goal area or after a certain amount of time steps.    

\textbf{Warehouse Tool Delivery}. In order to test if our approach is scalable to a larger domain requiring more complicated collaborations and long-term reasoning, we designed this Warehouse Tool Delivery problem (Fig.~\ref{wtdma}). 
This environment is a $5\times7$ continuous space, which involves one human working on an assembling task in the workshop. 
The progress bar on the top indicates the total number of steps in the task, the current step (green) the human is working on, and the completed step (black). 
Human always starts from step one and needs a particular tool for each future step to continue. A robot arm (gray circle) is responsible for searching for the correct tool on the tabletop (brown) and passing it to one of the mobile robots (green and blue circles) to complete the delivery to the human in time. 
In our experiments, the assembling task has $4$ steps in total, and the time cost on each is $18$. 
Note that, the human is only allowed to get the tool for the next one step from mobile robots. 
The correct tools needed by each human are unknown to robots, which have to be learned during training in order to perform efficient delivery.
Each episode ends after $H=150$ time-steps, or the human obtains the tool for the last step.

Each mobile robot has three available macro-actions: \emph{\textbf{Go-W}}, navigates the robot to the red waypoint at the workshop, and the length of this action depends on the robot's moving speed $v$ (0.6 in our case); 
\emph{\textbf{Go-TR}}, directs the robot to the waypoint located at the upper right of the tool room; 
\emph{\textbf{Get-Tool}}, leads the robot to the pre-allocated waypoint beside the table and waits there. 
This action will not terminate until either obtaining one tool from the robot arm or after 10 time-steps have passed. 
There are four macro-actions for the robot arm: \emph{\textbf{Wait-M}} lasts 1 time step to wait for mobile robots; \emph{\textbf{Search-Tool(i)}} takes 6 time steps to find a tool $i$ and place it in the staging area (lower left on the table where can hold at most two tools).
Running this action when the staging area
is fully occupied leads the robot to pause for 6 time-steps; 
\emph{\textbf{Pass-to-M(i)}} costs 4 time steps to pass one of the tools from the staging area, in first-in-first-out order, to mobile robot $i$. 

Each mobile robot can capture four different features in one macro-observation: its own location, the current step the human is working on (only observable in the workshop), the tools being carried by itself, and the number of tools at the waiting spots (only observable in the tool room). 
Fetch is allowed to observe the number of tools in the staging area, and which mobile robot is beside the table. 

The global rewards provide $-1$ each time step to encourage the robots to deliver the object(s) in a timely manner without causing the human to pause; a penalty of $-10$ is given when the robot arm executes {\bf\emph{Pass-to-M(i)}} but no mobile robots are beside the table; and a bonus of $+100$ is awarded to the entire team when the robots successfully deliver a correct tool to the human.

\begin{figure}[t!]
    \centering
    \captionsetup[subfigure]{labelformat=empty}
    \centering
    \subcaptionbox{\hspace{5mm}(a) 4$\times$4}
        [0.45\linewidth]{\includegraphics[height=4.5cm]{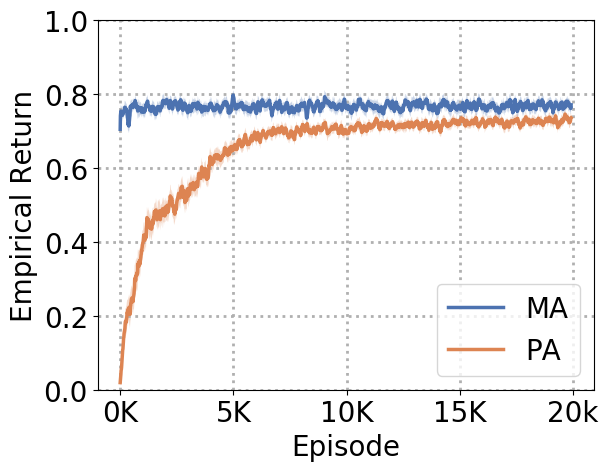}}
    ~
    \centering
    \subcaptionbox{\hspace{5mm}(b) 6$\times$6}
        [0.45\linewidth]{\includegraphics[height=4.5cm]{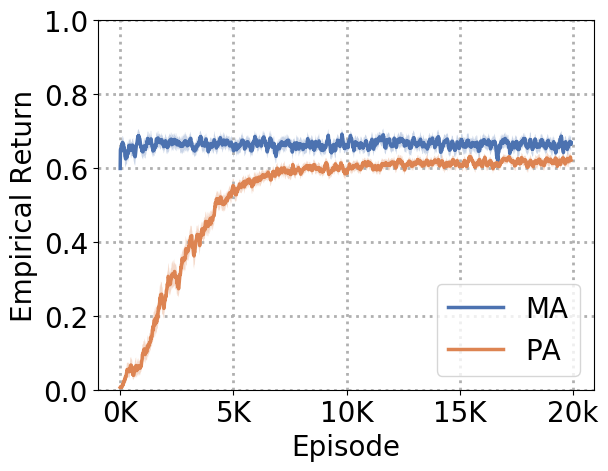}}
    ~
    \centering
    \subcaptionbox{\hspace{5mm}(c) 8$\times$8}
        [0.45\linewidth]{\includegraphics[height=4.5cm]{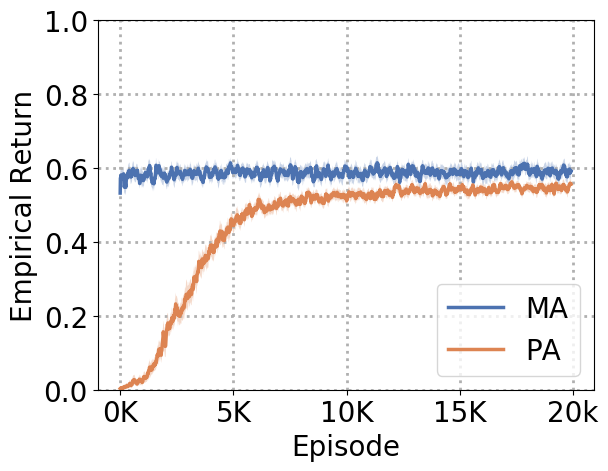}}
    ~
    \centering
    \subcaptionbox{\hspace{5mm}(d) 10$\times$10}
        [0.45\linewidth]{\includegraphics[height=4.5cm]{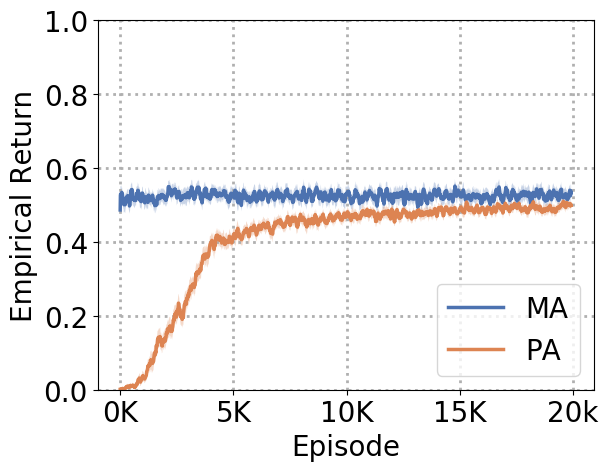}}
    ~
    \centering
    \subcaptionbox{\hspace{5mm}(e) 20$\times$20}
        [0.45\linewidth]{\includegraphics[height=4.5cm]{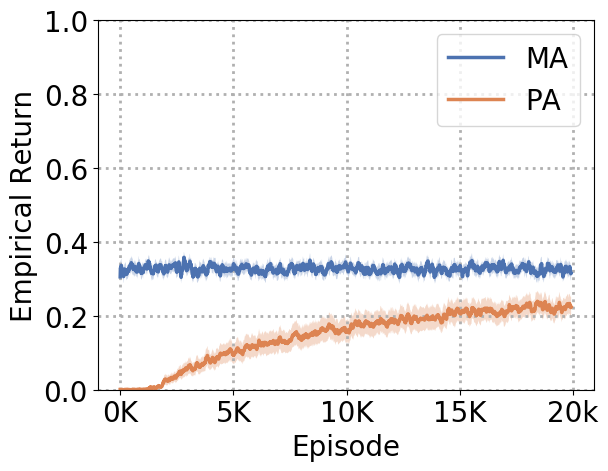}}
    ~
    \centering
    \subcaptionbox{\hspace{5mm}(f) 30$\times$30}
        [0.45\linewidth]{\includegraphics[height=4.5cm]{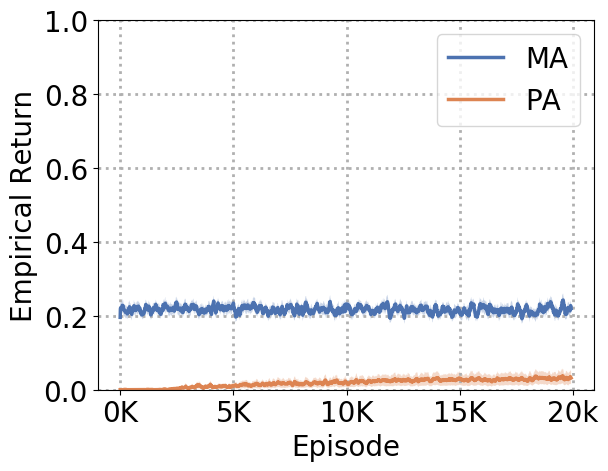}}
    \caption{Comparisons of learning decentralized macro-action (MA) policy and primitive-action (PA) policy in the capture target domain ($\gamma=0.95$) under variant grid world spaces.}
    \label{ct_ma_vs_pri}
\end{figure}

\begin{figure}[t!]
    \centering
    \captionsetup[subfigure]{labelformat=empty}
    \centering
    \subcaptionbox{\hspace{5mm}(a) 4$\times$4}
        [0.45\linewidth]{\includegraphics[height=4.5cm]{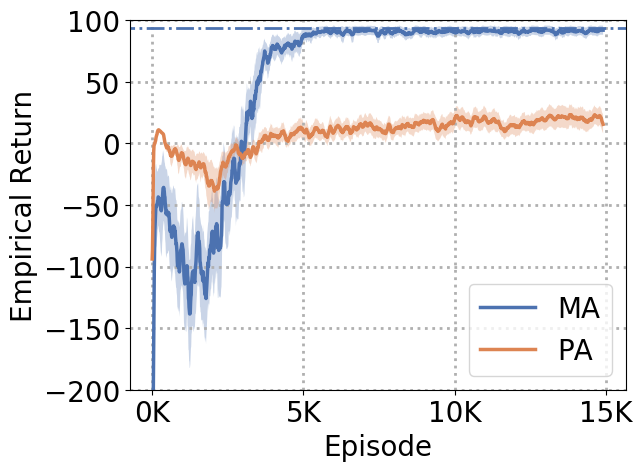}}
    ~
    \centering
    \subcaptionbox{\hspace{5mm}(b) 6$\times$6}
        [0.45\linewidth]{\includegraphics[height=4.5cm]{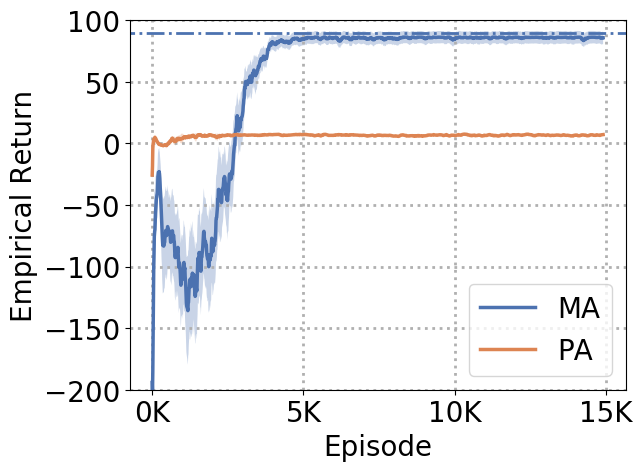}}
    ~
    \centering
    \subcaptionbox{\hspace{5mm}(c) 8$\times$8}
        [0.45\linewidth]{\includegraphics[height=4.5cm]{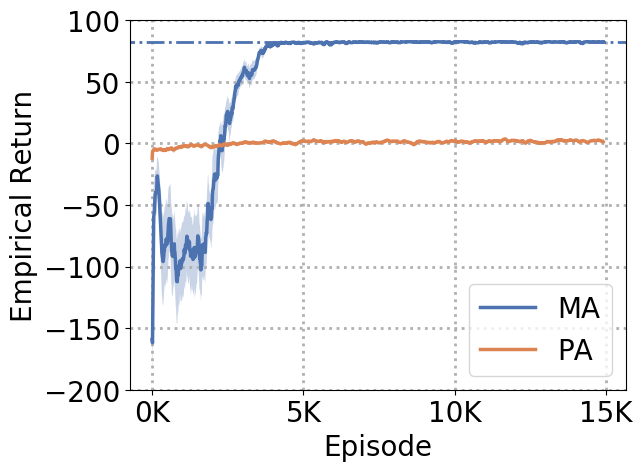}}
    ~
    \centering
    \subcaptionbox{\hspace{5mm}(d) 10$\times$10}
        [0.45\linewidth]{\includegraphics[height=4.5cm]{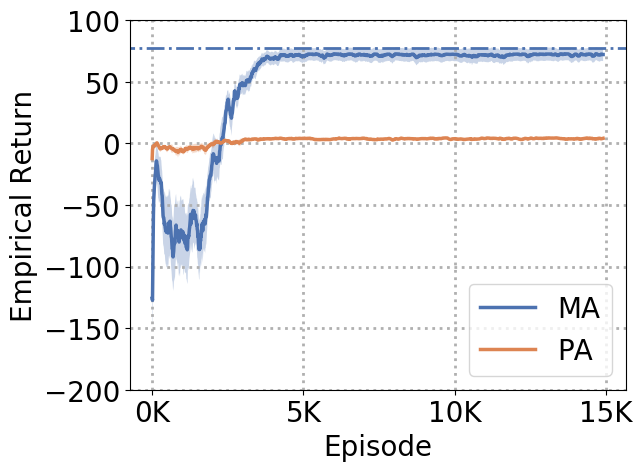}}
    ~
    \centering
    \subcaptionbox{\hspace{5mm}(e) 20$\times$20}
        [0.45\linewidth]{\includegraphics[height=4.5cm]{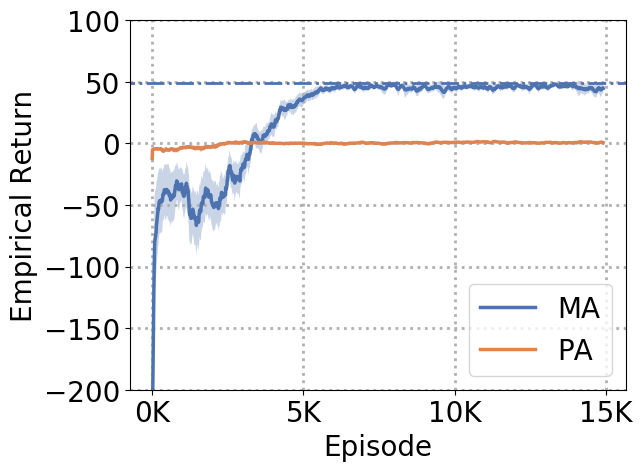}}
    ~
    \centering
    \subcaptionbox{\hspace{5mm}(f) 30$\times$30}
        [0.45\linewidth]{\includegraphics[height=4.5cm]{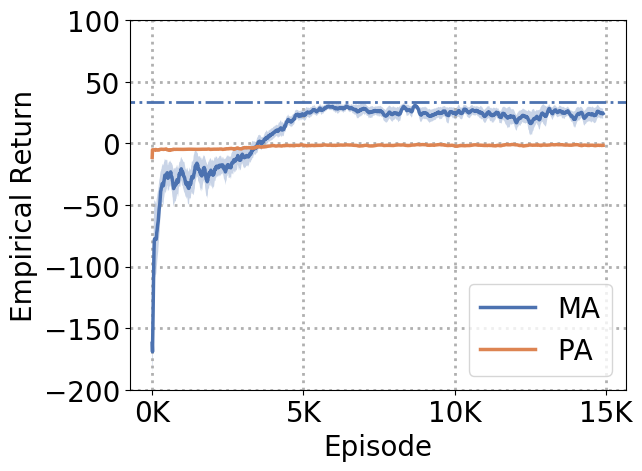}}
    \caption{Comparisons of learning decentralized macro-action (MA) policy versus primitive-action (PA) policy in the box pushing domain under variant grid world spaces.}
    \label{bp_ma_vs_pri}
\end{figure}

\subsection{Results}

\begin{figure}[t!]
    \centering
    \captionsetup[subfigure]{labelformat=empty}
    \centering
    \subcaptionbox{\hspace{5mm}(a) 4$\times$4}
        [0.45\linewidth]{\includegraphics[height=4.5cm]{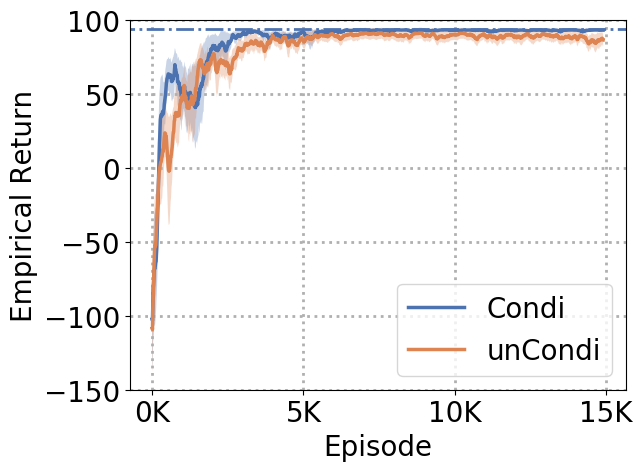}}
    ~
    \centering
    \subcaptionbox{\hspace{5mm}(b) 6$\times$6}
        [0.45\linewidth]{\includegraphics[height=4.5cm]{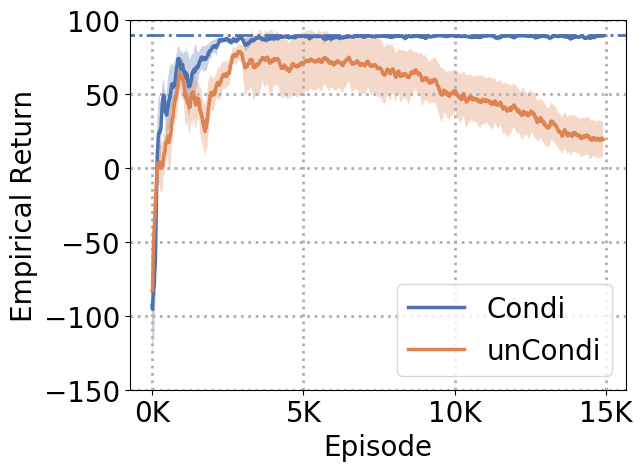}}
    ~
    \centering
    \subcaptionbox{\hspace{5mm}(c) 8$\times$8}
        [0.45\linewidth]{\includegraphics[height=4.5cm]{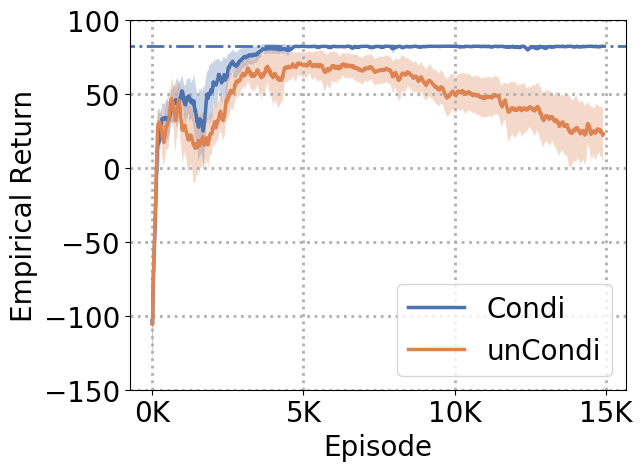}}
    ~
    \centering
    \subcaptionbox{\hspace{5mm}(d) 10$\times$10}
        [0.45\linewidth]{\includegraphics[height=4.5cm]{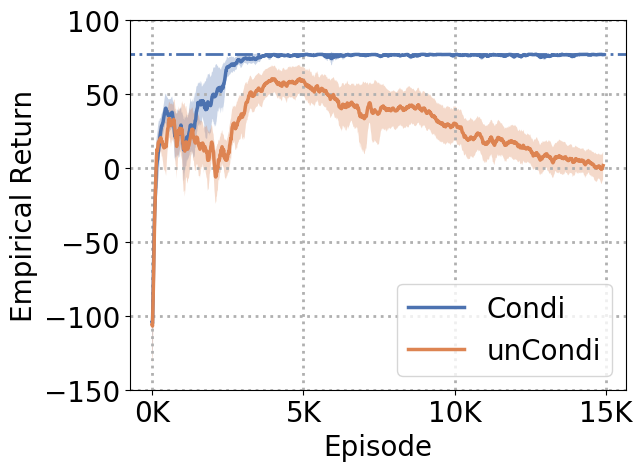}}
    ~
    \centering
    \subcaptionbox{\hspace{5mm}(e) 20$\times$20}
        [0.45\linewidth]{\includegraphics[height=4.5cm]{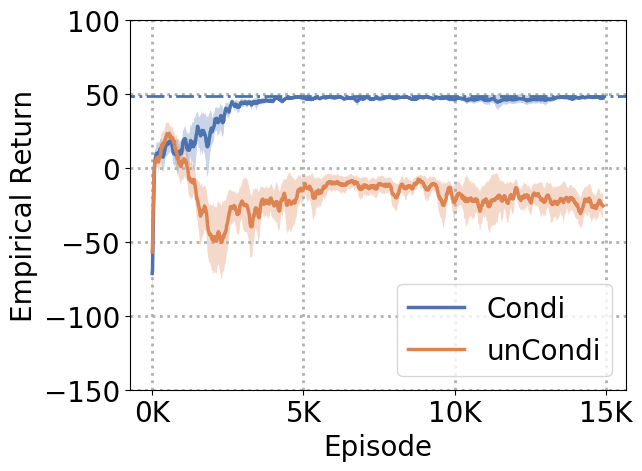}}
    ~
    \centering
    \subcaptionbox{\hspace{5mm}(f) 30$\times$30}
        [0.45\linewidth]{\includegraphics[height=4.5cm]{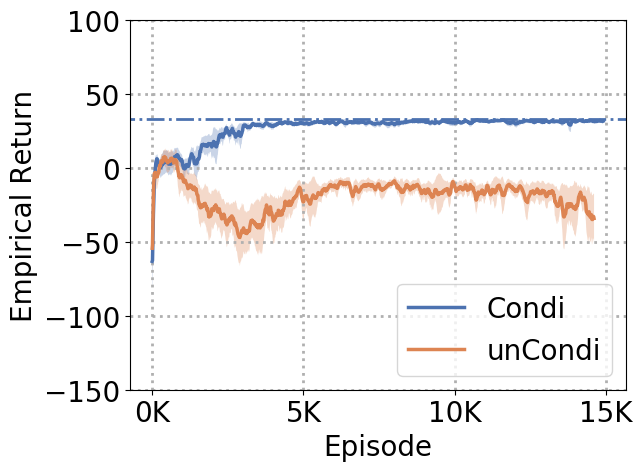}}
    \caption{Comparisons of learning macro-action-based centralized policy via \emph{conditional target prediction} (Condi) versus \emph{unconditional target prediction} (unCondi) in the Box Pushing domain under variant grid world sizes}
    \label{bpcondi_vs_uncondi}
\end{figure}

\begin{figure}[t!]
    \centering
    \captionsetup[subfigure]{labelformat=empty}
    \centering
    \subcaptionbox{\hspace{5mm}(a) 4$\times$4}
        [0.45\linewidth]{\includegraphics[height=4.5cm]{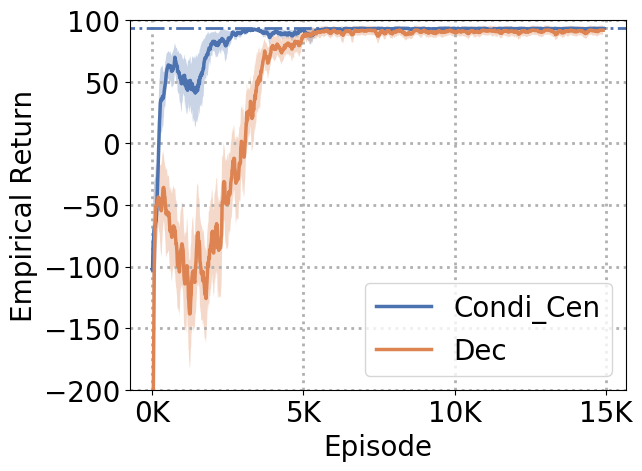}}
    ~
    \centering
    \subcaptionbox{\hspace{5mm}(b) 6$\times$6}
        [0.45\linewidth]{\includegraphics[height=4.5cm]{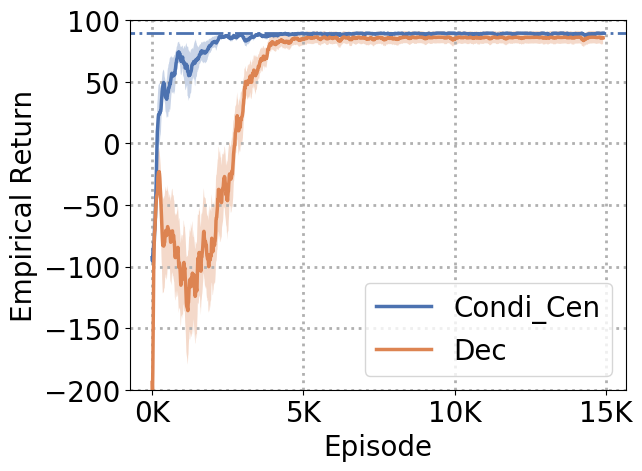}}
    ~
    \centering
    \subcaptionbox{\hspace{5mm}(c) 8$\times$8}
        [0.45\linewidth]{\includegraphics[height=4.5cm]{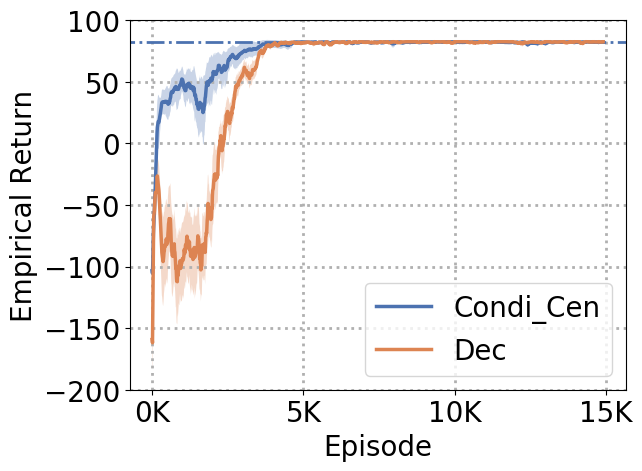}}
    ~
    \centering
    \subcaptionbox{\hspace{5mm}(d) 10$\times$10}
        [0.45\linewidth]{\includegraphics[height=4.5cm]{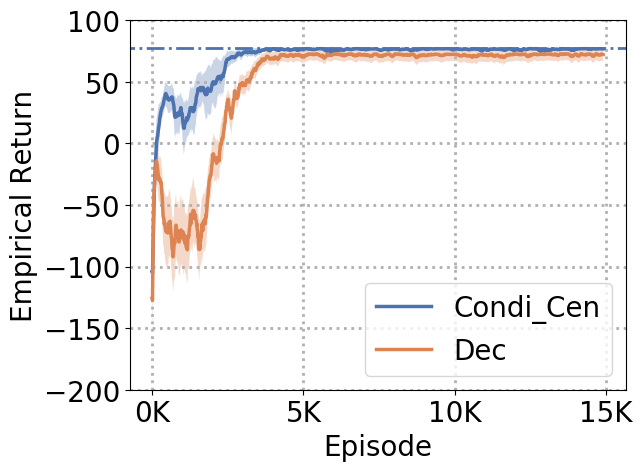}}
    ~
    \centering
    \subcaptionbox{\hspace{5mm}(e) 20$\times$20}
        [0.45\linewidth]{\includegraphics[height=4.5cm]{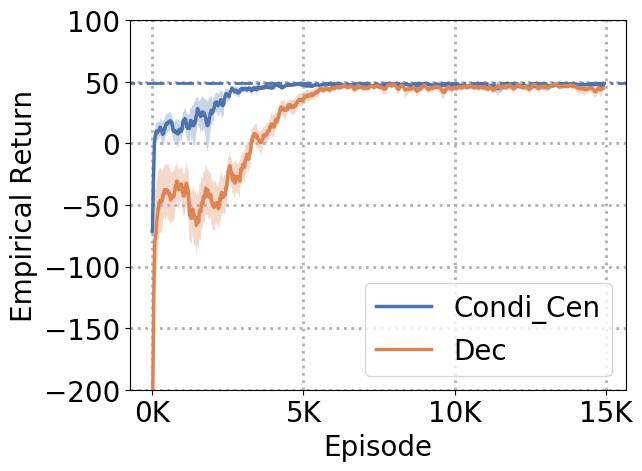}}
    ~
    \centering
    \subcaptionbox{\hspace{5mm}(f) 30$\times$30}
        [0.45\linewidth]{\includegraphics[height=4.5cm]{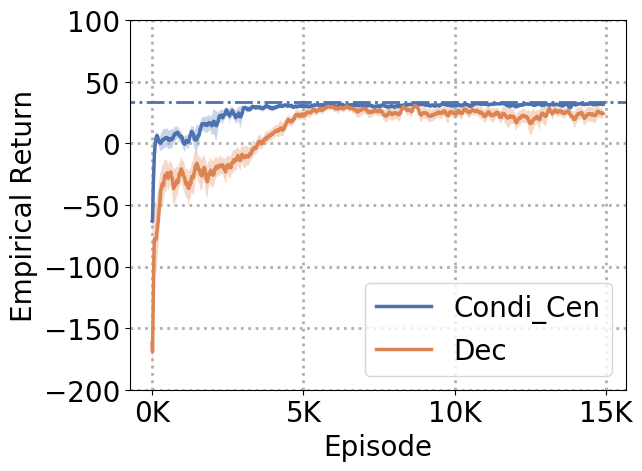}}
    \caption{Comparisons of macro-action-based centralized learning with \emph{conditional target prediction} versus decentralized learning in the Box Pushing domain under variant grid world sizes.}
    \label{bp_cen_vs_dec}
\end{figure}

\begin{figure}[h!]
    \centering
    \includegraphics[height=5cm]{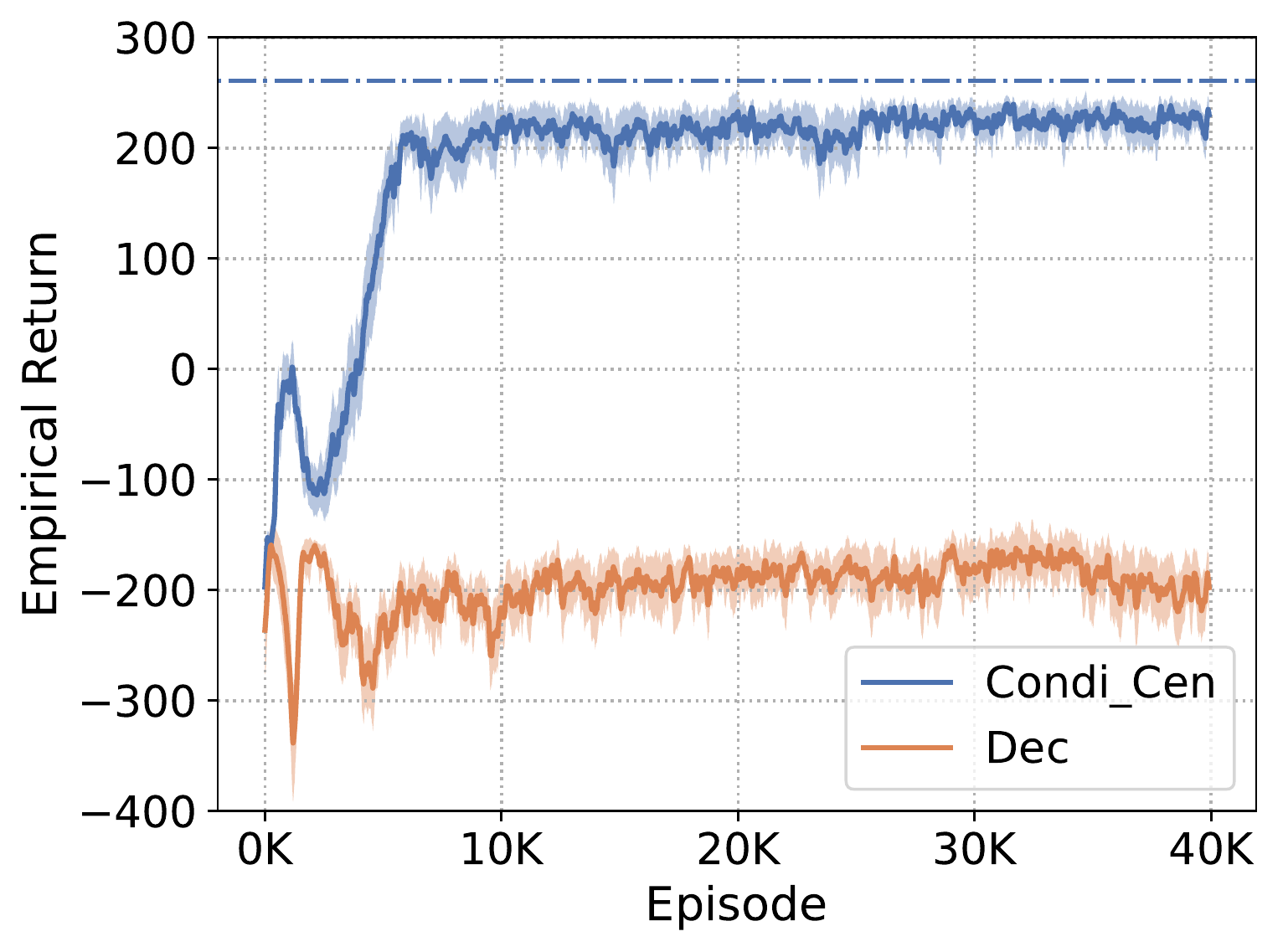}
    \caption{Performance of centralized learning versus decentralized learning under warehouse tool delivery domain.}
    \label{paper1:result_wtd}
\end{figure}

\noindent In this section, the performance of our approach on learning decentralized policies in the capture target and box pushing domains are first presented. 
Then, we show the evaluations on learning centralized policies in the box pushing domain, and also compare training via \textit{conditional target prediction} (Eq.~\ref{condi}) and the \textit{unconditional one} (Eq.~\ref{uncondi}). 
Finally, we demonstrate (as expected) that our centralized learning approach enables the robots to learn complex collaborative behaviors in the warehouse domain. 
The results shown below (Fig.~\ref{ct_ma_vs_pri} - Fig.~\ref{paper1:result_wtd}) are the mean of the episodic evaluation discounted returns (evaluation performed every 10 training episodes) over 40 runs with the standard error, and further smoothed by averaging over 10 neighbors. 
Optimal returns are shown as dash-dot lines. Readers are referred to the supplement for the full results. 

\textbf{Comparison on Learning Decentralized Policies}.
We first compare our decentralized approaches in the capture target and box pushing domains.
The experiments in the capture target domain use two MLP layers (32 neurons on each), one LSTM layer~\cite{LSTM} (64 hidden units), and another two MLP layers (32 neurons on each), which is the same architecture as seen in~\cite{DecHDRQN} except using a Leaky ReLU layer instead of the regular ReLU one as the activation function. In the box pushing domain, we tune the number of neurons in the LSTM layer down to 32. 

In the capture target domain, the macro-actions design provides a smaller action space than the primitive version, and makes the problem quite simple. Macro-actions facilitate agents to learn a good policy much faster to reach a return that the primitive learner takes a longer time to converge towards (Fig.~\ref{ct_ma_vs_pri}). 
With the growth of the grid world size, the better scalability of macro-actions than primitive-actions can be interpreted from the fact that macro-action-based learner always maintains an upper bound performance of the primitive one, especially, when the size increases to 20$\times$20 and 30$\times$30, macro-action-based learner achieves a clear outperformance.

In the box pushing domain, learning with macro-actions achieves near-optimal performance (Fig.~\ref{bp_ma_vs_pri}), such that two agents behave in cooperation to push the big box, rather than pushing the small one on each own learned under primitive actions setting. 
Also, near-optimal performance can always be achieved by the macro-actions learner even when the world space increases (e.g. 30$\times$30), but the primitive-actions learner gets stuck at a local optimum starting from the world size 6$\times$6.  

\textbf{Results on Learning Centralized Macro-Action Policies.}
Our approach of learning centralized macro-action-based policy is first evaluated in the box pushing domains. 
The centralized policies are parameterized by the same network architecture as mentioned above. 
Particularly, 32 neurons in each MLP layer and 64 neurons in LSTM are used for the grid world size smaller than $10 \times 10$, otherwise 64 neurons in each MLP layer.  

Fig.~\ref{bpcondi_vs_uncondi} indicates that, in the small grid world (4$\times$4), the performance of training centralized policy via \textit{unconditional prediction} (Eq.~\ref{uncondi}) can be as good as the \textit{conditional one} (Eq.~\ref{condi}). 
This is because the length of each macro-action (e.g., \emph{\textbf{Push}} and \emph{\textbf{Move-to-Small-Box(i)}}) is very short, so agents have a high chance to start or end the macro-actions simultaneously. 
In the middle size world (e.g., 6$\times$6 to 10$\times$10), the random exploration behavior ($\epsilon$-greedy) reduces the negative influence of the \emph{unconditional prediction} in the earlier training stage. 
However, the estimation error keeps getting accumulated and finally leads the learning to worse results.
In the larger domains, navigation macro-actions take a much longer time to complete, so that the asynchronous starting or ending of the macro-actions among the robots becomes more and more dominant. 
Under this case, \emph{conditional prediction} is able to provide a more accurate estimation on the target Q-value for training. 
This is why the conditional method outperforms the unconditional one under the grid world size 20$\times$20 and 30$\times$30. 

We also compare centralized learning with decentralized learning in the Box Pushing domain. 
As the results shown in Fig.~\ref{bp_cen_vs_dec}, the centralized learner can always learn the best policy and converge to the optimal value (dash-dot line) faster than the decentralized one. 
This is because the centralized policy receives all robots' observations as input and it is able to quickly capture good cooperative macro-actions. 

\textbf{Evaluations in Warehouse Tool Delivery Domain}. The optimal collaboration behaviors in this warehouse task depend not only on the time cost of each robot's macro-action execution, but also on how fast the human finishes each step of the task. 
Under the settings introduced in Section~\ref{chap:paper1:exp:domain}, we performed experiments (using the same network architecture as above) on learning both centralized (64 neurons in each MLP layer and 128 neurons in LSTM) and decentralized policies (half of the number of neurons as in the centralized one). 
The result, in Fig.~\ref{paper1:result_wtd}, shows that the centralized learner outperforms the decentralized learner, and converges to a value near the optimal one (dash-dot line). 

This is because, from the robot arm's perspective, the reward for delivering a correct tool is very delayed, which depends on mobile robots' choices and their moving speeds. Furthermore, a proper delivery requires the robot arm to reason about the correct tool even before 
performing cooperating (passing the tool) with mobile robots. This is difficult to learn under decentralized training using only local experiences. 

\begin{figure}[t!]
    \centering
    \begin{subfigure}{.45\textwidth}
        \centering
        \includegraphics[height=4.5cm]{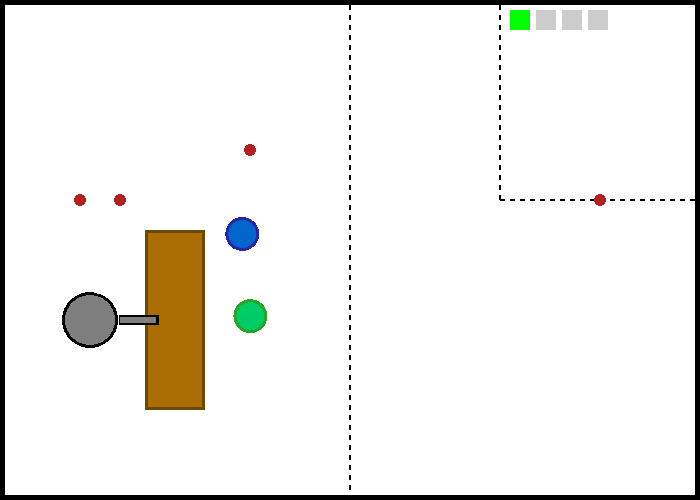}
        \caption{}
    \end{subfigure}
    ~
    \begin{subfigure}{.45\textwidth}
        \centering
        \includegraphics[height=4.5cm]{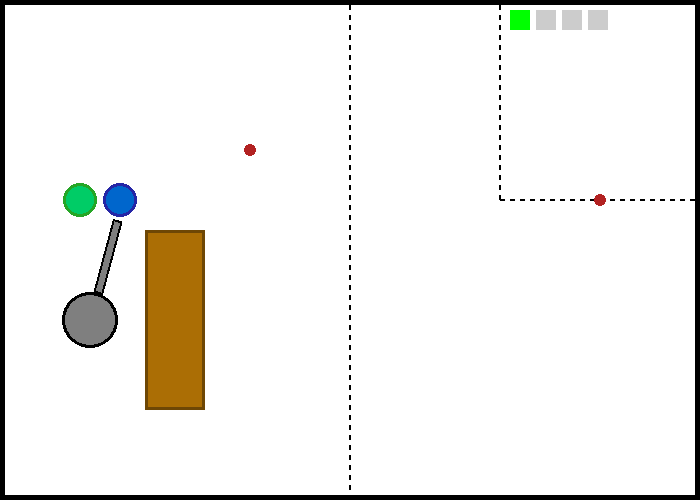}
        \caption{}
    \end{subfigure}
    ~
    \begin{subfigure}{.45\textwidth}
        \centering
        \includegraphics[height=4.5cm]{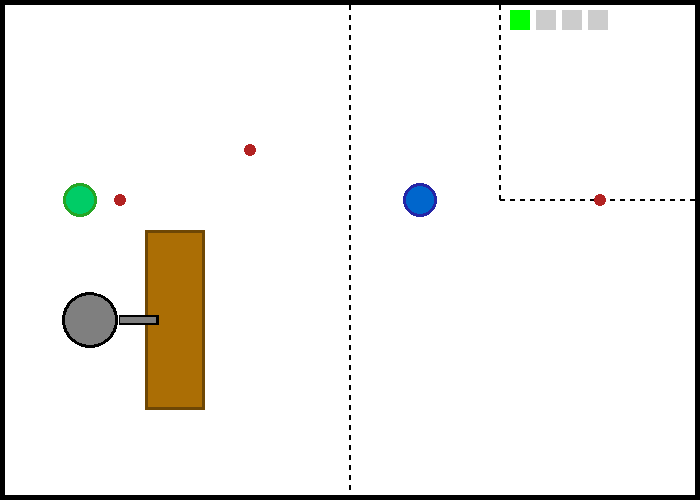}
        \caption{}
    \end{subfigure}
    ~
    \begin{subfigure}{.45\textwidth}
        \centering
        \includegraphics[height=4.5cm]{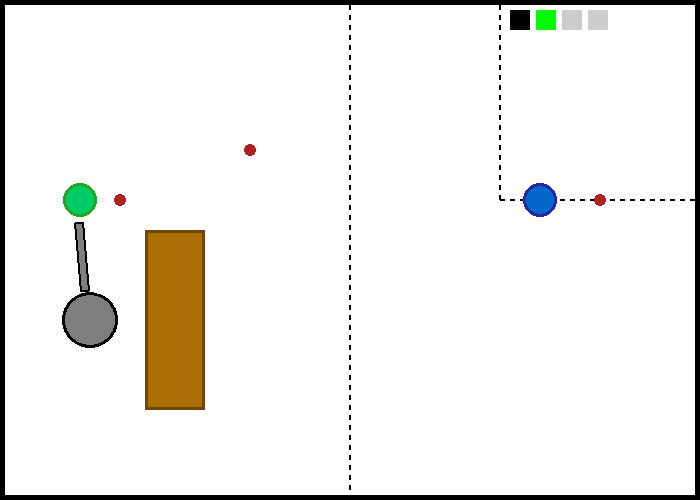}
        \caption{}
    \end{subfigure}
    ~
    \begin{subfigure}{.45\textwidth}
        \centering
        \includegraphics[height=4.5cm]{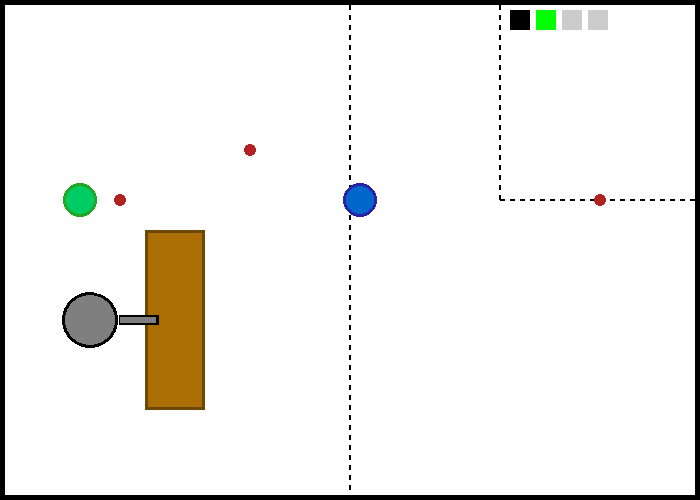}
        \caption{}
    \end{subfigure}
    ~
    \begin{subfigure}{.45\textwidth}
        \centering
        \includegraphics[height=4.5cm]{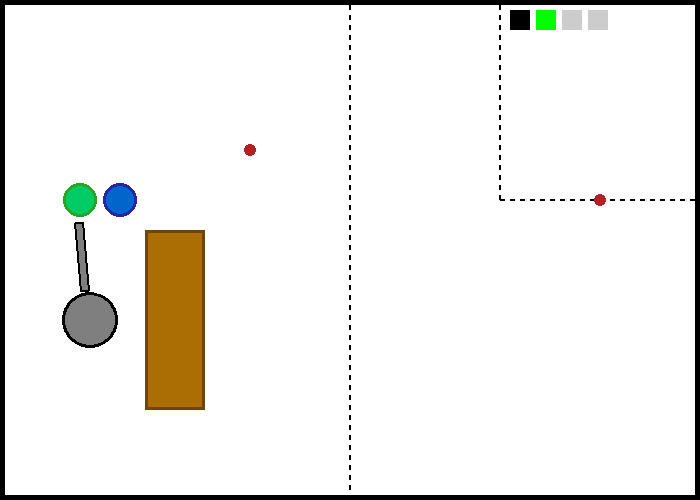}
        \caption{}
    \end{subfigure}
    \caption{Behaviors of running a centralized policy trained with mobile robots' velocity $v=0.6$. (a) The robot arm searches for the first tool for the human while mobile robots move towards the table; (b-c) Blue mobile robot gets the tool from the robot arm and then delivers it to the human, meanwhile, the robot arm starts to search for the second tool; (d) The human obtains the correct tool and moves on the next step, while the robot arm passes the second tool to another mobile robot; (e) The green mobile robot keeps staying there waiting for the last tool, because delivering two tools together on its own is quicker than letting the blue one deliver the third one. (f) The green mobile robot gets the third tool from the robot arm, and finishes the entire delivery task at the end.}
    \label{WTD1}
\end{figure}

We visualized the trained centralized policy in our simulator to better understand the robots' behaviors, which shows that the robot arm successfully reasons about the correct tool the human needs per step and cooperates with mobile robots to finish all deliveries in the optimal way (shown in Fig.~\ref{WTD1}). 

\begin{figure}[t!]
    \centering
    \begin{subfigure}{.45\textwidth}
        \centering
        \includegraphics[height=4.5cm]{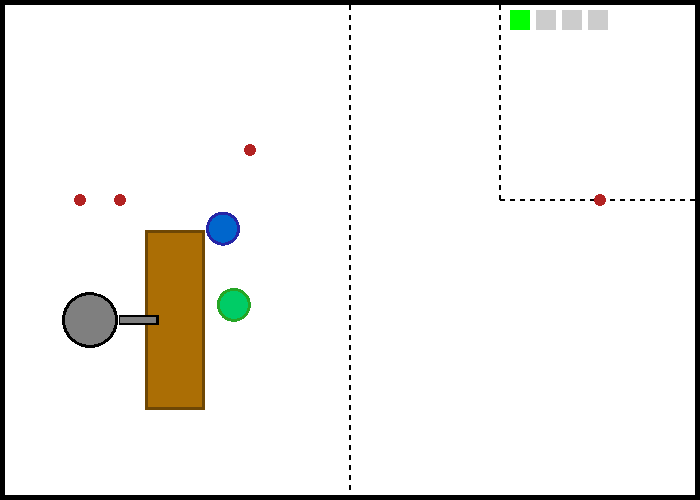}
        \caption{}
    \end{subfigure}
    ~
    \begin{subfigure}{.45\textwidth}
        \centering
        \includegraphics[height=4.5cm]{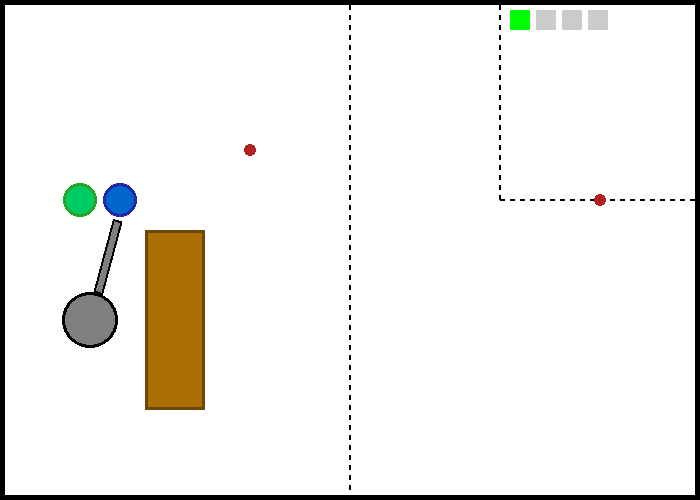}
        \caption{}
    \end{subfigure}
    ~
    \begin{subfigure}{.45\textwidth}
        \centering
        \includegraphics[height=4.5cm]{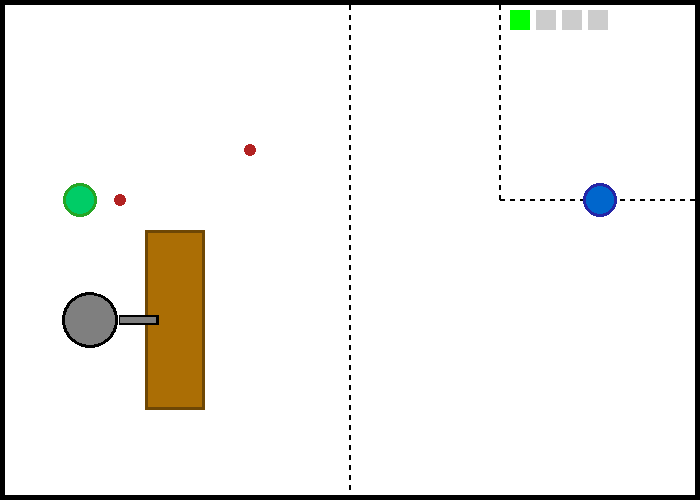}
        \caption{}
    \end{subfigure}
    ~
    \begin{subfigure}{.45\textwidth}
        \centering
        \includegraphics[height=4.5cm]{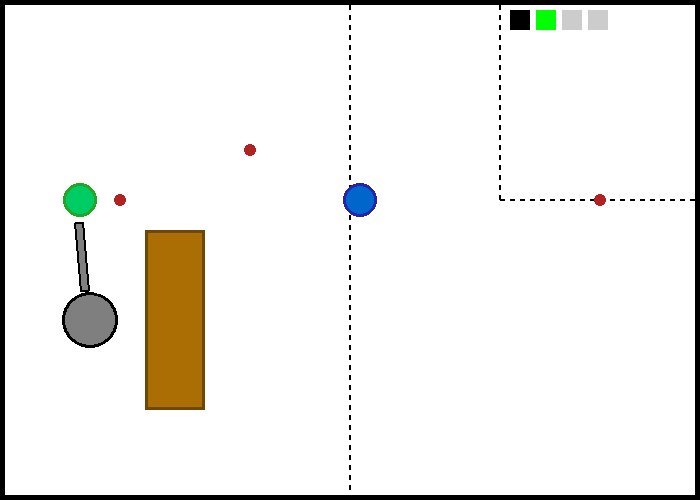}
        \caption{}
    \end{subfigure}
    ~
    \begin{subfigure}{.45\textwidth}
        \centering
        \includegraphics[height=4.5cm]{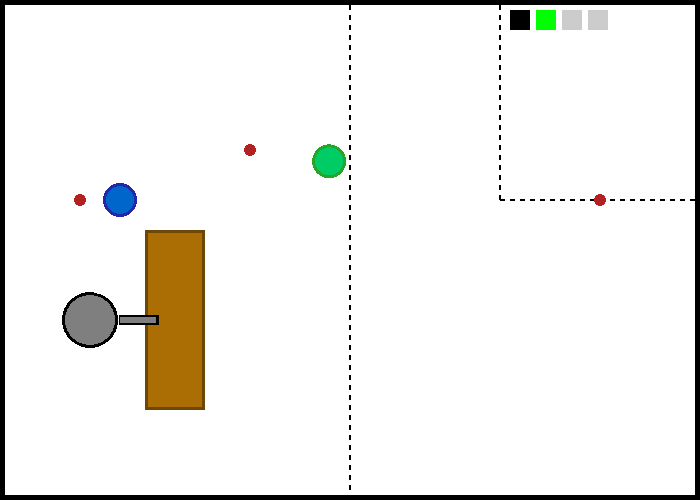}
        \caption{}
    \end{subfigure}
    ~
    \begin{subfigure}{.45\textwidth}
        \centering
        \includegraphics[height=4.5cm]{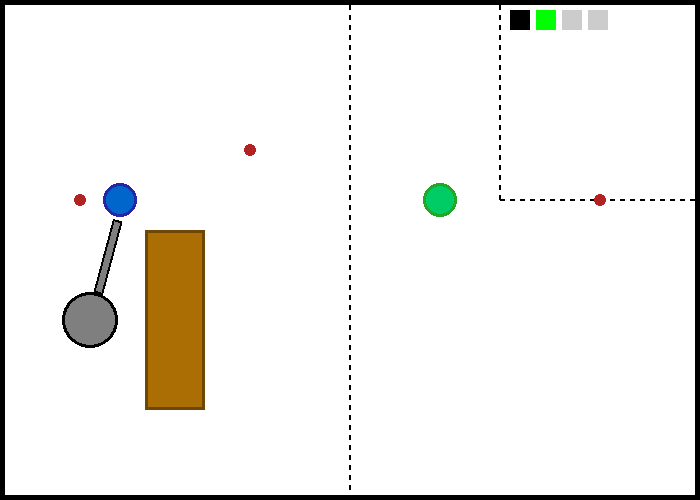}
        \caption{}
    \end{subfigure}
    \caption{Behaviors of running the same policy in Fig~\ref{WTD1} with a higher speed $v=0.8$. Differences happen on: (d) after getting the second tool from the robot arm, (e) the green mobile robot immediately goes to deliver it and the blue one has already come back to get the third tool. (f) The blue mobile robot then receives the last tool and finally completes the delivery task.}
    \label{WTD2}
\end{figure}

The high robustness of this centralized policy is further demonstrated by being examined under a higher velocity for mobile robots. 
Accordingly, the same policy generates a new collaborative strategy among robots, which is also optimal with respect to the speed change (shown in Fig.~\ref{WTD2}). 

\section{Conclusion}
\label{chap:paper1:con}
\noindent This chapter introduces the first formulation and approach for macro-action-based deep multi-agent reinforcement learning under partial observability. 
Both our decentralized and centralized learners achieve high-quality performance on two benchmark domains. Furthermore, the robots, in the warehouse domain, perform efficient and reasonable cooperation behaviors under the centralized policy. Importantly, the trained policy is naturally robust to the changes in macro-action execution. 


\chapter{Macro-Action-Based CTDE Q-Learning}
\label{chap:paper2}

\section{Introduction}
\label{chap:paper2:intro}

\noindent Multi-robot systems have become ubiquitous in our daily lives, such as drones for agricultural inspection, warehouse robots for cargo carrying, and self-driving cars~\cite{agricultural,waymo17,kiva}.
Although, in recent years, several multi-agent deep reinforcement learning (MARL) approaches have been proposed and have achieved excellent performance \cite{COMA,MADDPG,DecHDRQN,QMIX,vdn18}, these methods assume synchronized, primitive actions. 
However, in many real-world multi-robot tasks, high-quality solutions often require a team of robots to perform asynchronous actions under decentralized control without waiting for each other to continue.
In the chapter~\ref{chap:paper1}, we bridge this gap by proposing the first asynchronous macro-action-based multi-agent deep reinforcement learning frameworks. 
Macro-actions naturally represent temporally extended robot controllers that can be executed in an asynchronous manner~\cite{AAMAS14AKK,AmatoJAIR19}. 
The algorithms proposed in the previous chapter include both learning decentralized macro-action-value functions and centralized joint-macro-action-value functions. 
However, the decentralized method, using Decentralized Hysteretic DRQN with Double DQN (Dec-HDDRQN), performed quite poorly in the large and complex warehouse tool delivery domain. 
Nevertheless, decentralized execution is necessary for cases when there is limited or no communication between robots. 

In this chapter, we improve the learning of decentralized policies via two contributions: 
(a) A new macro-action-based decentralized deep double-Q learning approach, called MacDec-DDRQN, which adopts centralized training with decentralized execution by allowing each individual decentralized Q-net update to use a centralized Q-net; 
(b) MacDec-DDRQN introduces a choice of $\epsilon$-greedy exploration, either based on the centralized Q-net or the decentralized Q-nets. 
The best choice is often not clear without knowledge of domain properties. Therefore, a more general version of MacDec-DDRQN, called Parallel-MacDec-DDRQN, is proposed, in which, the centralized Q-net is trained purely based on the experiences generated by using centralized $\epsilon$-greedy exploration in one environment, simultaneously, agents perform decentralized exploration in a separate environment, and each decentralized Q-net is then optimized using the decentralized data and the centralized Q-net. 

We evaluate our methods both in simulation and on hardware. 
In simulation, our methods outperform the previous decentralized method by either converging to a much higher value or learning faster in both Box Pushing domain and Warehouse Tool Delivery domain with a single human involved. 
We also demonstrate the decentralized policies learned in simulation with real robots which shows high-quality cooperation to deliver the correct tools in an efficient way. 
To our knowledge, 
this is the first instance of decentralized macro-action-based policies trained via deep reinforcement learning run on a team of real robots.

\section{Approach}
\label{chap:paper2:approach}

\noindent In multi-agent environments, decentralized learning causes the environment to be non-stationary from each agent's perspective as other agents' policies change during learning. 
\emph{Centralized Training for Decentralized Execution} (CTDE) is the most popular framework that can naturally stabilize the learning of decentralized policies by overcoming the environmental non-stationarity issue and promote global cooperative behavior to be achieved via decentralized execution manner.  
One line of the implementation of CTDE is to use an \emph{actor-critic} framework, where a joint action-value function is learned and used to guide each agent's decentralized policy updating, originally proposed in COMA~\cite{COMA} and MADDPG~\cite{MADDPG}. 
VDN~\cite{vdn18} and QMIX~\cite{QMIX}, the pioneers of CTDE Q-learning, use centralized training by first training a joint, but factored, Q-net and later decomposing it into a decentralized Q-net for each agent to use in execution. 
Readers are referred to section~\ref{chap:BG:CTDE} for more details about these two means of CTDE. 

Differing from value-factorization-based approaches, in this section, we propose a new multi-agent Double DQN-based approach, called MacDec-DDRQN, to learn decentralized macro-action-value functions that are trained with a centralized macro-action-value function.

\subsection{Macro-Action-Based Decentralized Double Deep Recurrent Q-Net}

\noindent Double DQN has been implemented in multi-agent domains for learning either centralized or decentralized policies~\cite{xiao_corl_2019,MADDQN,WDDQN}. 
However, in the decentralized learning case, each agent independently adopts double Q-learning purely based on its own local information. 
Learning only from local information often impedes agents from achieving high-quality cooperation.    

In order to take advantage of centralized information for learning decentralized action-value functions, we train the centralized macro-action-value function $Q_{\phi}$ and each agent's decentralized macro-action-value function $Q_{\theta_i}$ simultaneously, 
and the target value for updating decentralized macro-action-value function $Q_{\theta_i}$ is then calculated by using the centralized $Q_{\phi}$ for macro-action selection and the decentralized target-net $Q_{\theta_i^-}$ for value estimation.  

More concretely, consider a domain with $N$ agents, and both the centralized Q-network $Q_{\phi}$ and decentralized Q-networks $Q_{\theta_i}$ for each agent $i$ are represented as DRQNs~\cite{DRQN}. 
Here, it is important to note that the centralized Q-network $Q_\phi$ does not involve any factorization architecture. As a result, $Q_\phi$ is not restricted by any constraint and it is able to represent the true centralized macro-action-values in any given domain.    

The experience replay buffer $\mathbf{\mathcal{D}}$, a merged version of Mac-CERTs and Mac-JERTs, contains the tuples $\langle\mathbf{z}, \mathbf{m}, \mathbf{z'}, \mathbf{r^c}, \vec{r}^{\,c}\rangle$, where $\mathbf{z} = \{z_0,..., z_N\}$, $\mathbf{m}=\{m_0,...,m_N\}$ and $\mathbf{r^c}=\{r^c_0,...,r^c_N\}$. 
In each training iteration, agents sample a mini-batch of sequential experiences to first optimize the centralized macro-action-value function $Q_\phi$ in the way proposed in Section~\ref{chap:paper1:app:MacCenQ}, and then update each decentralized macro-action-value function by minimizing the squared TD error:        

\begin{equation}
    \mathcal{L}(\theta_i)=\mathbb{E}_{<\mathbf{z}, \mathbf{m}, \mathbf{z'}, \mathbf{r^c}, \vec{r}^{\,c} >\sim\mathbf{\mathcal{D}}}\Big[\big(y_i - Q_{\theta_i}(h_i, m_i)\big)^2 \Big] 
\end{equation}

\noindent where,

\begin{equation}
    y_i=r^c_i + \gamma Q_{\theta^-_i}\biggr[h_i',\big[\argmax_{\mathbf{m'}}Q_{\phi}(\mathbf{h'}, \mathbf{m'})\big]_i\biggr]
    \label{ctde_unCondi}
    \vspace{4mm}
\end{equation}

In Eq.~\ref{ctde_unCondi}, $\big[\argmax_{\mathbf{m'}}Q_{\phi}(\mathbf{h'}, \mathbf{m'})\big]_i$ implies selecting the joint macro-action with the highest value and then selecting the individual macro-action for agent $i$. 
In this updating rule, not only double estimators $Q_{\theta_i^-}$ and $Q_{\phi}$ are applied to counteract overestimation on target Q-values, but also a centralized heuristic on action selection is embedded. 
Now, from each agent's perspective, the target Q-value is calculated by assuming all agents will behave based on the centralized Q-net next step (Eq.~\ref{ctde_unCondi}), in which the provided global info by the centralized Q-net will help each agent to avoid trapping in local optima and also facilitates them to learn cooperation behaviors.      

Additionally, similar to the idea of the conditional operation for training centralized joint macro-action-value function discussed in Section~\ref{chap:paper1:app:MacCenQ}, in order to obtain a more accurate prediction by taking each agent's macro-action executing status into account, Eq.~\ref{ctde_unCondi} can be rewritten as:

\begin{equation}
    y_i=r^c_i + \gamma Q_{\theta^-_i}\biggr[h_i',\big[\argmax_{\mathbf{m'}}Q_{\phi}(\mathbf{h'}, \mathbf{m'}\mid \mathbf{m^{undone}})\big]_i\biggr]
    \label{ctde_Condi}
    \vspace{4mm}
\end{equation}

\noindent Now, the agent who has not finished the macro-action at the target updating time step is considered to continue running the same macro-action in the target action-value computation.  

\begin{algorithm}[t!]
    \footnotesize
    \caption{Parallel-MacDec-DDRQN}
    \label{alg1}
        \begin{algorithmic}
            \State Initialize centralized Q-Networks: $Q_\phi$, $Q_\phi^-$
            \State Initialize decentralized Q-Networks for each agent $i$: $Q_{\theta_i}$, $Q_{\theta_i}^-$
            \State Initialize two parallel environments \emph{cen-env}, \emph{dec-env}
            \State Initialize two step counters $t_{\text{cen-env}}$, $t_{\text{dec-env}}$
            \State Initialize centralized buffer $\mathcal{D}_{\text{cen}} \leftarrow$ Mac-JERTs
            \State Initialize decentralized buffer $\mathcal{D}_{\text{dec}} \leftarrow$ Mac-CERTs
            \State Get initial joint-macro-observation $\vec{z}$ for agents in \emph{cen-env}
            \State Get initial macro-observation $z_i$ for each agent $i$ in \emph{dec-env}
            \For{\emph{dec-env-episode} = $1$ to $M$}
                \State Agents take joint-macro-action with cen-$\epsilon$-greedy using $Q_{\phi}$
                \State Store $\langle \vec{z}, \vec{m}, \vec{z}{\,}',\vec{r}^{\,c}\rangle$ in $\mathcal{D}_{\text{cen}}$
                \State $t_{\text{cen-env}}\leftarrow t_{\text{cen-env}} + 1$
                \State Each agent $i$ takes macro-action with dec-$\epsilon$-greedy using $Q_{\theta_i}$
                \State Store $\langle z_i, m_i, z_i',r^{c}_i\rangle$ in $\mathcal{D}_{\text{dec}}$
                \State $t_{\text{dec-env}}\leftarrow t_{\text{dec-env}} + 1$
                \If{$t_{\text{dec-env}}$ mod $I_{\text{train}} == 0$}
                    \State Sample a mini-batch $\mathcal{B}_{\text{cen}}$ of sequential experiences 
                    \State $\langle \vec{z}, \vec{m}, \vec{z}{\,}',\vec{r}^{\,c}\rangle$ from $\mathcal{D}_{\text{cen}}$
                    \State Perform a gradient decent step on $\bigr(y-Q_\phi(\vec{h}, \vec{m})\bigr)^2_{\mathcal{B}_{\text{cen}}}$, where
                    \State $y = \vec{r}\,^c + \gamma Q_{\phi-}\bigl(\vec{h}\,', \argmax_{\vec{m}\,'} Q_{\phi}(\vec{h}\,', \vec{m}\,'\mid \vec{m}^{\text{undone}})\bigr)$.
                    \State Sample a mini-batch $\mathcal{B}_{\text{dec}}$ of sequential experiences 
                    \State $\langle z_i, m_i, z_i',r^{\,c}_i\rangle$ from $\mathcal{D}_{\text{dec}}$ for each agent $i$ 
                    \State Perform a gradient decent step on $\bigr(y_i-Q_{\theta_i}(h_i, m_i)\bigr)^2_{\mathcal{B}_{\text{dec}}}$, where
                    \State $y_i=r^c_i + \gamma Q_{\theta^-_i}\biggr[h_i',\big[\argmax_{\mathbf{m'}}Q_{\phi}(\mathbf{h'}, \mathbf{m'}\mid \mathbf{m^{undone}})\big]_i\biggr]$

                \EndIf
                \If{$t_{\text{dec-env}}$ mod $I_{\text{TargetUpdate}} == 0$}
                    \State Update centralized target network $\phi^-\leftarrow\phi$
                    \State Update each agent's decentralized target network $\theta_i^-\leftarrow\theta_i$
                \EndIf
                \If{$t_{\text{cen-env}}==$max-episode-length \textbf{or} terminal state}
                    \State Reset \emph{cen-env} 
                    \State Get initial joint-macro-observation $\vec{z}$ for agents in \emph{cen-env}
                \EndIf
                \If{$t_{\text{dec-env}}==$max-episode-length \textbf{or} terminal state}
                    \State Reset \emph{dec-env} 
                    \State Get initial macro-observation $z_i$ for each agent $i$ in \emph{dec-env}
                \EndIf
            \EndFor
        \end{algorithmic}
\end{algorithm}

\subsection{Parallel-MacDec-DDRQN}

\noindent Exploration is also a difficult problem in multi-agent reinforcement learning. 
$\epsilon$-greedy exploration has been widely used in many methods such as Q-learning to generate training data~\cite{Sutton1998}. 
In DQN-based methods, as a hyper-parameter, $\epsilon$ often acts with a linear decay along with the training steps from $1.0$ to a lower value to achieve the trade-off between exploration and exploitation. And, exploration can be done either in a centralized way or in a decentralized way. 
Centralized exploration may help to choose cooperative actions more often that would have a low probability of being selected from decentralized policies, and decentralized exploration may provide more realistic data that is actually achievable by decentralized policies.      

Therefore, in our approach, besides tuning $\epsilon$, we introduce a hyper-selection for performing a $\epsilon$-greedy behavior policy that can perform either centralized exploration based on $Q_\phi$ or decentralized exploration using each agent's $Q_{\theta _i}$.  

However, without having enough knowledge about the properties of a given domain in the very beginning, it is not clear which exploration choice is the best. 
To cope with this, we propose a more generalized version of MacDec-DDRQN, called \emph{Parallel-MacDec-DDRQN}, summarized in Algorithm~\ref{alg1}. 
The core idea is to have two parallel environments with agents respectively performing centralized exploring (cen-$\epsilon$-greedy) and decentralized exploring (dec-$\epsilon$-greedy) in each. The centralized $Q_\phi$ is first trained purely using the centralized experiences, while each agent's decentralized $Q_{\theta_i}$ is then optimized using Eq.~\ref{ctde_Condi} with only decentralized experiences. The performance of this algorithm under the Warehouse domain is presented in Sectioin~\ref{chap:paper2:sim:re}.

\section{Simulated Experiments}
\label{chap:paper2:sim}

\subsection{Experimental Setup}

\noindent To evaluate the proposed approaches, we consider the same Box Pushing and Warehouse Tool Delivery domains introduced in Section~\ref{chap:paper1:exp:domain}.
We perform comparisons with the macro-action-based fully decentralized learning (Dec-HDDRQN) and fully centralized learning (Cen-DDRQN) introduced in Section~\ref{chap:paper1:app}. 
In the end, we also show the results of an ablation study to further demonstrate the importance of choosing target action from the centralized perspective to update decentralized Q-nets and the effectiveness of training two types of macro-action-value functions in parallel under separate environments.    

\begin{figure}[t!]
    \centering
    \captionsetup[subfigure]{labelformat=empty}
    \centering
    \subcaptionbox{\hspace{5mm}(a) 4$\times$4}
        [0.45\linewidth]{\includegraphics[height=4.7cm]{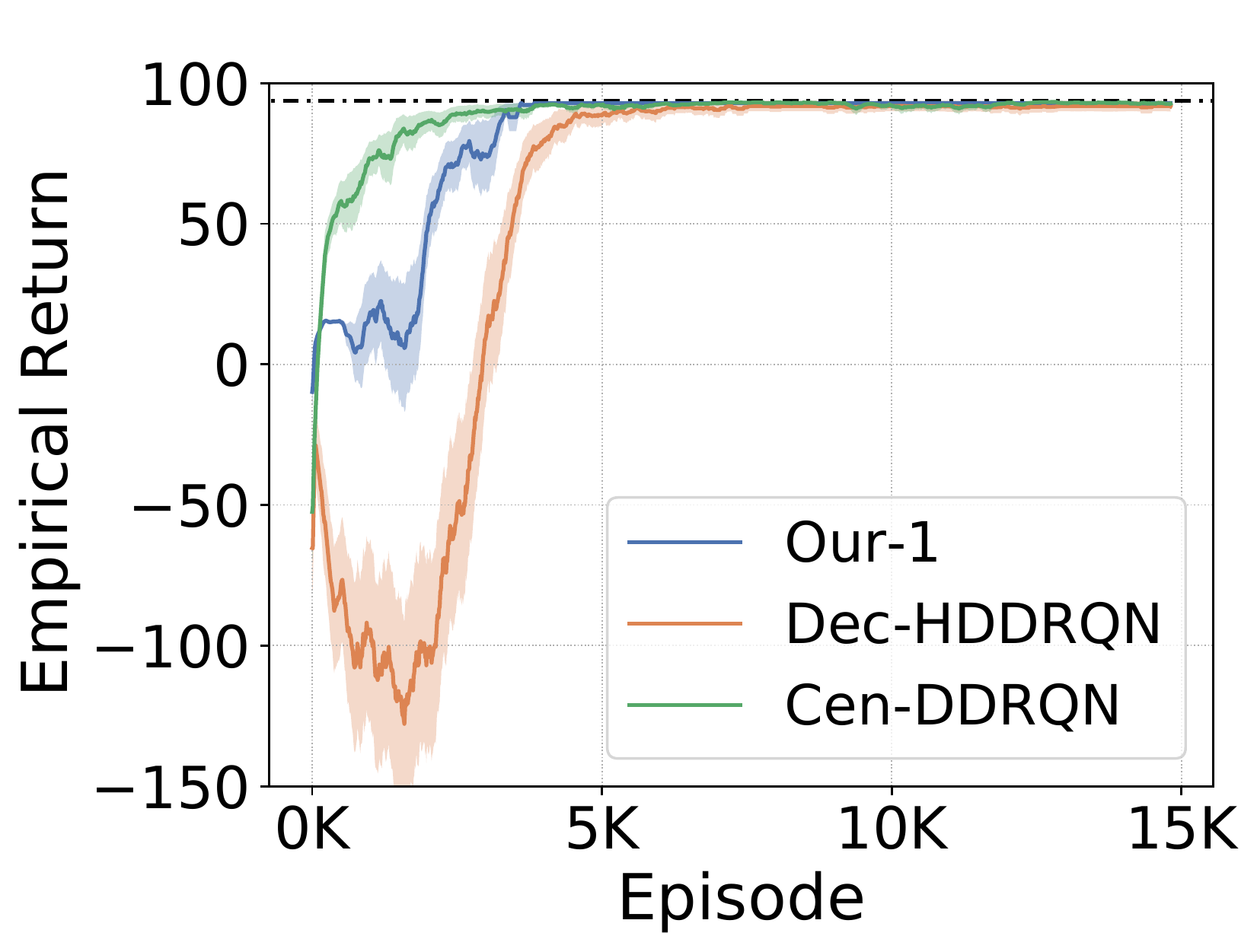}}
    ~
    \centering
    \subcaptionbox{\hspace{5mm}(b) 6$\times$6}
        [0.45\linewidth]{\includegraphics[height=4.5cm]{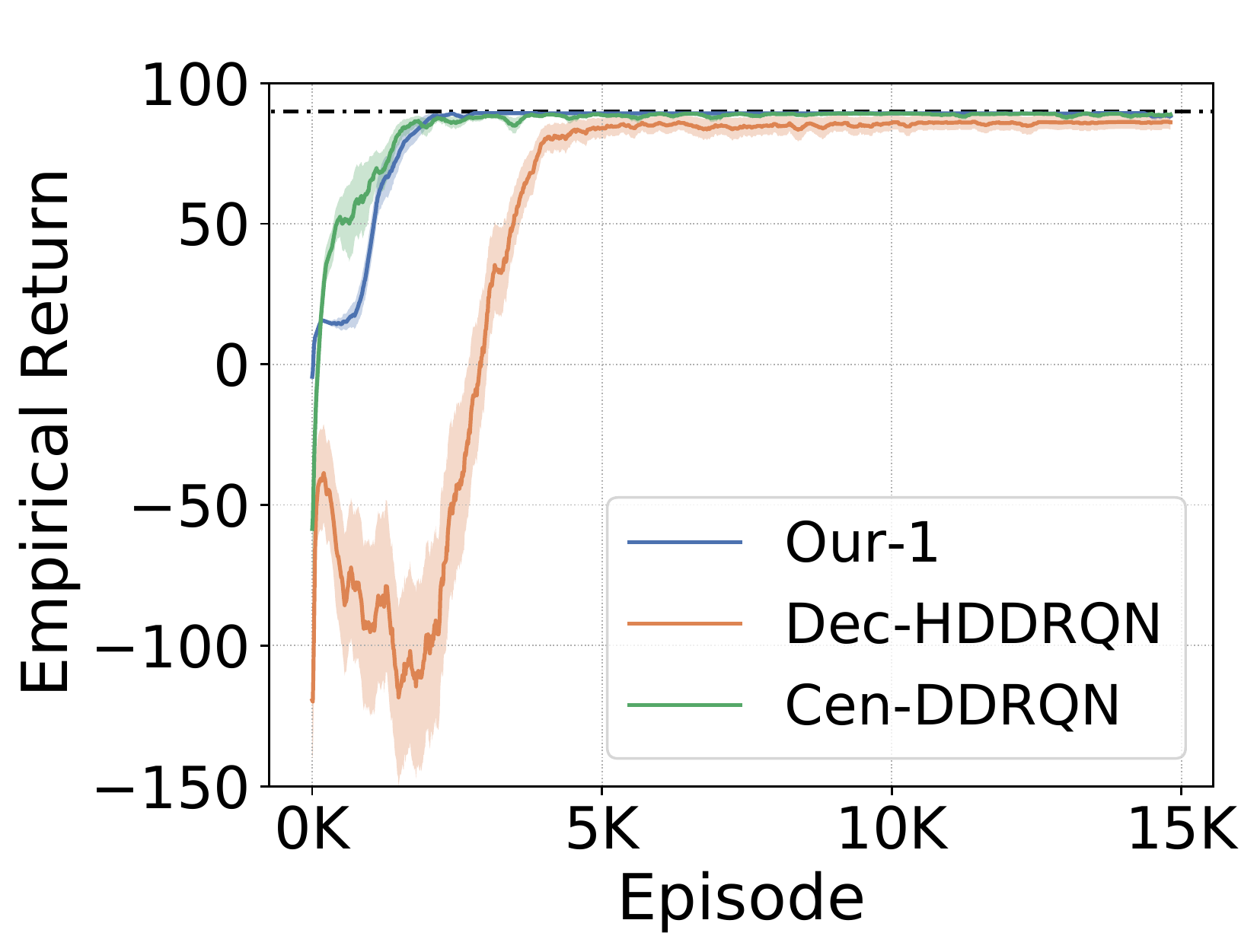}}
    ~
    \centering
    \subcaptionbox{\hspace{5mm}(c) 8$\times$8}
        [0.45\linewidth]{\includegraphics[height=4.5cm]{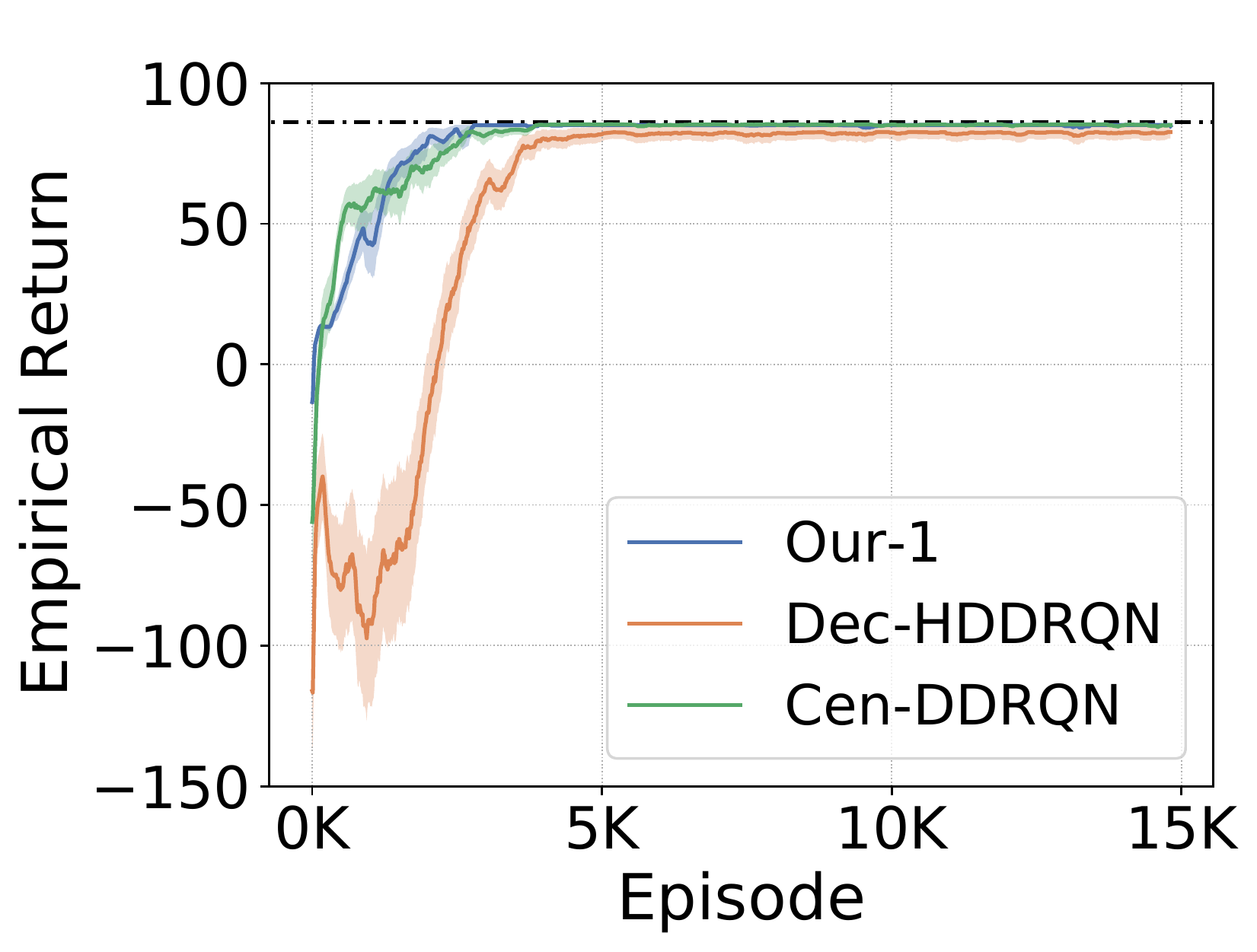}}
    ~
    \centering
    \subcaptionbox{\hspace{5mm}(d) 10$\times$10}
        [0.45\linewidth]{\includegraphics[height=4.5cm]{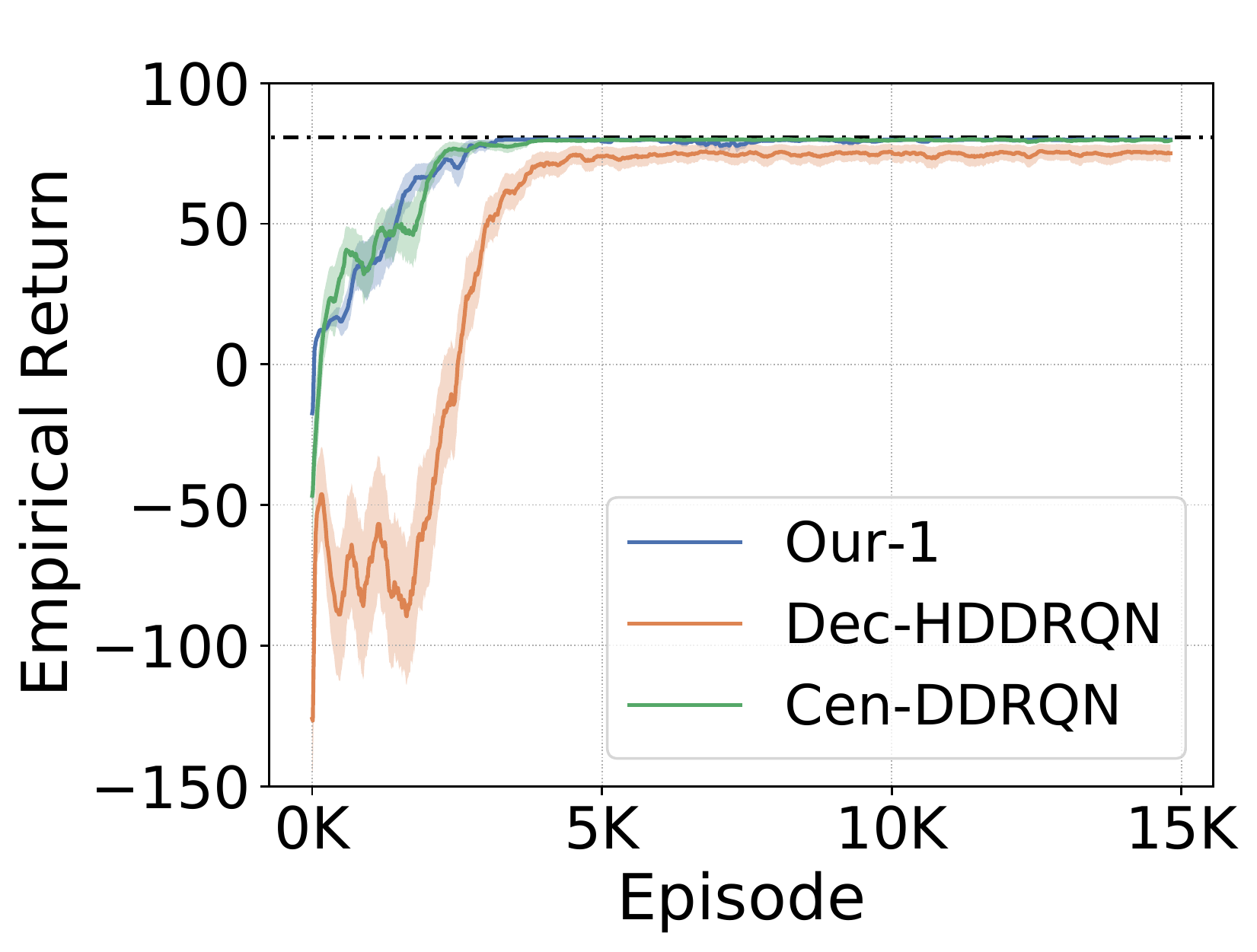}}
    ~
    \centering
    \subcaptionbox{\hspace{5mm}(e) 20$\times$20}
        [0.45\linewidth]{\includegraphics[height=4.5cm]{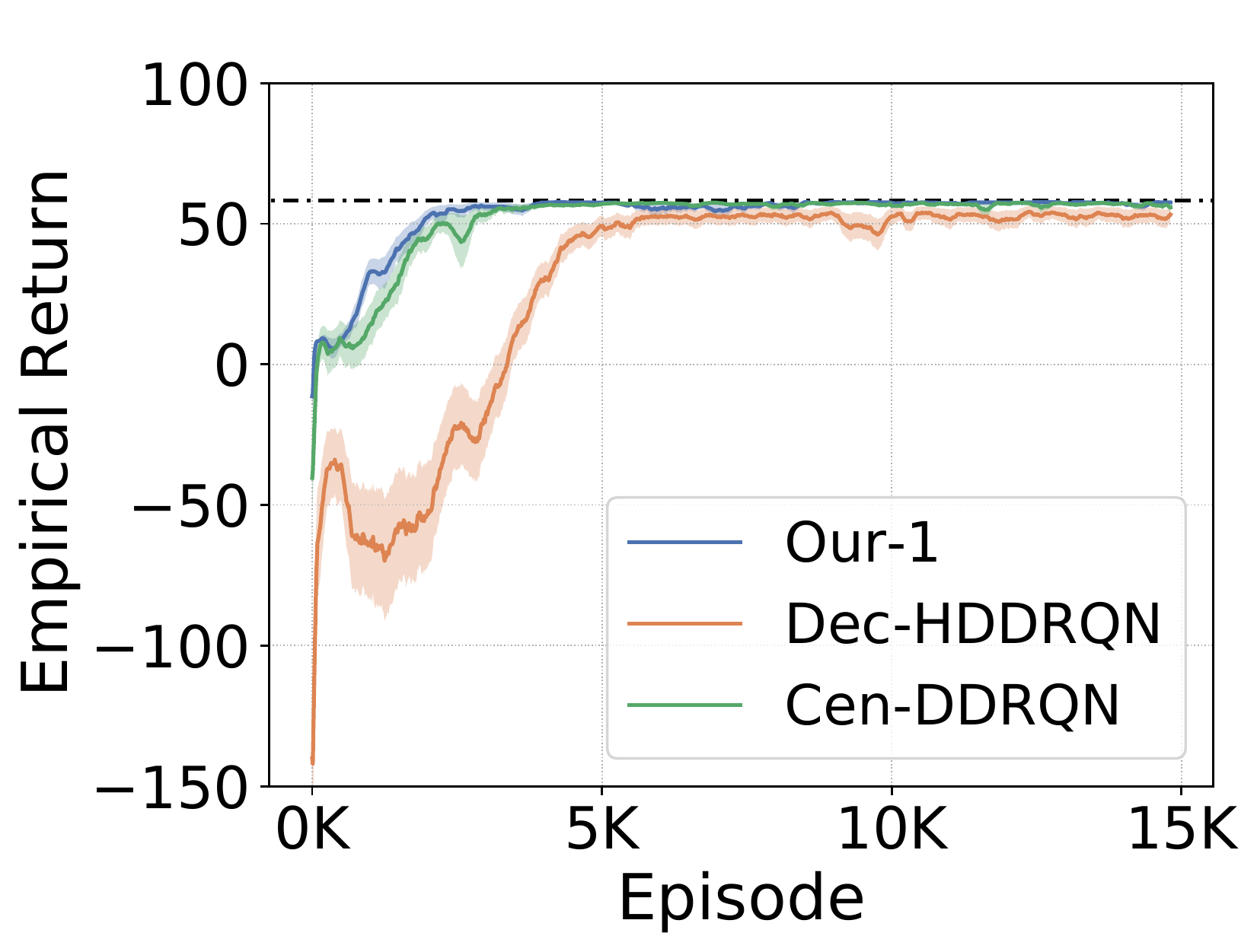}}
    ~
    \centering
    \subcaptionbox{\hspace{5mm}(f) 30$\times$30}
        [0.45\linewidth]{\includegraphics[height=4.5cm]{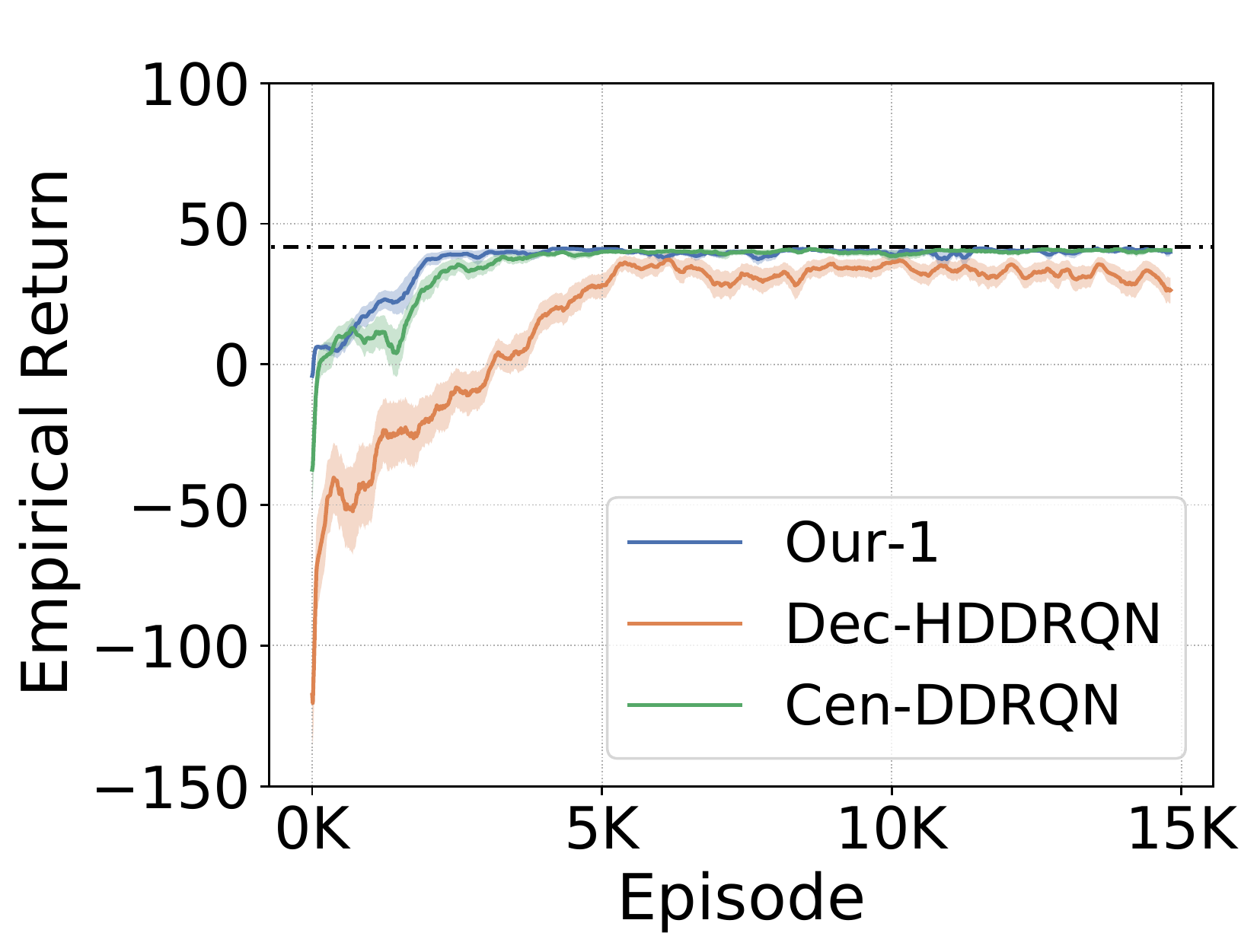}}
    \caption{Comparison of the average performance via three different learning approaches in the Box Pushing domain.}
    \label{bpmaresults}
\end{figure}

\subsection{Results}
\label{chap:paper2:sim:re}

\noindent \textbf{Results in the Box Pushing Domain}. We first evaluate our method MacDec-DDRQN (Our-1) with centralized $\epsilon$-greedy exploration in the box pushing domain, and compare its performance with Dec-HDDRQN and Cen-DDRQN. In all three methods, the decentralized Q-net consists of two MLP layers, one LSTM layer~\cite{LSTM} and another two MLP layers, in which there are 32 neurons on each layer with Leaky-Relu as the activation function for MLP layers. The centralized Q-net has the same architecture but 64 neurons in the LSTM layer. The performance for two sizes of the domain is shown in Fig~\ref{bpmaresults}, which is the mean of the episodic discounted returns ($\gamma=0.98$) over 40 runs with standard error and smoothed by 20 neighbors. The optimal returns are shown as red dash-dot lines. 

The results shown in Fig~\ref{bpmaresults} verify the significant advantage of having the centralized $Q_\phi$ in the double-Q updating (Eq.~\ref{ctde_Condi}) of MacDec-DDRQN such that it achieves similar performance to Cen-DDRQN and converges to the optimal returns earlier than Dec-HDDRQN. 
Furthermore, in the  bigger world space (eg., 20$\times$20 and 30$\times$30), our method even leads to slightly faster learning than the fully centralized approach. 
This is because centralized Q-learning deals with the joint macro-observation and joint macro-action space, which is much bigger than the decentralized spaces from each agent's perspective. 
MacDec-DDRQN has the key benefit of utilizing centralized information but learning decentralized policies over smaller observation and action spaces.   
 
\textbf{Results in the Warehouse Tool Delivery Domain}. We test our second proposed algorithm Parallel-MacDec-DDRQN (Our-2) in this warehouse domain using the same evaluation procedure mentioned above.
\begin{figure}[t!]
    \centering
    \includegraphics[height=5cm]{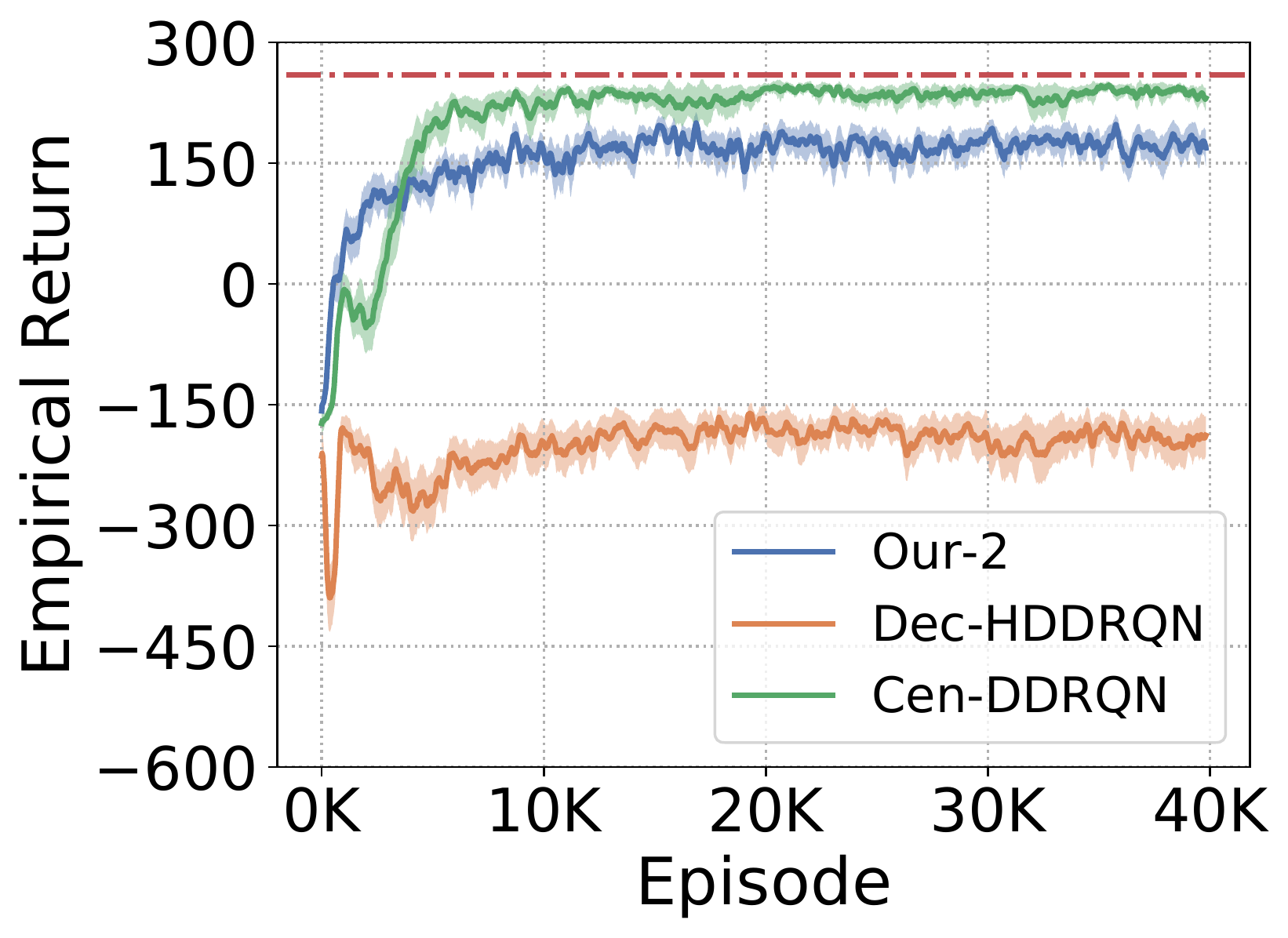}
    \caption{Comparison of the average performance via three different learning approaches in the Warehouse Tool Delivery domain.}
    \label{paper2:result_wtd}
\end{figure}
The results shown in Fig.~\ref{paper2:result_wtd} are generated by using the same neural network architecture as the one adopted in the BP domain but with 32 neurons in each MLP layer and 64 neurons in LSTM layer for both the centralized Q-net and each decentralized Q-net because of the bigger macro-action and macro-observation spaces.  

The most challenging part in this domain is that robots need to reason about collaborations among teammates and which tool the human will need next. However, the gray robot, who plays the key role in finding the correct tool first for delivery, does not have any knowledge about the human's needs nor any direct observation of the human's status. Also, the mobile robots cannot observe each other. From the gray robot's perspective, the reward for its selection is very delayed, which depends on the mobile robots' choice and their moving speeds. For these reasons, each robot individually learning from local signals (in Dec-HDDRQN) leads to much lower performance but the centralized learner can achieve near-optimal results. 
Our approach achieves significant improvements while learning decentralized policies, but due to the limitation of local information, it inherently cannot perform as well as the centralized learner in such a complicated domain. Nevertheless, the near-optimal behaviors are still learned by our Parallel-MacDec-DDRQN, which are presented in the real robot experiments (Section~\ref{chap:paper2:hardware}).

\begin{figure}[t!]
    \centering
    \includegraphics[height=4.6cm]{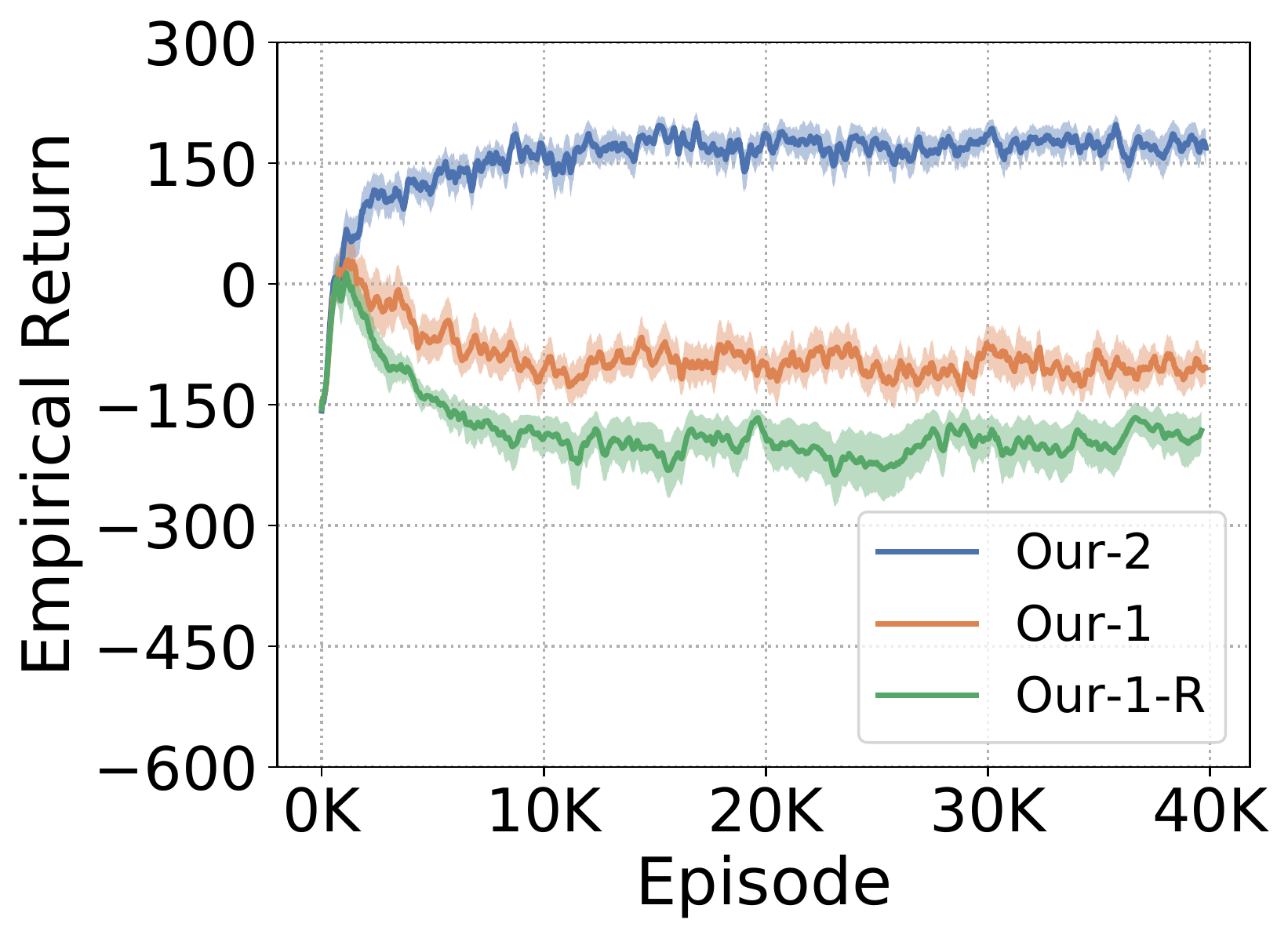}
    \caption{Results of ablation experiments in the Warehouse Tool Delivery Domain.}
    \label{ablation}
\end{figure}

We also conducted ablation experiments in WTD in order to investigate 1) the necessity of separately training the centralized Q-net and decentralized Q-nets in two environments by comparing Parallel-MacDec-DDRQN (Our-2) with MacDec-DDRQN with centralized exploration (Our-1); 2) the significance of including centralized $Q_\phi$ in double-Q updating to optimize each decentralized $Q_{\theta_i}$ (Eq.~\ref{ctde_Condi}) by performing Our-1 with regular deep double-Q learning (referred to Our-1-R). 
The results shown in Fig.~\ref{ablation} reveal that Our-2 outperforms another two ablations, which gives the affirmative answers to the above questions. 
We observe that Our-1 ends up with a significant drop in performance compared with Our-2. As the behavior policy in Our-1 is fully centralized, the poor performance of Our-1 indicates that the centralized sequential data is not practicable for training decentralized policies. 
In the parallel case, however, the centralized experiences in one environment guarantee the centralized Q-net can be well trained, and in the other environment, decentralized policies are trained based on decentralized data but have target actions suggested from the centralized perspective. The improved learning performance (the blue curve in Fig.~\ref{ablation}) confirms the efficiency of the parallel training paradigm. Also, without querying target actions from the centralized Q-net, Our-1-R's performance becomes worse than Our-1, which indicates that our new double-Q updating idea indeed plays a key role.  

\section{Hardware Experiments}
\label{chap:paper2:hardware}

To demonstrate that the learned decentralized policies in Parallel-MacDec-DDRQN can effectively control a team of robots to achieve high-quality results in practice, we recreated the warehouse domain using three real robots: one Fetch robot~\cite{FetchRobot} and two Turtlebots~\cite{Turtlebot}. The real-world environmental setup and learned collaborative behaviors are displayed in the following sections. 

\begin{figure}[t!]
    \centering
    \includegraphics[height=7cm]{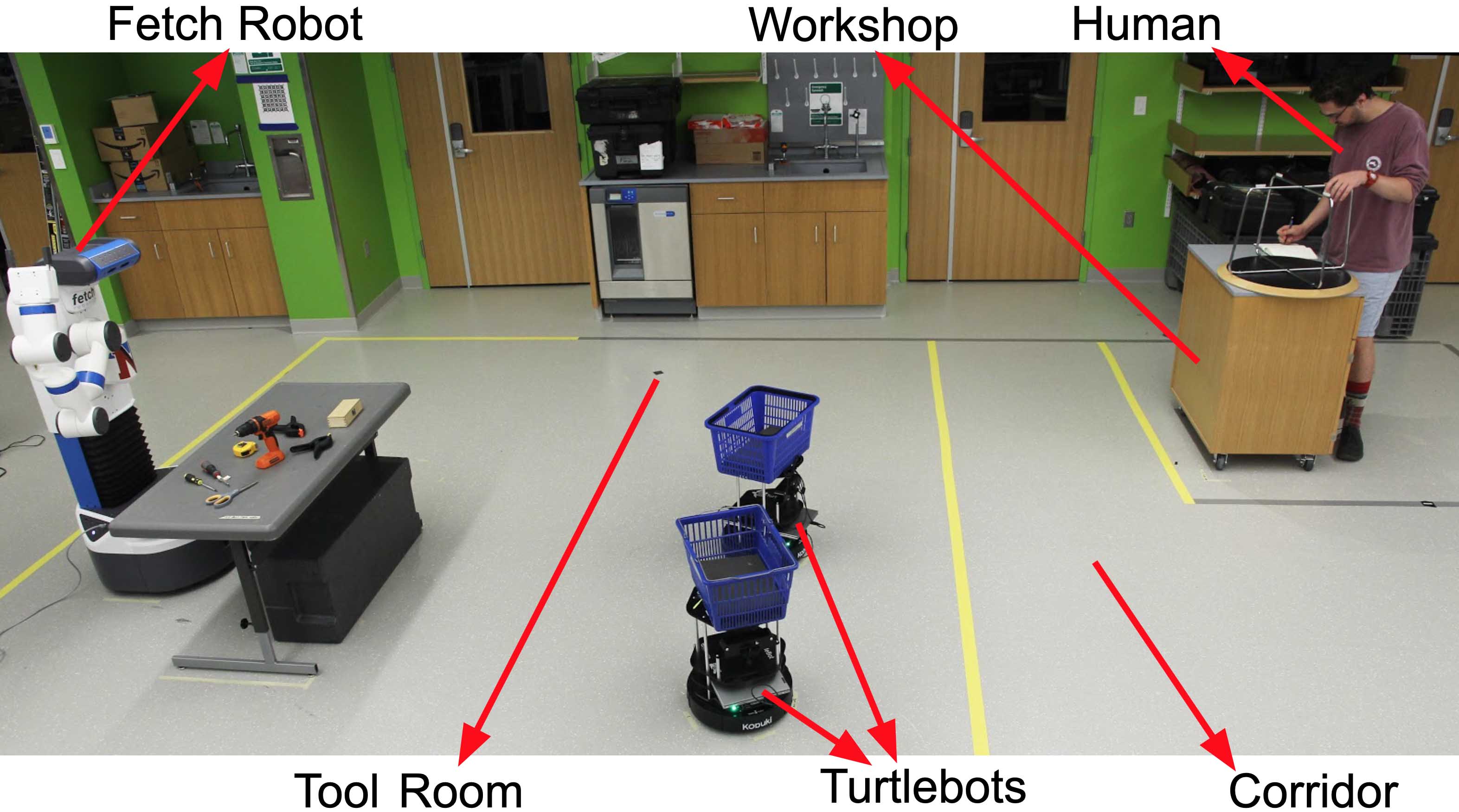}
    \caption{Hardware experiment setup.}
    \label{HW_Fig}
\end{figure}

\subsection{Experimental Setup}
\label{chap:paper2:hardware:setup}

As shown in Fig~\ref{HW_Fig}, a rectangle space with a dimension of 5.0m by 7.0m was taped to resemble the warehouse environment in the simulation (Fig~\ref{wtdma}). 
All the predefined waypoints and robots' initial positions were placed equally in ratio to the simulation. 
Also, the real-world human's task is to build a small table in the workshop, requiring three particular tools in the following order: a tape measure, a clamp, and an electronic drill (from YCB object set~\cite{YCB}).  

Each robot had its own decentralized macro-observation space designed over ROS~\cite{ROS} services that kept broadcasting the signals about Turtlebots' locations, human's state (only accessible to the Turtlebot when it is located in the workshop area), the status of each Turtlebot's basket, and the number of objects in the staging area (only observable in the tool room). Fetch's each manipulation macro-action is achieved by first projecting the point cloud data captured by Fetch's head camera into an OpenRAVE~\cite{Diankov:R} environment and performing motion planning using OMPL~\cite{OMPL} library. Turtlebot's movement macro-actions are controlled via the ROS navigation stack.

\begin{figure}[t!]
    \centering
    \subcaptionbox{Fetch searches and stages the tape measure as T-1 approaches the table.\label{realrobots_a}}
        [0.45\linewidth]{\includegraphics[width=7cm, height=3.8cm]{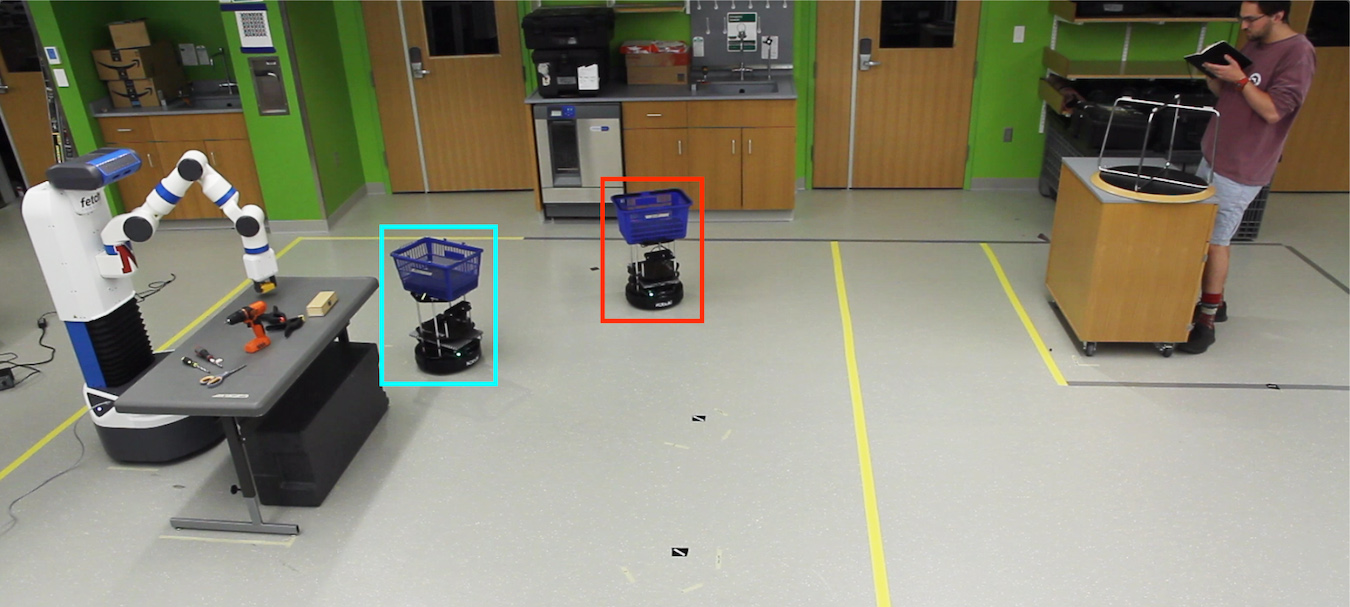}}
    ~
    \centering
    \subcaptionbox{Fetch sees T-1 arriving and passes it the tape measure, while T-0 reaches the workshop and observes the human's state.\label{realrobots_b}}
        [0.45\linewidth]{\includegraphics[width=7cm, height=3.8cm]{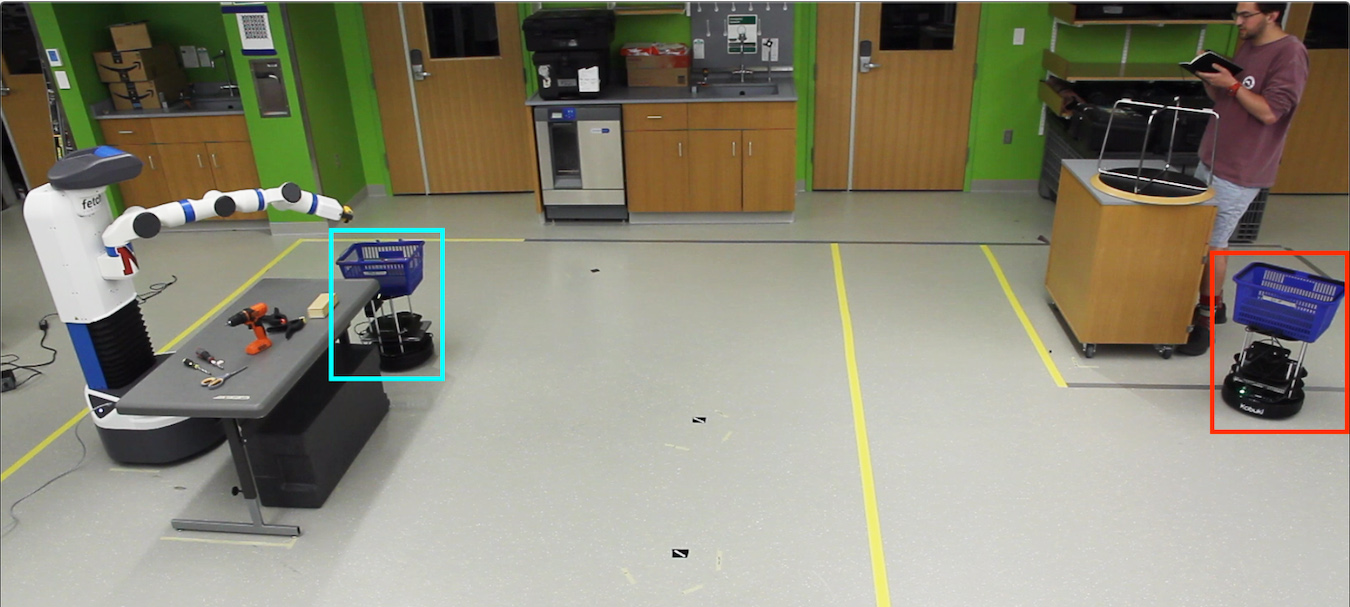}}
    ~
    \centering
    \subcaptionbox{T-1 observes the tape measure in its basket and moves to the workshop, while T-0 goes back tool room and Fetch finds the clamp. \label{realrobots_c}}
        [0.45\linewidth]{\includegraphics[width=7cm, height=3.8cm]{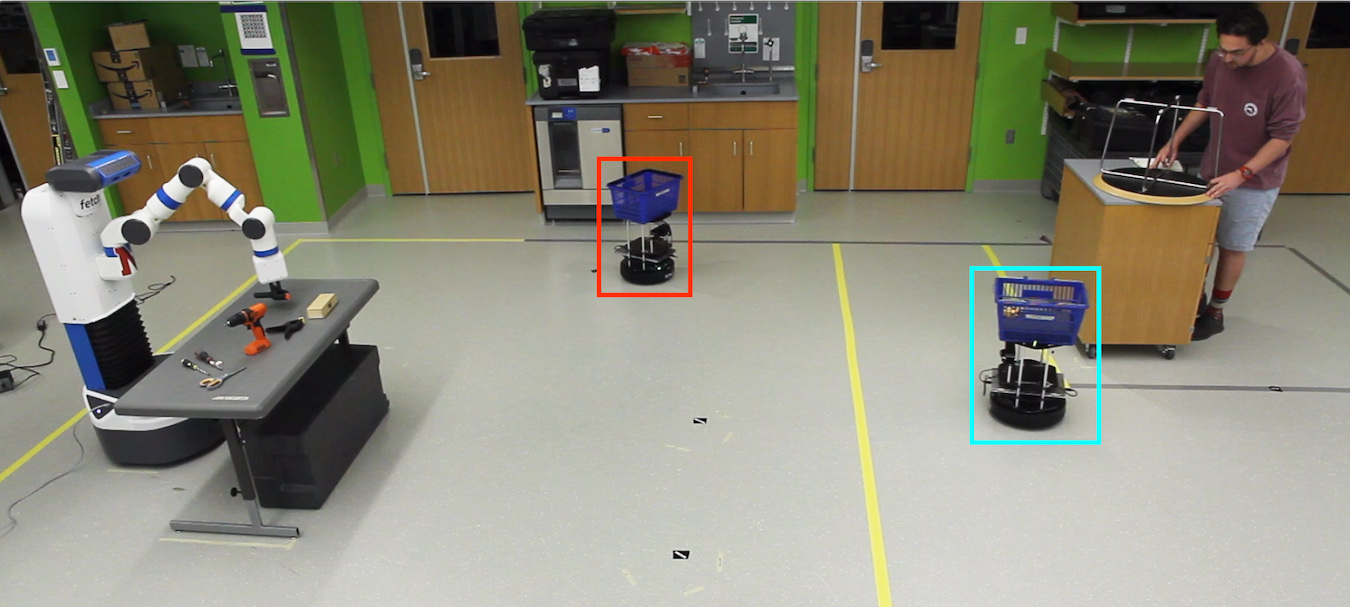}}
    ~
    \centering
    \subcaptionbox{T-1 deliveries the tape measure and T-0 runs to the table for the second tool, while Fetch notices no teammate around the table yet.   \label{realrobots_d}}
        [0.45\linewidth]{\includegraphics[width=7cm, height=3.8cm]{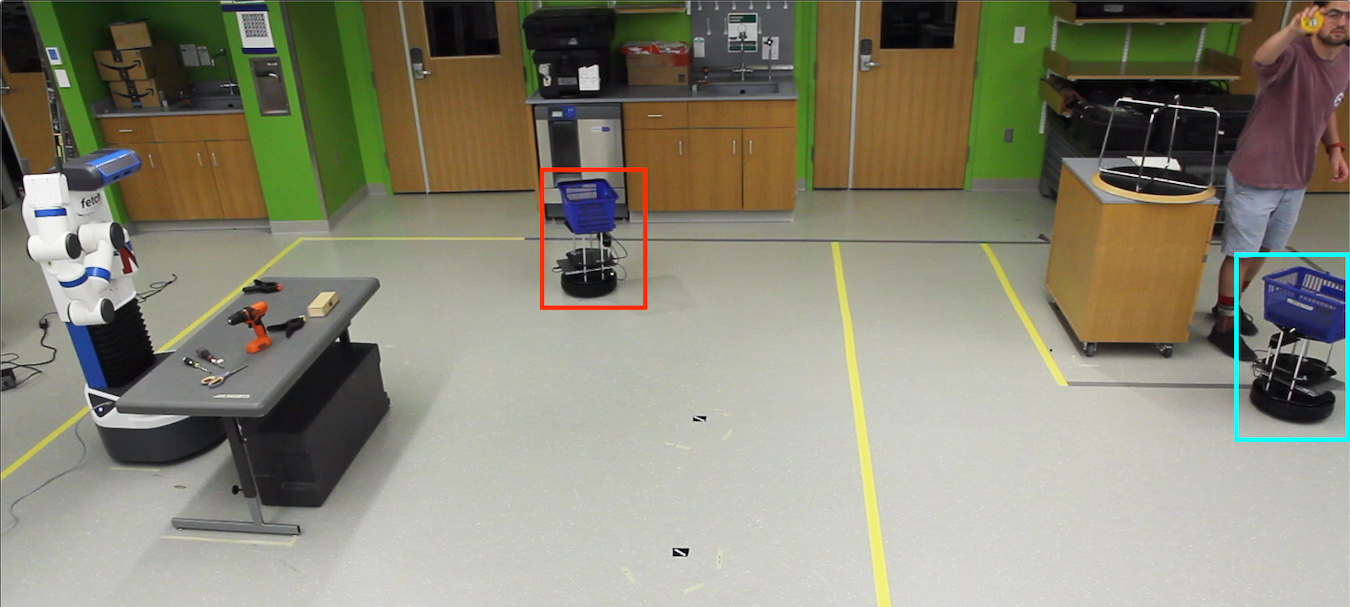}}
    ~
    \centering
    \subcaptionbox{Fetch grabs the electronic drill and stages it next to the clamp, while T-0 waits beside the table and T-1 is coming back.\label{realrobots_e}}
        [0.45\linewidth]{\includegraphics[width=7cm, height=3.8cm]{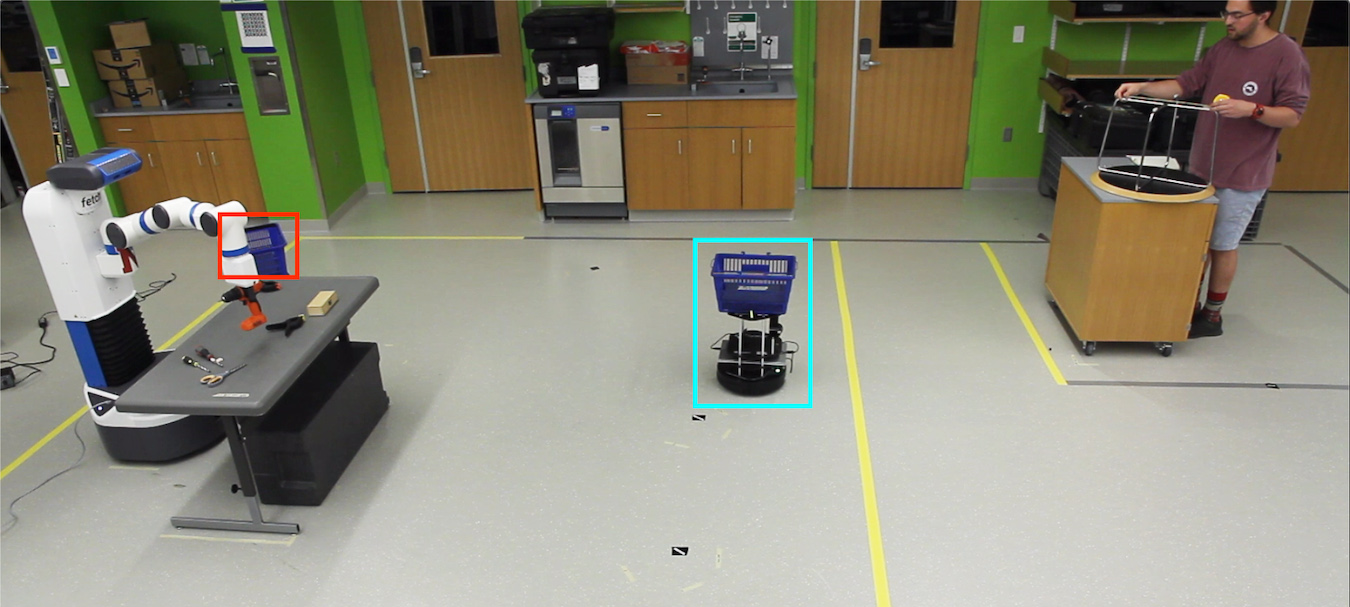}}
    ~
    \centering
    \subcaptionbox{Fetch observes T-0 has been ready there and passes clamp to it, in the meantime, T-1 arrives at the table.\label{realrobots_f}}
        [0.45\linewidth]{\includegraphics[width=7cm, height=3.8cm]{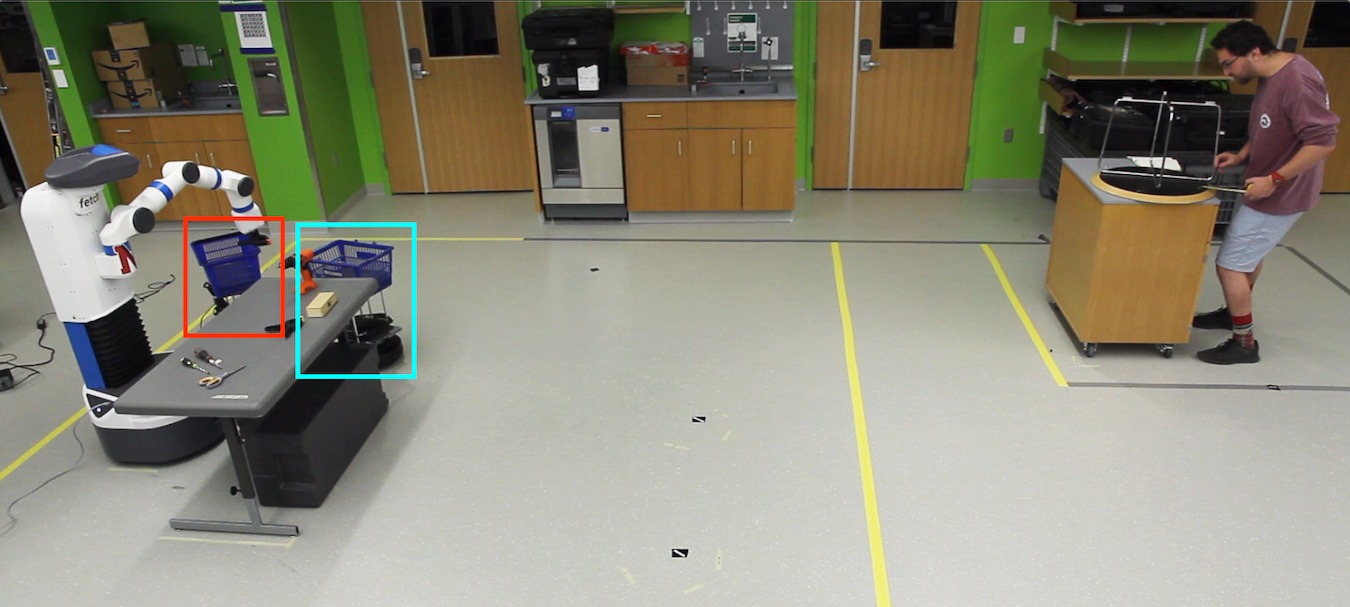}}
    \caption{Behaviors of robots running the decentralized policies (learned via Parallel-MacDec-DDRQN) in the warehouse domain, where Turtlebot-0 (T-0) is bounded in red and Turtlebot-1 (T-1) is bounded in blue.} 
    \label{hw_re1}
\end{figure}

\begin{figure}[t!]
    ~
    \centering
    \subcaptionbox{T-0 immediately goes to send the 2nd tool and Fetch passes the last tool to T-1.\label{realrobots_g}}
        [0.45\linewidth]{\includegraphics[width=7cm, height=3.8cm]{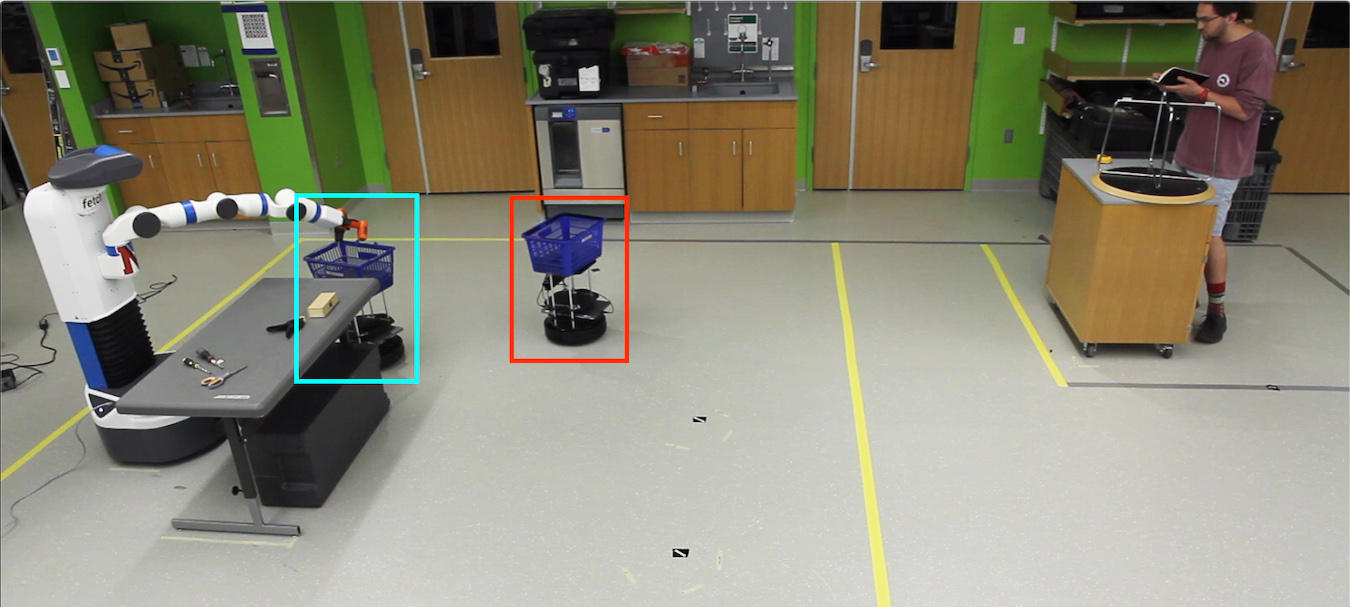}}
    ~
    \centering
    \subcaptionbox{Human gets the clamp from T-0, and T-1 is going to deliver the electronic drill.\label{realrobots_h}}
        [0.45\linewidth]{\includegraphics[width=7cm, height=3.8cm]{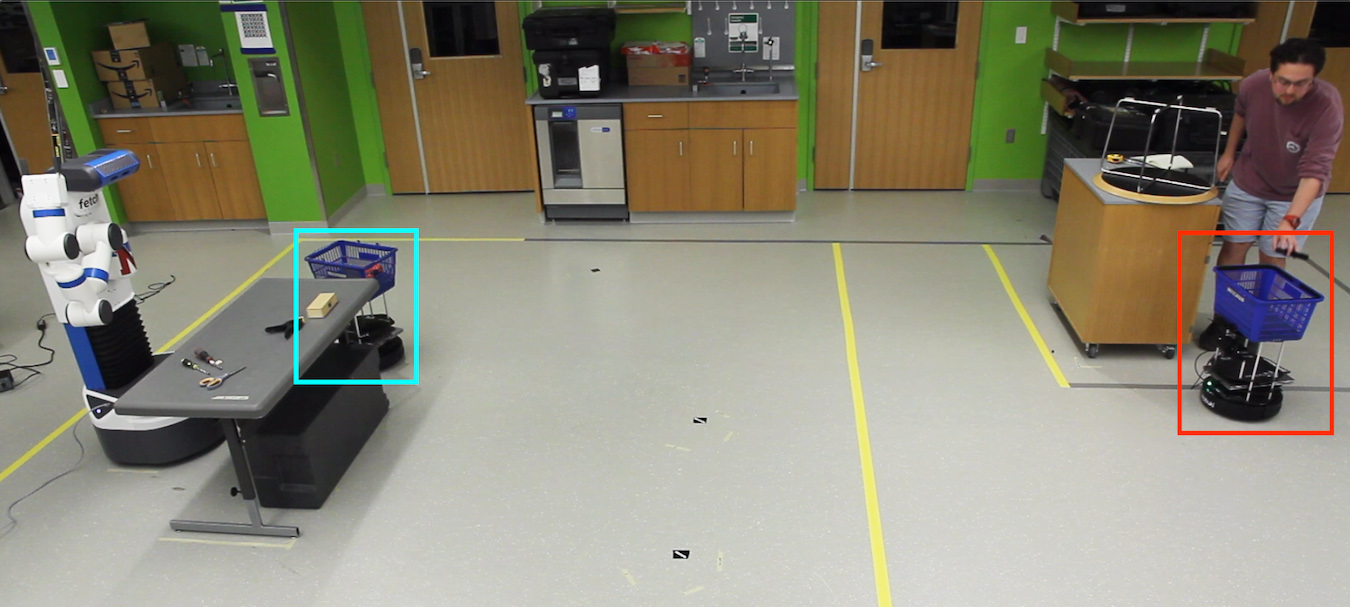}}
    ~
    \centering
    \subcaptionbox{The last tool is passed to the human by T-1 and the entire delivery task is completed.\label{realrobots_i}}
        [0.9\linewidth]{\includegraphics[width=7cm, height=3.8cm]{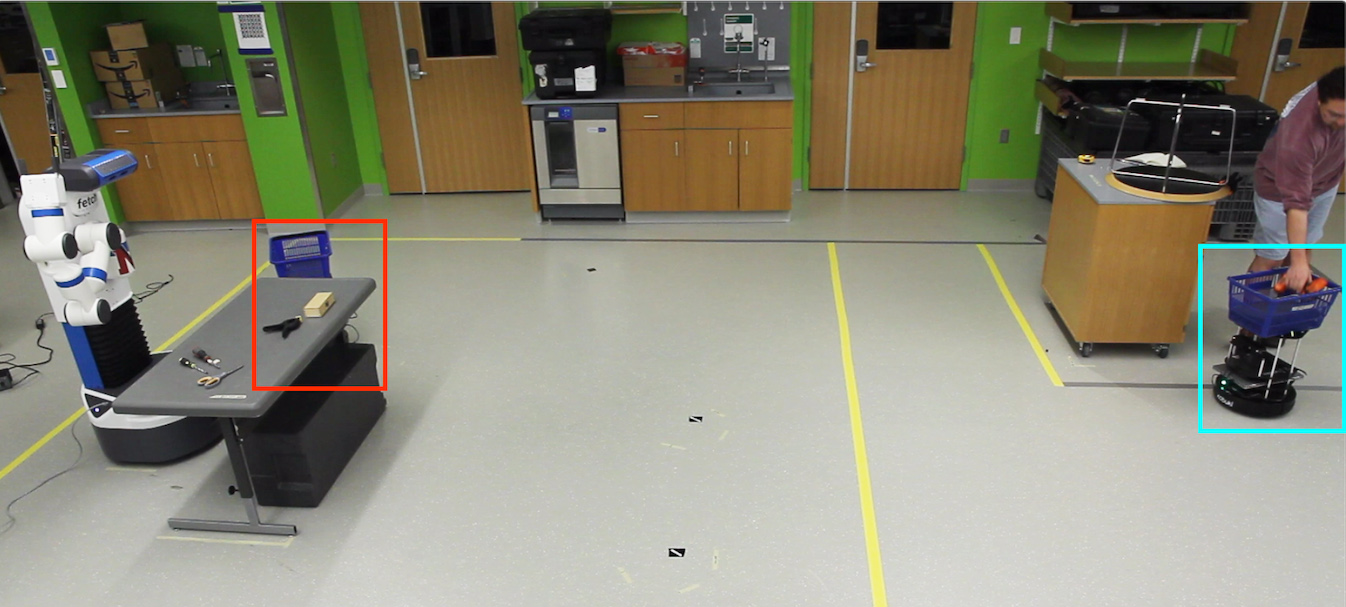}}
    \caption{Behaviors of robots running the decentralized policies (learned via Parallel-MacDec-DDRQN) in the warehouse domain followed with Fig.~\ref{hw_re1}.} 
    \label{hw_re2}
\end{figure}

\subsection{Results}
\label{chap:paper2:hardware:re}

\noindent Fig.~\ref{HW_Fig} shows the sequential cooperative behaviors performed by the robots. 
Although there is no direct interaction between the Fetch robot and the human, the trained policy indeed learned the correct tools that the human need for each future step and commanded Fetch to find them in the proper order. 
Furthermore, Fetch behaved well such that: 
(a) Fig.~\ref{realrobots_c}-\ref{realrobots_e}, after placing the clamp into the staging area followed by observing no Turtlebot beside the table, it continued to look for the third object instead of waiting for Turtlebot-0 (bounded in red) to come over; 
(b) Fig.~\ref{realrobots_e}-\ref{realrobots_f}, after finding the electronic drill, it first passed the clamp (the correct second object that the human needed) to Turtlebot-0 who arrived at the table ahead of Turtlebot-1(bounded in blue). 
In the meantime, Turtlebots were also clever in the way that: 
(a) they delivered the three tools in turn, instead of either letting one of them send all the tools or performing delivery only after having all the tools in the basket which actually would lead to human pausing; 
(b) they directly went to the human for delivery after obtaining a tool from Fetch without any redundant movement, e.g. going to the tool room waypoint again.

\section{Conclusion}
\label{chap:paper2:con}

\noindent This chapter introduces MacDec-DDRQN and Parallel-MacDec-DDRQN: two new macro-action-based multi-agent deep reinforcement learning methods with decentralized execution. These methods enable each agent's decentralized Q-net to be trained end-to-end while capturing the effects of other agents' actions by using a centralized Q-net and decentralized policy updating. The results in the benchmark Box Pushing domain demonstrate the advantage of our methods where the decentralized training achieves equally good performance as the centralized one. Furthermore, in the warehouse domain, our method outperforming Dec-HDDRQN confirms the benefits and the efficiency of our new double-Q updating rule. Importantly, a team of real robots running the decentralized policies learned via our method performed efficient and reasonable behaviors in the warehouse domain, which validates the usefulness of our macro-action-based deep RL frameworks in practice.



\chapter{Macro-Action-Based Actor-Critic Policy Gradients}
\label{chap:paper3}

\section{Introduction}
\label{chap:paper3:intro}

\noindent In recent years, multi-agent policy gradient methods using the actor-critic framework have achieved impressive success in solving a variety of cooperative and competitive domains~\cite{COMA,MAAC,MADDPG,VDAC,DOP,SQDDPG,LIIR,LICA,CM3,VinyalsAlphaStar,M3DDPG,hideseek,du2021learning}. However, as these methods assume synchronized primitive-action execution over agents, they struggle to solve large-scale real-world multi-agent problems that involve long-term reasoning and asynchronous behavior. 

As previous chapters introduced, the idea of temporally-extended actions has been incorporated into multi-agent settings. 
In particular, we consider the \emph{Macro-Action Decentralized Partially Observable Markov Decision Process} (MacDec-POMDP)~\cite{AAMAS14AKK,AmatoJAIR19}.
The MacDec-POMDP is a general model for cooperative multi-agent problems with partial observability and (potentially) different action durations. As a result, agents can start and end macro-actions at different time steps so decision-making can be asynchronous.

The MacDec-POMDP framework has shown strong scalability with planning-based methods (where the model is given)~\cite{ICRA15MacDec,RSS15,IJRR17DecPOSMDP,GDICE,xiao_icra_2018}.
In terms of multi-agent reinforcement learning (MARL), 
there have been many hierarchical approaches, 
they don't typically address asynchronicity since they assume agents' have  high-level decisions with the same  duration~\cite{schroeder:nips19,MAIntro-OptionQ,NachumAPGK19,WangK0LZITF20,wang:iclr2021,HAVEN,JiachenMAHRL}.
Only limited studies have considered asynchronicity~\cite{MendaCGBTKW19,DOC,xiao_corl_2019}, yet, none of them provides a general formulation for multi-agent policy gradients that allows agents to asynchronously learn and execute. 

In this chapter, we also assume a set of macro-actions has been predefined for each domain. This is well-motivated by the fact that, in real-world multi-robot systems, each robot is already equipped with certain controllers (e.g., a navigation controller, and a manipulation controller) that can be modeled as macro-actions~\cite{RSS15,IJRR17DecPOSMDP,wu2021spatial, xiao_corl_2019}. Similarly, as it is common to assume primitive actions are given in a typical RL domain, we assume the macro-actions are given in our case.  
The focus of the policy gradient methods is then on learning high-level policies over macro-actions.\footnote{Our approach could potentially also be applied to other models with temporally-extended actions \cite{IJRR17DecPOSMDP}.}

Our contributions include a set of macro-action-based multi-agent actor-critic methods that generalize their primitive-action counterparts. 
First, we formulate a \emph{macro-action-based independent actor-critic} (Mac-IAC) method. Although independent learning suffers from a theoretical curse of environmental non-stationarity, it allows fully online learning and may still work well in certain domains. Second, we introduce a \emph{macro-action-based centralized actor-critic} (Mac-CAC) method, for the case where full communication is available during execution.
We also formulate a centralized training for decentralized execution (CTDE) paradigm~\cite{OliehoekSV08,KraemerB16} variant of our method. 
CTDE has gained popularity since such methods can learn better decentralized policies by using centralized information during training. 
Current primitive-action-based multi-agent actor-critic methods typically use a centralized critic to optimize each decentralized actor. However, the asynchronous joint macro-action execution from the centralized perspective could be very different with the completion time being very different from each agent's decentralized perspective. 
To this end, we first present a \emph{Naive Independent Actor with Centralized Critic} (Naive IACC) method that naively uses a joint macro-action-value function as the critic for each actor's policy gradient estimation; and then propose a novel \emph{Independent Actor with Individual Centralized Critic} (Mac-IAICC) method that learns individual critics using centralized information to address the above challenge.    

We evaluate our proposed methods on diverse macro-action-based multi-agent problems: a benchmark Box Pushing domain~\cite{xiao_corl_2019}, a variant of the Overcooked domain~\cite{wu_wang2021too} and a larger warehouse service domain~\cite{xiao_corl_2019}. 
Experimental results show that our methods are able to learn high-quality solutions while primitive-action-based methods cannot, and show the strength of Mac-IAICC for learning decentralized policies over Naive IAICC and Mac-IAC. 
Decentralized policies learned by using Mac-IAICC are successfully deployed on real robots to solve a warehouse tool delivery task in an efficient way.

\section{Approach}
\label{chap:paper3:app}

\noindent Multi-agent deep reinforcement learning (MARL) with asynchronous decision-making and macro-actions is more challenging as it is difficult to determine \emph{when} to update each agent's policy and \emph{what} information to use. 
Although the macro-action-based deep Q-learning methods proposed in Chapter~\ref{chap:paper1} give us the base to learn macro-action value functions, they do not directly extend to the policy gradient case, particularly in the case of centralized training for decentralized execution (CTDE). 
In this section, we propose principled formulations of on-policy macro-action-based multi-agent actor-critic methods for decentralized learning (Section~\ref{chap:paper3:app:mac-iac}), centralized learning (Section~\ref{chap:paper3:app:mac-cac}), and CTDE (Section~\ref{chap:paper3:app:mac-iacc}). 
In each case, we first introduce the version with a Q-value function as the critic and then present the variance reduction version in our implementation. 
We use $h_i$ to represent an agent's local macro-observation-action history, and $\vec{h}$ to represent the joint history.

\begin{tcolorbox}[float=t!,colback=gray!15!white,colframe=gray!60!black,title=Theorem 1: Macro-Action-Based Policy Gradient Theorem (episodic case)]
As POMDPs can always be transformed to history-based MDPs, we can directly adapt the general Bellman equation for the state values of a hierarchical policy~\cite{Sutton:1999} to a macro-action-based POMDP by replacing the state $s$ with a history $h$ as follows: 
    \vspace{-3mm}
\begin{equation*}
    V^\Psi(h) = \sum_{m}\Psi(m|h)Q^{\Psi}(h,m) 
\end{equation*}
    \vspace{-5mm}
\begin{equation*}
    Q^\Psi(h,m) = r^c(h,m) + \sum_{h'}P(h'| h, m)V^{\Psi}(h') 
\end{equation*}
Next, we follow the proof of the policy gradient theorem~\cite{sutton2000policy}:
    \vspace{-3mm}
\begin{align*}
    \nabla_\theta V^{\Psi_\theta}(h)&=\nabla_\theta\Big[\sum_{m}\Psi_{\theta}(m| h)Q^{\Psi_\theta}(h,m)\Big] \\
    &=\sum_m\Big[\nabla_{\theta}\Psi_\theta(m| h)Q^{\Psi_\theta}(h,m) + \Psi_{\theta}(m| h)\nabla_{\theta}Q^{\Psi_\theta}(h,m)\Big]\\
    &=\sum_m\Big[\nabla_{\theta}\Psi_\theta(m| h)Q^{\Psi_\theta}(h,m) + \\
    &\,\,\,\,\,\,\,\,\,\Psi_{\theta}(m| h)\nabla_{\theta}\big(r^c(h,m) +\sum_{h'}P(h'| h, m)V^{\Psi_\theta}(h')\big)\Big]\\
    &=\sum_m\Big[\nabla_{\theta}\Psi_\theta(m| h)Q^{\Psi_\theta}(h,m) + \Psi_{\theta}(m| h)\sum_{h'}P(h'| h, m)\nabla_{\theta}V^{\Psi_\theta}(h')\big)\Big]\\
    &=\sum_{\hat{h}\in H}\sum_{k=0}^\infty P(h\rightarrow \hat{h}, k, \Psi_\theta)\sum_{m}\nabla_\theta\Psi_\theta(m|\hat{h})Q^{\Psi_\theta}(\hat{h},m)
\end{align*}
Then, we can have:
    \vspace{-3mm}
\begin{align*}
    \nabla_\theta J(\theta) &= \nabla_\theta V^{\Psi_\theta}(h_0)\\
    &=\sum_{h\in H}\sum_{k=0}^\infty P(h_0\rightarrow h, k, \Psi_\theta)\sum_{m}\nabla_\theta\Psi_\theta(m|h)Q^{\Psi_\theta}(h,m)\\
    &=\sum_{h'}\eta(h')\sum_{h}\frac{\eta(h)}{\sum_{h'}\eta(h')}\sum_{m}\nabla_\theta\Psi_\theta(m|h)Q^{\Psi_\theta}(h,m)\\
    &\propto\sum_{h}\mu^{\Psi_\theta}(h)\sum_{m}\nabla_\theta\Psi_\theta(m|h)Q^{\Psi_\theta}(h,m)
\end{align*}
    \label{theorem:1}
\end{tcolorbox}

\subsection{Macro-Action-Based Independent Actor-Critic}
\label{chap:paper3:app:mac-iac}

\noindent Similar to the idea of IAC with primitive-actions (Section~\ref{chap:BG:MARL:Dec}), a straightforward extension is to have each agent independently optimize its own macro-action-based policy (actor) using a local macro-action-value function (critic). 
Hence, we start with deriving a \emph{macro-action-based policy gradient theorem} in Theorem 1 by incorporating the general Bellman equation for the state values of a macro-action-based policy~\cite{Sutton:1999} into the \emph{policy gradient theorem} in MDPs~\cite{sutton2000policy} (assuming $\gamma=1$ for the episodic case), and then we extend it to MacDec-POMDPs so that each agent can have the following policy gradient w.r.t.~the parameters of its macro-action-based policy $\Psi_{\theta_i}(m_i|h_i)$: 

\begin{equation}
        \nabla_{\theta_i} J (\theta_i)=\mathbb{E}_{\vec{\Psi}_{\vec{\theta}}}\biggr[\nabla_{\theta_i}\log\Psi_{\theta_i}(m_i\mid h_i)Q^{\Psi_{\theta_i}}_{\phi_i}(h_i,m_i)\biggr]
        \label{Mac-IAC-Q}
\end{equation}

\begin{figure}[t!]
    \centering
    \includegraphics[height=6.5cm]{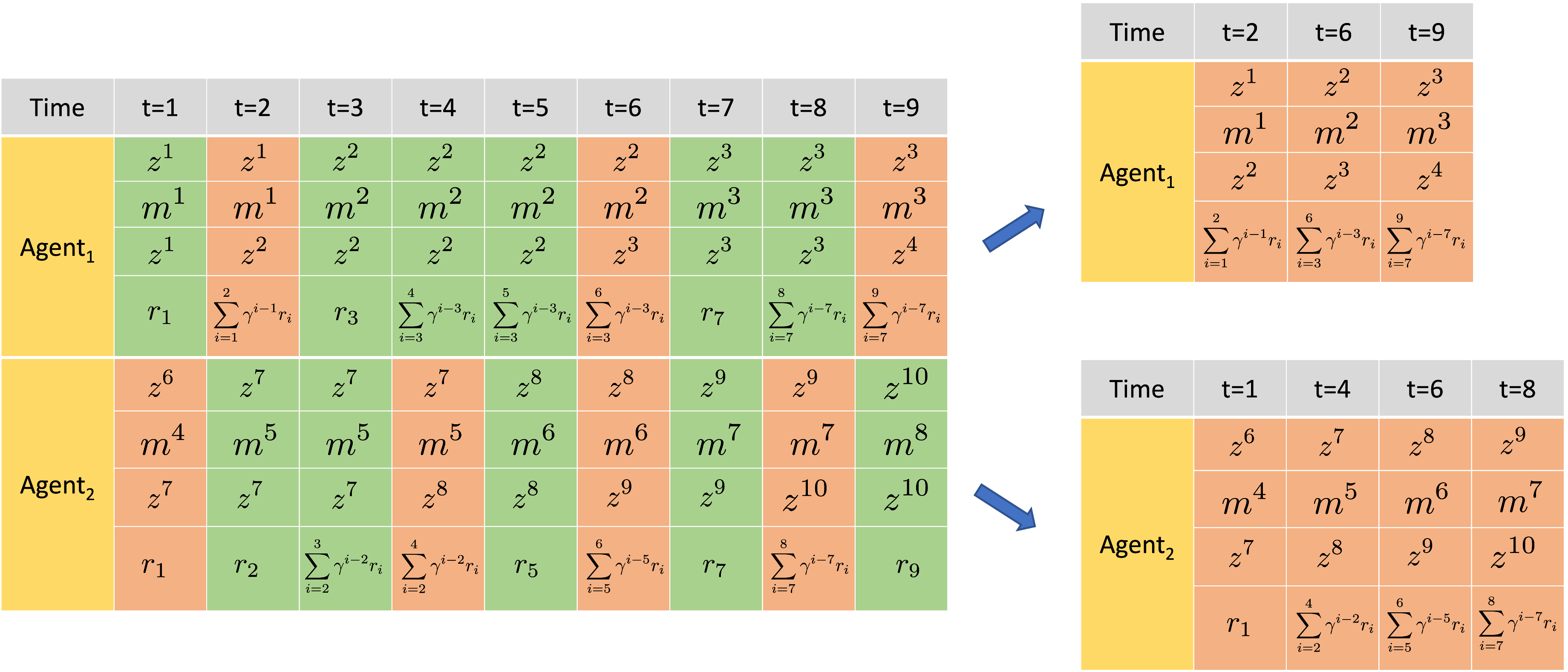}
    \caption{An example of the trajectory squeezing process in Mac-IAC. We collect each agent's high-level transition tuple at every primitive-step. Each agent is allowed to obtain a new macro-observation if and only if the current macro-action terminates, otherwise, the next macro-observation is set as the same as the previous one. Each agent separately squeezes its sequential experiences by picking out the transitions when its macro-action terminates (red cells). Each agent independently trains the critic and the policy using the squeezed trajectory.}
    \label{Mac-IAC-Traj}
\end{figure}

During training, each agent accesses its own trajectories and squeezes them in the way as shown in Fig.~\ref{Mac-IAC-Traj} to train the critic $Q_{\phi_i}^{\Psi_{\theta_i}}(h_i,m_i)$ via on-policy TD learning and perform gradient ascent using Eq.~\ref{Mac-IAC-Q} to update the policy when the agent's macro-action terminates. In our case, we train a local history value function $V^{\Psi_{\theta_i}}_{\mathbf{w}_i}(h_i)$ as each agent's critic and use it as a baseline to achieve variance reduction. The corresponding policy gradient is as follows: 

\begin{equation}
        \nabla_{\theta_i} J (\theta_i)=\mathbb{E}_{\vec{\Psi}_{\vec{\theta}}}\biggr[\nabla_{\theta_i}\log\Psi_{\theta_i}(m_i\mid h_i)A(h_i,m_i)\biggr]
        \label{Mac-IAC-V}
\end{equation}
\begin{equation}
        A(h_i,m_i) = r^c_i + \gamma^{\tau_{m_i}}V^{\Psi_{\theta_i}}_{\mathbf{w}_i}(h_i')-V^{\Psi_{\theta_i}}_{\mathbf{w}_i}(h_i)
\end{equation}
where, the cumulative reward $r^c_i$ is w.r.t. the execution of agent $i$'s macro-action $m_i$. We show the pseudocode of Mac-IAC in Algorithm~\ref{Mac-IAC-Code}. 

\begin{algorithm}[h!]
    \footnotesize
    \caption{Mac-IAC}
    \label{alg1}
        \begin{algorithmic}[1]
            \State Initialize a decentralized policy network for each agent $i$: $\Psi_{\theta_i}$
            \State Initialize decentralized critic networks for each agent $i$: $V_{\mathbf{w}_i}^{\Psi_{\theta_i}}$, $V_{\mathbf{w}_i^-}^{\Psi_{\theta_i}}$ 
            \State Initialize a buffer $\mathcal{D}$
            \For{\emph{episode} = $1$ to $M$}
                \State $t=0$
                \State Reset env
                \While{not reaching a terminal state \textbf{and} $t < \mathbb{H}$}
                    \State $t \leftarrow t + 1$
                    \For{each agent $i$}
                        \If{the macro-action $m_i$ is terminated}
                            \State $m_{i} \sim \Psi_{\theta_i}(\cdot \mid h_i; \epsilon)$
                        \Else
                            \State Continue running current macro-action $m_i$
                        \EndIf
                    \EndFor
                    \For{each agent $i$}
                        \State Get cumulative reward $r^c_i$, next macro-observation $z'_i$ 
                        \State Collect $\langle z_i,m_i, z'_i, r^c_i \rangle$ into the buffer $\mathcal{D}$
                    \EndFor
                \EndWhile
                \If{\emph{episode} mod $I_{\text{train}} = 0$}
                    \For{each agent $i$}
                        \State Squeeze agent $i$'s trajectories in the buffer $\mathcal{D}$
                        \State Perform a gradient descent step on $L(\mathbf{w}_i)=\big(y-V^{\Psi_{\theta_i}}_{\mathbf{w}_i}(h_i)\big)^2_\mathcal{D}$, where  $y = r^c_i + \gamma^{\tau_{m_i}} V^{\Psi_{\theta_i}}_{\mathbf{w}_i^-}(h_i')$ 
                        \State Perform a gradient ascent on:
                        \State $\nabla_{\theta_i} J(\theta_i) = \mathbb{E}_{\vec{\Psi}_{\vec{\theta}}}\Big[\nabla_{\theta_i}\log\Psi_{\theta_i}(m_i|h_i)\big(r^c_i + \gamma^{\tau_{m_i}} V^{\Psi_{\theta_i}}_{\mathbf{w}_i^-}(h_i')-V^{\Psi_{\theta_i}}_{\mathbf{w}_i}(h_i)\big)\Big]$
                    \EndFor
                    \State Reset buffer $\mathcal{D}$
                \EndIf
                \If{\emph{episode} mod $I_{\text{TargetUpdate}} = 0$}
                    \For{each agent $i$}
                        \State Update the critic target network $\mathbf{w}_i^-\leftarrow\mathbf{w}_i$
                    \EndFor
                \EndIf
            \EndFor
        \end{algorithmic}
        \label{Mac-IAC-Code}
\end{algorithm}

\subsection{Macro-Action-Based Centralized Actor-Critic}
\label{chap:paper3:app:mac-cac}

\begin{figure}[t!]
    \centering
    \includegraphics[height=4cm]{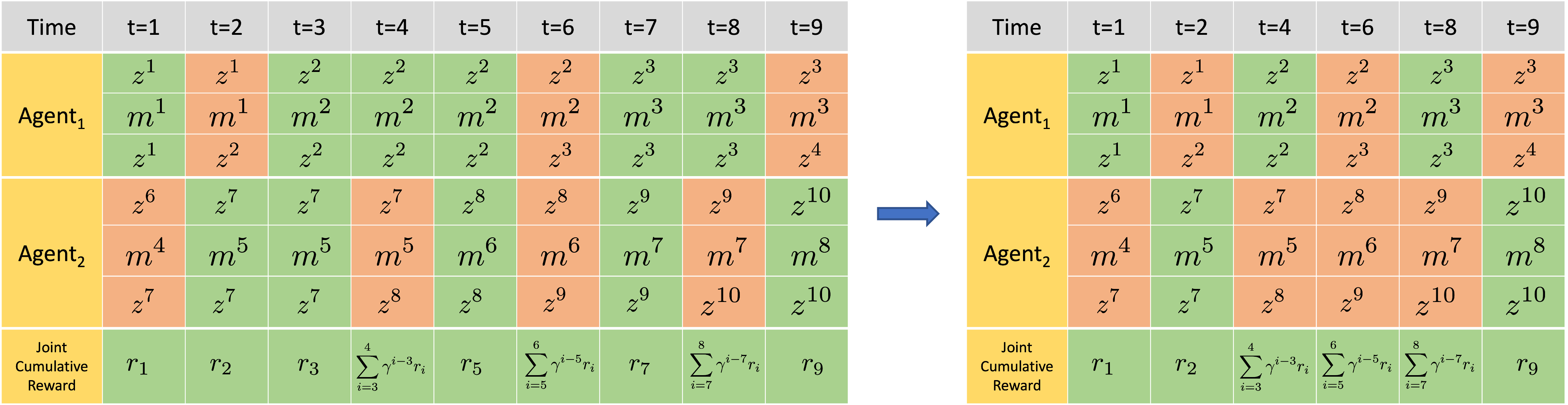}
    \caption{An example of the trajectory squeezing process in Mac-CAC. Joint sequential experiences are squeezed by picking out joint transition tuples when the joint macro-action terminates, in that, $any$ agent's macro-action termination (marked in red) ends the joint macro-action at the timestep. For example, at $t=1$, agents execute a joint macro-action $\vec{m}=\langle m^1, m^4\rangle$ for one timestep; at $t=2$, the joint macro-action becomes $\langle m^1,m^5\rangle$ as $\text{Agent}_2$ finished $m^4$ at last step and chooses a new macro-action $m^5$; $\text{Agent}_1$ finished its macro-action $m_1$ at $t=2$ and selects a new macro-action $m^2$ at $t=3$ so that the joint macro-action switches to $\langle m^2, m^5 \rangle$ which keeps running until the 4th timestep. Therefore, the first two joint macro-actions have two single-step rewards respectively, and the reward of joint macro-action $\langle m^2, m^5 \rangle$ is an accumulative reward over two consecutive timesteps.} 
    \label{Mac-CAC-Traj}
\end{figure}

\noindent In the fully centralized learning case, we treat all agents as a single joint agent to learn a centralized actor $\Psi_\theta(\vec{m}\mid\vec{h})$ with a centralized critic $Q^{\Psi_\theta}_\phi(\vec{h},\vec{m})$, and the policy gradient can be expressed as:
\begin{equation}
        \nabla_{\theta} J (\theta)=\mathbb{E}_{\Psi_\theta}\biggr[\nabla_{\theta}\log\Psi_{\theta}(\vec{m}\mid \vec{h})Q^{\Psi_{\theta}}_{\phi}(\vec{h},\vec{m})\biggr]
        \label{Mac-CAC-Q}
\end{equation}

Similarly, in order to achieve low variance optimization for the actor, we learn a centralized history value function $V^{\Psi_\theta}_{\mathbf{w}}(\vec{h})$ by minimizing a TD-error loss over joint trajectories that are squeezed w.r.t. each joint macro-action termination (as long as one of the agents terminates its macro-action, defined in Section~\ref{chap:paper1:app:MacCenQ}). 
One example of the squeezing process for a joint trajectory is shown in Fig.~\ref{Mac-CAC-Traj}.
Accordingly, the policy's updates are performed when each joint macro-action is completed by ascending the following gradient:

\begin{equation}
        \nabla_{\theta} J (\theta)=\mathbb{E}_{\Psi_\theta}\biggr[\nabla_{\theta}\log\Psi_{\theta}(\vec{m}\mid \vec{h})A(\vec{h},\vec{m})\biggr]
        \label{Mac-IAC-V}
\end{equation}
\begin{equation}
        A(\vec{h},\vec{m}) = \vec{r\,}^c + \gamma^{\vec{\tau}_{\vec{m}}}V^{\Psi_\theta}_{\mathbf{w}}(\vec{h}')-V^{\Psi_\theta}_{\mathbf{w}}(\vec{h})
\end{equation}
where the cumulative reward $\vec{r\,}^c$ is w.r.t.~the execution of the joint macro-action $\vec{m}$. The pseudocode for Mac-CAC is presented below: 

\begin{algorithm}[t!]
    \footnotesize
    \caption{Mac-CAC}
    \label{alg2}
        \begin{algorithmic}[1]
            \State Initialize a centralized policy network: $\Psi_{\theta}$
            \State Initialize centralized critic networks: $V_{\mathbf{w}}^{\Psi_{\theta}}$, $V_{\mathbf{w}^-}^{\Psi_{\theta}}$ 
            \State Initialize a centralized buffer $\mathcal{D}\leftarrow\text{Mac-JERTs}$,
            \For{\emph{episode} = $1$ to $M$}
                \State $t=0$
                \State Reset env
                \While{not reaching a terminal state \textbf{and} $t < \mathbb{H}$}
                    \State $t \leftarrow t + 1$
                    \If{the joint macro-action $\vec{m}$ is terminated}
                        \State $\vec{m} \sim \Psi_{\theta}(\cdot \mid \vec{h}, \vec{m}^{\text{undone}}; \epsilon)$
                    \Else
                        \State Continue running current joint macro-action $\vec{m}$
                    \EndIf
                    \State Get a joint cumulative reward $\vec{r\,}^c$, next joint macro-observation $\vec{z\,}'$
                    \State Collect $\langle \vec{z},\vec{m}, \vec{z\,}', \vec{r\,}^c \rangle$ into the buffer $\mathcal{D}$
                \EndWhile
                \If{\emph{episode} mod $I_{\text{train}} = 0$}
                    \State Squeeze joint macro-level trajectories in the buffer $\mathcal{D}$ according to joint macro-action terminations
                    \State Perform a gradient descent step on $L(\mathbf{w})=\big(y-V^{\Psi_{\theta}}_{\mathbf{w}}(\vec{h})\big)^2_\mathcal{D}$, where  $y = \vec{r\,}^c + \gamma^{\vec{\tau}_{\vec{m}}} V^{\Psi_{\theta}}_{\mathbf{w}^-}(\vec{h}')$ 
                    \State Perform a gradient ascent on $\nabla_{\theta} J(\theta) = \mathbb{E}_{\Psi_\theta}\Big[\nabla_{\theta}\log\Psi_{\theta}(\vec{m}\mid \vec{h})\big(\vec{r\,}^c + \gamma^{\vec{\tau}_{\vec{m}}} V^{\Psi_{\theta}}_{\mathbf{w}^-}(\vec{h}')-V^{\Psi_{\theta}}_{\mathbf{w}}(\vec{h})\big)\Big]$
                    \State Reset buffer $\mathcal{D}$
                \EndIf
                \If{\emph{episode} mod $I_{\text{TargetUpdate}} = 0$}
                    \State Update the critic target network $\mathbf{w}^-\leftarrow\mathbf{w}$
                \EndIf
            \EndFor
        \end{algorithmic}
\end{algorithm}

\subsection{Macro-Action-Based Independent Actor with Centralized Critic}
\label{chap:paper3:app:mac-iacc}

\noindent As mentioned earlier, fully centralized learning requires perfect online communication that is often hard to guarantee, and fully decentralized learning suffers from environmental non-stationarity due to agents' changing policies. 
In order to learn better decentralized macro-action-based policies, in this section, we propose two macro-action-based actor-critic algorithms using the CTDE paradigm. 
Typically, the difference between a joint macro-action termination from the centralized perspective and a macro-action termination from each agent's local perspective gives rise to a new challenge: \emph{what kind of centralized critic to learn and how to use it to optimize decentralized policies under such an asymmetric asynchrony from the two perspectives}, which we mainly investigate below.

\begin{figure}[t!]
    \centering
    \includegraphics[height=5.8cm]{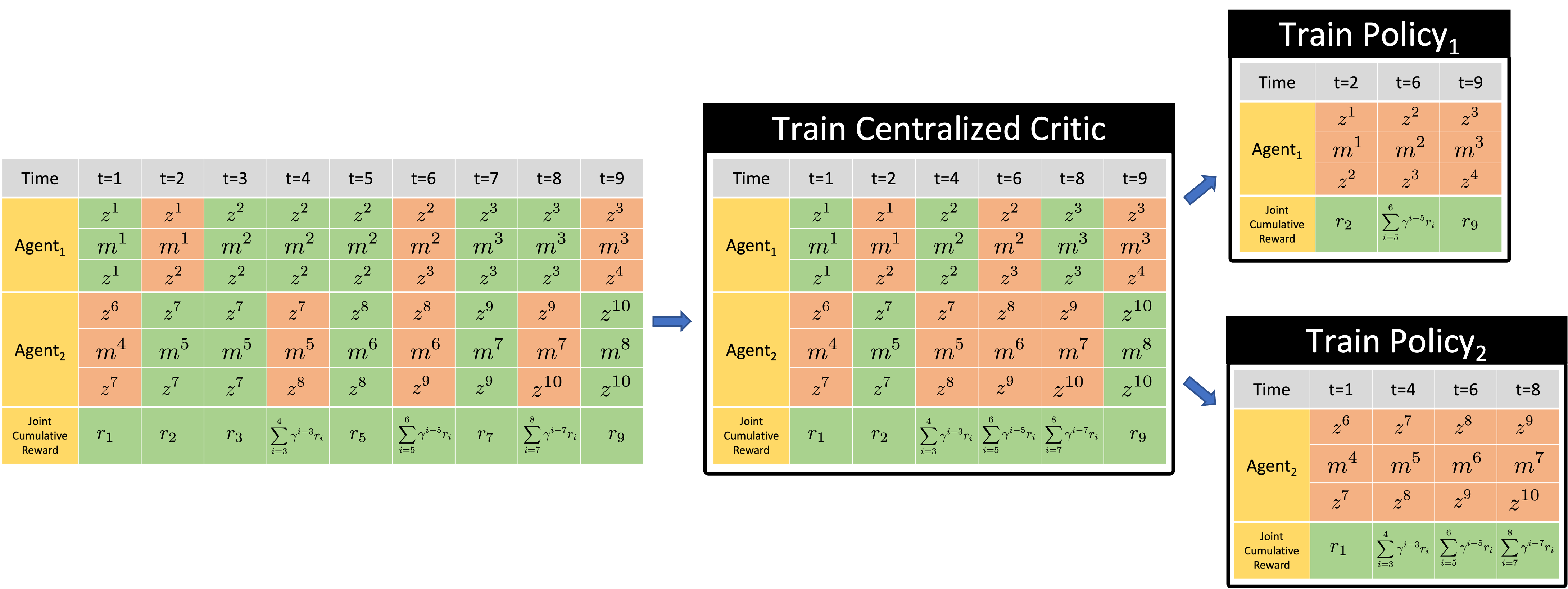}
    \caption{An example of the trajectory squeezing process in Navie Mac-IACC.The joint trajectory is first squeezed depending on joint macro-action termination for training the centralized critic (line 18-19 in Algorithm~\ref{Mac-NIACC-Code}). Then, the trajectory is further squeezed for each agent depending on each agent's own macro-action termination for training the decentralized policy (line 20-23 in Algorithm~\ref{Mac-NIACC-Code}.}
    \label{Mac-NIACC-Traj}
\end{figure}

\textbf{Naive Mac-IACC.}
\label{N-Mac-IACC}
A naive way of incorporating macro-actions into a CTDE-based actor-critic framework is to directly adapt the idea of the primitive-action-based IACC (Section~\ref{chap:BG:CTDE}) to have a shared joint macro-action-value function $Q^{\vec{\Psi}_{\vec{\theta}}}_\phi(\mathbf{x},\vec{m})$ in each agent's decentralized macro-action-based policy gradient as:
\begin{equation}
        \nabla_{\theta_i} J (\theta_i)=\mathbb{E}_{\vec{\Psi}_{\vec{\theta}}}\biggr[\nabla_{\theta_i}\log\Psi_{\theta_i}(m_i\mid h_i)Q^{\vec{\Psi}_{\vec{\theta}}}_\phi(\mathbf{x},\vec{m})\biggr]
        \label{N-Mac-IACC-Q}
\end{equation}
To reduce variance, with a value function $V^{\vec{\Psi}_{\vec{\theta}}}_\mathbf{w}(\mathbf{x})$ as the centralized critic, the policy gradient w.r.t.~the parameters of each agent's high-level policy can be rewritten as: 
\begin{equation}
        \nabla_{\theta_i}J(\theta_i) = \mathbb{E}_{\vec{\Psi}_{\vec{\theta}}}\biggr[\nabla_{\theta_i}\log\Psi_{\theta_i}(m_i \mid h_i)A(\mathbf{x}, \vec{m})\biggr]
\end{equation}
\begin{equation}
        A(\mathbf{x}, \vec{m}) = \vec{r}^{\,c} + \gamma^{\vec{\tau}_{\vec{m}}}V_{\mathbf{w}}^{\vec{\Psi}_{\vec{\theta}}}(\mathbf{x}') - V_{\mathbf{w}}^{\vec{\Psi}_{\vec{\theta}}}(\mathbf{x})
            \label{N-Mac-IACC-V}
\end{equation}
Here, the critic 
is trained in the fully centralized manner described in Section~\ref{chap:paper3:app:mac-cac} while allowing it to access additional global information (e.g., joint macro-observation-action history, ground truth state, or both) represented by the symbol $\mathbf{x}$. 
However, updates of each agent's policy $\Psi_{\theta_i}(m_i\mid h_i)$ only occur at the agent's own macro-action termination time steps rather than depending on joint macro-action terminations in the centralized critic training. One example of how the trajectories are manipulated for training and the corresponding pseudocode are summarized in Fig.~\ref{Mac-NIACC-Traj} and Algorithm~\ref{Mac-NIACC-Code} respectively.

\begin{algorithm}[h!]
    \footnotesize
    \caption{Naive Mac-IACC}
        \begin{algorithmic}[1]
            \State Initialize a decentralized policy network for each agent $i$: $\Psi_{\theta_i}$
            \State Initialize centralized critic networks: $V_{\mathbf{w}}^{\vec{\Psi}_{\vec{\theta}}}$, $V_{\mathbf{w}^-}^{\vec{\Psi}_{\vec{\theta}}}$ 
            \State Initialize a decentralized buffer $\mathcal{D}\leftarrow\text{Mac-JERTs}$,
            \For{\emph{episode} = $1$ to $M$}
                \State $t=0$
                \State Reset env
                \While{not reaching a terminal state \textbf{and} $t < \mathbb{H}$}
                    \State $t \leftarrow t + 1$
                    \For{each agent $i$}
                        \If{the macro-action $m_i$ is terminated}
                            \State $m_{i} \sim \Psi_{\theta_i}(\cdot \mid h_i; \epsilon)$
                        \Else
                            \State Continue running current macro-action $m_i$
                        \EndIf
                    \EndFor
                    \State Get a reward $\vec{r\,}^c$ accumulated based on current joint macro-action termination
                    \State Get next joint macro-observations $\vec{z\,}'$
                    \State Collect $\langle \vec{z},\vec{m}, \vec{z\,}', \vec{r\,}^c \rangle$ into the buffer $\mathcal{D}$
                \EndWhile
                \If{\emph{episode} mod $I_{\text{train}} = 0$}
                    \State Squeeze joint macro-level trajectories in the buffer $\mathcal{D}$ according to joint macro-action terminations
                    \State Perform a gradient descent step on $L(\mathbf{w})=\big(y-V^{\vec{\Psi}_{\vec{\theta}}}_{\mathbf{w}}(\vec{h})\big)^2_\mathcal{D}$, where  $y = \vec{r\,}^c + \gamma^{\vec{\tau}_{\vec{m}}} V^{\vec{\Psi}_{\vec{\theta}}}_{\mathbf{w}^-}(\vec{h}')$ 
                    \For{each agent $i$}
                        \State Squeeze agent $i$'s trajectories in the buffer $\mathcal{D}$ according to its own macro-action terminations
                        \State Perform a gradient ascent on: 
                        \State $\nabla_{\theta_i} J(\theta_i) = \mathbb{E}_{\vec{\Psi}_{\vec{\theta}}}\Big[\nabla_{\theta_i}\log\Psi_{\theta_i}(m_i|h_i)\big(\vec{r\,}^c + \gamma^{\vec{\tau}_{\vec{m}}} V^{\vec{\Psi}_{\vec{\theta}}}_{\mathbf{w}^-}(\vec{h}')-V^{\vec{\Psi}_{\vec{\theta}}}_{\mathbf{w}}(\vec{h})\big)\Big]$
                    \EndFor
                    \State Reset buffer $\mathcal{D}$
                \EndIf
                \If{\emph{episode} mod $I_{\text{TargetUpdate}} = 0$}
                    \State Update the critic target network $\mathbf{w}^-\leftarrow\mathbf{w}$
                \EndIf
            \EndFor
        \end{algorithmic}
        \label{Mac-NIACC-Code}
\end{algorithm}

\textbf{Independent Actor with Individual Centralized Critic (Mac-IAICC).}
Note that naive Mac-IACC is technically incorrect. 
The cumulative reward $\vec{r\,}^c $ in Eq~\ref{N-Mac-IACC-V} is based on the corresponding joint macro-action's termination that is defined as when \emph{any} agent finishes its own macro-action, which produces two potential issues: a) $\vec{r}^{\,c} +\gamma^{\vec{\tau}_{\vec{m}}} V_{\mathbf{w}}^{\vec{\Psi}_{\vec{\theta}}}(\mathbf{x}')$ may not estimate the value of the macro-action $m_i$ well as the reward does not depend on $m_i$'s termination; b) from agent $i$'s perspective, its policy gradient estimation may involve higher variance associated with the asynchronous macro-action terminations of other agents.  

\begin{figure}[t!]
    \centering
    \includegraphics[height=7.4cm]{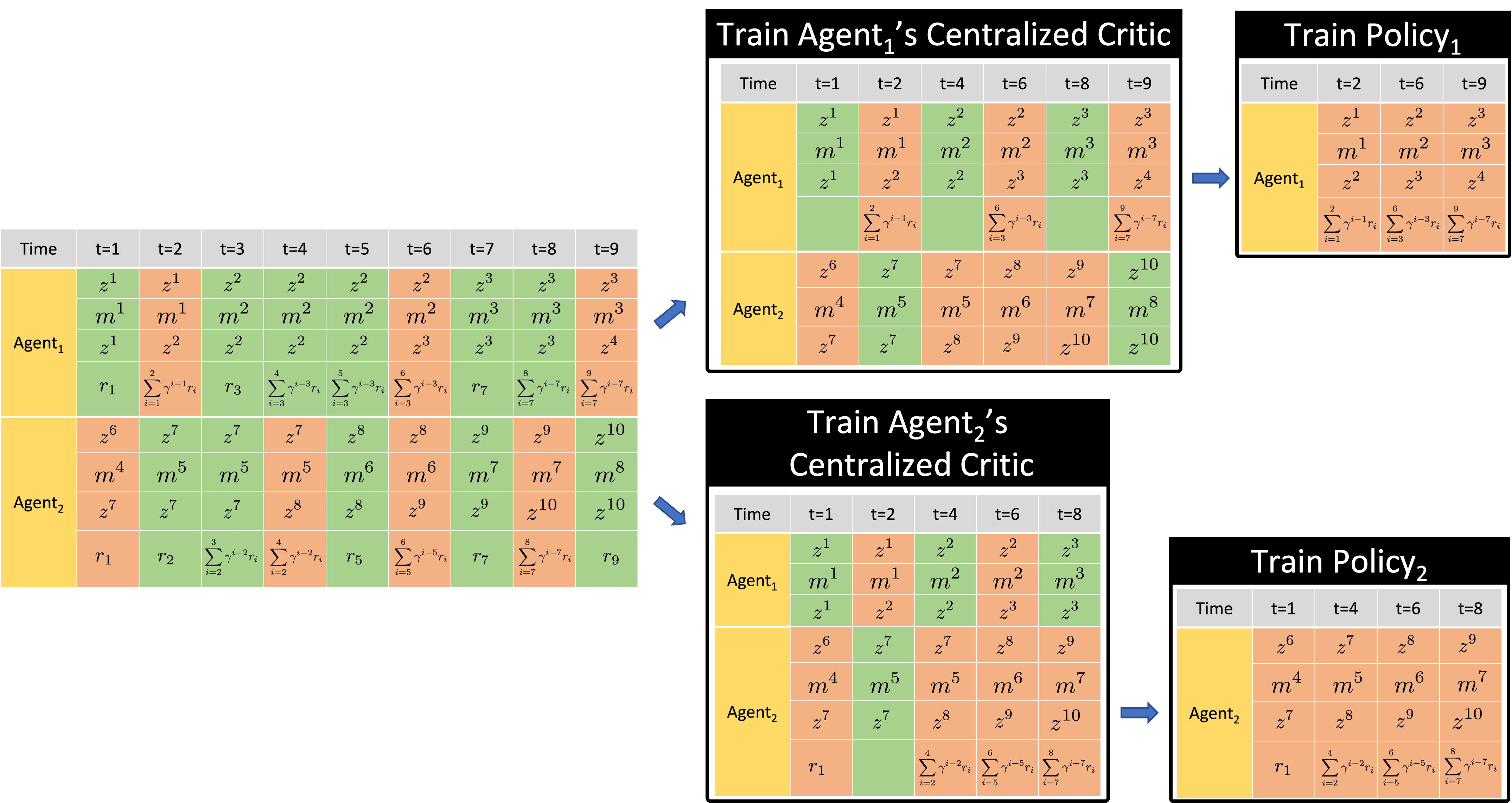}
    \caption{An example of the trajectory squeezing process in Mac-IAICC: each agent learns an individual centralized critic for the decentralized policy optimization. In order to achieve better use of centralized information, the recurrent layer in each critic's neural network should receive all the valid joint macro-observation-action information (when $any$ agent terminates its macro-action (lines 20-22) and obtains a new joint macro-observation). However, the critic's TD updates and the policy's updates still rely on each agent's individual macro-action termination and the accumulative reward at the corresponding timestep (lines 23-26). Hence, the trajectory squeezing process for training each critic still depends on joint-macro-action termination but only retaining the accumulative rewards w.r.t. the corresponding agent's macro-action termination for computing the TD loss (the middle part in the above picture). Then, each agent's trajectory is further squeezed depending on its macro-action termination to update the decentralized policy.}
    \label{Mac-IAICC-Traj}
\end{figure}

To tackle aforementioned issues, we propose to learn a separate centralized critic $V_{\mathbf{w}_i}^{\vec{\Psi}_{\vec{\theta}}}(\mathbf{x}')$ for each agent via TD-learning. In this case, each TD-error for updating $V_{\mathbf{w}_i}^{\vec{\Psi}_{\vec{\theta}}}(\mathbf{x}')$ is computed by using the reward $r^c_i$ that is accumulated purely based on the execution of the agent $i$'s macro-action $m_i$. With this TD-error estimation, each agent's decentralized macro-action-based policy gradient becomes: 

\begin{algorithm}[t!]
    \footnotesize
    \caption{Mac-IAICC}
        \begin{algorithmic}[1]
            \State Initialize a decentralized policy network for each agent $i$: $\Psi_{\theta_i}$
            \State Initialize centralized critic networks for each agent $i$: $V_{\mathbf{w}_i}^{\vec{\Psi}_{\vec{\theta}}}$, $V_{\mathbf{w}^-_i}^{\vec{\Psi}_{\vec{\theta}}}$ 
            \State Initialize a decentralized buffer $\mathcal{D}$
            \For{\emph{episode} = $1$ to $M$}
                \State $t=0$
                \State Reset env
                \While{not reaching a terminal state \textbf{and} $t < \mathbb{H}$}
                    \State $t \leftarrow t + 1$
                    \For{each agent $i$}
                        \If{the macro-action $m_i$ is terminated}
                            \State $m_{i} \sim \Psi_{\theta_i}(\cdot \mid h_i; \epsilon)$
                        \Else
                            \State Continue running current macro-action $m_i$
                        \EndIf
                    \EndFor
                    \For{each agent $i$}
                        \State Get a reward $r^c_i$ accumulated based on agent $i$'s macro-action termination
                    \EndFor
                    \State Get next joint macro-observations $\vec{z\,}'$
                    \State Collect $\langle \vec{z},\vec{m}, \vec{z\,}', \{r^c_1,\dots, r^c_n \} \rangle$ into the buffer $\mathcal{D}$
                \EndWhile
                \If{\emph{episode} mod $I_{\text{train}} = 0$}
                    \For{each agent $i$}
                        \State Squeeze trajectories in the buffer $\mathcal{D}$ according to joint macro-action terminations
                        \State Compute the TD-error of each timestep in the squeezed experiences:
                        \State $L(\mathbf{w}_i)=\big(y-V^{\vec{\Psi}_{\vec{\theta}}}_{\mathbf{w}_i}(\vec{h})\big)^2_\mathcal{D}$, where  $y = r^c_i + \gamma^{\tau_{m_i}} V^{\vec{\Psi}_{\vec{\theta}}}_{\mathbf{w}^-_i}(\vec{h}')$ 
                        \State Perform a gradient descent only over the TD-errors when agent $i$'s macro-action is terminated
                        \State Squeeze agent $i$'s trajectories in the buffer $\mathcal{D}$ according to its own macro-action terminations
                        \State Perform a gradient ascent on: 
                        \State $\nabla_{\theta_i} J(\theta_i) = \mathbb{E}_{\vec{\Psi}_{\vec{\theta}}}\Big[\nabla_{\theta_i}\log\Psi_{\theta_i}(m_i|h_i)\big(r^c_i + \gamma^{\tau_{m_i}} V^{\vec{\Psi}_{\vec{\theta}}}_{\mathbf{w}^-_i}(\vec{h}')-V^{\vec{\Psi}_{\vec{\theta}}}_{\mathbf{w}_i}(\vec{h})\big)\Big]$
                    \EndFor
                    \State Reset buffer $\mathcal{D}$
                \EndIf
                \If{\emph{episode} mod $I_{\text{TargetUpdate}} = 0$}
                    \For{each agent $i$}
                        \State Update the critic target network $\mathbf{w}_i^-\leftarrow\mathbf{w}_i$
                    \EndFor
                \EndIf
            \EndFor
        \end{algorithmic}
        \label{Mac-IAICC-Code}
\end{algorithm}

\begin{equation}
        \nabla_{\theta_i}J(\theta_i) = \mathbb{E}_{\vec{\Psi}_{\vec{\theta}}}\biggr[\nabla_{\theta_i}\log\Psi_{\theta_i}(m_i \mid h_i)A(\mathbf{x}, m_i)\biggr]
            \label{Mac-IAICC}
\end{equation}
\begin{equation}
        A(\mathbf{x}, m_i) = r^{c}_i + \gamma^{\tau_{m_i}}V_{\mathbf{w}_i}^{\vec{\Psi}_{\vec{\theta}}}(\mathbf{x}') - V_{\mathbf{w}_i}^{\vec{\Psi}_{\vec{\theta}}}(\mathbf{x})
\end{equation}

Now, from agent $i$'s perspective, $r^{c}_i + \gamma^{\tau_{m_i}}V_{\mathbf{w}_i}^{\vec{\Psi}_{\vec{\theta}}}(\mathbf{x}')$ is capable of offering a more accurate value prediction for the macro-action $m_i$, since both the reward, $r^{c}_i$ and the value function $V_{\mathbf{w}_i}^{\vec{\Psi}_{\vec{\theta}}}(\mathbf{x}') $ depend on agent $i$'s macro-action termination. Also, unlike the case in Naive Mac-IACC, other agents' terminations cannot lead to extra noisy estimated rewards w.r.t. $m_i$ anymore so that the variance on policy gradient estimation gets reduced. 
Then, updates for both the critic and the actor occur when the corresponding agent's macro-action ends and take the advantage of information sharing. 
The detailed trajectory squeezing process for Mac-IAICC is visualized in Fig.~\ref{Mac-IAICC-Traj}, and the pseudocode is displayed in Algorithm~\ref{Mac-IAICC-Code}.

\begin{figure}[t!]
    \centering
    \captionsetup[subfigure]{labelformat=empty}
    \centering
    \subcaptionbox{(a) Box Pushing\vspace{2mm}\label{domain_BP}}
        [0.3\linewidth]{\includegraphics[height=3.8cm]{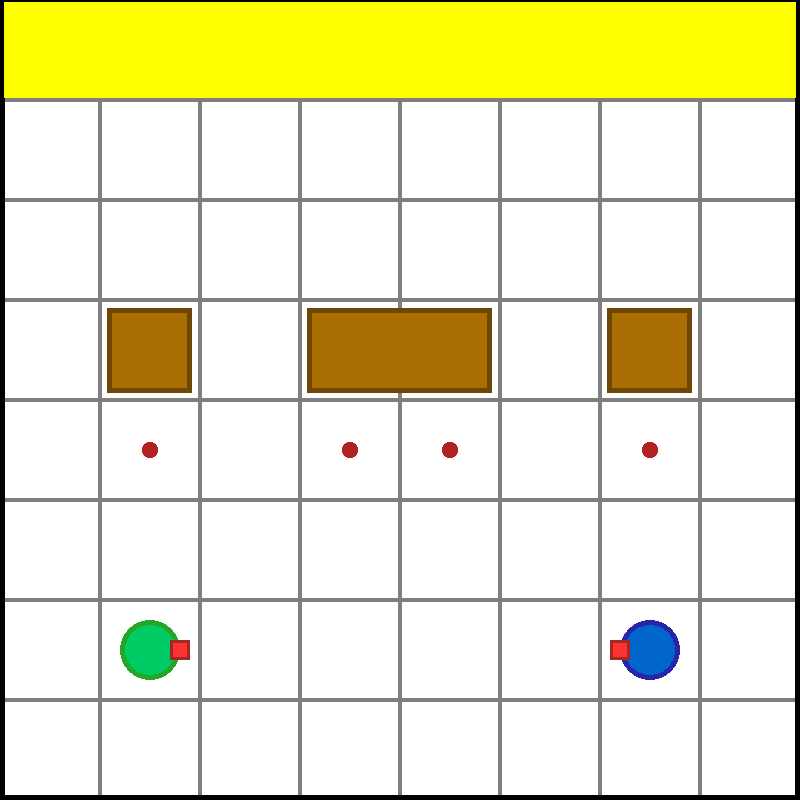}}
    ~
    \centering
    \subcaptionbox{(b) Overcooked-A\vspace{2mm}\label{domain_OA}}
        [0.3\linewidth]{\includegraphics[height=3.8cm]{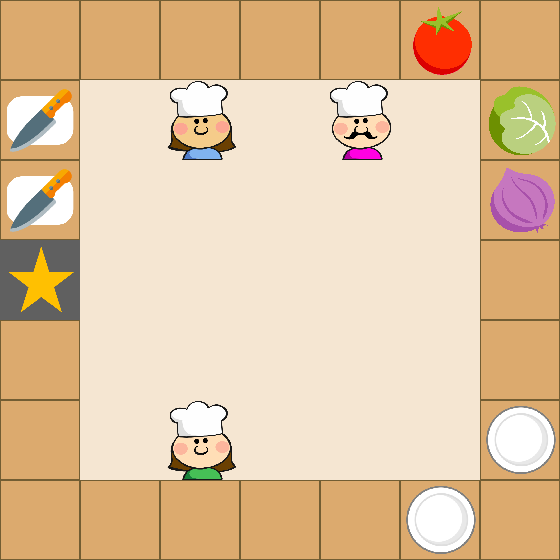}}
    ~
    \centering
    \subcaptionbox{(c) Overcooked-B\vspace{2mm}\label{domain_OB}}
        [0.3\linewidth]{\includegraphics[height=3.8cm]{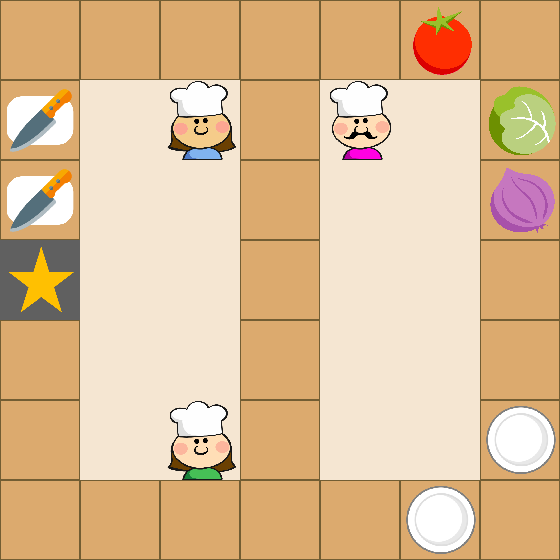}}
    ~
    \centering
    \subcaptionbox{(d) Warehouse-A\vspace{2mm}\label{domain_wtdA}}
        [0.4\linewidth]{\includegraphics[height=3.8cm]{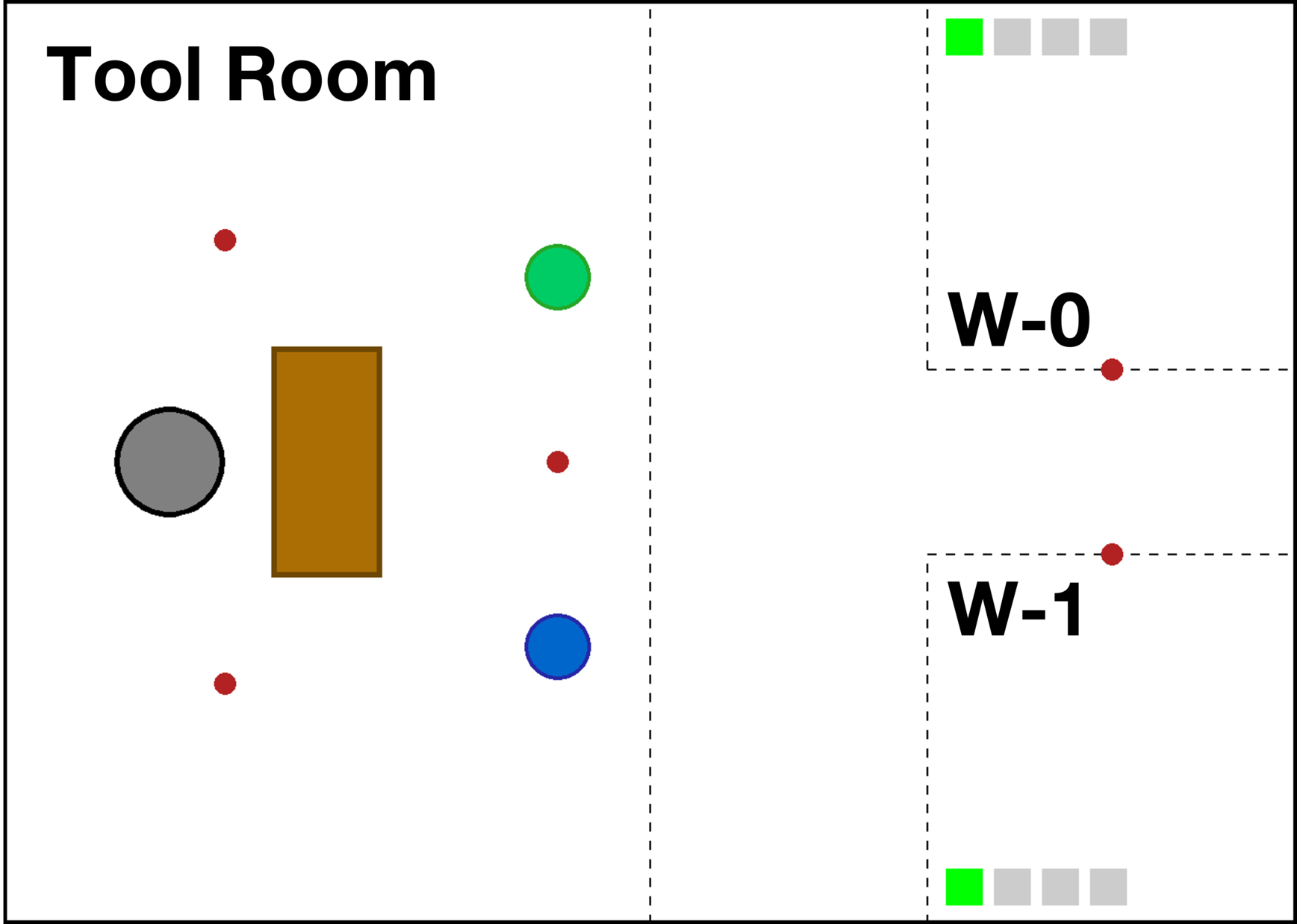}}
    ~
    \centering
    \subcaptionbox{(e) Warehouse-B\vspace{2mm}\label{domain_wtdB}}
        [0.4\linewidth]{\includegraphics[height=3.8cm]{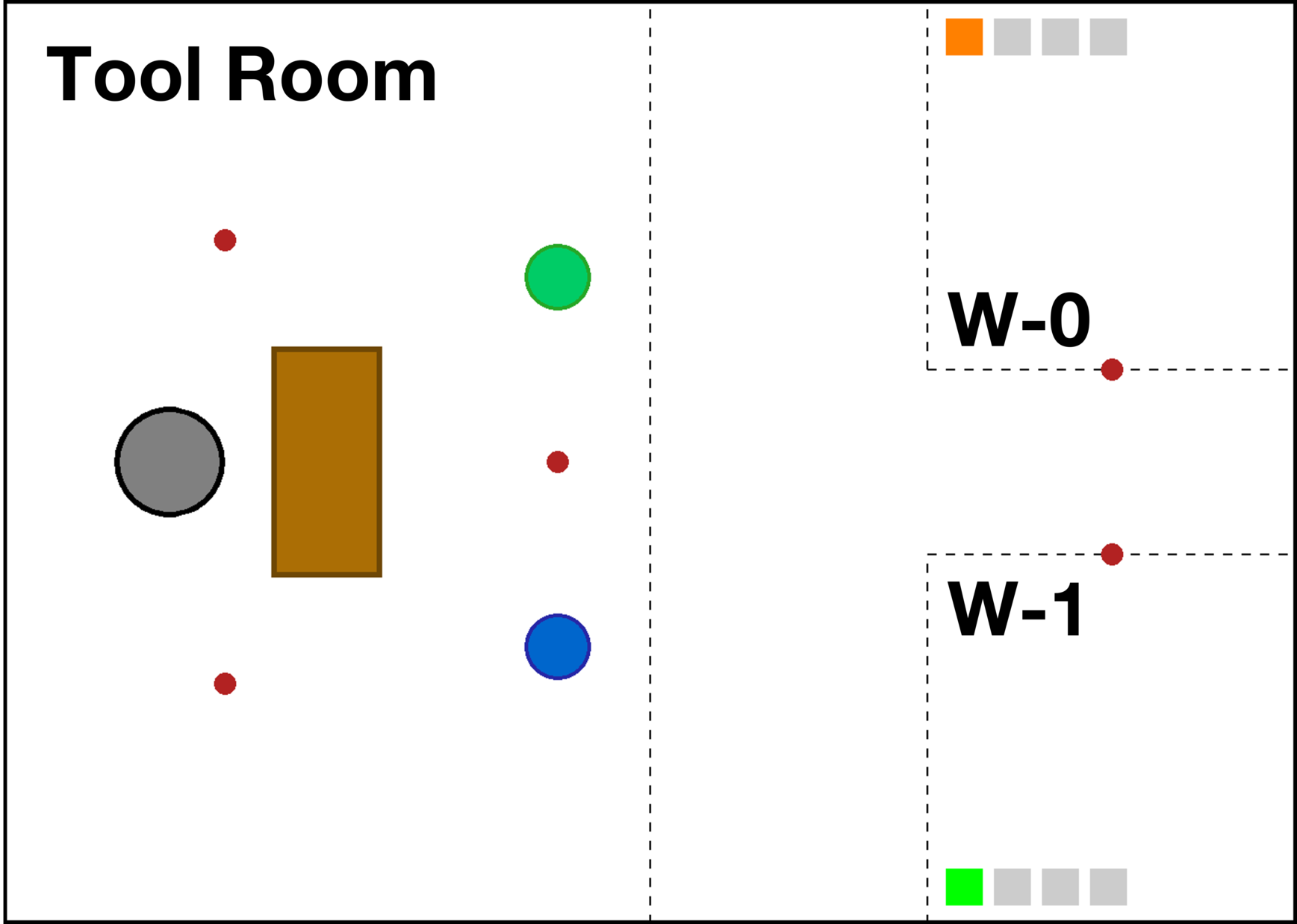}}
     ~
    \centering
    \subcaptionbox{(f) Warehouse-C\vspace{2mm}\label{domain_wtdC}}
        [0.48\linewidth]{\includegraphics[height=3.8cm]{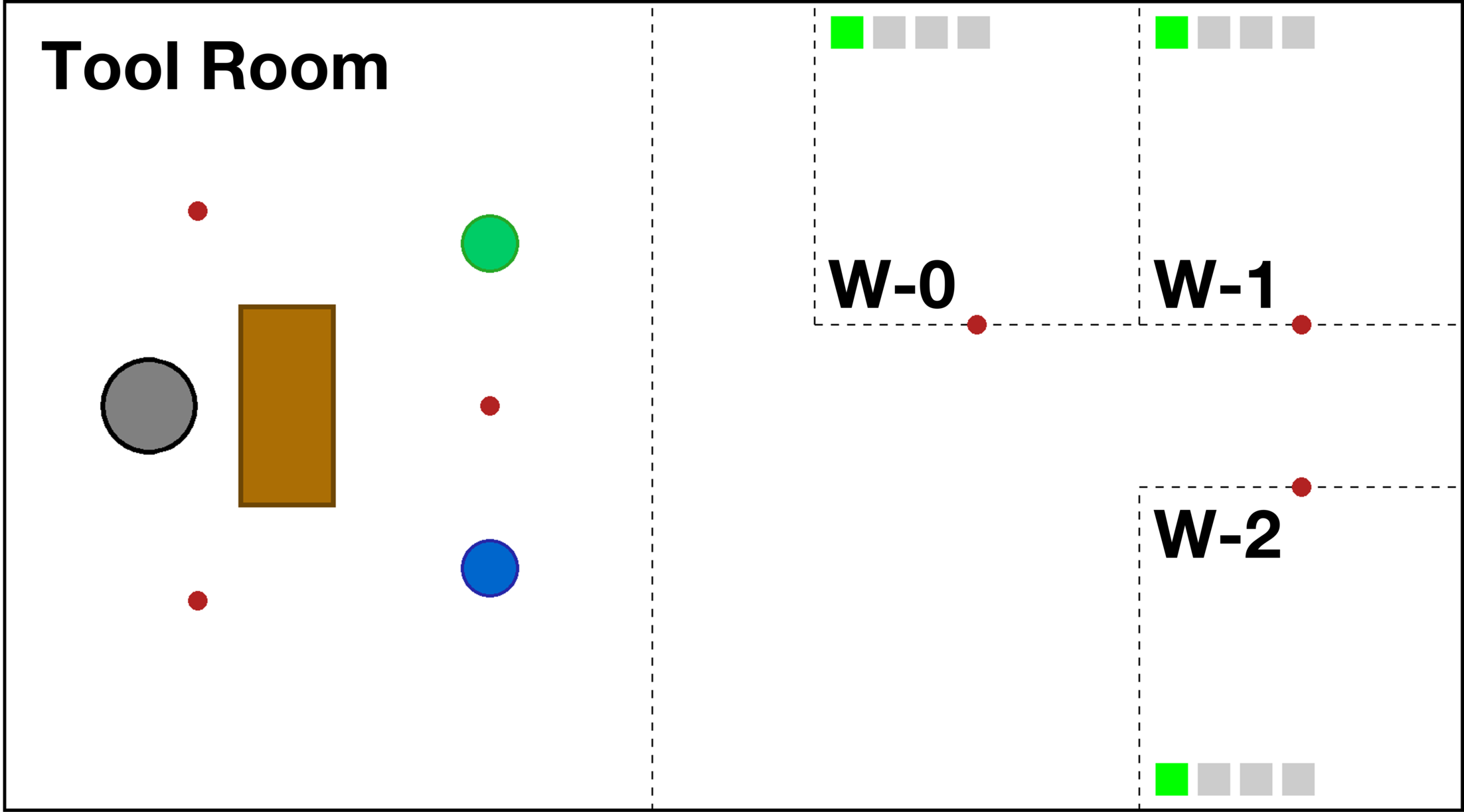}}
     ~
    \centering
    \subcaptionbox{(g) Warehouse-D\vspace{2mm}\label{domain_wtdD}}
        [0.48\linewidth]{\includegraphics[height=3.8cm]{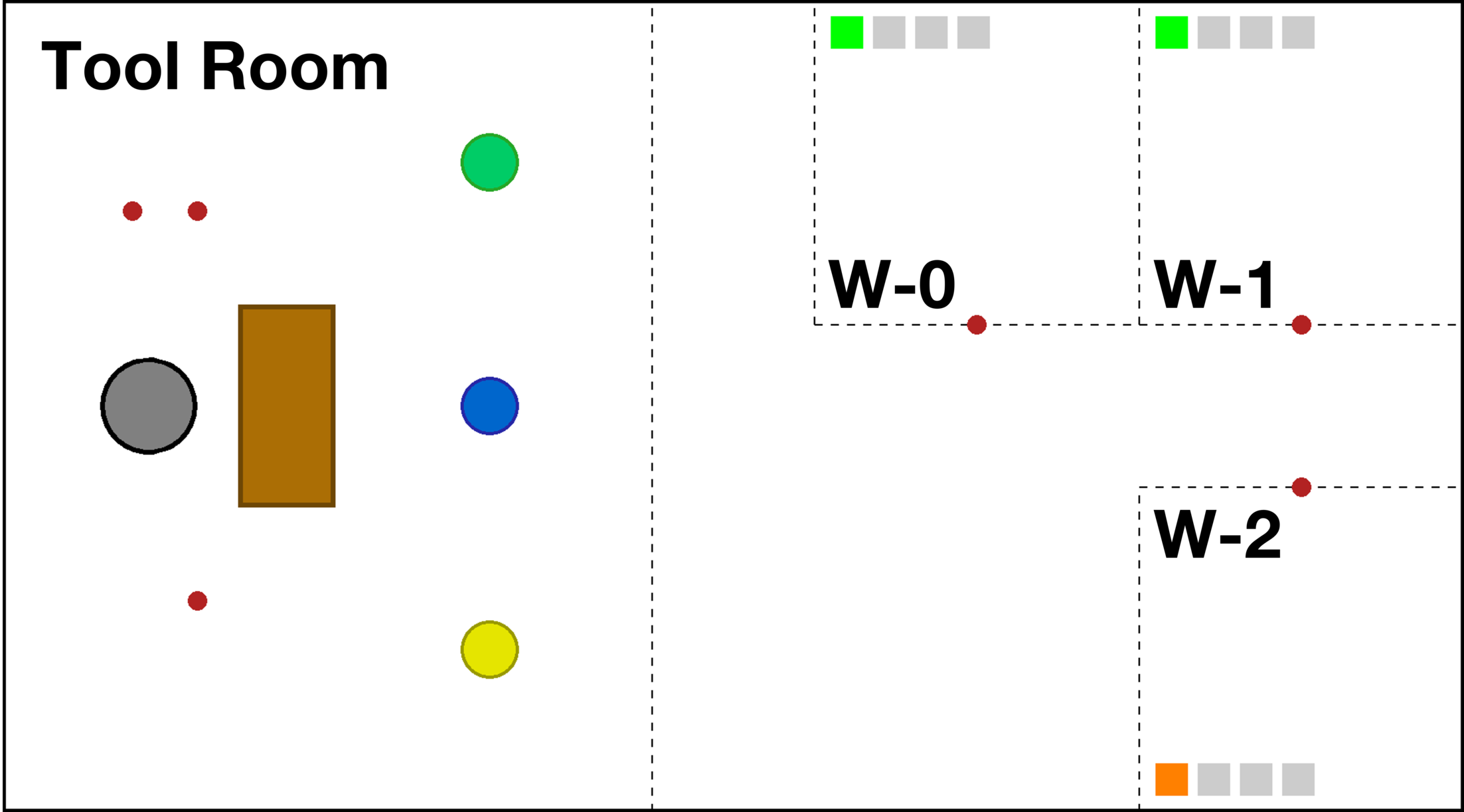}}
    ~
    \centering
    \subcaptionbox{(h) Warehouse-E\label{domain_wtdE}}
        [0.45\linewidth]{\includegraphics[height=3.8cm]{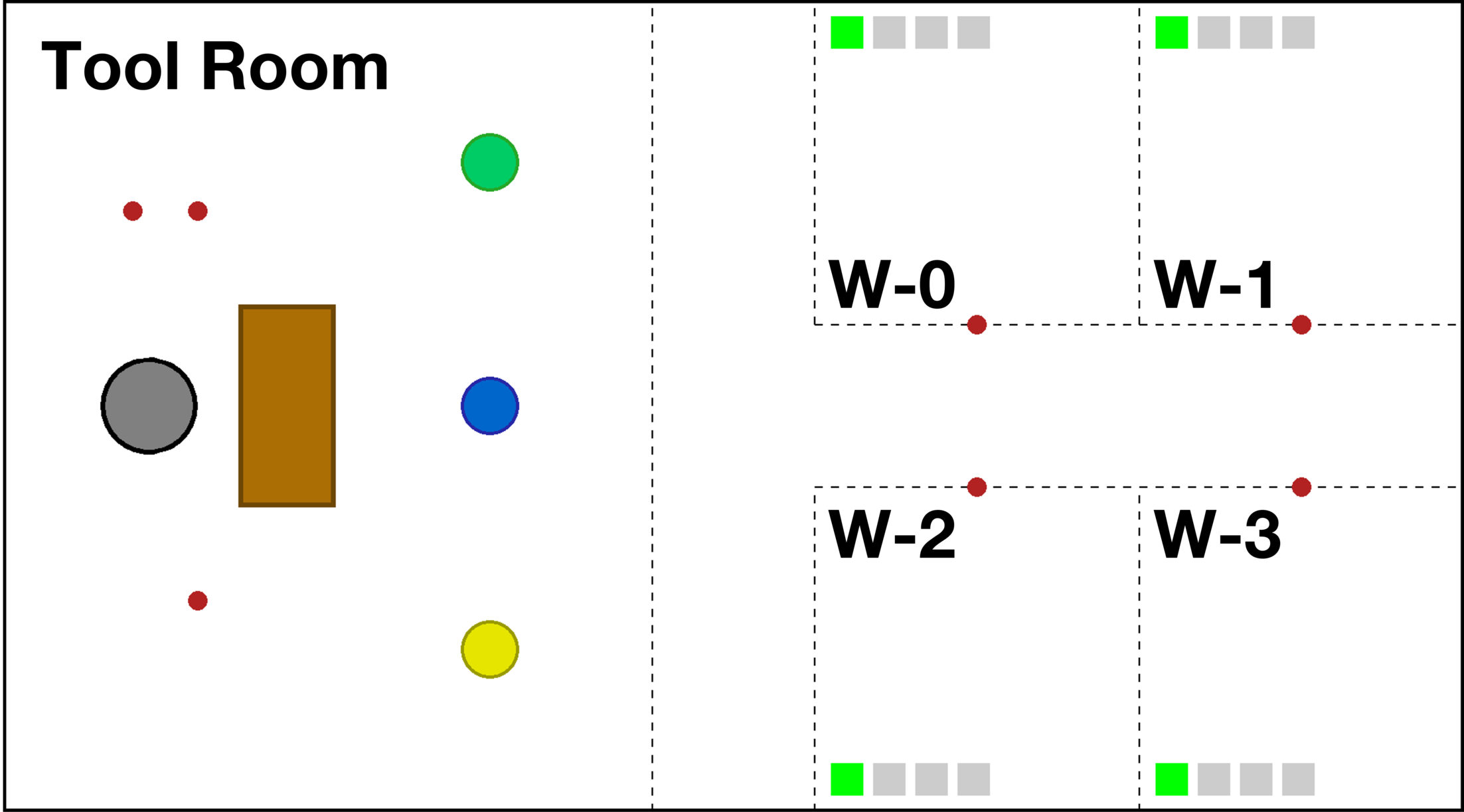}}
    \caption{Experimental environments.}
    \label{mac_pg_domains}
\end{figure}

\section{Simulated Experiments}
\label{chap:paper3:sim}

\subsection{Experimental Setup}
\label{chap:paper3:sim:setup}

\noindent We investigate the performance of our algorithms over a variety of multi-agent problems with macro-actions (Fig.~\ref{mac_pg_domains}): Box Pushing, Overcooked~\cite{wu_wang2021too},
and a larger Warehouse Tool Delivery domain. 
Macro-actions are defined by using prior domain knowledge as they are straightforward in these tasks. 
Typically, we also include primitive-actions in the macro-action set (as one-step macro-actions), which gives agents the chance to learn more complex policies that use both when it is necessary.

\textbf{Box Pushing} (Fig.~\ref{domain_BP}). Two robots are in an environment with two small boxes and one large box. The optimal solution is to cooperatively push the big box to the yellow goal area, but partial observability makes this difficult. Specifically, robots have four primitive-actions: \emph{move forward}, \emph{turn-left}, \emph{turn-right} and \emph{stay}. In the macro-action case, each robot has three one-step macro-actions: \emph{\textbf{Turn-left}}, \emph{\textbf{Turn-right}}, and \emph{\textbf{Stay}}, as well as three multi-step macro-actions: \emph{\textbf{Move-to-small-box(i)}} and
\emph{\textbf{Move-to-big-box(i)}} which navigate the robot to the red spot below the corresponding box and terminate with the robot facing the box; \emph{\textbf{Push}} causes the robot to keep moving forward until arriving at the world's boundary (potentially pushing the small box or trying to push the big one). The big box only moves if both agents push it at the same time. 
Each robot can only capture the status (\emph{empty}, \emph{teammate}, \emph{boundary}, \emph{small or big box}) of the cell in front of it as one observation or macro-observation. 
When any box is pushed to the goal, the team receives a terminal reward ($+300$ for the big box and $+20$ for each small  box). A penalty $-10$ is issued when any robot hits the boundary or pushes the big box on its own. Each episode terminates when any box is pushed to the goal area, or when reaching the task horizon (100 time steps for the grid world smaller than 10$\times$10, otherwise 200 time steps). 

\begin{wrapfigure}{t!}{.5\textwidth}
    \centering
    \includegraphics[height=3cm]{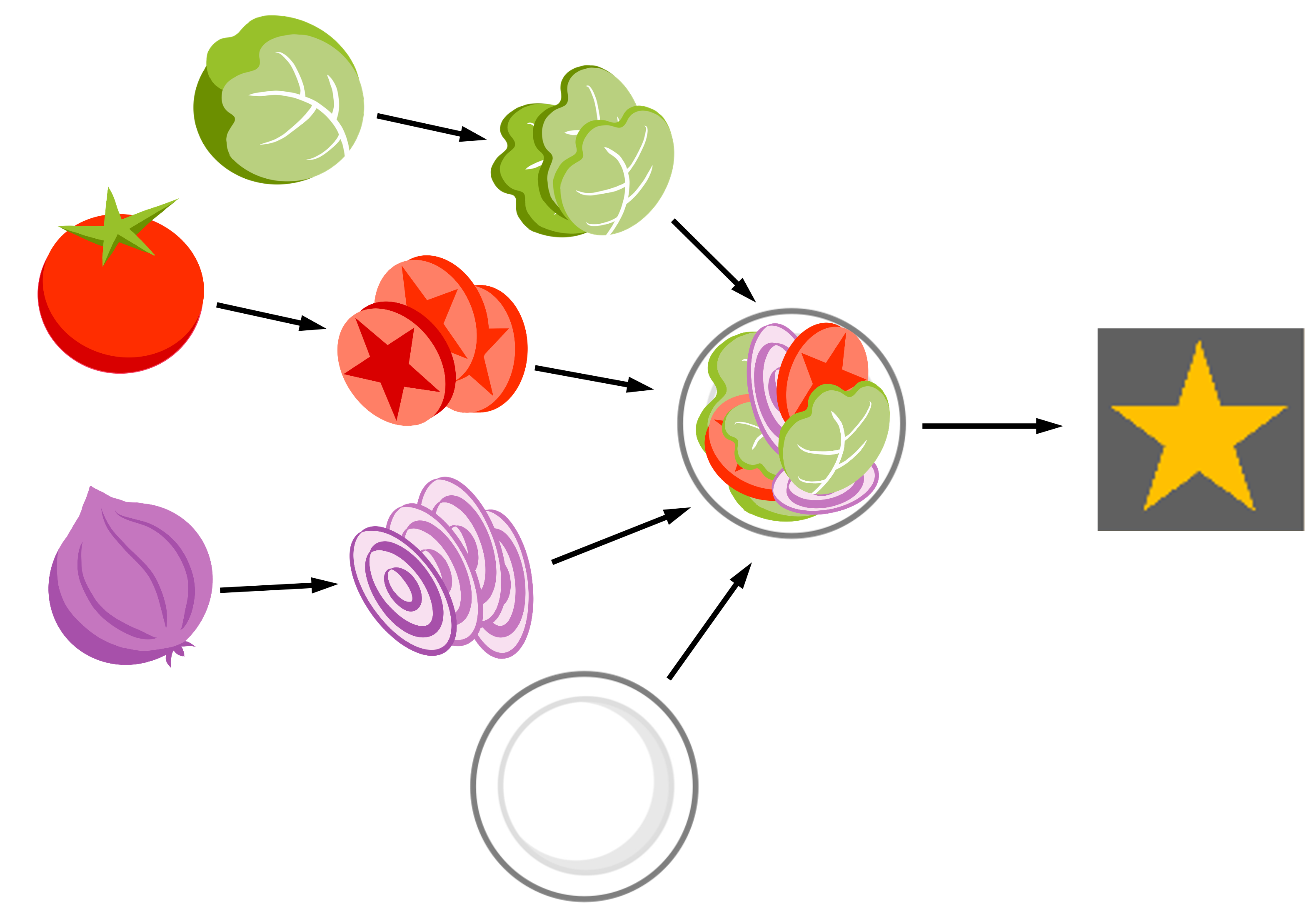}
    \caption{Salad recipe.}
    \label{recipe}
\end{wrapfigure}
\textbf{Overcooked} (Fig.~\ref{domain_OA} - \ref{domain_OB}).
Three agents work in a kitchen (7$\times$7 grid world) and must learn to cooperatively prepare a lettuce-tomato-onion salad and deliver it to the `star' cell as soon as possible.
The challenge is that the recipe for making a lettuce-tomato-onion salad (Fig.~\ref{recipe}) is unknown to the agents.
The agents have to learn the correct procedure in terms of picking up raw vegetables, chopping, and merging on a plate as well as delivering.   

{\bf (a) State Space}. The environment includes three agents, one tomato, one lettuce, one onion, two plates, two cutting boards, and one delivery cell (the 'star' one). The global state information consists of the positions of the above entities, and the status of each vegetable (chopped, unchopped, or the progress under chopping).

{\bf (b) Primitive-Action Space}. By using primitive-actions (\emph{move up}, \emph{down}, \emph{left}, \emph{right}, and \emph{stay}), agents can move around and achieve picking, placing, chopping and delivering by standing next to the corresponding cell and moving against it (e.g., in Fig.~\ref{domain_OA}, the pink agent can \emph{move right} and then \emph{move up} to pick up the tomato).

{\bf (c) Macro-Action Space}. Here, we first describe the main function of each macro-action and then list the corresponding termination conditions.
\begin{itemize}
        \item{Five one-step macro-actions that are the same as the primitive ones;} 
            \vspace{-1mm}
        \item{\textbf{\emph{Chop}}, cuts a raw vegetable into pieces (taking three time steps) when the agent stands next to a cutting board and an unchopped vegetable is on the board, otherwise it does nothing; and it terminates when:
            \begin{itemize}
                    \vspace{-1mm}
                \item{The vegetable on the cutting board has been chopped into pieces;}
                    \vspace{-1mm}
                \item{The agent is not next to a cutting board;}
                    \vspace{-1mm}
                \item{There is no unchopped vegetable on the cutting board;}
                    \vspace{-1mm}
                \item{The agent holds something in hand.}
            \end{itemize}}
            \vspace{-1mm}
        \item{\emph{\textbf{Get-Lettuce}}, \emph{\textbf{Get-Tomato}}, and \emph{\textbf{Get-Onion}}, navigate the agent to the latest observed position of the vegetable, and pick the vegetable up if it is there; otherwise, the robot moves to check the initial position of the vegetable. The corresponding termination conditions are listed below:
            \begin{itemize}
                    \vspace{-1mm}
                \item{The agent successfully picks up a chopped or unchopped vegetable;}
                    \vspace{-1mm}
                \item{The agent observes the target vegetable is held by another agent or itself;}
                    \vspace{-1mm}
                \item{The agent is holding something else in hand;}
                    \vspace{-1mm}
                \item{The agent's path to the vegetable is blocked by another agent;}
                    \vspace{-1mm}
                \item{The agent does not find the vegetable either at the latest observed location or the initial location;}
                    \vspace{-1mm}
                \item{The agent attempts to enter the same cell with another agent, but has a lower priority than another agent.}
            \end{itemize}}
            \vspace{-1mm}
        \item{\emph{\textbf{Get-Plate-1/2}}, navigates the agent to the latest observed position of the plate, and picks the vegetable up if it is there; otherwise, the robot moves to check the initial position of the vegetable. The corresponding termination conditions are listed below:
            \begin{itemize}
                    \vspace{-1mm}
                \item{The agent successfully picks up a plate;}
                    \vspace{-1mm}
                \item{The agent observes the target plate is held by another agent or itself;}
                    \vspace{-1mm}
                \item{The agent is holding something else in hand;}
                    \vspace{-1mm}
                \item{The agent's path to the plate is blocked by another agent;}
                    \vspace{-1mm}
                \item{The agent does not find the plate either at the latest observed location or at the initial location;}
                    \vspace{-1mm}
                \item{The agent attempts to enter the same cell with another agent but has a lower priority than another agent.}
            \end{itemize}}
            \vspace{-1mm}
        \item{\emph{\textbf{Go-Cut-Board-1/2}}, navigates the agent to the corresponding cutting board with the following termination conditions:
            \begin{itemize}
                    \vspace{-1mm}
                \item{The agent stops in front of the corresponding cutting board, and places an in-hand item on it if the cutting board is not occupied;}
                    \vspace{-1mm}
                \item{If any other agent is using the target cutting board, the agent stops next to the teammate;}
                    \vspace{-1mm}
                \item{The agent attempts to enter the same cell with another agent but has a lower priority than another agent.}
            \end{itemize}}
            \vspace{-1mm}
        \item{\emph{\textbf{Go-Counter}} (only available in Overcook-B, Fig.~\ref{domain_OB}), navigates the agent to the center cell in the middle of the map when the cell is not occupied, otherwise it moves to an adjacent cell. If the agent is holding an object the object will be placed. If an object is in the cell, the object will be picked up.} 
            \vspace{-1mm}
        \item{\emph{\textbf{Deliver}}, navigates the agent to the `star' cell for delivering with several possible termination conditions:
            \begin{itemize}
                    \vspace{-1mm}
                \item{The agent places the in-hand item on the cell if it is holding any item;}
                    \vspace{-1mm}
                \item{If any other agent is standing in front of the `star' cell, the agent stops next to the teammate;}
                    \vspace{-1mm}
                \item{The agent attempts to enter the same cell with another agent, but has a lower priority than another agent.}
            \end{itemize}}
\end{itemize}

{\bf (d) Macro-Observation Space}. Each robot only observes the \emph{positions} and \emph{status} of the entities within a $5\times5$ square  centered on the robot. The macro-observation space for each agent is the same as the primitive-observation space. 

{\bf (e) Rewards}. The reward mechanism involves: $+10$ for chopping a vegetable into pieces, $+200$ terminal reward for delivering a lettuce-tomato-onion salad, $-5$ reward for delivering any wrong item that is then reset to its initial position, and $-0.1$ for every time step.

{\bf (f) Episode Termination}. Each episode terminates either when agents successfully deliver a tomato-lettuce-onion salad or the horizon, 200 time steps, is reached.

\textbf{Warehouse Tool Delivery} (Fig.~\ref{domain_wtdA} - \ref{domain_wtdE}). 
In each workshop (e.g., W-0), a human is working on an assembly task (involving 4 sub-tasks that each takes a number of time steps to complete) and requires three different tools for future sub-tasks to continue. 
A robot arm (gray) must find tools for each human on the table (brown) and pass them to mobile robots (green, blue and yellow) who are responsible for delivering tools to humans. 
Note that, the correct tools needed by each human are unknown to robots, which has to be learned during training in order to perform efficient delivery without letting humans wait.
We consider variants with two or three mobile robots and two to four humans to examine the scalability of our methods and the effectiveness of Mac-IAICC at handling more intricate asynchronous terminations over robots. We also consider one faster human (orange) to check if robots can learn a priority for assisting this human (Fig.~\ref{domain_wtdB}). 

{\bf (a) State Space}. The environment is either a 5$\times$7 (Fig.~\ref{domain_wtdA} and ~\ref{domain_wtdB}) or a 5$\times$9 (Fig.~\ref{domain_wtdC} - \ref{domain_wtdE}) continuous space. 
A global state consists of the 2D position of each mobile robot, the execution status of the arm robot's current macro-action (e.g how many steps are left for completing the macro-action, but in real-world, this should be the angle and speed of each arm's joint), the subtask each human is working with a percentage indicating the progress of the subtask, and the position of each tool (either on the brown table or carried by a mobile robot). The initial state of every episode is deterministic as shown in Fig.~\ref{domain_wtdA} - \ref{domain_wtdE}, where humans always start from the first step.\\ 

{\bf (b) Mobile Robot's Macro-Action Space}. 

\begin{itemize}
        \vspace{-1mm}
    \item{\emph{\textbf{Go-W(i)}}, navigates to the waypoint (red) at the corresponding workshop;}
        \vspace{-1mm}
    \item{\emph{\textbf{Go-TR}}, goes to the waypoint at the right side of the tool room (covered by the blue robot in Fig.~\ref{domain_wtdD} and \ref{domain_wtdE});}
        \vspace{-1mm}
    \item{\emph{\textbf{Go-Tool}}, navigates to a pre-allocated waypoint (that is different for each robot to avoid collisions) next to the robot arm and waits there until either receiving a tool or 10 time steps have passed.}
\end{itemize}

{\bf (c) Robot arm's Macro-Action Space}. 

\begin{itemize}
        \vspace{-1mm}
    \item{\emph{\textbf{Search-Tool(i)}}, takes 6 time steps to find tool $i$ and place it in a staging area (containing at most two tools) on the table, and otherwise, it freezes the robot for the amount of time the action would take when the area is fully occupied;}
        \vspace{-1mm}
    \item{\emph{\textbf{Pass-to-M(i)}}, takes 4 time steps to pass the first staged tool to mobile robot $i$}
        \vspace{-1mm}
    \item{\emph{\textbf{Wait-M}, waits for 1 time step.}}
\end{itemize}

{\bf (d) Macro-Observation Space}. The robot arm only observes the \emph{type} of each tool in the staging area and \emph{which mobile robot} is waiting at the adjacent waypoints. 
Each mobile robot always knows its \emph{position} and the \emph{type} of tool that it is carrying, and can observe the \emph{number} of tools in the staging area or the \emph{sub-task} a human is working on only when at the tool room or the workshop respectively. 

{\bf (e) Environmental Dynamics}. Transitions are deterministic. Each mobile robot moves at a fixed velocity of 0.8 and is only allowed to receive tools from the arm robot rather than from humans. 
Note that each human is only allowed to possess the tool for the next subtask from a mobile robot when the robot locates at the corresponding workshop and carries the correct tool. 
Humans are not allowed to pass tools back to mobile robots. There are enough tools for humans on the table in the tool room, such that the number of each type of tool exactly matches the number of humans in the environment.
Humans cannot start the next subtask without obtaining the correct tool. Humans' dynamics in their tasks are shown in Table~\ref{table:humanDyn}.

\begin{table}[h!]
\caption {The number of time steps taken by each human on each subtask in scenarios.}
\label{table:humanDyn}
\centering
\begin{tabular}{lccccc}
\toprule
Scenarios       & Warehouse-A     & Warehouse-B   & Warehouse-C     & Warehouse-D  & Warehouse-E \\
\cmidrule(r){1-6}
Human-0         & $[27,20,20,20]$   & $[18,15,15,15]$ & $[40,40,40,40]$   & $[38,38,38,38]$  & $[40,40,40,40]$ \\ 
Human-1         & $[27,20,20,20]$   & $[48,18,15,15]$ & $[40,40,40,40]$   & $[38,38,38,38]$  & $[40,40,40,40]$ \\
Human-2         & N/A             & N/A           & $[40,40,40,40]$   & $[27,27,27,27]$  & $[40,40,40,40]$ \\
Human-3         & N/A             & N/A           & N/A             & N/A            & $[40,40,40,40]$ \\
\bottomrule
\end{tabular}
\end{table}

{\bf (f) Rewards}. The team receives: $+100$ for delivering a correct tool to a human on time, $-20$ for delayed delivery, $-10$ for the arm robot running \emph{\textbf{Pass-to-M(i)}} without the mobile robot $i$ being next to it, and $-1$ every time step. 

{\bf (g) Episode Termination}. Each episode terminates when all humans obtained all the correct tools for all subtasks, otherwise, the episode will run until reaching the maximal time steps (200 for Warehouse-A and B, 250 for Warehouse-C and D, 300 for Warehouse-E).

\subsection{Results}
\label{chap:paper3:sim:re}

\noindent All methods apply the same neural network architecture for the networks in both actor-critic and value-based approaches, which consists of two fully connected (FC) layers with Leaky-ReLU activation function, one GRU layer~\cite{GRU} and one more FC layer followed by an output layer. 
In all methods, the centralized networks have more neurons than the decentralized ones in order to deal with larger joint macro-observation and macro-action spaces, as shown in Table~\ref{table:architecture}. 
Hyper-parameter tuning uses grid search over a wide range of candidates. 
Exploration is done with a linear decaying $\epsilon$-\text{soft} policy~\cite{COMA}. 
The performance metric of one training trial is a mean discounted return measured by periodically (every 100 episodes) evaluating the learned policies over 10 testing episodes. 
We plot the averaged performance of each method over 20 independent trials with one standard error and smooth the curves over 10 neighbors. We show the optimal expected return in the Box Pushing domain as a dash-dot line. 

\begin{table}[t!]
\caption {Number of neurons on each layer in networks for all methods}
\label{table:architecture}
\centering
\begin{tabular}{ccccccc}
\toprule
Domain          & \multicolumn{2}{c}{Box Pushing} & \multicolumn{2}{c}{Overcooked} & \multicolumn{2}{c}{Warehouse} \\
\midrule
Actor \& Critic \& Q-network & Dec   & Cen   & Dec   & Cen  & Dec  & Cen  \\
\cmidrule{1-7}
MLP-1           & 32              & 32            & 32              & 128          & 32             & 32           \\
MLP-2           & 32              & 32            & 32              & 128           & 32             & 32           \\
GRU             & 32              & 64            & 32              & 64           & 32             & 64           \\
MLP-3           & 32              & 32            & 32              & 64           & 32             & 32           \\ 
\bottomrule
\end{tabular}
\end{table}

\begin{figure}[t!]
    \centering
    \captionsetup[subfigure]{labelformat=empty}
    \centering
    \subcaptionbox{}
        [0.9\linewidth]{\includegraphics[scale=0.18]{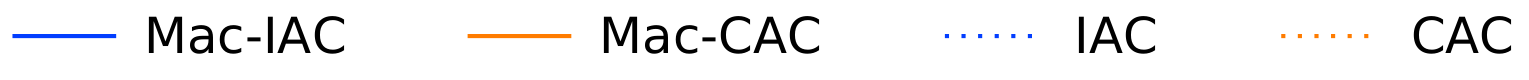}}
    ~
    \centering
    \subcaptionbox{}
        [0.32\linewidth]{\includegraphics[height=4cm]{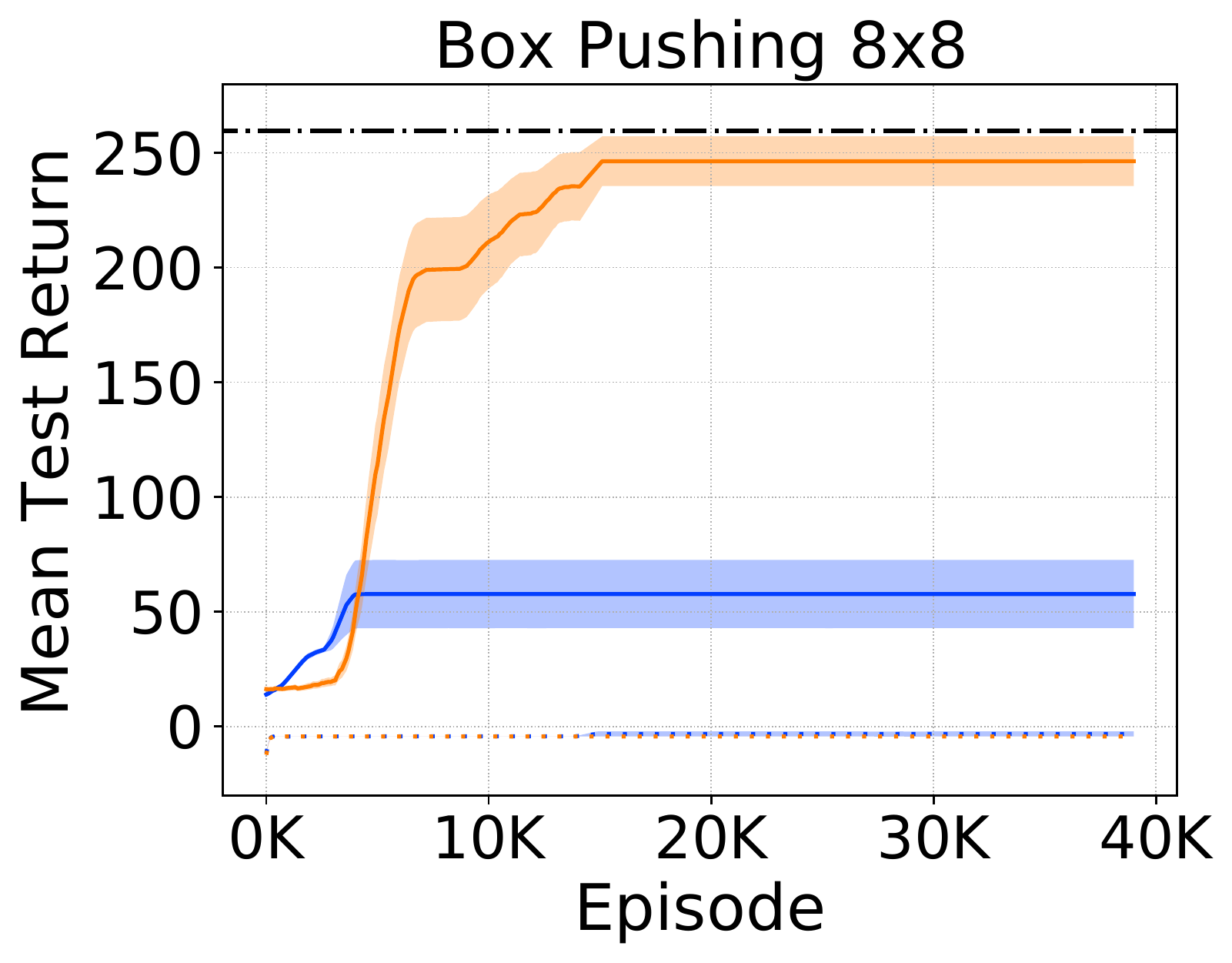}}
    ~
    \centering
    \subcaptionbox{}
        [0.32\linewidth]{\includegraphics[height=4cm]{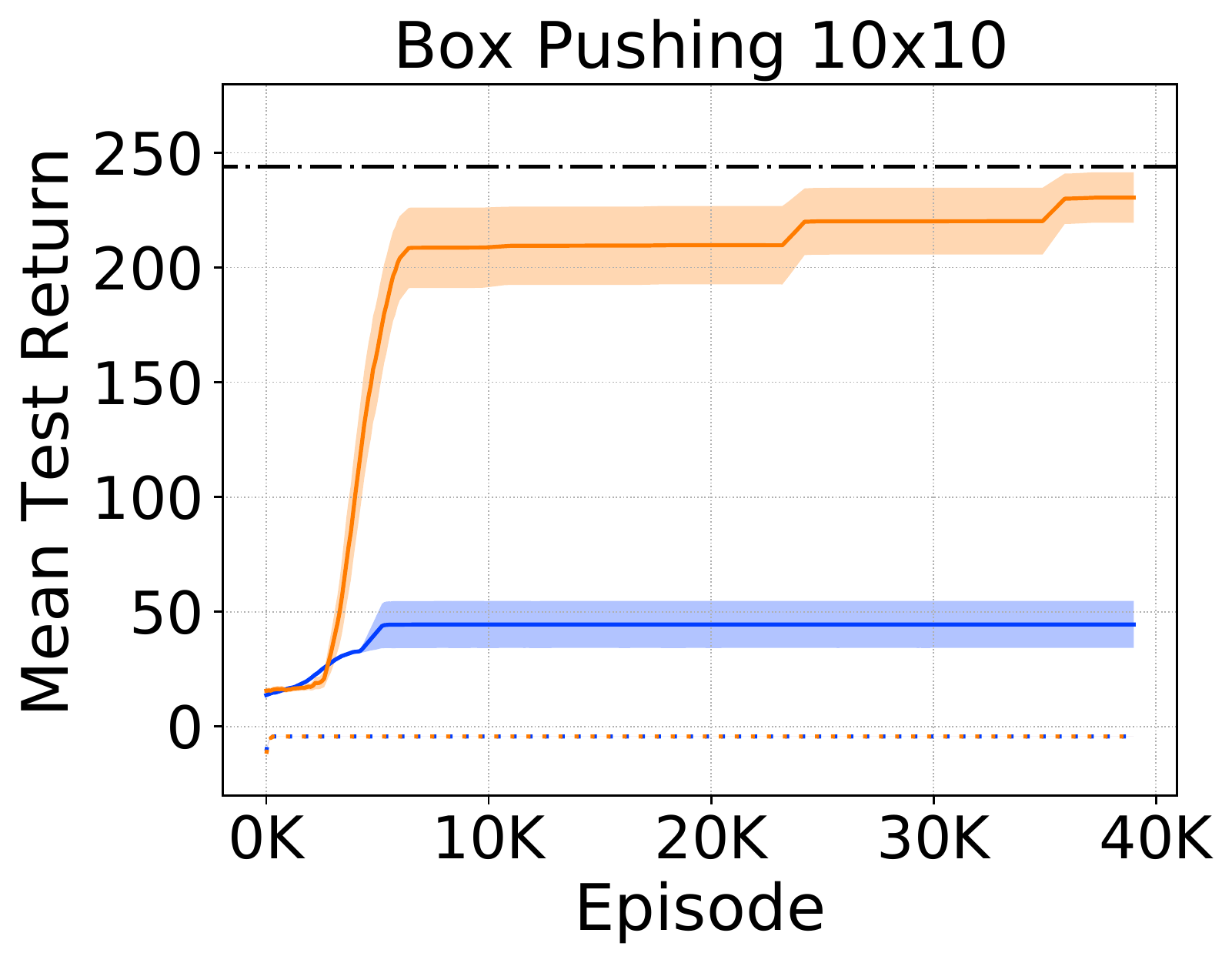}}
    ~
    \centering
    \subcaptionbox{}
        [0.32\linewidth]{\includegraphics[height=4cm]{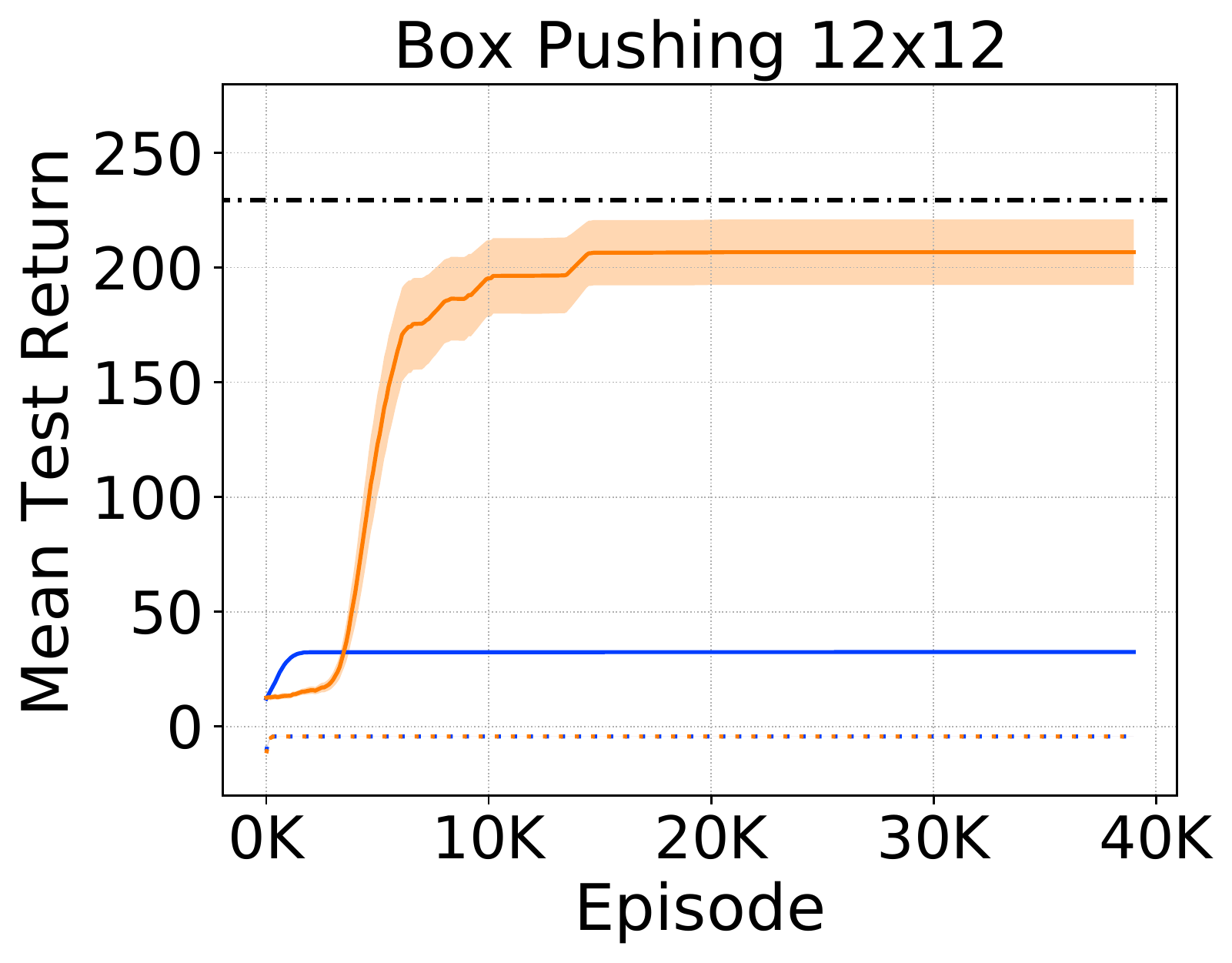}}
    ~
    \centering
    \subcaptionbox{}
        [0.32\linewidth]{\includegraphics[height=4cm]{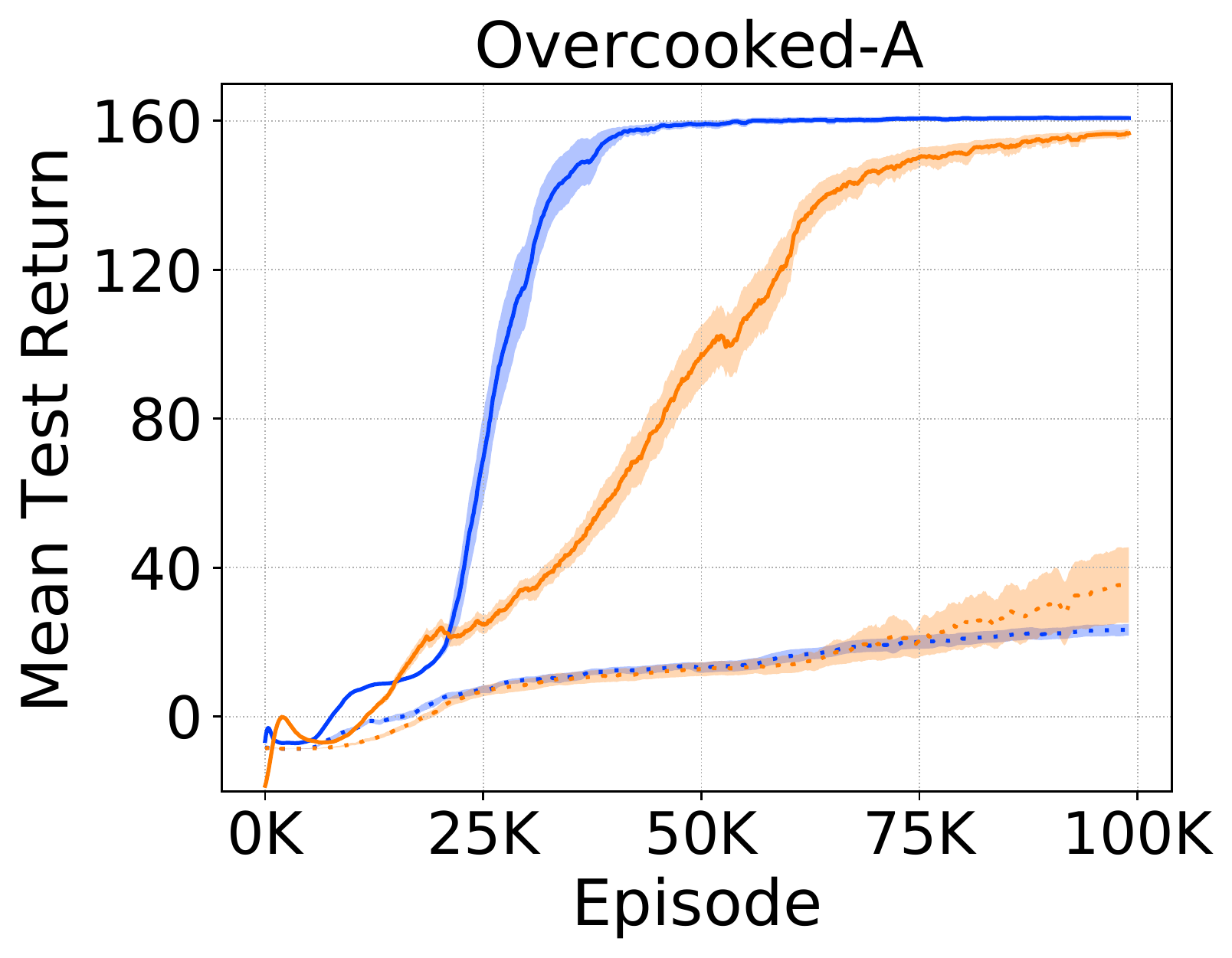}}
    ~
    \centering
    \subcaptionbox{}
        [0.32\linewidth]{\includegraphics[height=4cm]{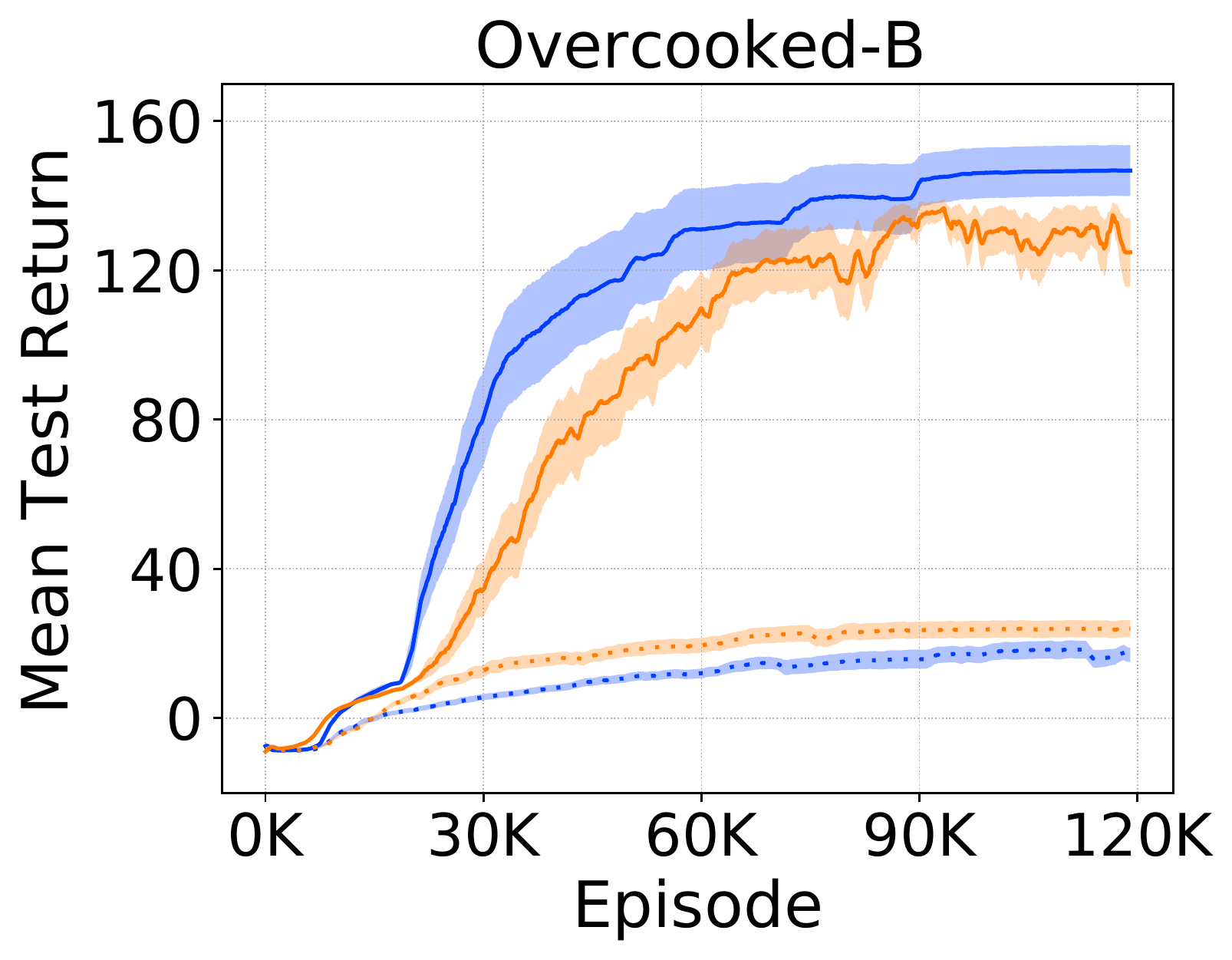}}
    \caption{Decentralized learning and centralized learning with macro-actions vs primitive-actions.}
    \label{mac_adv}
\end{figure}

\textbf{Advantages of learning with macro-actions}. We first present a comparison of our macro-action-based actor-critic methods against the primitive-action-based methods in fully decentralized and fully centralized cases. 
We consider various grid world sizes of the  Box Pushing domain (top row in Fig.~\ref{mac_adv} and two Overcooked scenarios (bottom row in Fig.~\ref{mac_adv}). The results show  significant performance improvements of using macro-actions over primitive-actions. More concretely, in the Box Pushing domain, reasoning about primitive movements at every time step 
makes the problem intractable so the robots cannot learn any good behaviors in primitive-action-based approaches other than to keep moving around. 
Conversely, 
Mac-CAC reaches near-optimal performance, enabling the robots to push the big box together.
Unlike the centralized critic which can access joint information, even in the macro-action case, it is hard for each robot's decentralized critic to correctly measure the responsibility for a penalty caused by a teammate pushing the big box alone. Mac-IAC thus converges to a local-optima of pushing two small boxes in order to avoid getting the penalty. 

In the Overcooked domain, an efficient solution requires the robots to asynchronously work on independent subtasks (e.g., in scenario A, one robot gets a plate while another two robots pick up and chop vegetables; and in scenario B, the right robot transports items while the left two robots prepare the salad). This large amount of independence explains why Mac-IAC can solve the task well. 
This also indicates that using local information is enough for robots to achieve high-quality behaviors. As a result,  Mac-CAC learns slower because it must figure out the redundant part of joint information in much larger joint macro-level history and action spaces than the spaces in the decentralized case. 
The primitive-action-based methods begin to learn, but perform poorly in such long-horizon tasks.

\begin{figure*}[t!]
    \centering
    \captionsetup[subfigure]{labelformat=empty}
    \centering
    \subcaptionbox{}
        [0.9\linewidth]{\includegraphics[scale=0.2]{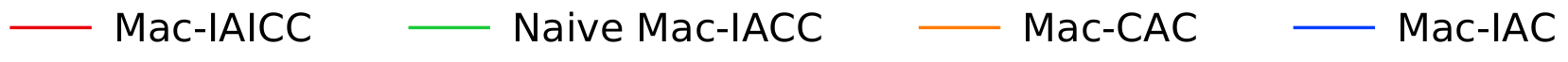}\vspace{-3mm}}
    ~
    \centering
    \subcaptionbox{}
        [0.31\linewidth]{\includegraphics[height=4cm]{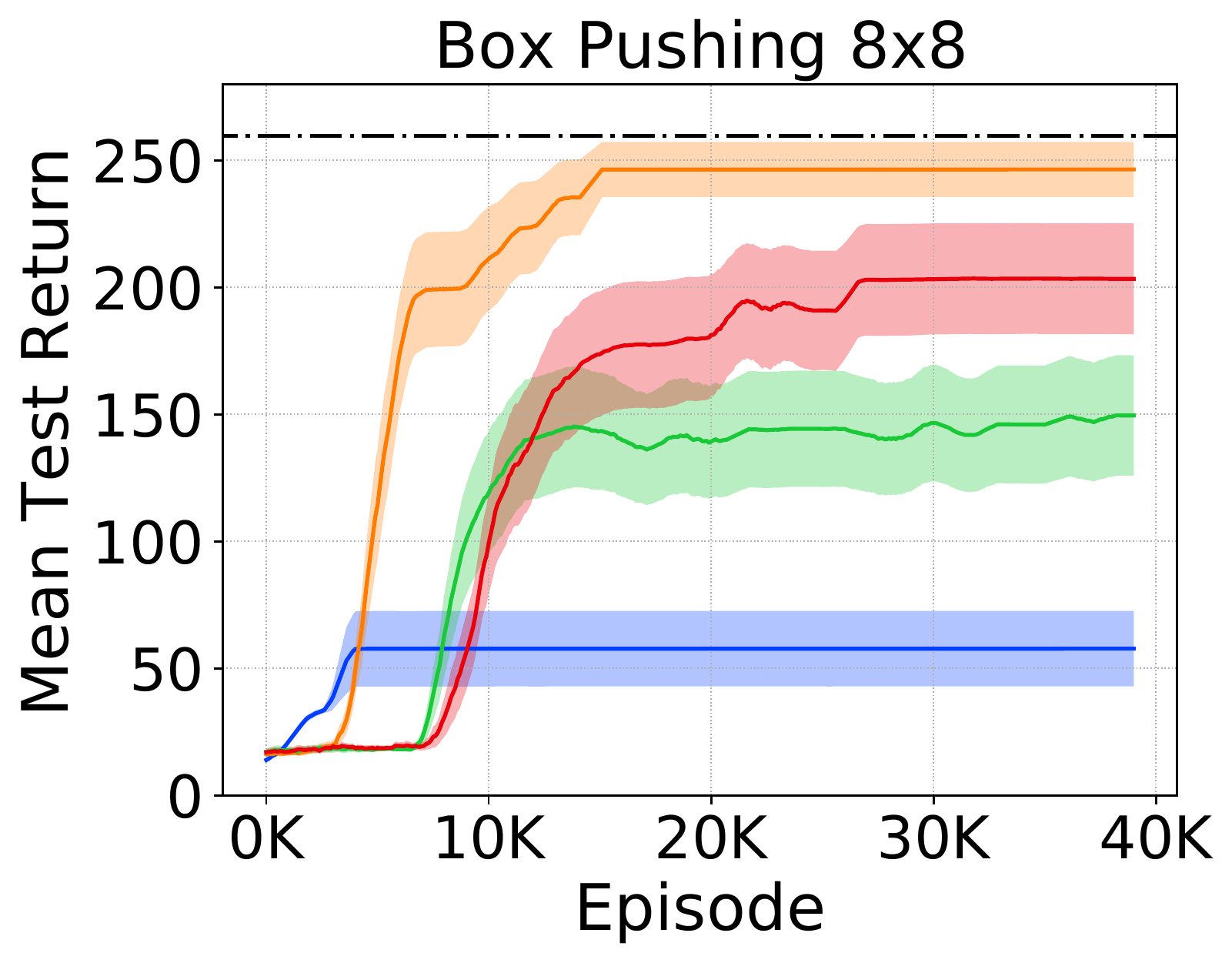}\vspace{-3mm}}
    ~
    \centering
    \subcaptionbox{}
        [0.31\linewidth]{\includegraphics[height=4cm]{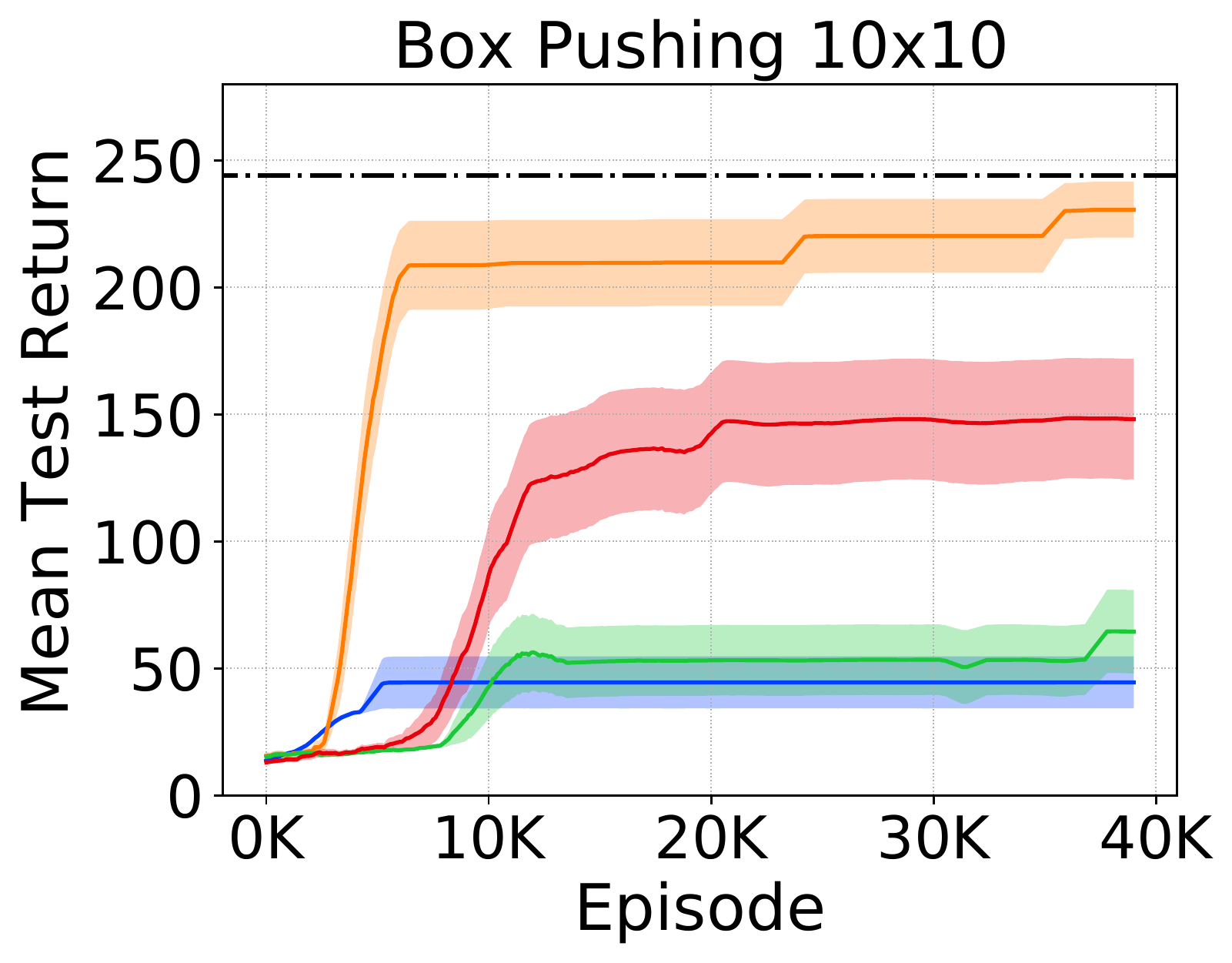}\vspace{-3mm}}
    ~
    \centering
    \subcaptionbox{}
        [0.31\linewidth]{\includegraphics[height=4cm]{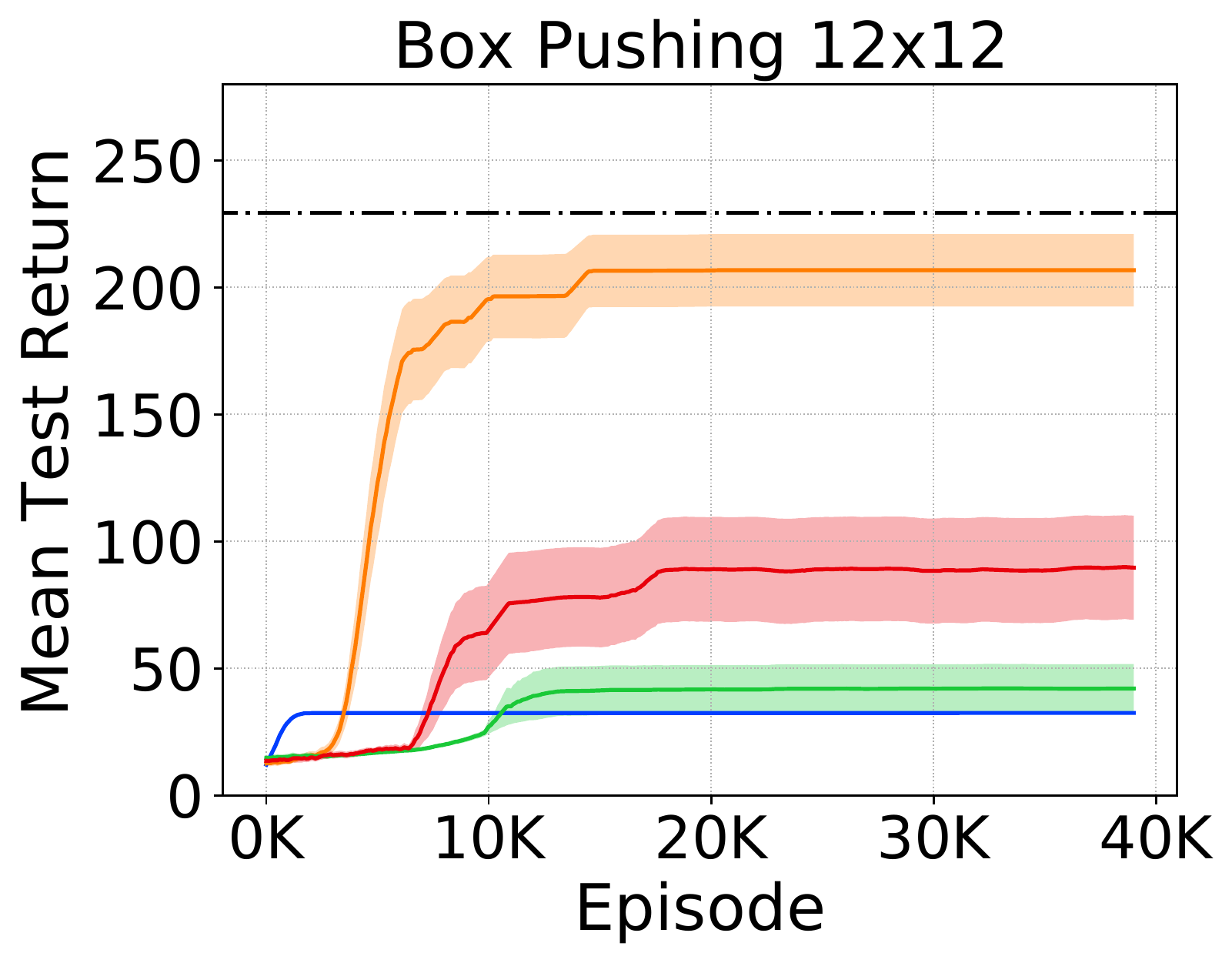}\vspace{-3mm}}
    ~
    \centering
    \subcaptionbox{}
        [0.35\linewidth]{\includegraphics[height=4cm]{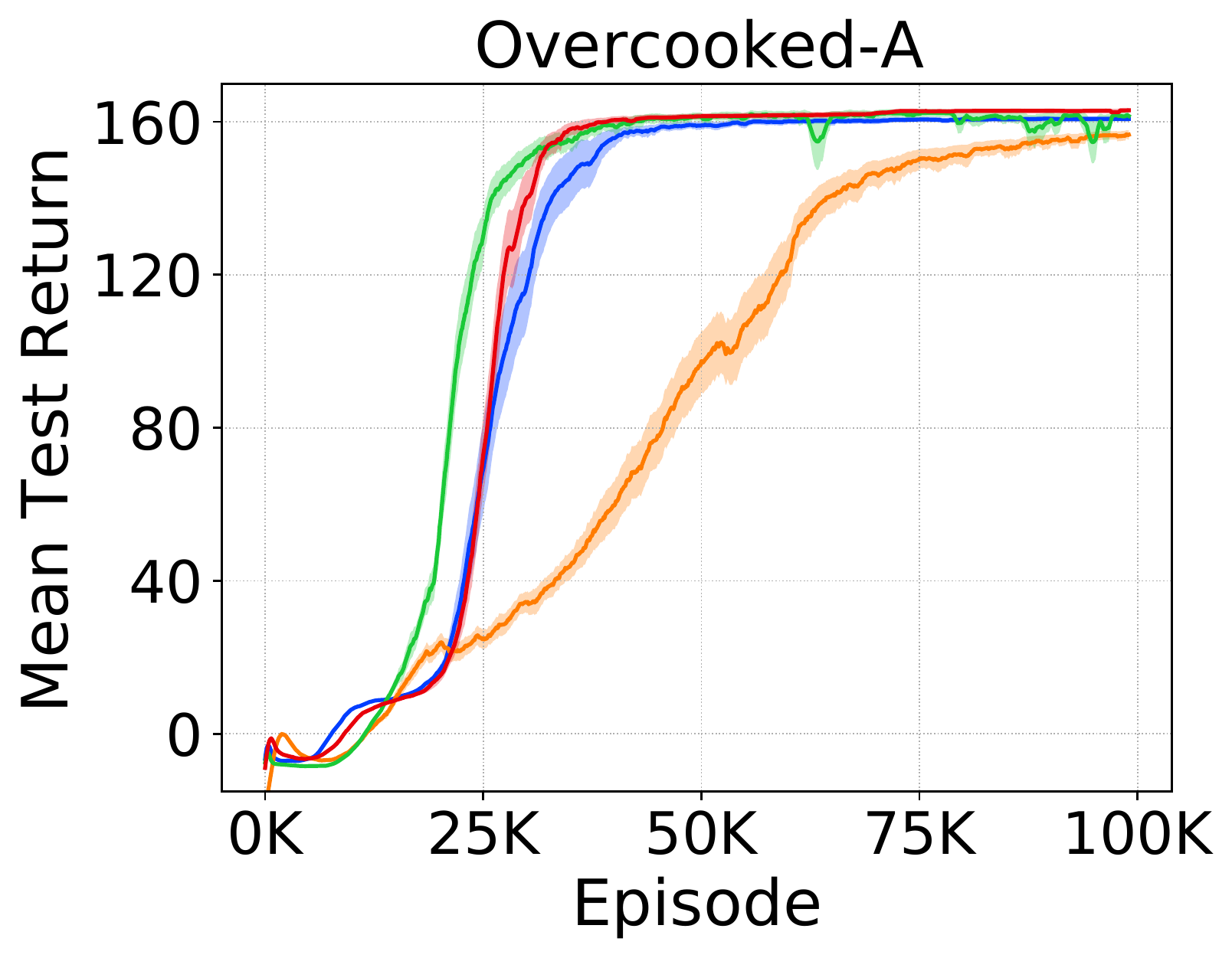}\vspace{-3mm}}
    ~
    \centering
    \subcaptionbox{}
        [0.35\linewidth]{\includegraphics[height=4cm]{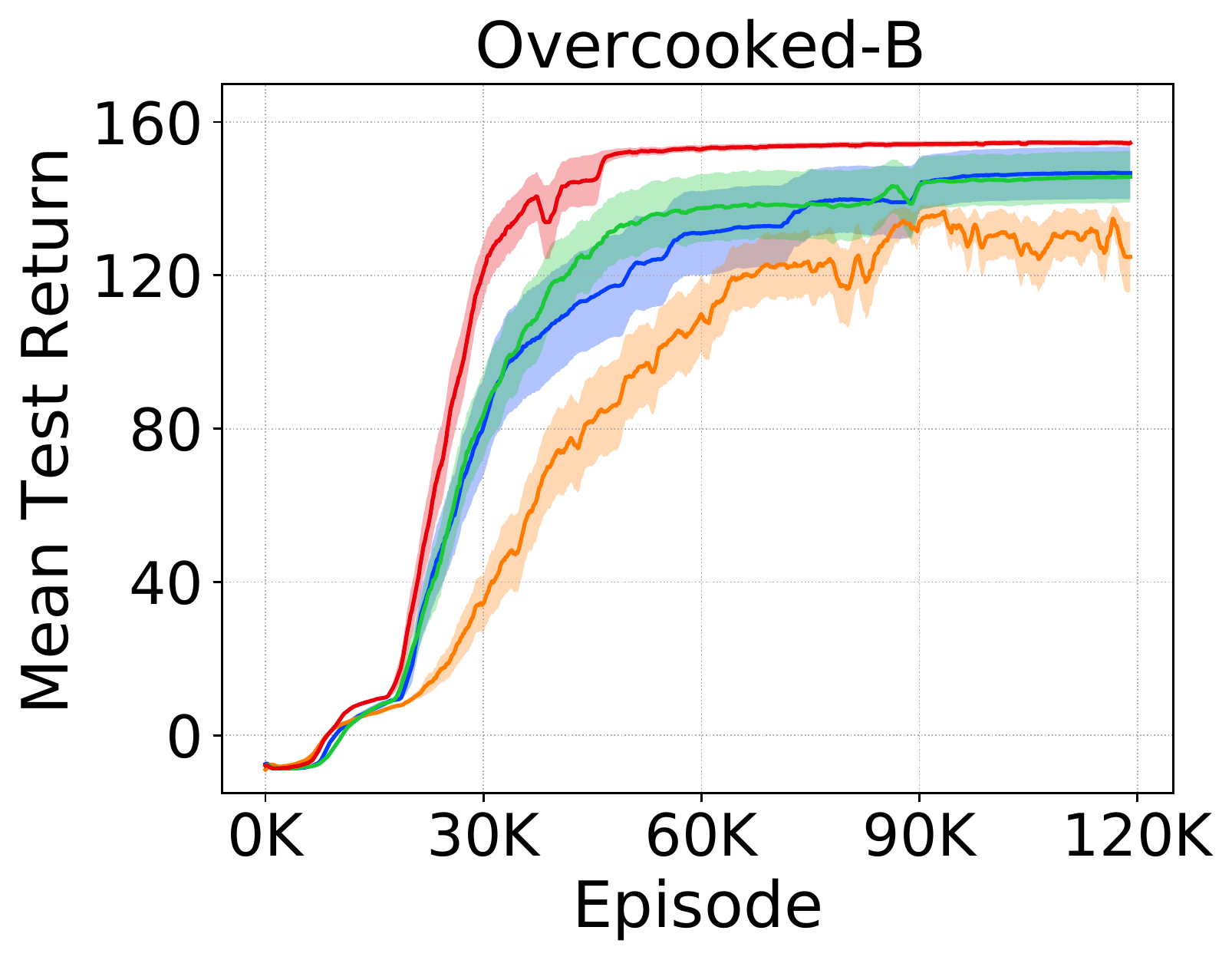}\vspace{-3mm}}
    ~
    \centering
    \subcaptionbox{}
        [0.31\linewidth]{\includegraphics[height=4cm]{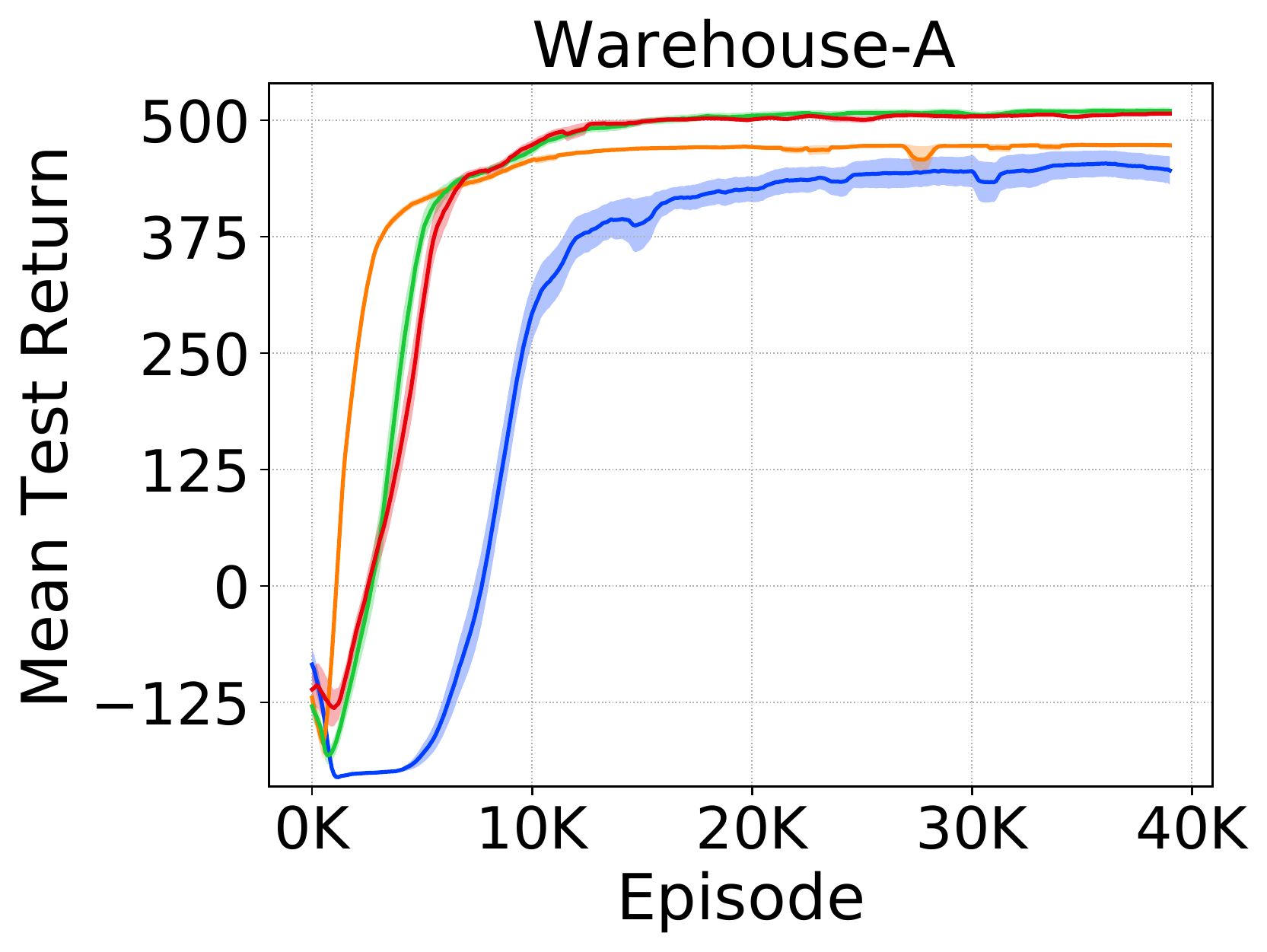}\vspace{-3mm}}
    ~
    \centering
    \subcaptionbox{}
        [0.31\linewidth]{\includegraphics[height=4cm]{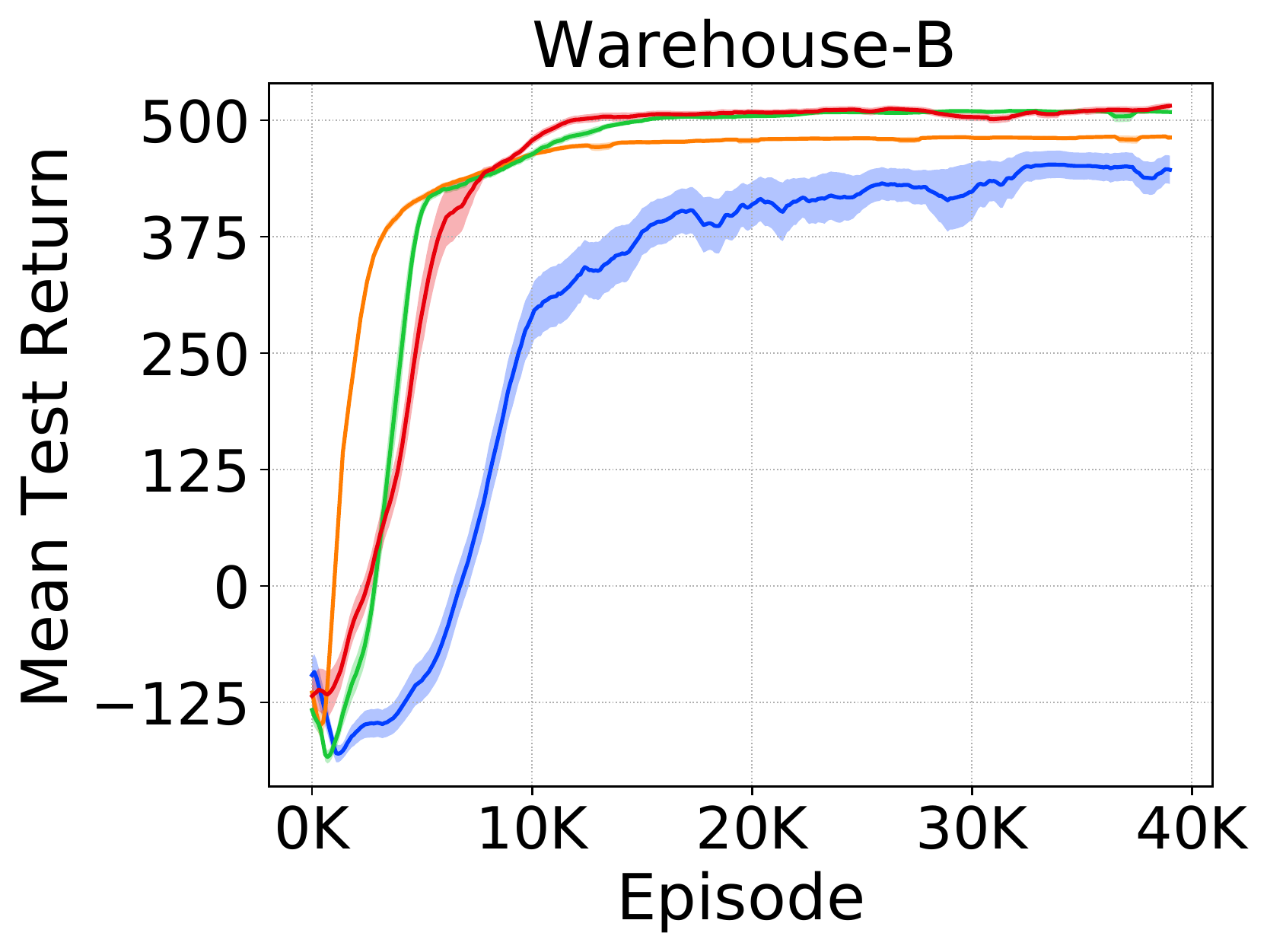}\vspace{-3mm}}
    ~    
    \centering
    \subcaptionbox{}
        [0.31\linewidth]{\includegraphics[height=4cm]{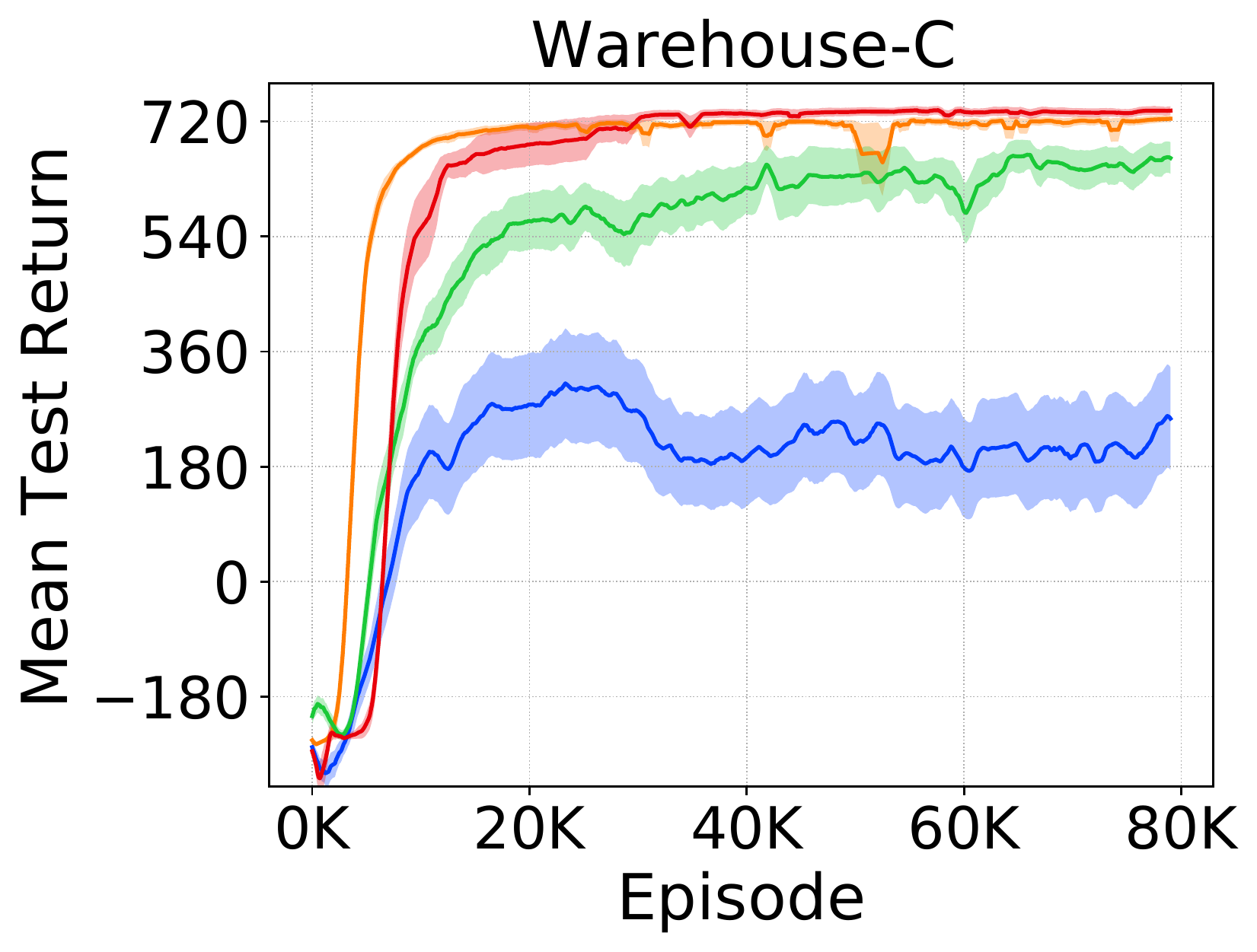}\vspace{-3mm}}
    ~
    \centering
    \subcaptionbox{}
        [0.35\linewidth]{\includegraphics[height=4cm]{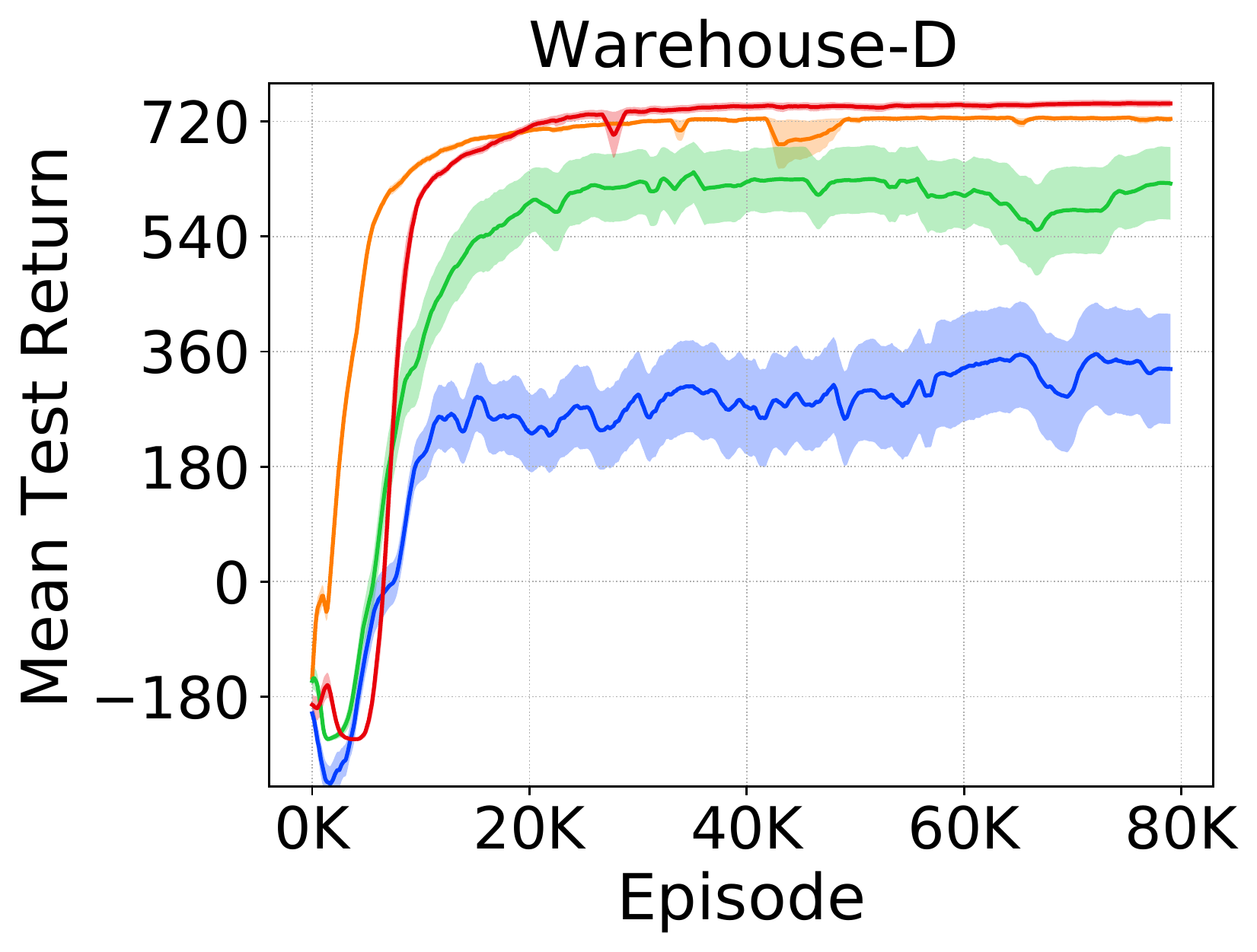}\vspace{-3mm}}
    ~
    \centering
    \subcaptionbox{}
        [0.35\linewidth]{\includegraphics[height=4cm]{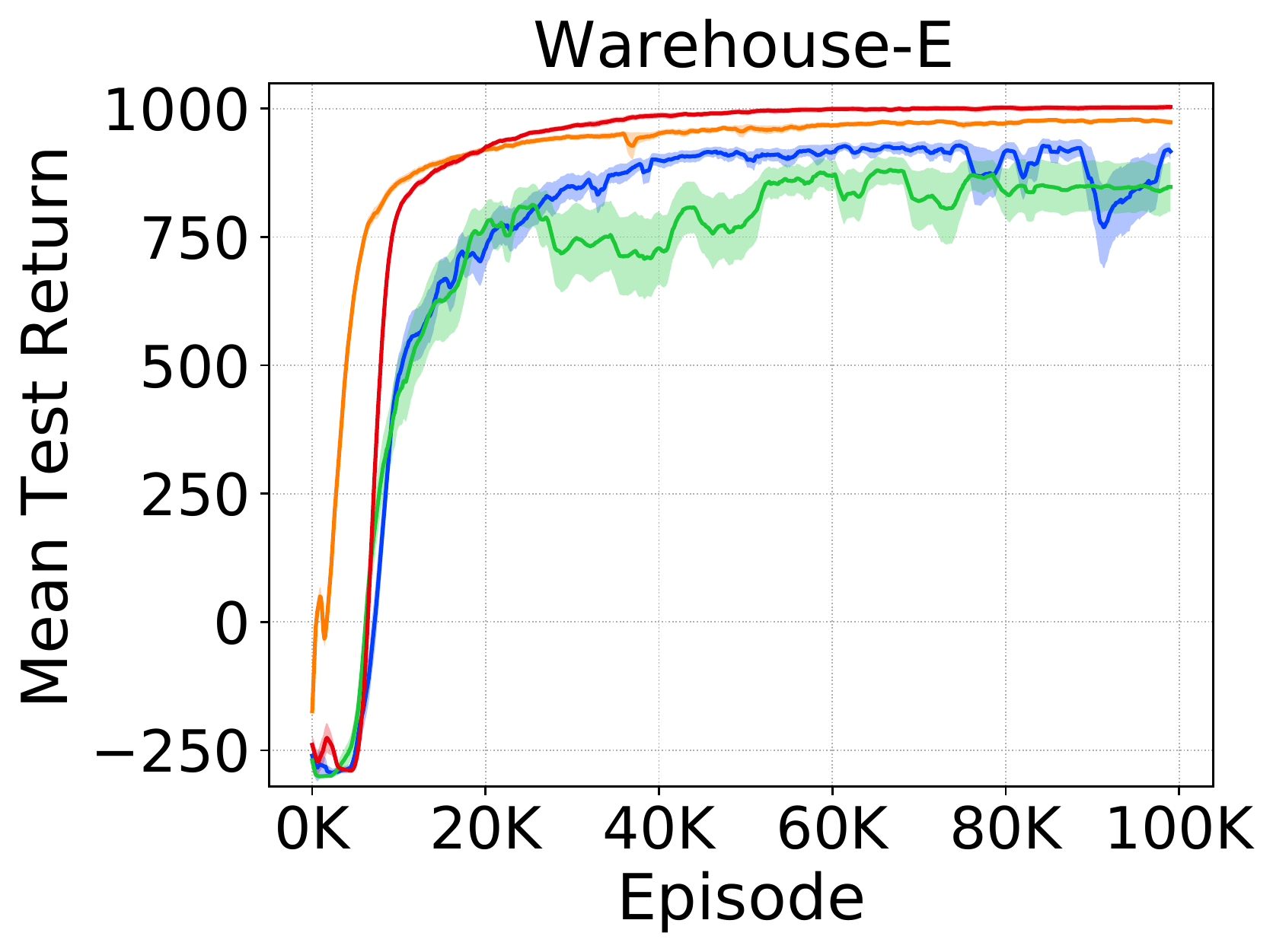}\vspace{-3mm}}
    \caption{Comparison of macro-action-based asynchronous actor-critic methods.}
    \label{mac_ctde}
\end{figure*}

\textbf{Advantages of having individual centralized critics.} Fig.~\ref{mac_ctde} shows the evaluation of our methods in all three domains.  
As each agent's observation is extremely limited in Box Pushing, we allow centralized critics in both Mac-IAICC and Naive Mac-IACC to access the state (agents' poses and boxes' positions), but use the joint macro-observation-action history in the other two domains.

In the Box Pushing task (the top row in Fig.~\ref{mac_ctde}), 
Naive Mac-IACC (green) and Mac-IAICC (red) both show improvement on the quality of decentralized policies.  
As the grid world size grows, the performance of Naive Mac-IACC drops much faster than Mac-IAICC. 
From each agent's perspective, the bigger the world size is, the more time steps a macro-action could take, and the less accurate the critic of Naive Mac-IACC becomes since it is trained depending on any agent's macro-action termination. 
Conversely, Mac-IAICC gives each agent a separate centralized critic to achieve more proper value estimation with respect to individual macro-action execution. 
As a result, despite Mac-IAICC's performance also gets reduced as the domain scaled up, agents can still learn to push the big box in most of the trials. 
Instead, Naive Mac-IACC almost traps into a local-optimal behavior such as only pushing the small boxes.   

In Overcooked-A (the left one in the second row in Fig.~\ref{mac_ctde}), as Mac-IAICC's performance is determined by the training of three agents' critics, it learns slower than Naive Mac-IACC in the early stage but converges to a slightly higher value and has better learning stability than Naive Mac-IACC in the end. 
The result of scenario B (the right one in the second row in Fig.~\ref{mac_ctde}) shows that Mac-IAICC outperforms other methods in terms of achieving  better sample efficiency, a higher final return, and a lower variance. 
The middle wall in scenario B limits each agent's moving space and leads to a higher frequency of  macro-action terminations. The shared centralized critic in Naive Mac-IACC thus provides more noisy value estimations for each agent's actions. Because of this, Naive Mac-IACC performs worse with more variance. Mac-IAICC, however, does not get hurt by such environmental dynamics change. 
Both Mac-CAC and Mac-IAC are not competitive with Mac-IAICC in this domain.  

In the Warehouse scenarios (the third row and the last row in Fig.~\ref{mac_ctde}), Mac-IAC (blue) performs the worst due to its natural limitations and the domain's partial observability. 
In particular, it is difficult for the gray robot (arm) to learn an efficient way to find the correct tools purely based on local information and very delayed rewards that depend on the mobile robots' behaviors. In contrast, in the fully centralized Mac-CAC (orange), both the actor and the critic have global information, so Mac-CAC can learn faster in the early training stage. However, Mac-CAC eventually gets stuck at a local-optima in all five scenarios due to the exponential dimensionality of joint history and action spaces over robots.   
By leveraging the CTDE paradigm, both Mac-IAICC and Naive Mac-IACC perform the best in warehouse A. Yet, the weakness of Naive Mac-IACC is clearly exposed when the problem is scaled up in Warehouse B, C, and D. In these larger cases, the robots' asynchronous macro-action executions (e.g., traveling between rooms) become more complex and cause more mismatching between the termination from each agent's local perspective and the termination from the centralized perspective, and therefore, Naive Mac-IACC's performance significantly deteriorates, even getting worse than Mac-IAC in Warehouse-D. 
In contrast, Mac-IAICC can maintain its outstanding performance, converging to a higher value with much lower variance, compared to other methods. 
This outcome confirms not only Mac-IAICC's scalability but also the effectiveness of having an individual critic for each agent to handle variable degrees of asynchronicity in agents' high-level decision-making. 
\vspace{10mm}

\begin{minipage}{\textwidth}

  \begin{minipage}[h]{0.52\textwidth}
    \centering
        \begin{tabular}{lcccccc}
        \toprule
        Scenarios & Ablation Experiment \\
        \cmidrule(r){1-1}
        Human-0         & $[18,18,18,18]$ \\ 
        Human-1         & $[18,18,18,18]$\\
        Human-2         & N/A \\
        Human-3         & N/A\\
        \bottomrule
        \end{tabular}
      \captionof{table}{The number of time steps taken by each human in the ablation study. }
      \end{minipage}
  \hfill
  \begin{minipage}[h]{0.45\textwidth}
    \centering
    \includegraphics[height=4cm]{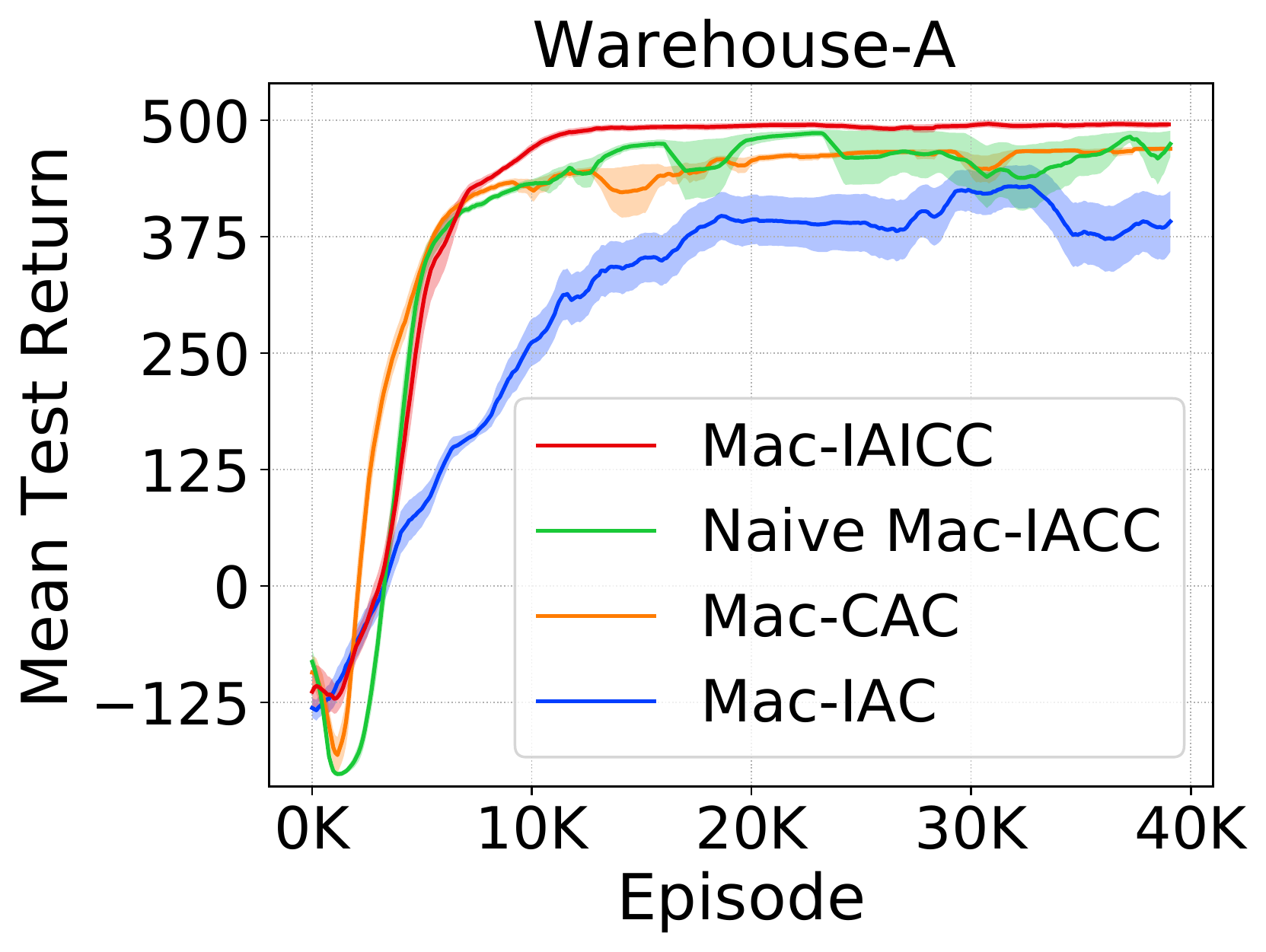}
    \captionof{figure}{Results of an ablation study.}
    \label{ablation}
  \end{minipage}
\end{minipage}
\vspace{10mm}

\begin{figure}[b!]
    \centering
    \captionsetup[subfigure]{labelformat=empty}
    \centering
    \subcaptionbox{}
        [0.9\linewidth]{\includegraphics[scale=0.2]{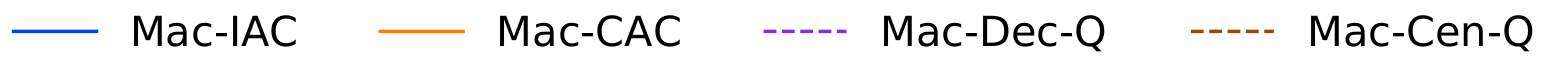}}
    ~
    \centering
    \subcaptionbox{}
        [0.31\linewidth]{\includegraphics[height=4cm]{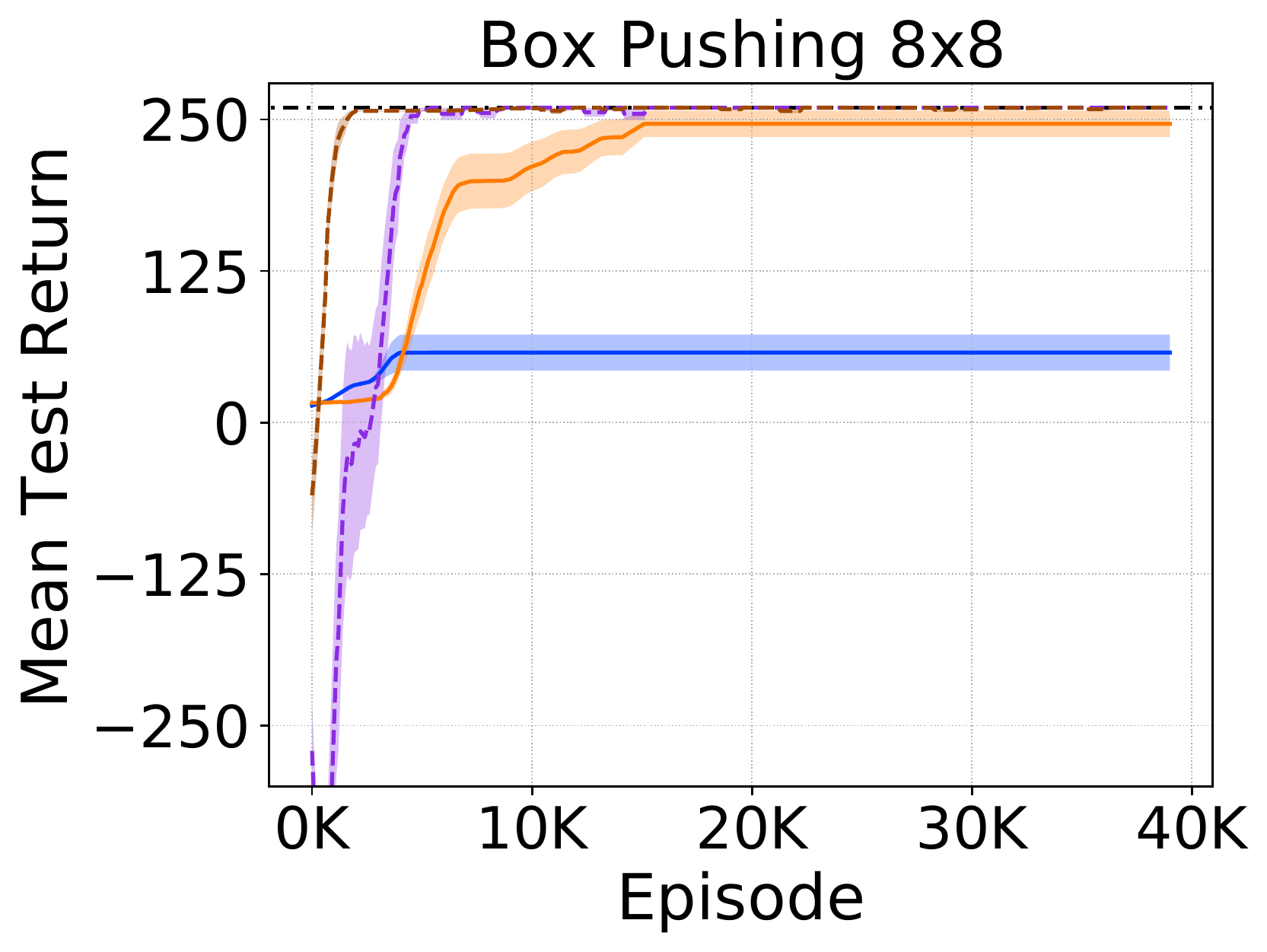}\vspace{-3mm}}
    ~
    \centering
    \subcaptionbox{}
        [0.31\linewidth]{\includegraphics[height=4cm]{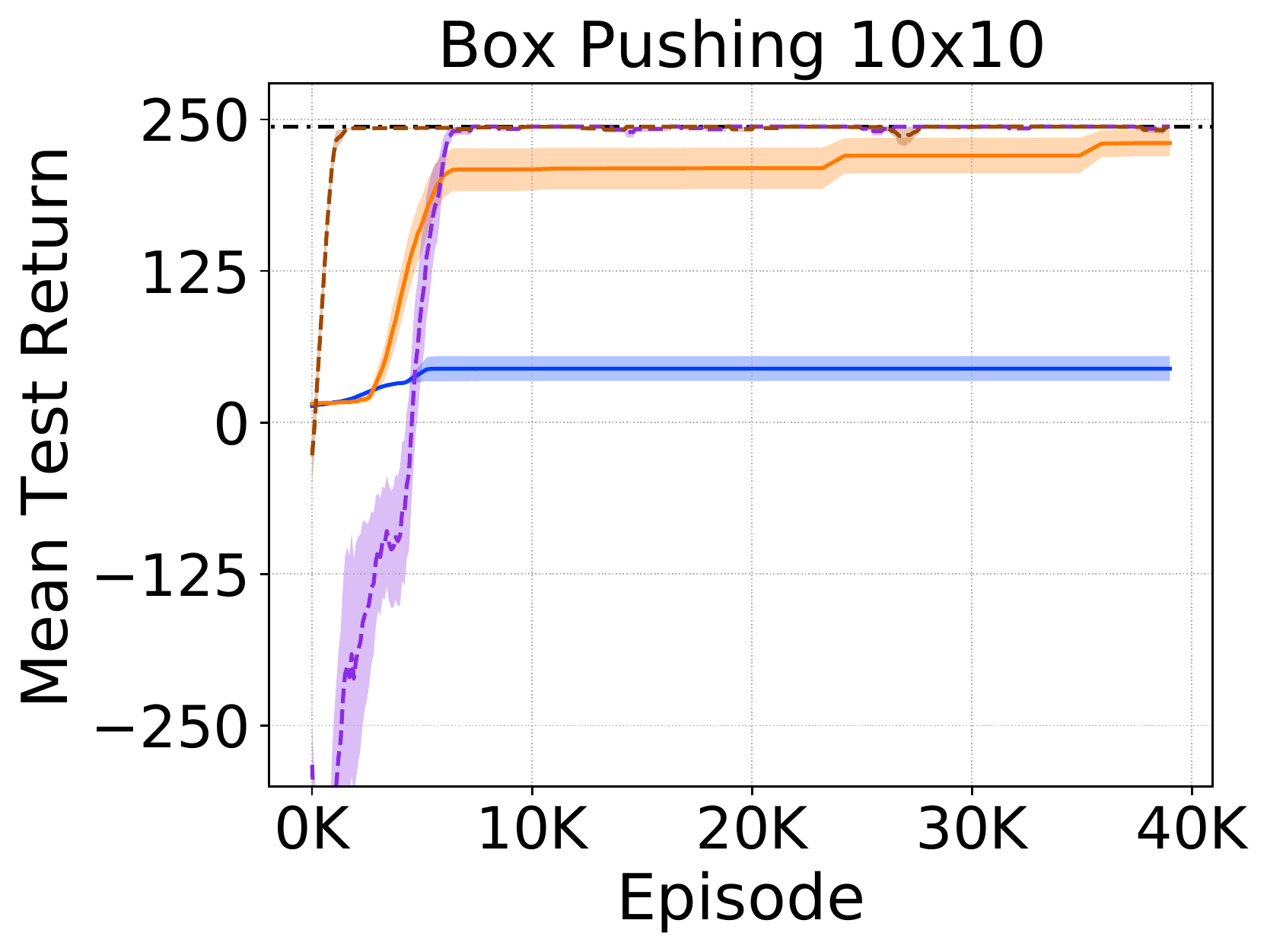}\vspace{-3mm}}
     ~
    \centering
    \subcaptionbox{}
        [0.31\linewidth]{\includegraphics[height=4cm]{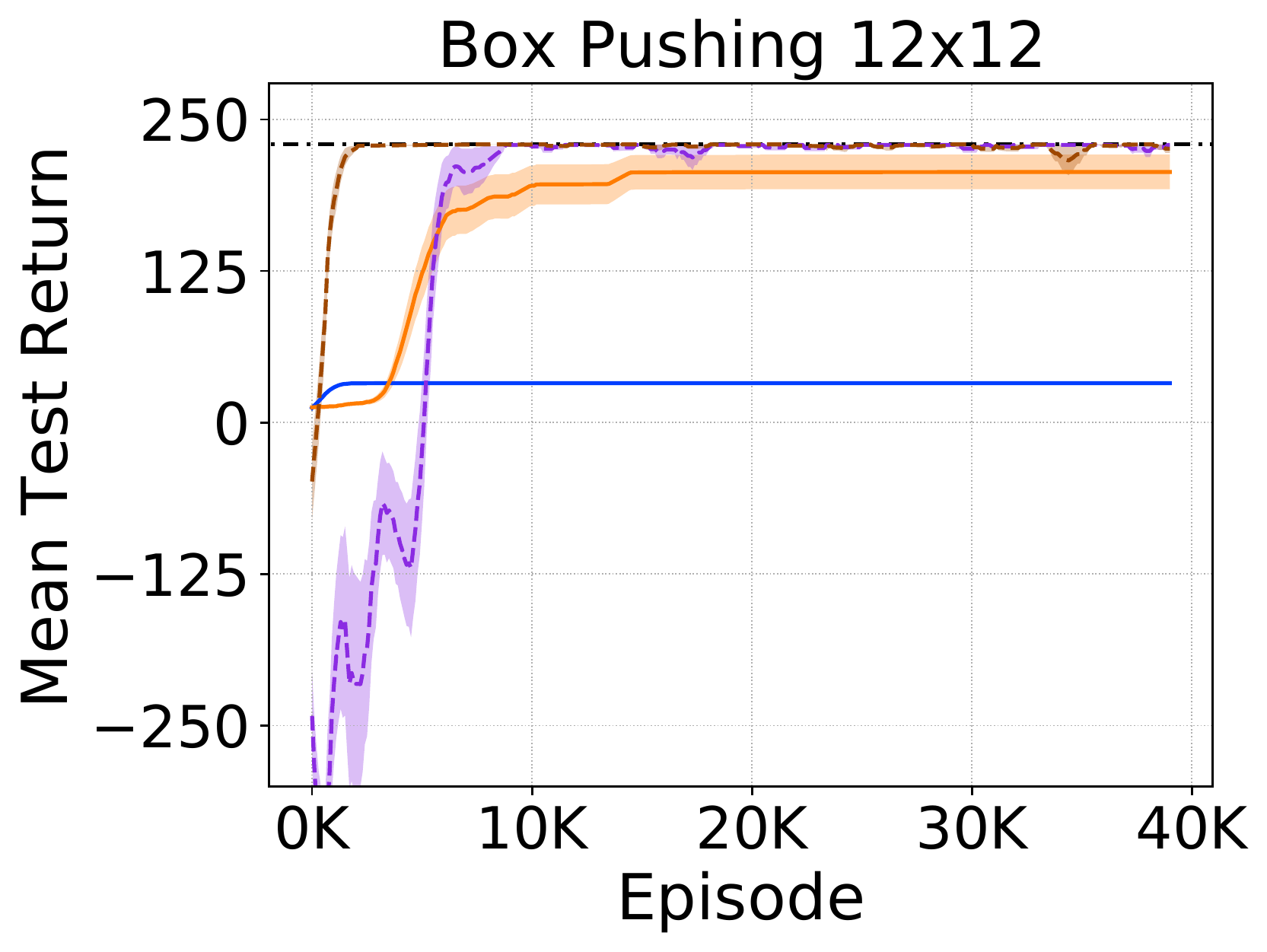}\vspace{-3mm}}
    ~
    \centering
    \subcaptionbox{}
        [0.4\linewidth]{\includegraphics[height=4cm]{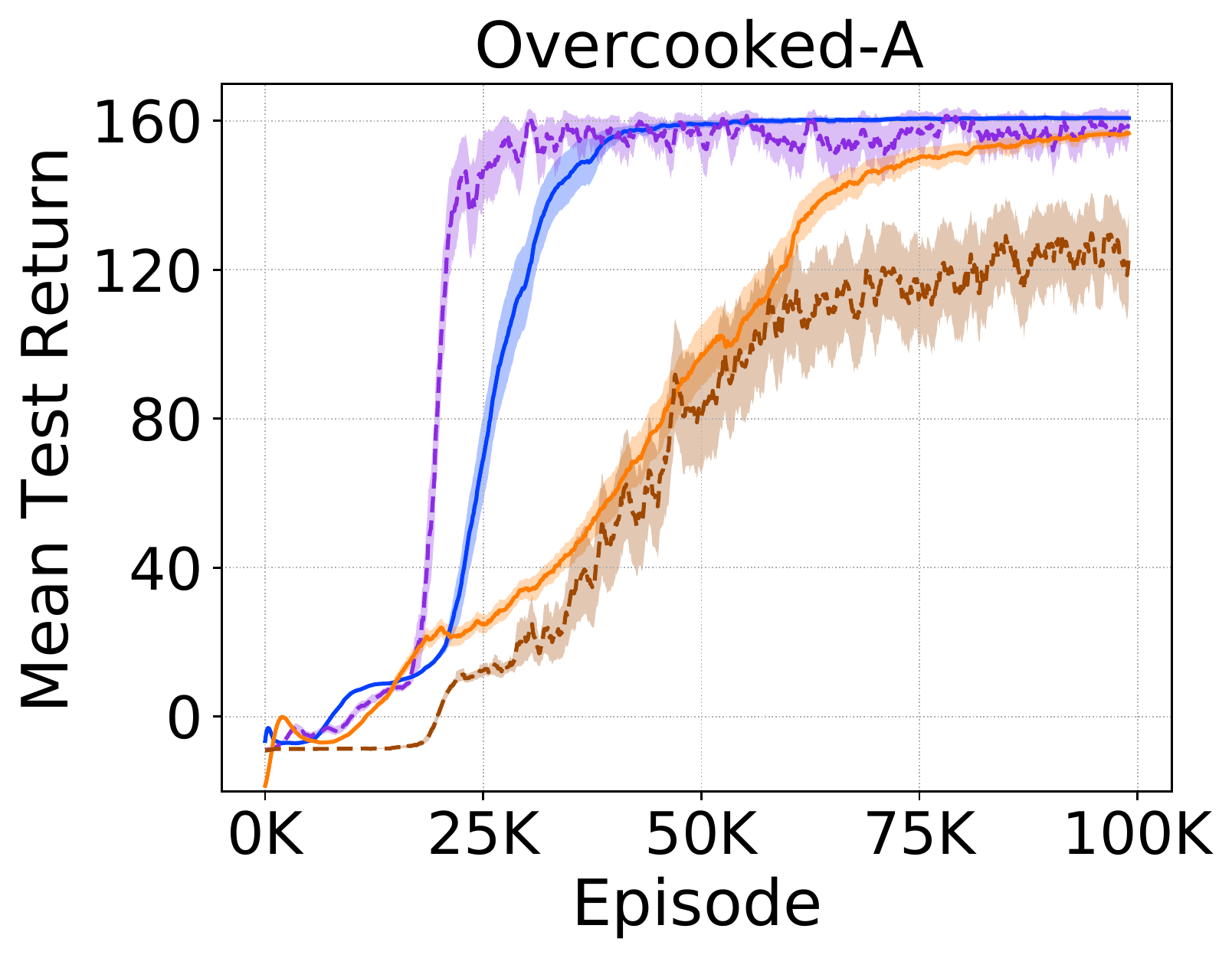}\vspace{-3mm}}
    ~
    \centering
    \subcaptionbox{}
        [0.4\linewidth]{\includegraphics[height=4cm]{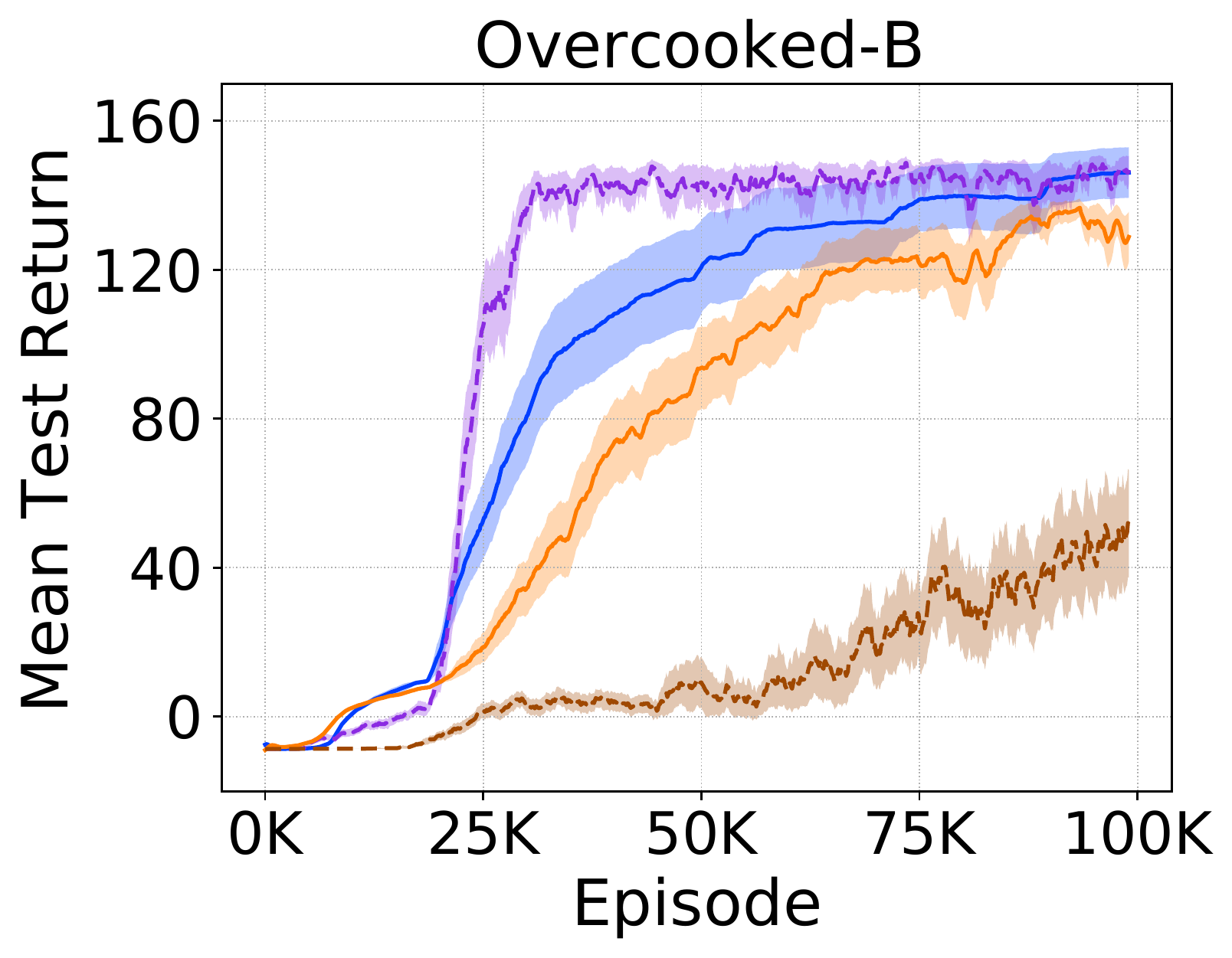}\vspace{-3mm}}
    ~
    \centering
    \subcaptionbox{}
        [0.31\linewidth]{\includegraphics[height=4cm]{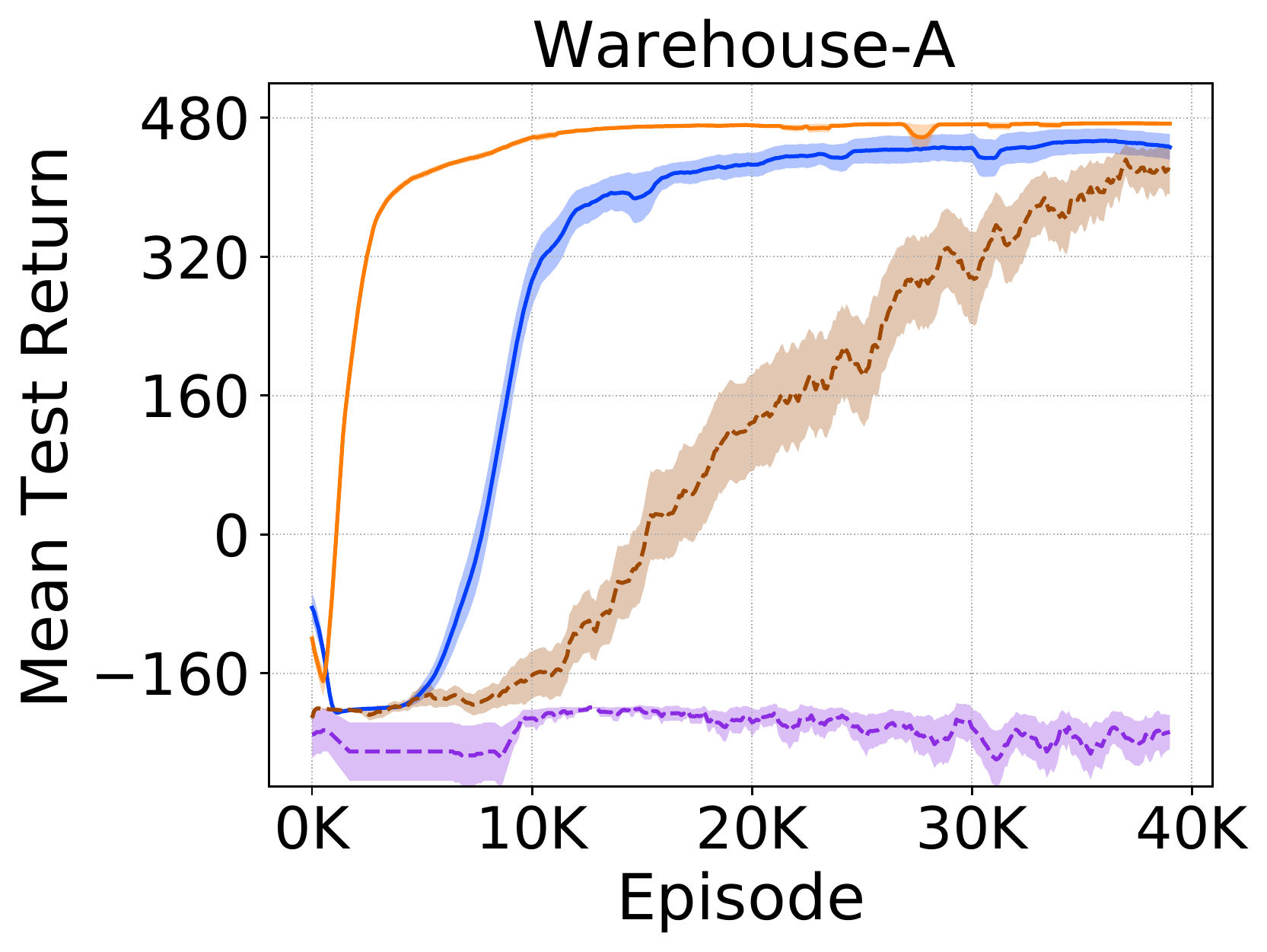}\vspace{-3mm}}
    ~
    \centering
    \subcaptionbox{}
        [0.31\linewidth]{\includegraphics[height=4cm]{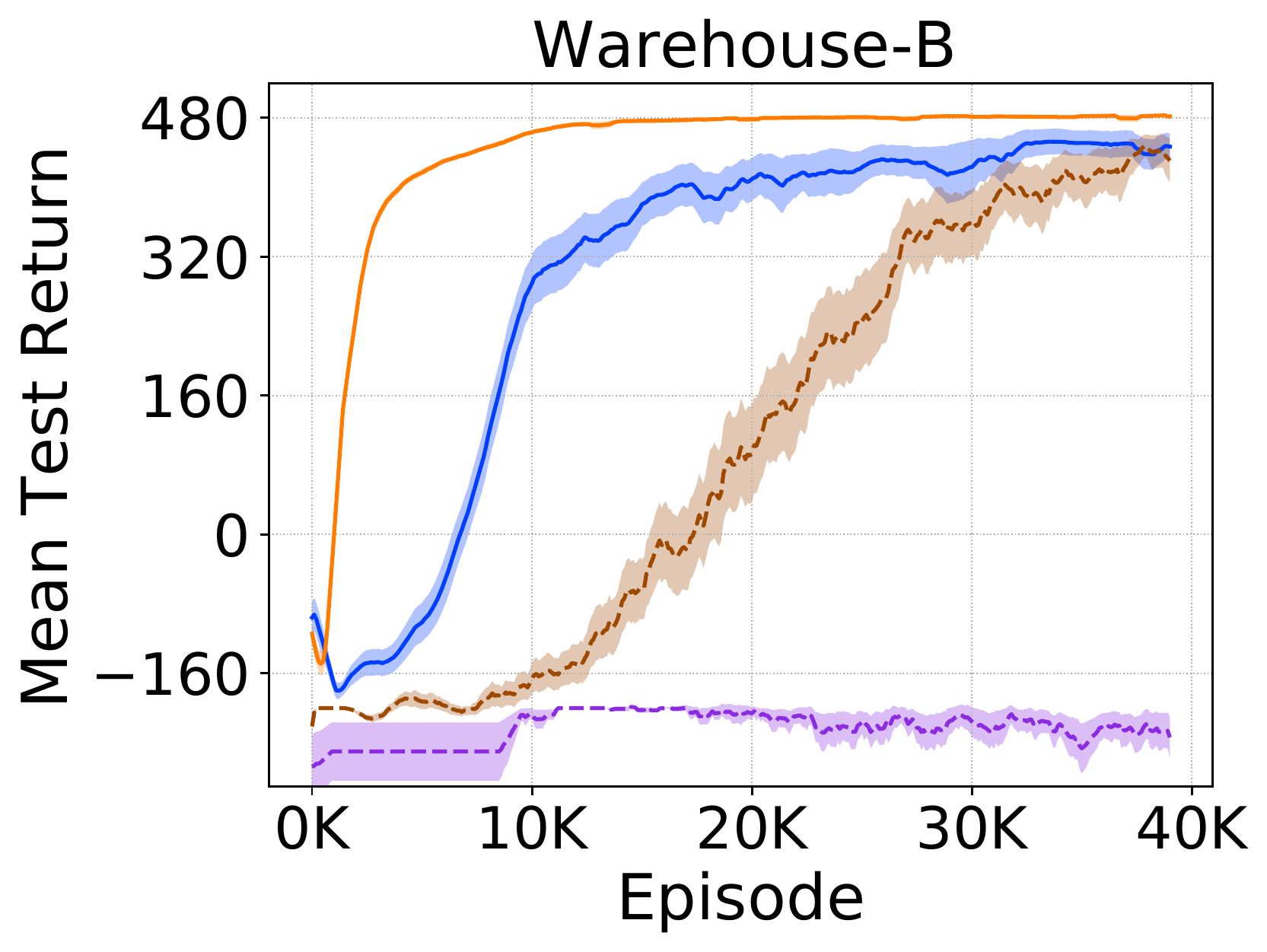}\vspace{-3mm}}
    ~
    \centering
    \subcaptionbox{}
        [0.31\linewidth]{\includegraphics[height=4cm]{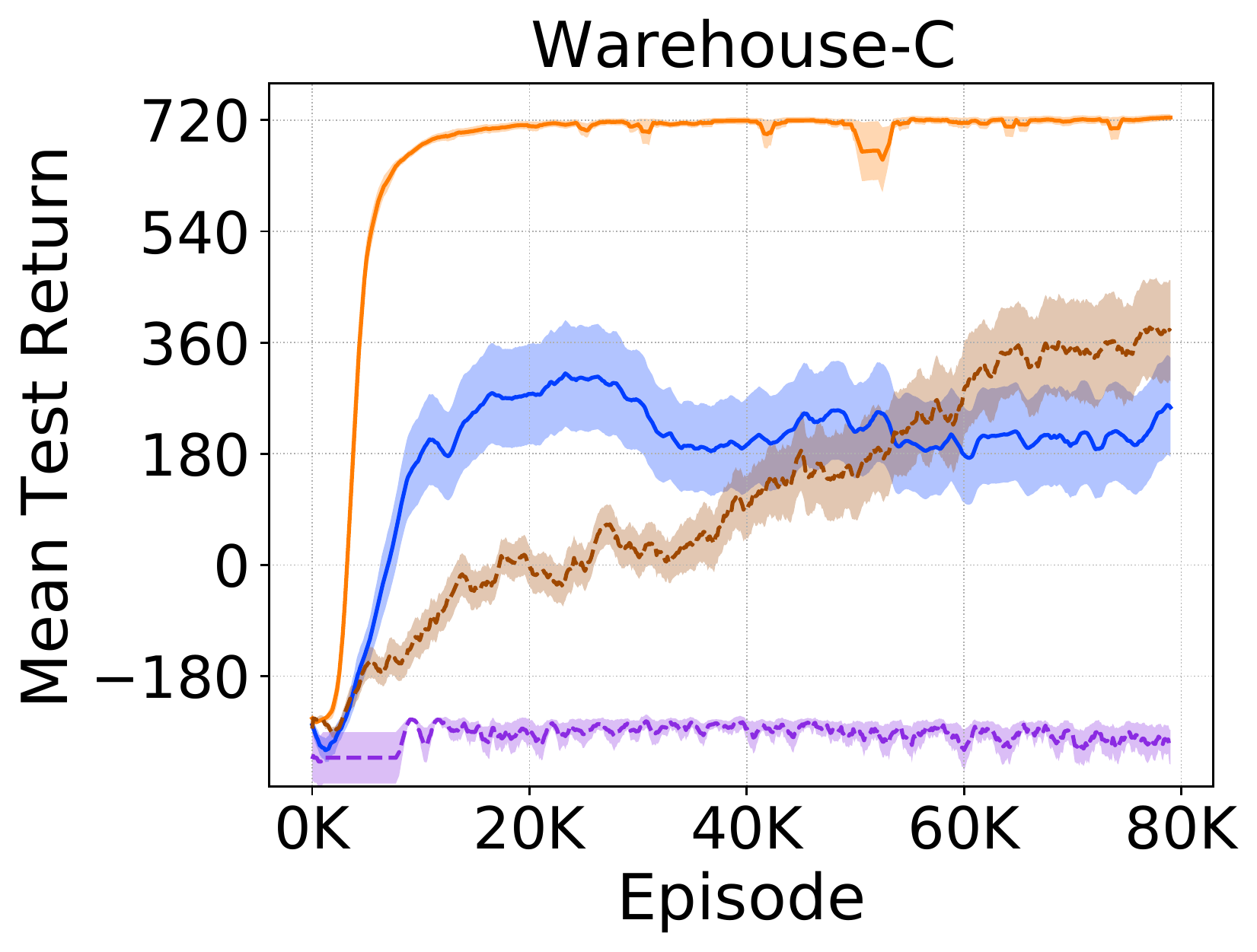}\vspace{-3mm}}
    \caption{Comparisons of macro-action-based actor-critic methods and value-based methods.}
    \label{pg_vs_ac}
\end{figure}

We also conducted an ablation experiment in Warehouse-A, where two humans still operate at the same speed on their tasks but faster than the original setting. Such a change makes agents' learning more difficult, because the probability of having a delayed delivery for each tool grows,
especially when agents are exploring. Agents likely receive more penalties during training.    
Fig.~\ref{ablation} shows the learning quality of Naive Mac-IACC degrades markedly and becomes much less stable with higher variance than its performance in the original domain configuration (shown in Fig.~\ref{domain_wtdA}). In contrast, Mac-IAICC retains its high-quality performance, which reveals its robustness to noisy penalty signals and further proves the advantage of separately training a centralized critic depending on each agent's own macro-action terminations. Both Mac-CAC and Mac-IAC still cannot rival Mac-IAICC. 

\textbf{Comparative analysis between actor-critic and value-based approaches in decentralized and centralized training paradigms}.
Here, we compare our actor-critic methods (Mac-IAC and Mac-CAC) with the value-based approaches (Mac-Dec-Q and Mac-Cen-Q) proposed in Chapter~\ref{chap:paper1}, shown in Fig.~\ref{pg_vs_ac}. 
The Box Pushing task requires agents to simultaneously reach the big box and push it together. 
This consensus is rarely achieved when agents independently sample actions using stochastic policies in Mac-IAC and it is hard to learn from pure on-policy data. 
By having a replay buffer, value-based approaches show much stronger sample efficiency than on-policy actor-critic approaches in this domain with a small action space (left figure).   
Such an advantage is sustained by the decentralized value-based method (Mac-Dec-Q) but gets lost in the centralized one (Mac-Cen-Q) in the Overcooked domains due to a huge joint macro-action space ($15^3$).
On the contrary, our actor-critic methods can scale to large domains and learn high-quality solutions. 
This is particularly noticeable in Warehouse-A, where the policy gradient methods quickly learn a high-quality policy while the centralized Mac-Cen-Q is slow to learn and the decentralized Mac-Dec-Q is unable to learn. 
In addition, the stochastic policies in actor-critic methods potentially have a better exploration property so that, in Warehouse domains, Mac-IAC can bypass an obvious local-optima that Mac-Dec-Q falls into, where the robot arm greedily chooses \emph{\textbf{Wait-M}} to avoid more penalties.

\begin{figure}[t!]
    \centering
    \captionsetup[subfigure]{labelformat=empty}
    \centering
    \subcaptionbox{}
        [0.9\linewidth]{\includegraphics[scale=0.26]{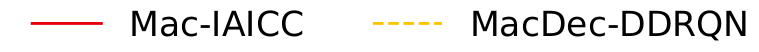}}
    ~
    \centering
    \subcaptionbox{}
        [0.31\linewidth]{\includegraphics[height=4cm]{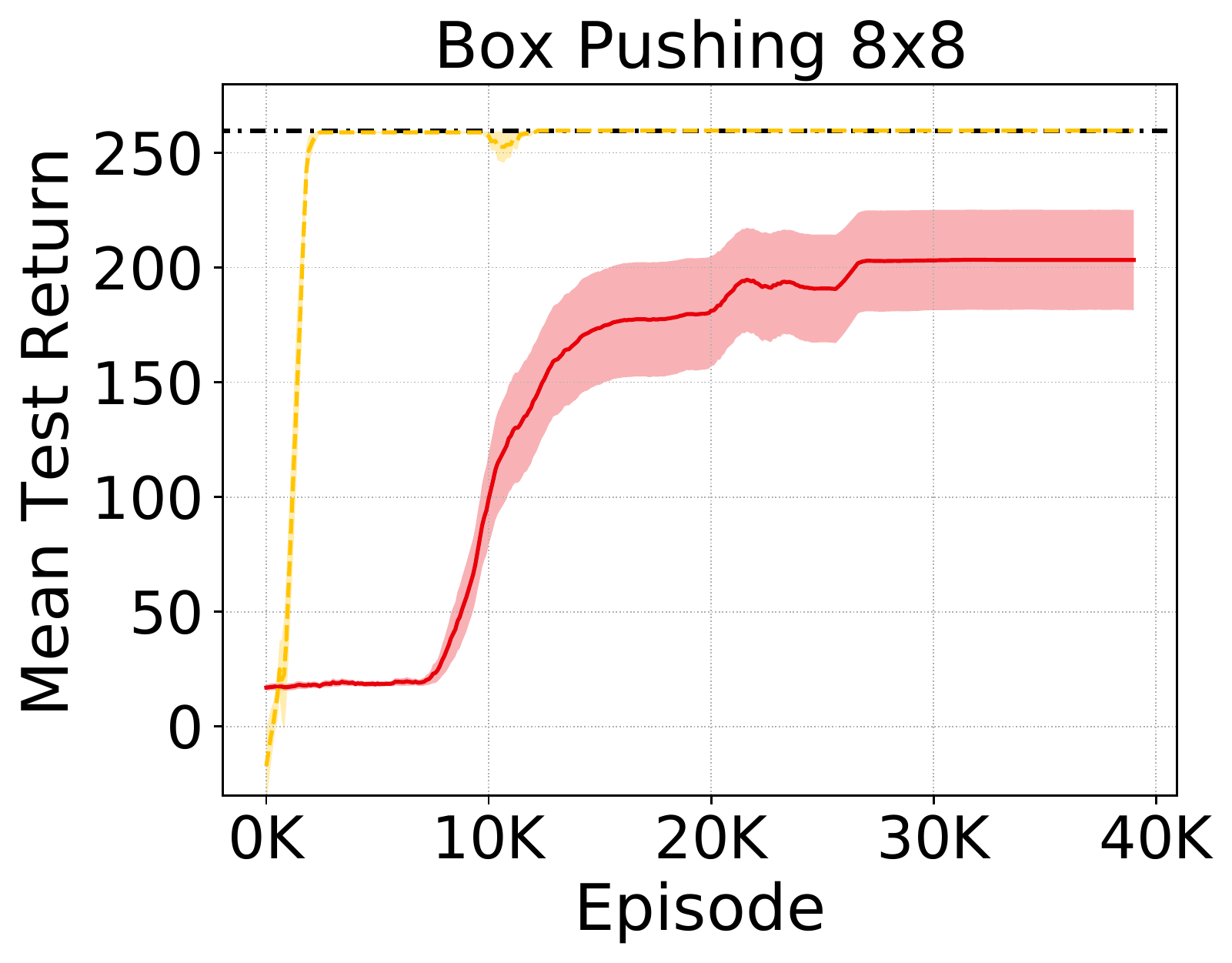}\vspace{-3mm}}
    ~
    \centering
    \subcaptionbox{}
        [0.31\linewidth]{\includegraphics[height=4cm]{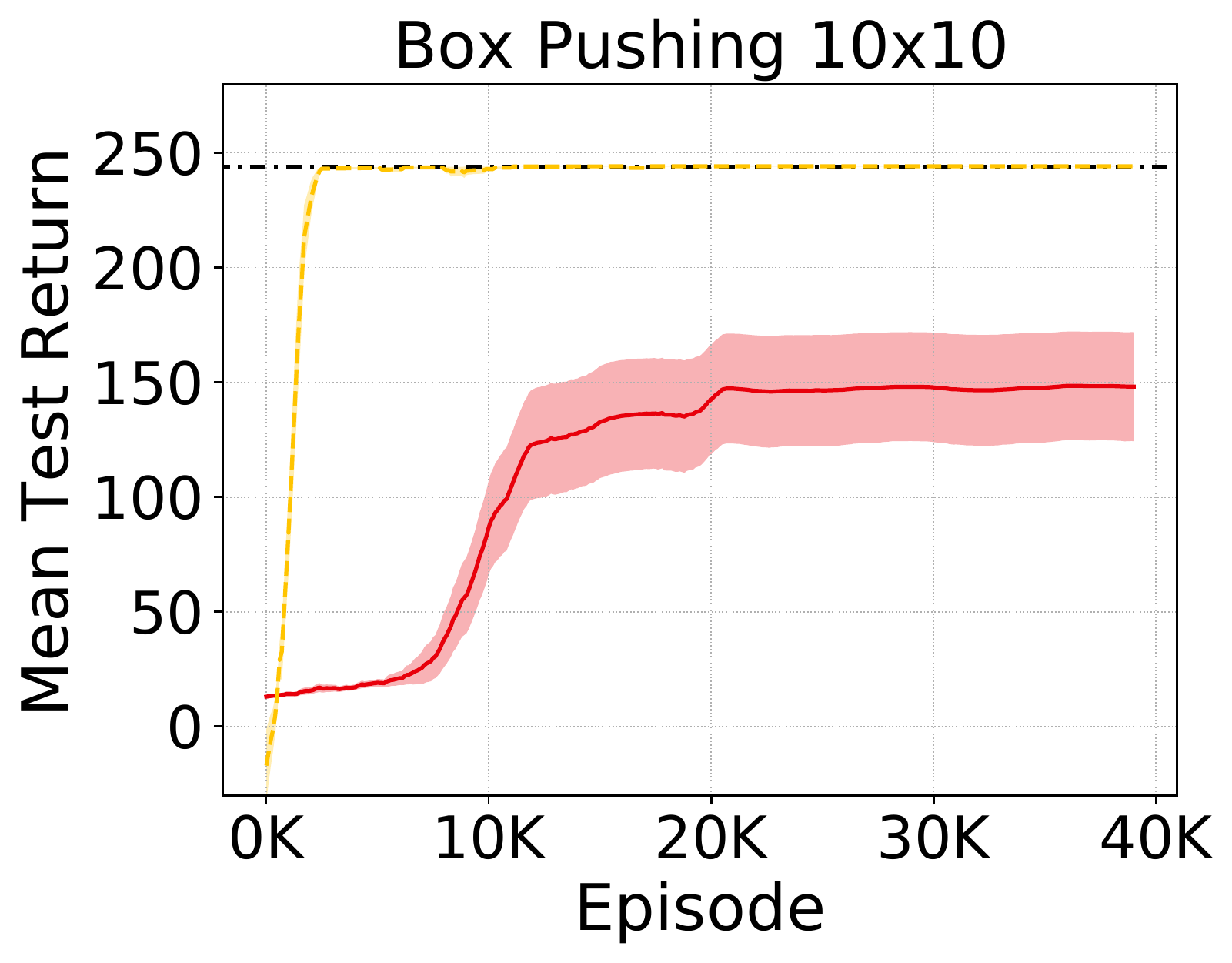}\vspace{-3mm}}
     ~
     \centering
    \captionsetup[subfigure]{labelformat=empty}
    \subcaptionbox{}
        [0.31\linewidth]{\includegraphics[height=4cm]{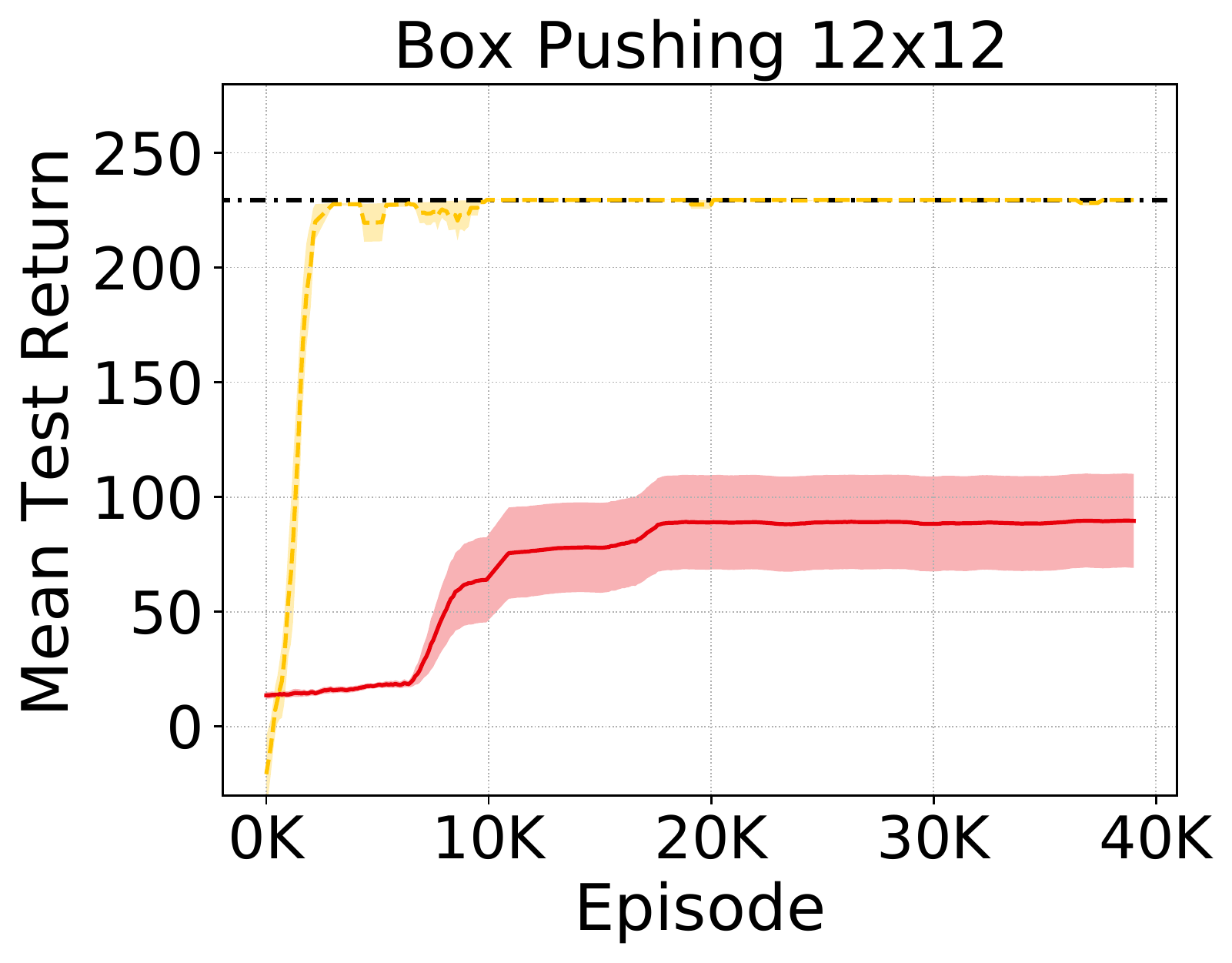}\vspace{-3mm}}
    ~
    \centering
    \subcaptionbox{}
        [0.4\linewidth]{\includegraphics[height=4cm]{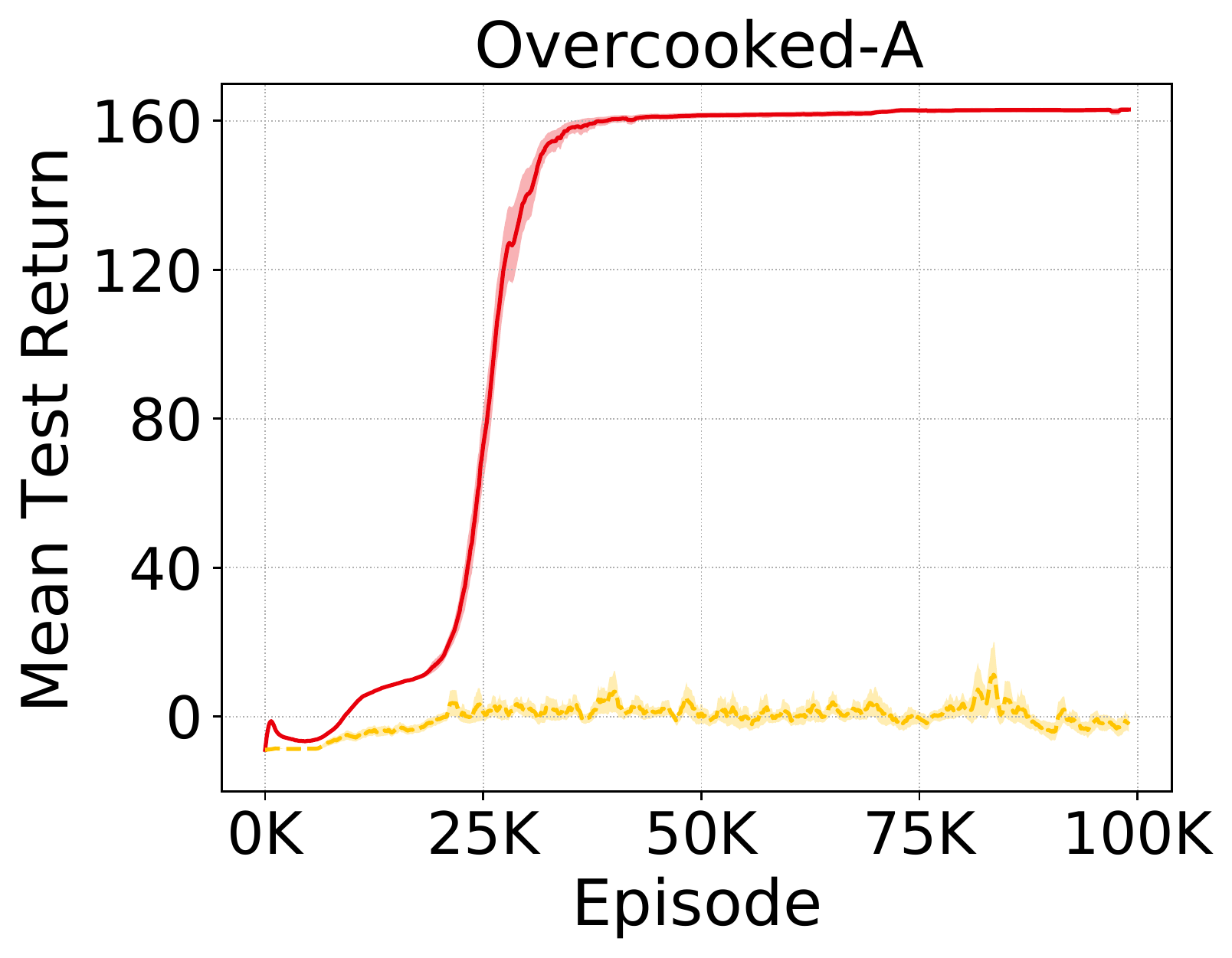}\vspace{-3mm}}
    ~
    \centering
    \subcaptionbox{}
        [0.4\linewidth]{\includegraphics[height=4cm]{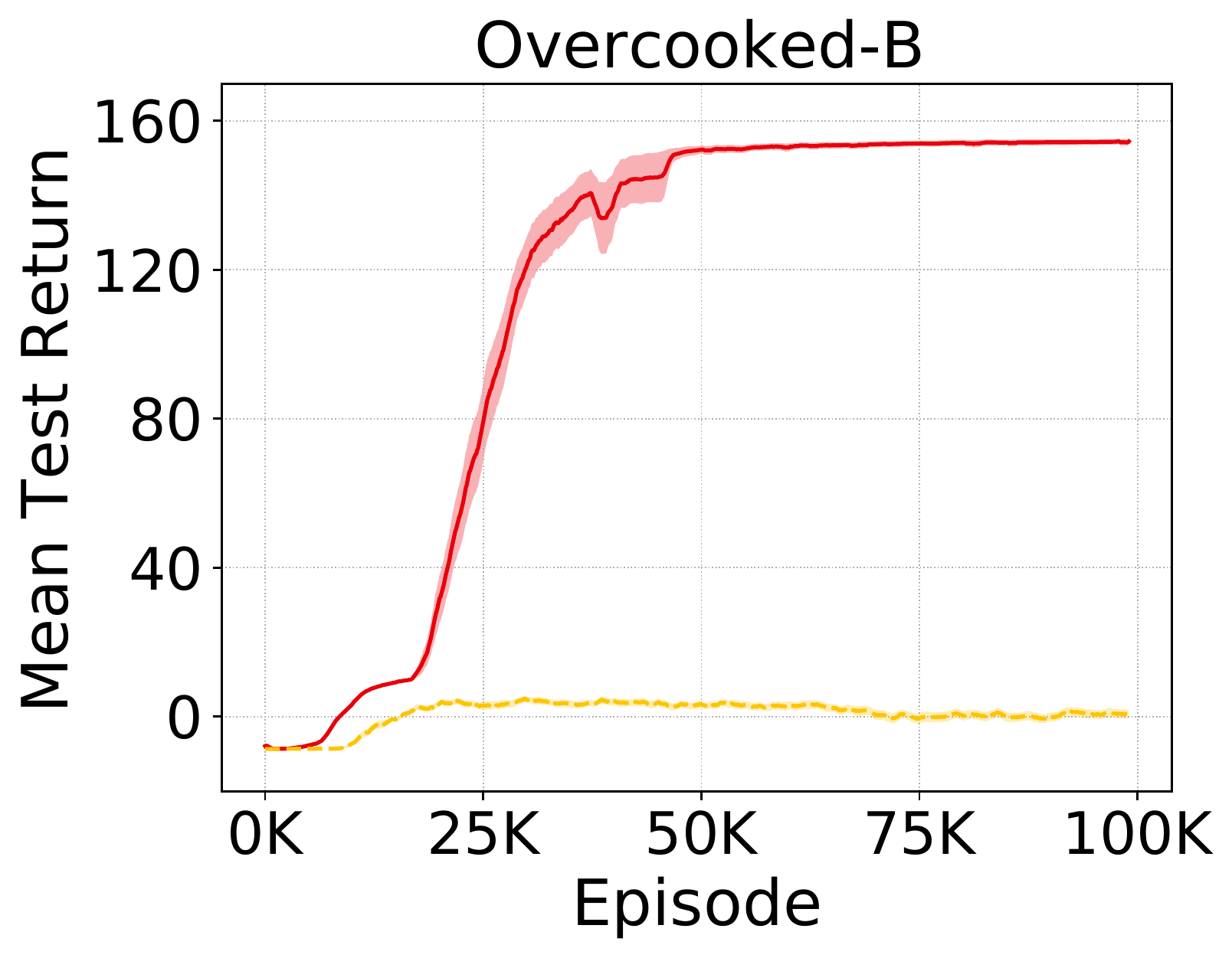}\vspace{-3mm}}
    \caption{Comparisons of Mac-IAICC and MacDec-DDRQN.}
    \label{ctde_qvspg_1}
\end{figure}

\begin{figure}[t!]
    \centering
    \captionsetup[subfigure]{labelformat=empty}
    \centering
    \subcaptionbox{}
        [0.9\linewidth]{\includegraphics[scale=0.26]{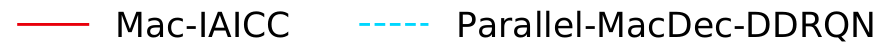}}
    ~
    \centering
    \subcaptionbox{}
        [0.31\linewidth]{\includegraphics[height=4cm]{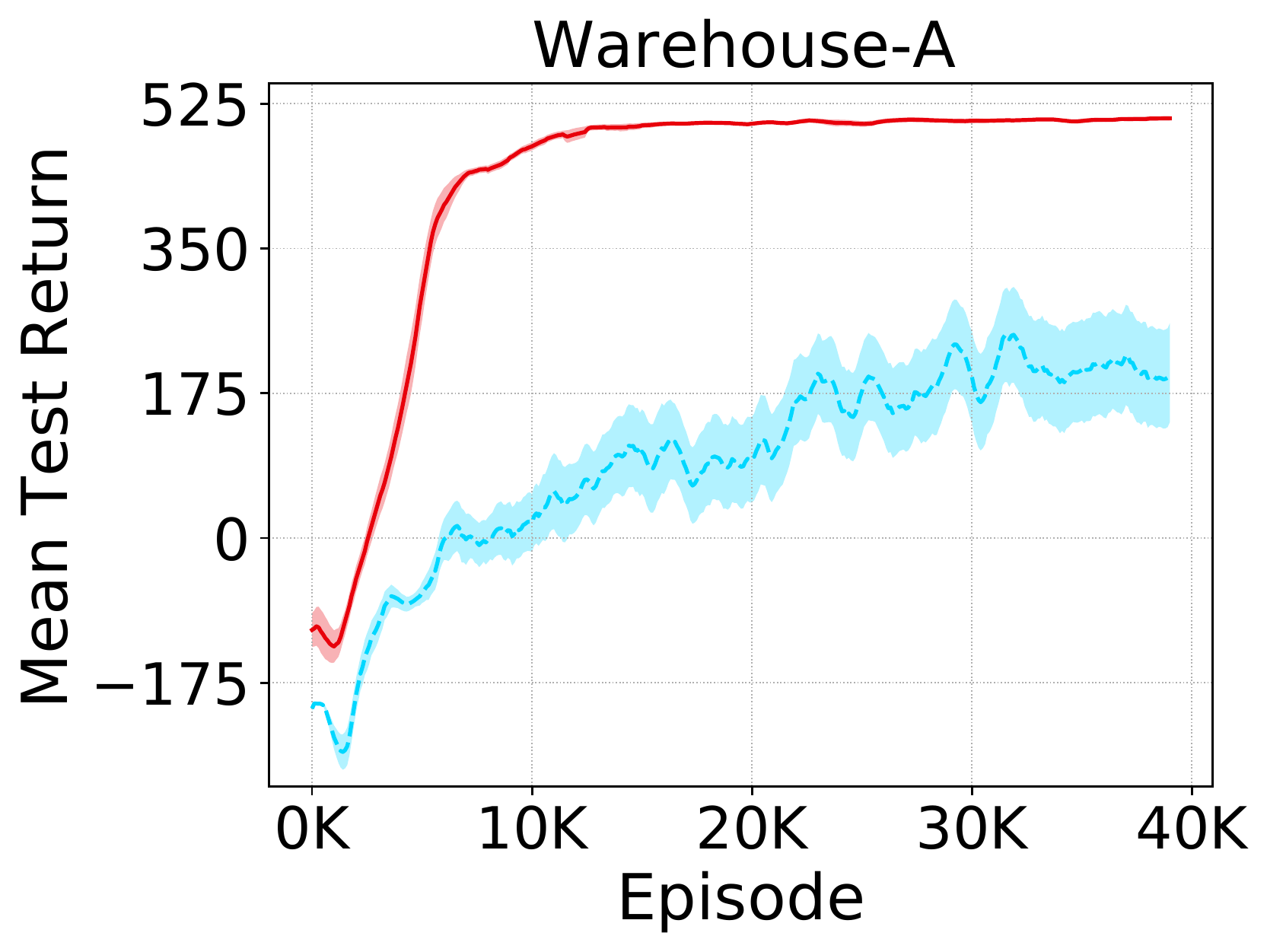}\vspace{-3mm}}
    ~
    \centering
    \subcaptionbox{}
        [0.31\linewidth]{\includegraphics[height=4cm]{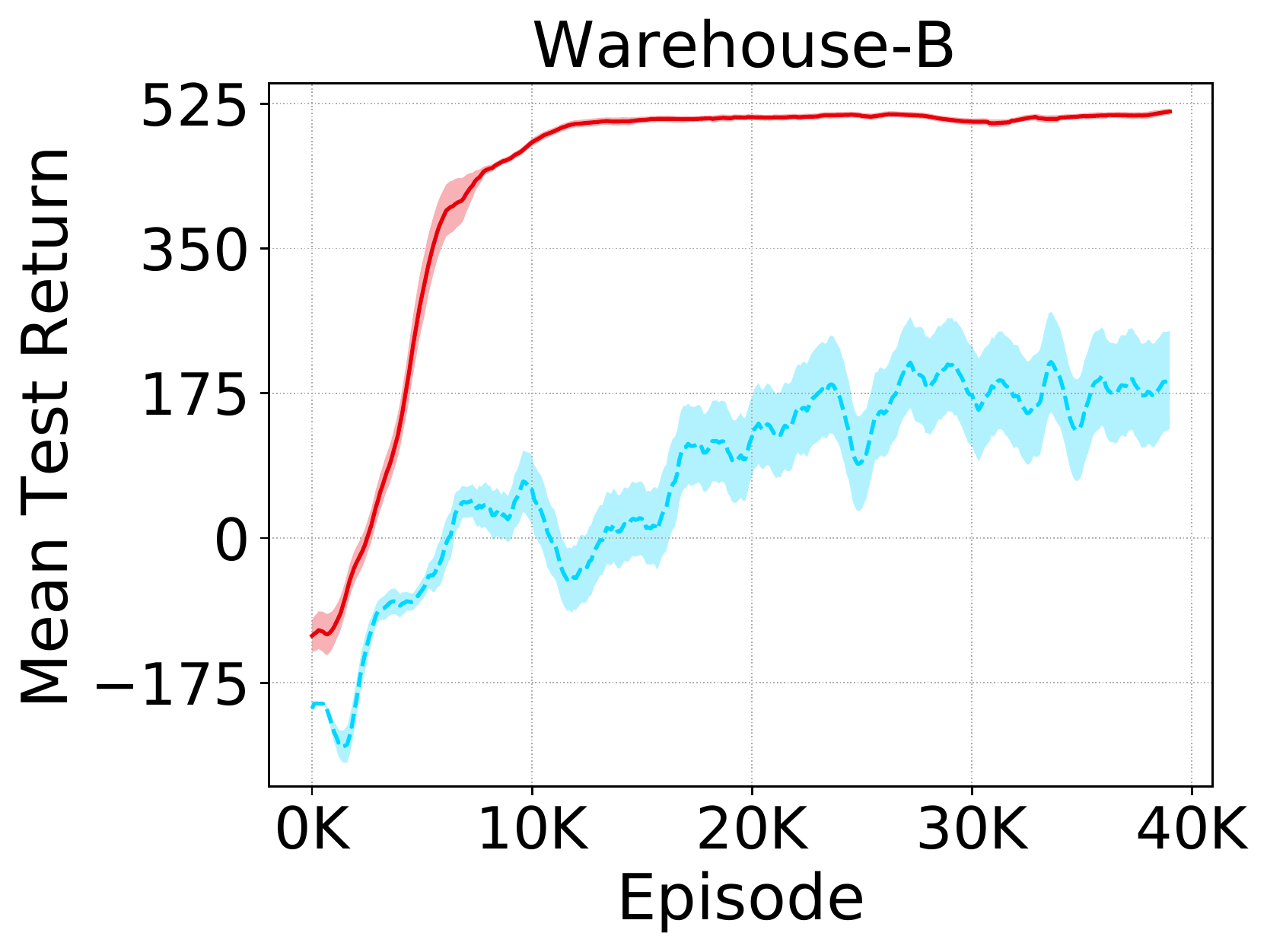}\vspace{-3mm}}
    ~
    \centering
    \subcaptionbox{}
        [0.31\linewidth]{\includegraphics[height=4cm]{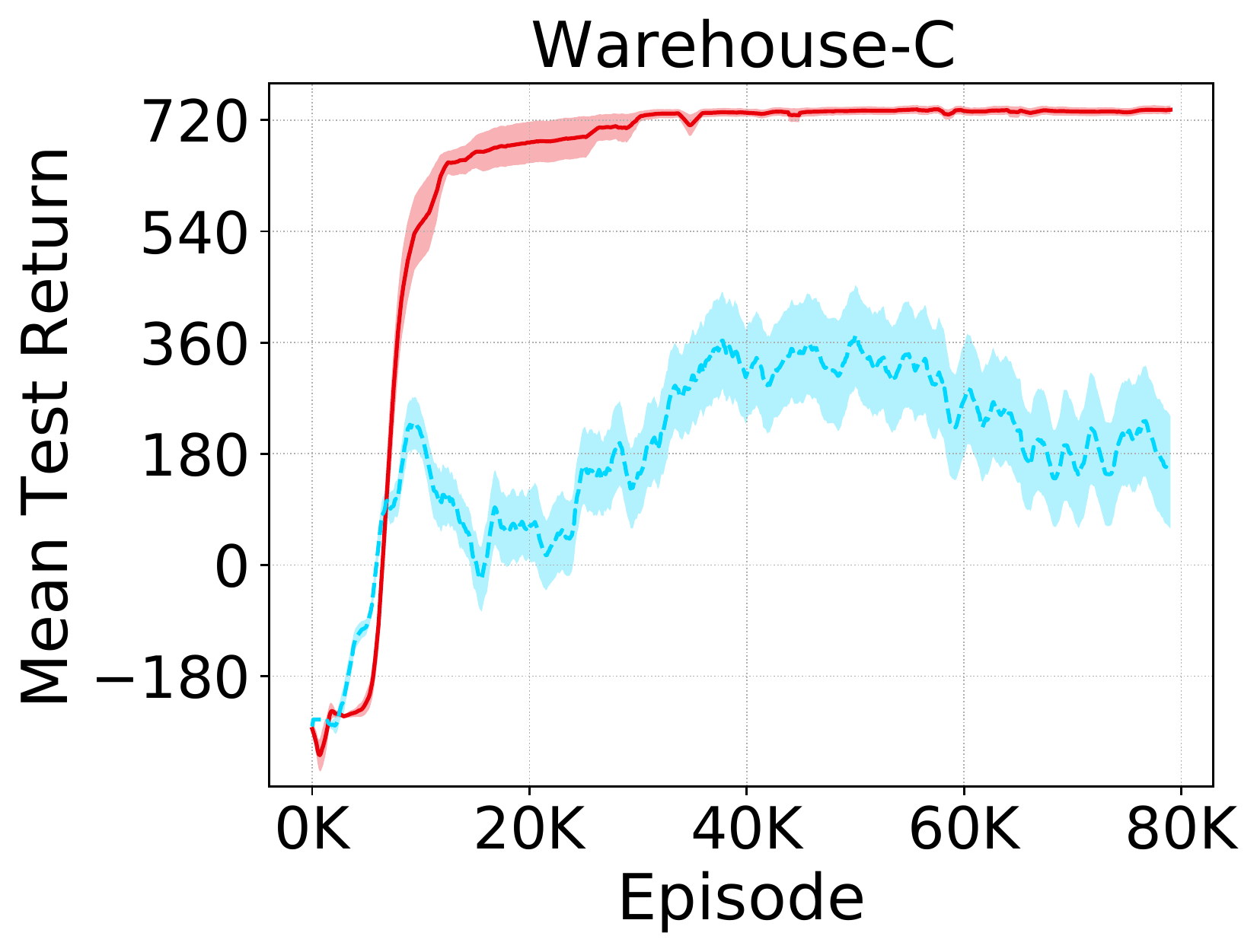}\vspace{-3mm}}
    \caption{Comparisons of Mac-IAICC and Parallel-MacDec-DDRQN.}
    \label{ctde_qvspg_2}
\end{figure}

\textbf{Comparative analysis between actor-critic and value-based approaches in CTDE paradigm}. 
We also conduct comparisons between our CTDE-based actor-critic method (Mac-IAICC) and our CTDE-base Q-learning methods (MacDec-DDRQN and Parallel-MacDec-DDRQN) proposed in Chapter~\ref{chap:paper2}. 
In the Box Pushing task, we train agents using MacDec-DDRQN with a centralized $\epsilon$-greedy policy for exploration since the cooperation for pushing the big box together is more likely to be generated from the centralized perspective.
Meanwhile, taking advantage of a replay buffer, MacDec-DDRQN learns much faster than Mac-IAICC (shown in the top row in Fig~\ref{ctde_qvspg_1}). 
Although Mac-IAICC possesses a centralized critic to provide a global action-value estimation, its overall performance is still limited by the on-policy data generated only in decentralized execution, where the bigger the world is, the lower the probability for sampling the aforementioned cooperation would be.         
In the Overcooked domain, we conduct one experiment to train agents using MacDec-DDRQN with a decentralized $\epsilon$-greedy policy and the other one with a centralized $\epsilon$-greedy policy. 
They both cannot learn any good behaviors, and in Fig.~\ref{ctde_qvspg_1}, we show the slightly better one achieved by using decentralized exploration, which makes sense as fully decentralized learning has solved this problem very well (as shown in Fig.~\ref{pg_vs_ac}). 
However, the centralized Q-function in MacDec-DDRQN is learned quite slowly due to the huge joint macro-action space, and becomes the bottleneck such that it cannot offer a good target action for optimizing decentralized action-value functions and hurts the learning (the bottom row in Fig.~\ref{ctde_qvspg_1}).  
Mac-IAICC successfully avoids the dilemma of the exponential joint action space by letting each agent learn an individual joint history-value function as the critic. 
As the ablation study in Section~\ref{chap:paper2:sim:re} has proved the necessity of having parallel environments to learn different Q-value functions in solving the warehouse task, we run Parallel-MacDec-DDRQN in three larger warehouse scenarios and show the comparison with Mac-IAICC in Fig~\ref{ctde_qvspg_2}. 
Considering the results of Mac-Dec-Q (purple curve) shown in Fig.~\ref{pg_vs_ac}, we can conclude that the centralized Q-value function involved in Parallel-MacDec-DDRQN prevents the decentralized policies from a very bad local-optimum. 
But eventually, the learned decentralized policies converge to another local-optima. 
We suspect the way of using the centralized Q-net to optimize decentralized policies (as in Eq.~\ref{ctde_Condi}) limits the improvement. 
One hypothesis is that the target action suggested by the centralized Q-net conditioning on joint information actually cannot always be reproduced by the decentralized Q-nets conditioning on only local information. 
Another hypothesis is that, besides having separate environments to train decentralized and centralized Q-nets, it is necessary to have separate hyper-parameters for them. 
A set of hyper-parameters may be good for the centralized learner while not fitting the optimization for decentralized learners well.
As a result, even though the centralized learner can suggest good target actions, the updates for decentralized learners still get trapped into a local-optimum due to the hyper-parameters. 
On the other hand, if the hyper-parameter tuning is purely based on the performance of decentralized learners, the convergence to a local-optimum can be caused by the poor centralized learner trained with the same set of hyper-parameters.
Finally, Mac-IAICC's leading performance over three scenarios further demonstrates its strong scalability to large and long-horizon problems.

\section{Hardware Experiments}
\label{chap:paper3:hw}

\subsection{Experimental Setup}
\label{chap:paper3:hw:setup}

\begin{figure}[h!]
    \centering
    \includegraphics[scale=1.4]{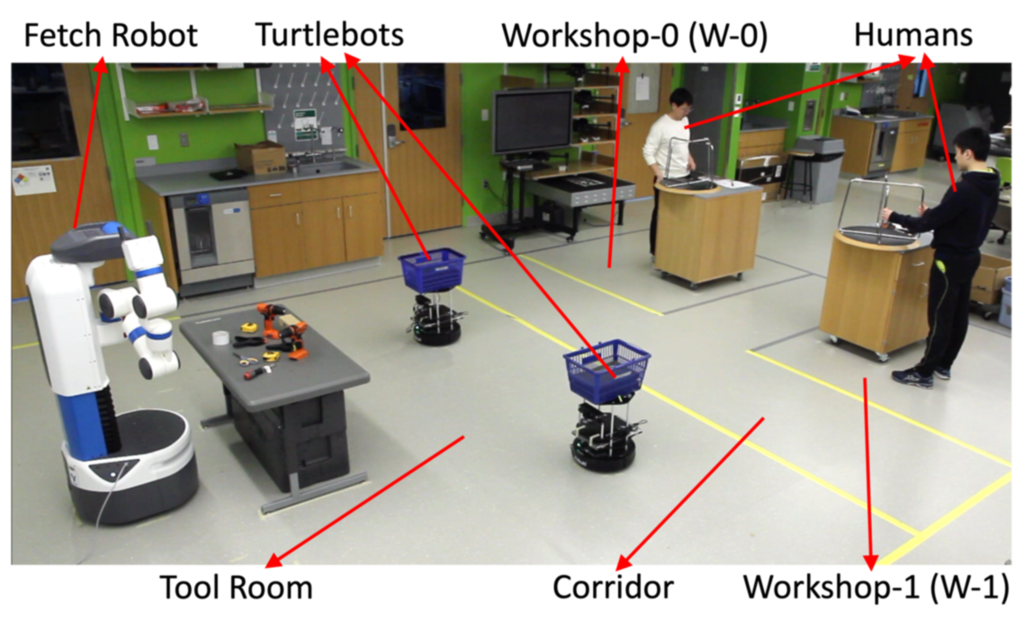}
    \caption{Overview of Warehouse-A hardware domain.}
    \label{twohuman_hd_domain}
\end{figure}

\noindent While simulation results validate that the proposed Mac-IAICC approach achieves the best performance for learning decentralized policies in various macro-action-based domains, we also extend scenario A of the Warehouse Tool Delivery task to a hardware domain. Fig.~\ref{twohuman_hd_domain} provides an overview of the real-world experimental setup. An open area is divided into regions, a tool room, a corridor, and two workshops, to resemble the configuration shown in Fig.~\ref{domain_wtdA}. This mission involves one Fetch Robot~\cite{FetchRobot} and two Turtlebots~\cite{Turtlebot} to cooperatively find and deliver three YCB tools~\cite{YCB}, in the order: a tape measure, a clamp, and an electric drill, required by each human in order to assemble an IKEA table. 

The Turtlebot's navigation macro-actions were executed by using the ROS navigation stack~\cite{ROSNavigation}. For Fetch's manipulation macro-actions, we combined PCL bindings for Python~\cite{PCLPython}, MoveIt~\cite{MoveIt}, and the OpenRave simulator~\cite{Diankov:R} with an OMPL~\cite{OMPL} plugin to achieve the picking and placing of tools. The information about the number of tools in staging areas and each human's working status was tracked and broadcast by ROS services but were only observable in the tool room and the corresponding workshop area respectively (to simulate possible visual information).

\subsection{Results}
\label{chap:paper3:hw:re}

\begin{figure*}[h!]
    \centering
    \captionsetup[subfigure]{labelformat=empty}
    \centering
    \subcaptionbox{(a) Fetch finds and stages a tape measure while Turtlebots reach W-1 and observe the human's state.\label{real_a}}
        [0.45\linewidth]{\includegraphics[height=3.6cm]{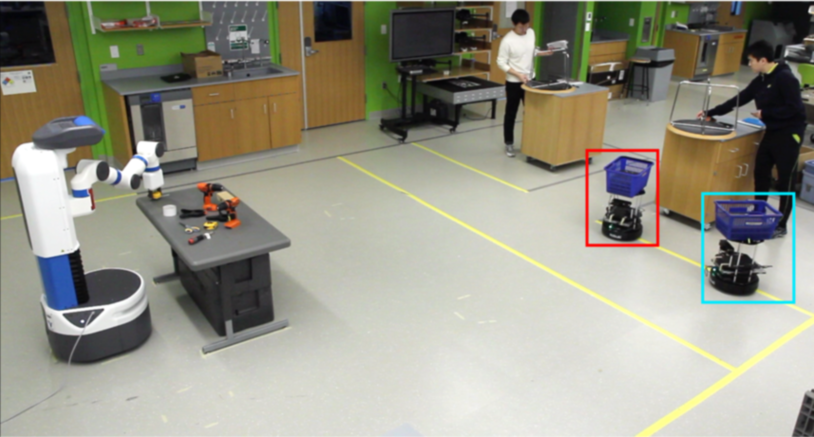}}
    ~
    \centering
    \subcaptionbox{(b) As no Turtlebot beside, Fetch looks for the 2nd tape measure. Meanwhile, Turtlebots approach the table.\label{real_b}}
        [0.45\linewidth]{\includegraphics[height=3.6cm]{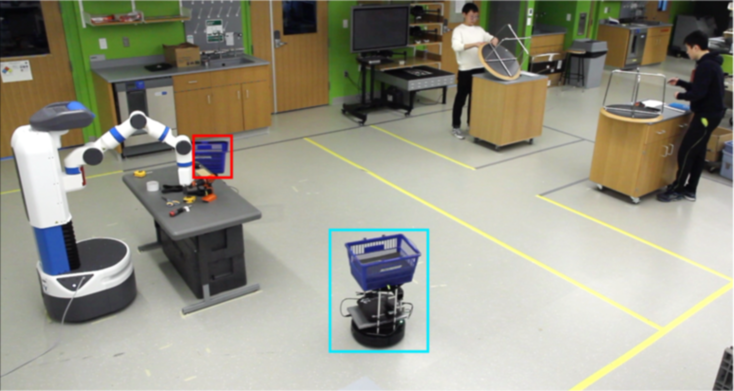}}
    ~
    \centering
    \subcaptionbox{(c) T-1 delivers a tap measure to W-1 and T-0 carries the other tape measure, while Fetch retrievals a clamp.\label{real_e}}
        [0.45\linewidth]{\includegraphics[height=3.6cm]{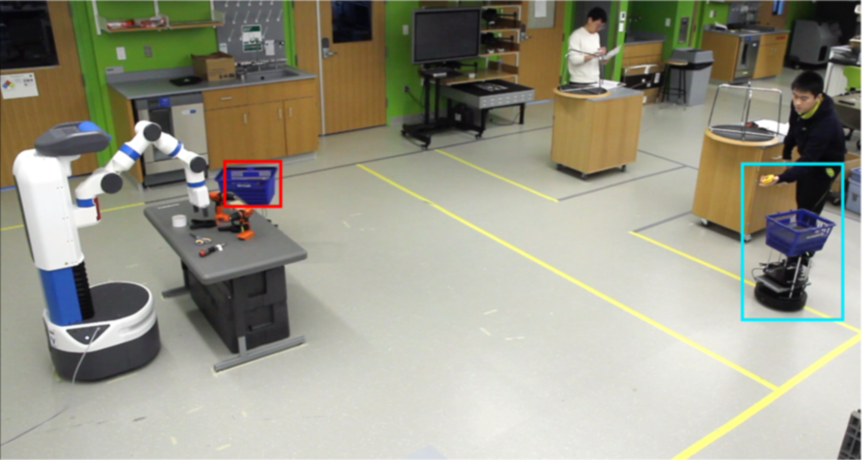}}
    ~
    \centering
    \subcaptionbox{(d) Fetch goes for the 2nd clamp as no Turtlebot ready beside and T-0 moves to W-0, while T-1 approaches the table.\label{real_f}}
        [0.45\linewidth]{\includegraphics[height=3.6cm]{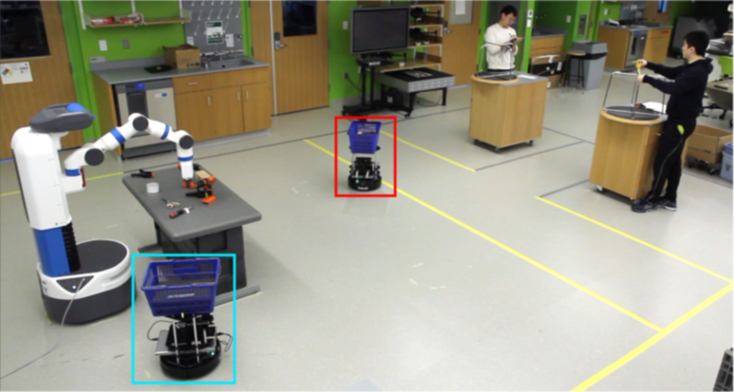}}
    ~
    \centering
    \subcaptionbox{(e) T-0 delivers a tap measure to W-0 and T-1 waits beside the table, while Fetch stages the 2nd clamp.\label{real_g}}
        [0.45\linewidth]{\includegraphics[height=3.6cm]{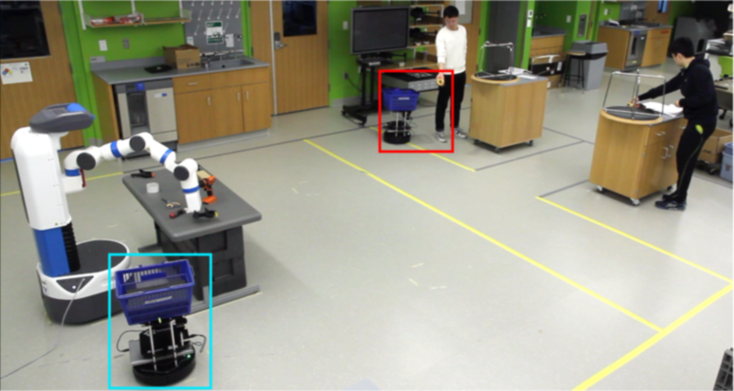}}
    ~
    \centering
    \subcaptionbox{(f) Fetch passes a clamp to T-1, while T-0 just arrives at the table. \label{real_h}}
        [0.45\linewidth]{\includegraphics[height=3.6cm]{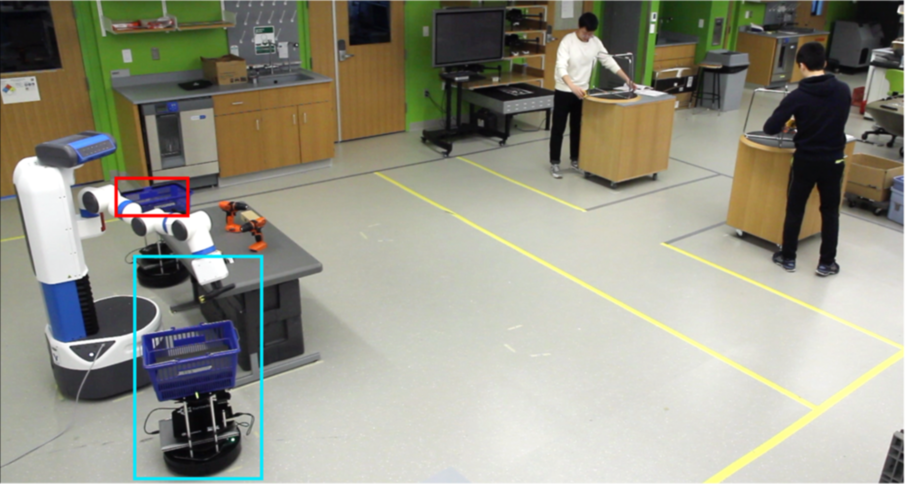}}
    ~

    \caption{Robots' sequential collaborative behaviors generated by running the decentralized policies learned by Mac-IAICC in a warehouse domain, where Turtlebot-0 (T-0) is bounded in red and Turtlebot-1 (T-1) is bounded in blue.}
    \label{real_exp-1}
\end{figure*}

\begin{figure*}[t!]
    \centering
    \captionsetup[subfigure]{labelformat=empty}
    \centering
    \subcaptionbox{(a) T-1 delivers a clamp to W-1, while T-0 carries the other clamp and goes to W-0. Fetch searches for an electric drill.\label{real_j}}
        [0.45\linewidth]{\includegraphics[height=3.6cm]{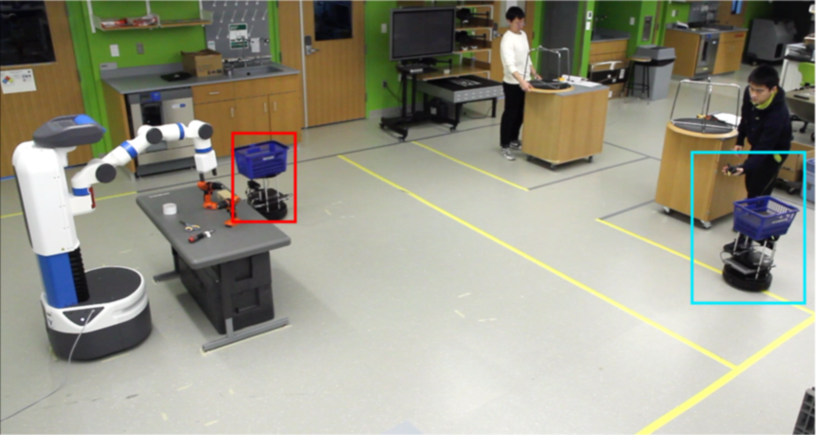}}
    ~
    \centering
    \subcaptionbox{(b) T-0 delivers a clamp to W-0 and T-1 returns tool room, while Fetch finishes staging an electric drill and notices no teammate around the table yet.\label{real_k}}
        [0.45\linewidth]{\includegraphics[height=3.6cm]{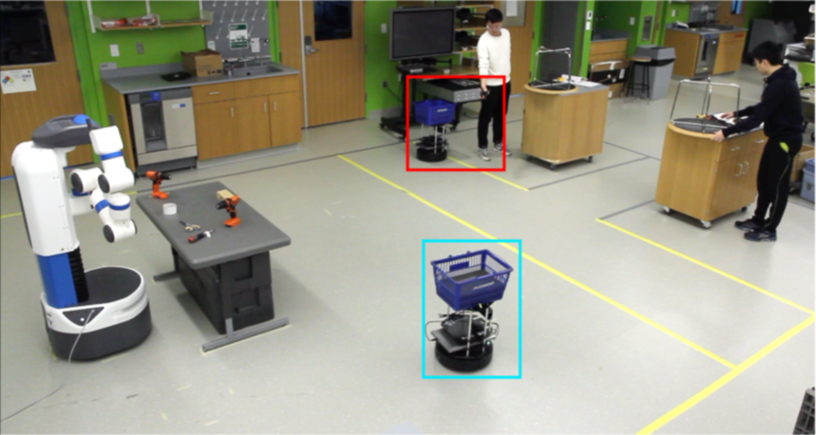}}
    ~
    \centering
    \subcaptionbox{(c) Fetch continues to look for the 2nd electric drill while T-1 moves close to the table.\label{real_l}} 
        [0.45\linewidth]{\includegraphics[height=3.6cm]{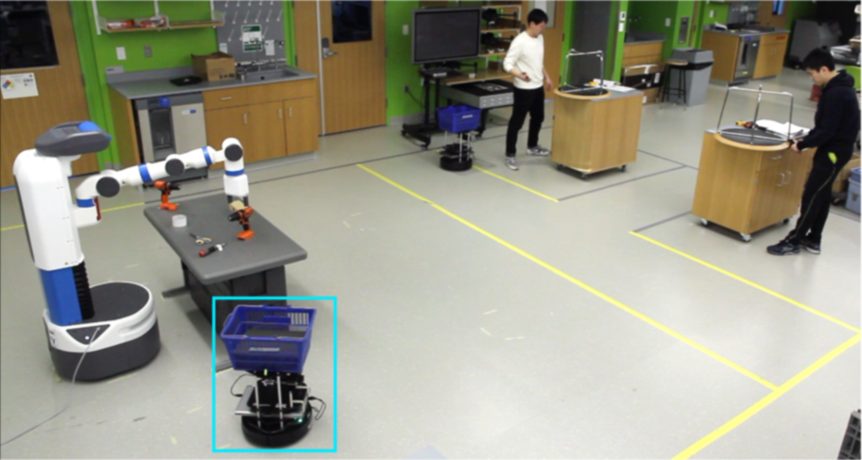}}
     ~
    \centering
    \subcaptionbox{(d) Fetch passes an electric drill to T-1, while T-0 comes back.\label{real_m}}
        [0.45\linewidth]{\includegraphics[height=3.6cm]{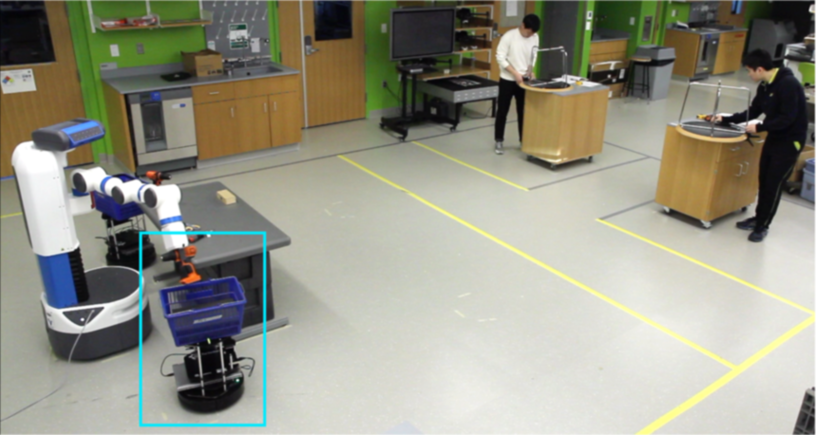}}
     ~
    \centering
    \subcaptionbox{(e) T-1 delivers an electric drill to W-1, while Fetch finishes passing the other electric drill to T-0.\label{real_n}}
        [0.45\linewidth]{\includegraphics[height=3.6cm]{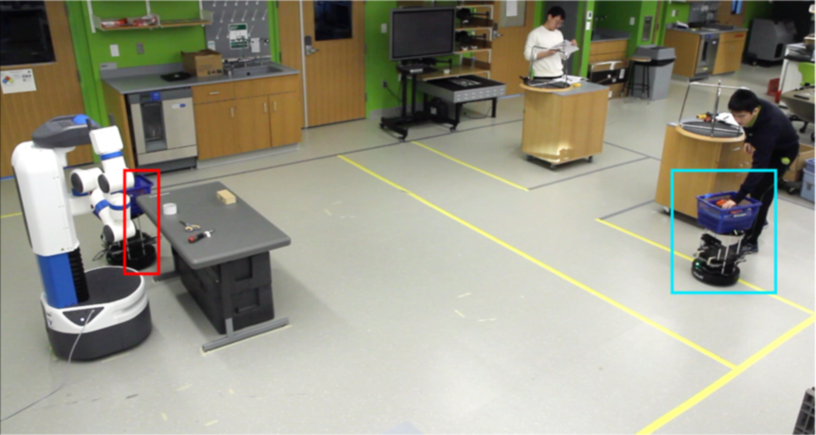}}
    ~    
    \centering
    \subcaptionbox{(f) T-0 delivers an electric drill (the last tool) to W-0 and the entire delivery task is completed.\label{real_o}}
        [0.45\linewidth]{\includegraphics[height=3.6cm]{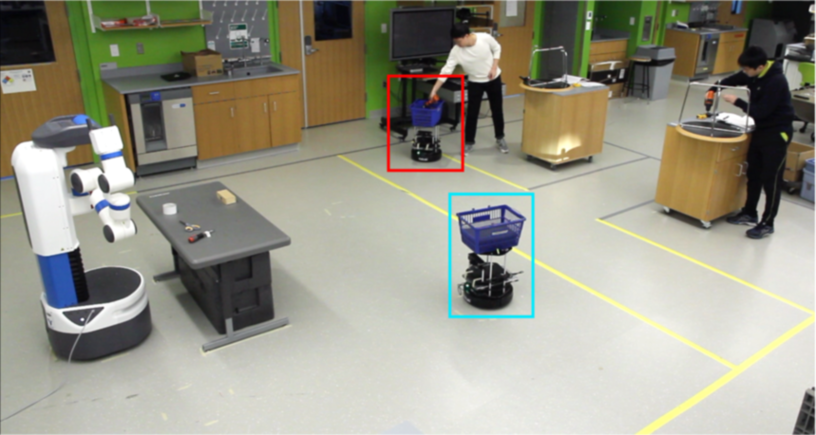}}
    \caption{Robots' sequential collaborative behaviors generated by running the decentralized policies learned by Mac-IAICC in a warehouse domain, where Turtlebot-0 (T-0) is bounded in red and Turtlebot-1 (T-1) is bounded in blue.}
    \label{real_exp-2}
\end{figure*}

\noindent Fig.~\ref{real_exp-1} and Fig.~\ref{real_exp-2} show the sequential collaborative behaviors of the robots in one hardware trial, where the robots successfully reasoned about the correct tools needed by each human and efficiently delivered them in the proper order. 
More concretely, the Fetch was smart to find tools in parallel such that two tape measures (Fig.~\ref{real_a} - \ref{real_b}), two clamps (Fig.~\ref{real_e} - \ref{real_g}), and two electric drills (Fig.~\ref{real_j} - \ref{real_l}), were found instead of finding all three types of tool for one human and then moving on to the other which would result in one of the humans waiting. 
Furthermore, the Fetch's efficiency is also reflected in the behaviors such that it continued to find the next tool when there was no Turtlebot waiting beside it (Fig.~\ref{real_b}, \ref{real_f} and \ref{real_l}) and immediately passed a tool to the Turtelbot that was ready beside it (Fig.~\ref{real_h} and \ref{real_m}). Meanwhile, Turtlebots (T-0 and T-1) performed efficient delivery  that successfully avoids delayed delivery by sending tools one by one to the nearby workshop respectively (e.g., T-0 focused on W-0 shown in Fig.~\ref{real_g}, \ref{real_k} and \ref{real_o}, and T-1 focused on W-1 shown in Fig.~\ref{real_e}, \ref{real_j} and \ref{real_n}), rather than waiting for all tools before delivering, traveling a longer distance to serve the human at the diagonal, or prioritizing one of the humans altogether.

\section{Conclusion}
\label{chap:paper3:con}

\noindent In this chapter, we introduce the first general formulation for asynchronous multi-agent macro-action-based policy gradients under partial observability along with proposing a decentralized actor-critic method (Mac-IAC), a centralized actor-critic method (Mac-CAC), and two CTDE-based actor-critic methods (Naive Mac-IACC and Mac-IAICC). 
These are the first approaches to be able to incorporate robot controllers that may require different amounts of time to complete (macro-actions) in a general asynchronous multi-agent actor-critic framework. 
Empirically, our methods are able to learn high-quality macro-action-based policies allowing agents to perform asynchronous collaborations in large and long-horizon problems. 
Importantly, our most advanced method, Mac-IAICC, shows the strength of  allowing agents to have individual centralized critics tailored to the agent's own macro-action executions. 
Additionally, the practicality of our approach is validated  in a real-world multi-robot setup based on a warehouse domain.  


\chapter{Conclusion}
\label{chap:conclude}

\noindent Realistic multi-agent problems are often long-horizon with large state and observation spaces and various uncertainties, in which agents are naturally required to be able to asynchronously operate without waiting for each other to terminate, and are expected to be capable of making decisions in a hierarchical fashion with different levels of abstraction.   
Asynchronous and hierarchical multi-agent reinforcement learning under partial observability thus becomes an essential and promising research topic for providing practical solution methods to real-world multi-agent systems.
Along this line of research, in this thesis, we formulated the first set of macro-action-based model-free deep MARL frameworks involving both value-based algorithms and actor-critic policy gradient algorithms. 
In this chapter, we begin by discussing our contributions and then describe some remaining open challenges for future work.

\section{Summary of Contributions}

\noindent In this thesis, we consider fully cooperative multi-agent scenarios with a set of predefined macro-actions for each agent, where each macro-action can be either hand-coded based on prior domain knowledge or obtained via single-agent learning or planning approaches. 
The objective of all proposed methods then focuses on training high-level policies over macro-actions.
As macro-actions naturally require multiple time steps to complete, multi-agent decision-making with macro-actions is allowed to be asynchronous.  
Such asynchronicity exactly matches the nature of real-world multi-agent behavior, but it also raises the key challenge about when to perform updates and what information to maintain in MARL with macro-actions for different training purposes.   
Prior to our work, existing deep MARL frameworks cannot directly work with macro-actions since they rely on synchronous learning and execution across all agents.

To address the above challenge, in Chapter~\ref{chap:paper1}, we first developed principled methods to extend deep Q-net for learning decentralized and centralized macro-action-value functions.
In the decentralized case, we designed a new replay-buffer, called Macro-Action Concurrent Experience Replay Trajectories (Mac-CERTs), to allow agents currently collect high-level information (macro-action and macro-observation), and meanwhile, each agent independently accumulates a reward depending on its own macro-action execution status and the global reward signal. 
In the training, each agent only accesses its own experiences and squeezes them by removing repeated information in the duration of each macro-action. 
As a result, each agent's macro-action-value function updates only take place at the completion time steps of its own macro-actions.
Asynchronous learning over agents in this decentralized case is attained. 
In the centralized case, agents use the other new replay buffer, named Macro-Action Joint Experience Replay Trajectories (Mac-JERTs), to maintain joint high-level information and a joint cumulative reward for each joint macro-action at every time step.   
Centralized macro-action-value function updates are performed only at the time step when \emph{any} agent finishes its macro-action (defined as the termination of a joint macro-action).
Importantly, we proposed a \emph{conditional target-value prediction} in the TD loss for centralized learning in order to correctly capture agents' asynchronous macro-action execution, which obeyed the fact that only the agent who had terminated the current macro-actions could switch to a new one at next time step.   
These two frameworks build up the base for developing MARL algorithm with macro-actions, and accept any variants of DQN to learn decentralized and centralized macro-action-value functions.  

In order to learn better decentralized macro-action-based policies to solve complex tasks, in Chapter~\ref{chap:paper2}, we introduced the first CTDE-based MARL algorithm with macro-actions, called Macro-Action Decentralized Double Deep Recurrent Q-Network (MacDec-DDRQN). 
The key idea of MacDec-DDRQN is to train each agent's decentralized macro-action value function by querying a centralized macro-action-value function for target-action selection in TD updates. 
This particular adaptation of single-agent double Q-learning to multi-agent settings potentially reduces the hurt of environmental non-stationarity and promotes the decentralized behavior to be as cooperative as the centralized one. 
We also presented a Parallel-MacDec-DDRQN that ensures the centralized macro-action-value function is well trained using purely centralized data and then updates each decentralized macro-action-value function purely based on decentralized data but obtaining heuristic from the centralized one.  
It is important to note that, in practice, neither of these two versions always works better than the other. 
In the domains where the optimal centralized behavior can be achieved in a decentralized manner but pivotal cooperative choices are rare in decentralized data, we suggest using MacDec-DDRQN with centralized exploration; and the Parallel version may better fit the cases where the decentralized optimization has to rely on realistic decentralized data but requiring centralized guidance to avoid local-optimums. 

Finally, in Chapter~\ref{chap:paper3}, we formulated a set of macro-action-based actor-critic algorithms that allow agents to asynchronously optimize parameterized policies via policy gradients.  
We first described a macro-action-based independent actor-critic (Mac-IAC) algorithm that adapts the decentralized value-based approach to learn a local history value function as the critic for each agent and then updates each agent's macro-action-based policy via decentralized policy gradients when its macro-action terminates. 
Secondly, we presented a macro-action-based centralized actor-critic (Mac-CAC) algorithm that adapts the centralized value-based approach to learn a joint history value function as the centralized critic and then optimizes a parameterized centralized policy over joint macro-actions via centralized policy gradients. 
Lastly, we showed the technical flaws of directly incorporating macro-actions into the primitive-action-based independent actor with a centralized critic framework (referred to as Naive IACC), and addressed the corresponding issues by proposing a macro-action-based independent actor with individual centralized critic (Mac-IAICC). 
In Mac-IAICC, agents are allowed to train an individual centralized critic to obtain a more accurate value estimation with respect to each own macro-actions with much less noise than Naive IACC.  
We demonstrated Mac-IAICC is scalable to long horizons and able to generate high-quality asynchronous solutions for large multi-agent domains both in simulation and on hardware.  

Our formalism and methods open the door for other macro-action-based multi-agent reinforcement learning methods ranging from extensions of other current methods to new approaches and domains. 
We expect even more scalable learning methods that are feasible and flexible enough in solving realistic multi-robot problems.

\section{Future Work}

\noindent \textbf{Continuous Macro-Action MARL}. All the proposed methods in this thesis are based on a discrete macro-action space. 
However, some problems can also involve continuous macro-actions. 
For example, in navigation related tasks, a macro-action can be parameterized as either a pair of position and orientation to reach or a group of Bezier points to determine a trajectory in a 2D/3D continuous space. 
However, so far, there has been no work considering continuous macro-actions in multi-agent settings.
Investigating a better way for macro-action parameterization as well as developing corresponding asynchronous MARL methods is thus an interesting future topic. 

\textbf{Macro-Action Value Factorization Networks}. Our macro-action-based CTDE Q-learning methods involve two major flaws: one is that the argmax over the joint macro-action space severely limits the scalability; and the other is that the methods may work poorly in the case where the centralized behavior is naturally not fully decentralizable.     
A promising way to tackle these two limitations is to learn decentralized macro-action-value functions in the way of value factorization. 
The question of how best to deal with agents' asynchronous macro-action execution through factorization networks remains open.

\textbf{Macro-Action Improvement}. The completed work in this thesis relies on an essential assumption that a set of macro-actions has been predefined. 
However, given a problem, the predefined macro-actions based on human knowledge may not be the best fit and also perhaps impede agents to learn the global optimal behavior. 
We expect principled methods that allow agents to leverage predefined macro-actions for fast learning while enabling agents to keep improving the macro-actions to generate better solutions.
This area is currently rarely studied in the context of having agents with asynchronous high-level decision-making, which is indeed an open challenge. 

\textbf{Asynchronous and Hierarchical Communication}. In this thesis, we do not consider any explicit communication. 
However, communication is an important skill for autonomous agents to efficiently collaborate in solving complex tasks with partial observability.
Communications can be part of the action space, and agents are potentially able to have various modes of communication. 
Decision-making on communications will be in terms of \emph{how}, \emph{when}, \emph{what} to communicate and \emph{who} to communicate with.
Imagine in human-in-the-loop multi-agent settings, autonomous agents can use a limited-bandwidth channel to talk to each other and communicate with humans via large-language models.  
Different means of communication may cost different amounts of time or energy.
Thus, asynchronous communication is also naturally requirement real-world multi-agent systems.
Furthermore, agents can communicate at different levels for different purposes, such as high-level communication for sub-task allocation over agents while low-level communication for better cooperation during execution. 
How to explicitly model communications in asynchronous and hierarchical MARL is an open question. 
Additionally, communication is likely an important mechanism to help with learning better termination functions in order to improve predefined macro-actions.

\textbf{Multi-Agent Multi-Level Decision-Making}. Decision-making in hierarchies has been demonstrated as a promising way to improve the scalability of solution methods and make intractable problems to be solvable.     
The state-of-the-art single-agent and multi-agent hierarchical RL methods mainly focus on two-level hierarchies.
However, as we have experienced, a remarkable characteristic of human cooperative behavior is the ability to make decisions across multiple levels of abstraction. 
Given a task to a team, coordinative decision-makings among team members first happen at the high-level over a set of abstractions, which produces a set of subtasks or mid-level skills that each, as an instance of an abstraction, is performed by either a single member alone or several members together via low-level skills that achieve a sequence of body movements through the control of muscle and joints. 
For example, when a team of chefs is preparing dishes according to orders, their collaboration may have a very clear subtasks allocation at the top (e.g., making a particular type of food, preparing ingredients, or decorating plates) but likely get more complex such as each chef switching between several subtasks when orders become crowded. 
At the next level, each subtask requires a chef to act a sequence of basic skills, such as chopping, pouring, stirring, shaking or moving, etc, where these skills are finally operated via human body control at the low-level. 
Hence, to better solve more complex and larger multi-agent problems, whether it is necessary to have a deeper hierarchical decision-making structure or not is an open-ended question. 
Accordingly, how to design and learn the different pieces in a multi-level decision-making architecture as well as considering more intricate asynchronous interaction over agents and environmental uncertainties is an important challenge. 

\textbf{Real-World Multi-Robot Applications}. Last but not least, it is important, appealing, and valuable to use the developed methods under the aforementioned topics to solve a wider range of realistic multi-robot tasks. 
Interesting domains include (but are not limited to): robots in sports, multi-robot service in offices, hospitals, and warehouses, kitchen robots, collaborative manipulation, and self-driving cars. 
Validating solution methods on real-world multi-robot systems will not only create tremendous value for making robots truly useful for human society but also can further inspire future research topics. 



\chapter*{Appendix}
\addcontentsline{toc}{chapter}{Appendix}  
\label{chap:appendix}

\section*{Behavior Visualization in Simulation}
\addcontentsline{toc}{section}{Behavior Visualization in Simulation}  

In this section, we display the decentralized behaviors learned by using Mac-IAICC under all considered domains.

\subsection*{Box Pushing}

\begin{figure*}[h!]
    \centering
    \captionsetup[subfigure]{labelformat=empty}
    \centering
    \subcaptionbox{(a) Green robot executes \emph{\textbf{Move-to-big-box(1)}} to move to the left waypoint below the big box while the blue robot runs \emph{\textbf{Move-to-big-box(2)}} to move to the right waypoint below the big box.}
        [0.30\linewidth]{\includegraphics[scale=0.13]{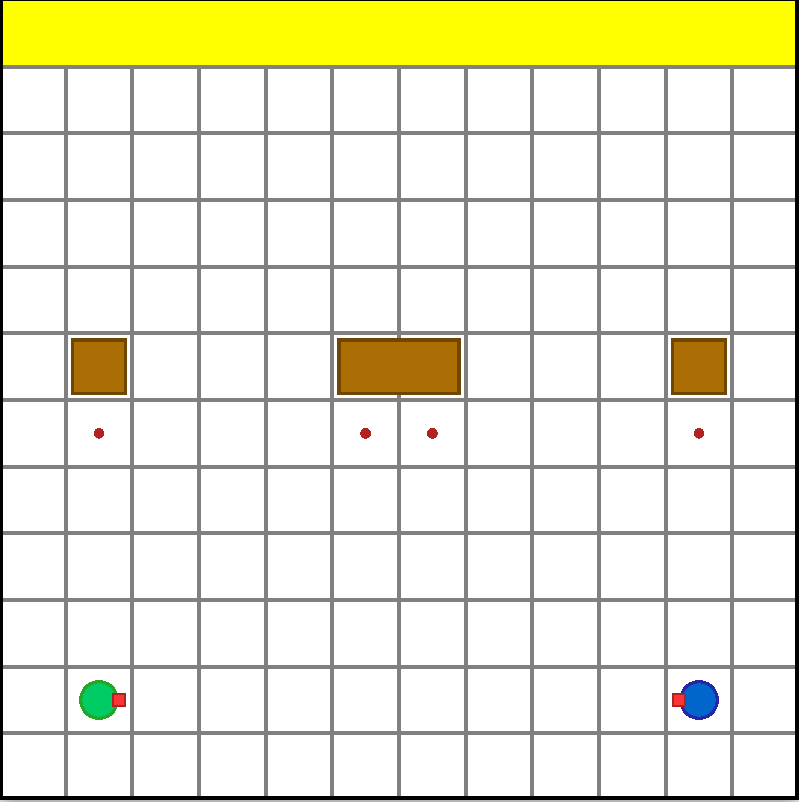}}
    \quad
    \centering
    \subcaptionbox{(b) After completing the previous macro-actions, robots choose \emph{\textbf{Push}} to move the big box towards the goal together.}
        [0.30\linewidth]{\includegraphics[scale=0.13]{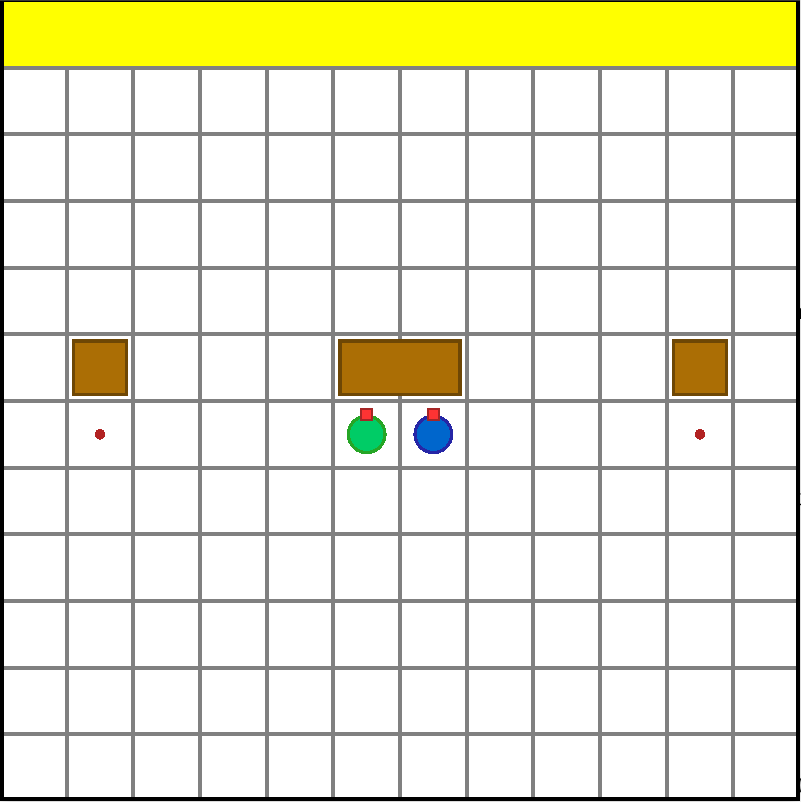}}
    \quad
    \centering
    \subcaptionbox{(c) Robots finish the task by pushing the big box to the goal area.}
        [0.30\linewidth]{\includegraphics[scale=0.13]{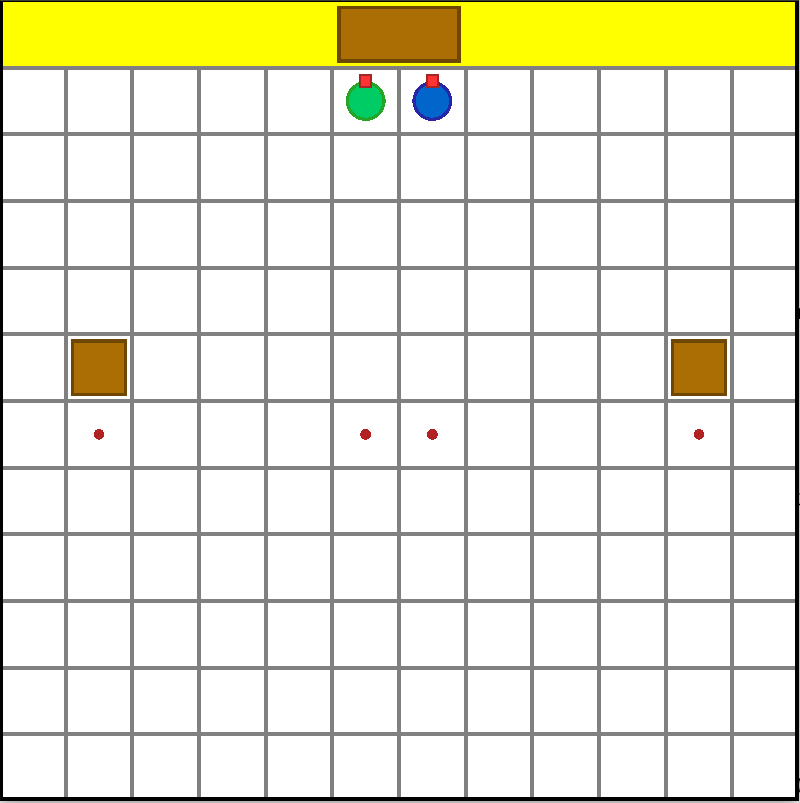}}
    \caption{Visualization of the optimal macro-action-based behaviors learned using Mac-IAICC in the Box Pushing domain under a $12\times12$ grid world.}
    \label{bp_behavior}
\end{figure*}

We show the optimal behaviors learned under the grid world size $12\times12$ in Fig.~\ref{bp_behavior}. 

\subsection*{Overcooked}
\label{B-Overcooked}

\begin{figure*}[h!]
    \centering
    \captionsetup[subfigure]{labelformat=empty}
    \centering
    \subcaptionbox{(i) After placing the onion on the cutting board, the green robot executes \textbf{\emph{Chop}}.}
        [0.23\linewidth]{\includegraphics[scale = 0.20]{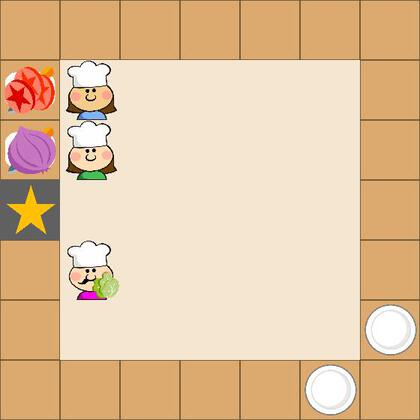}}
    ~    
    \subcaptionbox{(j) The green robot executes \textbf{\emph{Get-Plate-2}}, and the blue robot keep running \textbf{\emph{Move-Down}} to make room for the pink robot to merge the chopped vegetables later on.}
        [0.23\linewidth]{\includegraphics[scale = 0.20]{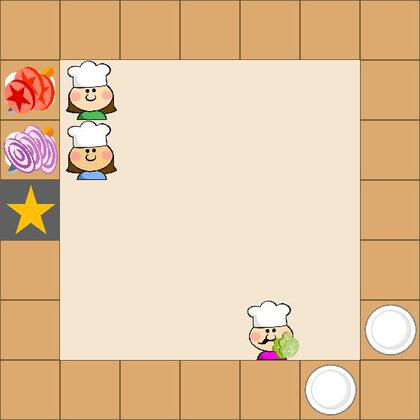}}
    ~    
    \subcaptionbox{(k) The pink robot reaches the plate and it is going to put the lettuce on the plate.}
        [0.23\linewidth]{\includegraphics[scale = 0.20]{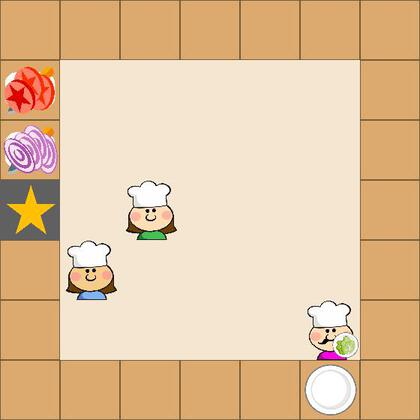}}
    ~    
    \subcaptionbox{(l) After putting the lettuce on the plate, the pink robot merges the onion in the plate by executing executes \textbf{\emph{Get-Onion}}.}
        [0.23\linewidth]{\includegraphics[scale = 0.20]{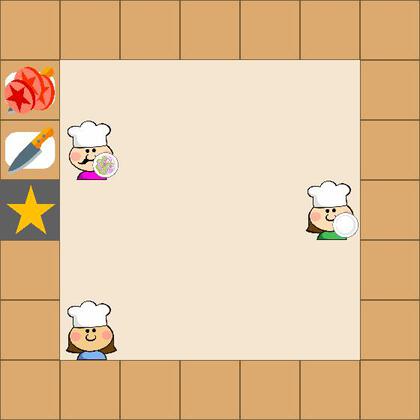}}
    ~
    \subcaptionbox{(m) The pink robot gets the chopped tomato into the plate by executing \emph{Get-Tomato}.}
        [0.23\linewidth]{\includegraphics[scale = 0.20]{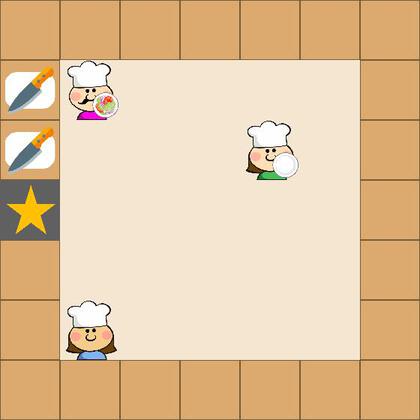}}
    ~    
    \subcaptionbox{(n) The pink robot successfully delivers the tomato-lettuce-onion salad by running \emph{Deliver}.}
        [0.23\linewidth]{\includegraphics[scale = 0.20]{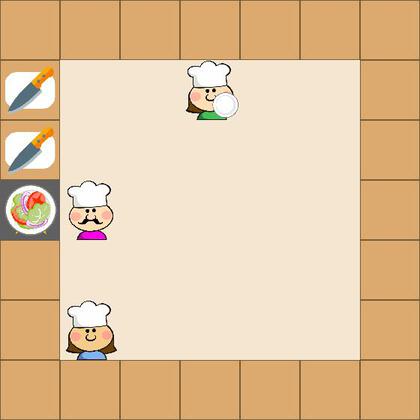}}
    ~
    \caption{Visualization of running decentralized policies learned by Mac-IAICC in Overcooked-A.}
    \label{fig:Overcooked_visual_mapA}
\end{figure*}

\textbf{\emph{Map A:}} In this map, our method learns an efficient collaboration such as three robots separately get three different vegetables, and then go to the cutting board and chop them respectively. Especially, the pink robot leans to take away the chopped lettuce in order to make room for the incoming green robot to chop the onion (Fig.~\ref{fig:Overcooked_visual_mapA}h - \ref{fig:Overcooked_visual_mapA}i). 
  
\begin{figure*}[h!]
    \centering
    \captionsetup[subfigure]{labelformat=empty}
    \centering
    \subcaptionbox{(a) The blue robot executes \textbf{\emph{Get-Lettuce}}. The pink robot executes \textbf{\emph{Get-Tomato}}. The green robot executes \textbf{\emph{Get-Onion}}.}
        [0.23\linewidth]{\includegraphics[scale = 0.15]{fig/paper3/Overcooked/3_agent_D.png}}
    ~    
    \subcaptionbox{(b) After getting the lettuce, the pink robot executes \textbf{\emph{Go-Cut-Board-2}}.}
        [0.23\linewidth]{\includegraphics[scale = 0.20]{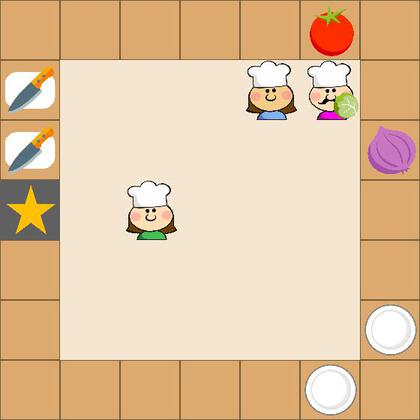}}
    ~    
    \subcaptionbox{(c) After getting the tomato, the blue robot executes \textbf{\emph{Go-Cut-Board-1}}.}
        [0.23\linewidth]{\includegraphics[scale = 0.20]{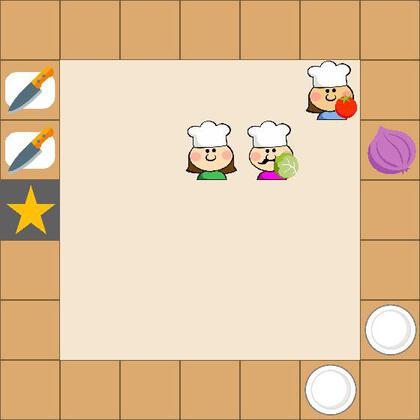}}
    ~    
    \subcaptionbox{(d) After getting the onion, the green robot executes \textbf{\emph{Go-Cut-Board-2}}.}
        [0.23\linewidth]{\includegraphics[scale = 0.20]{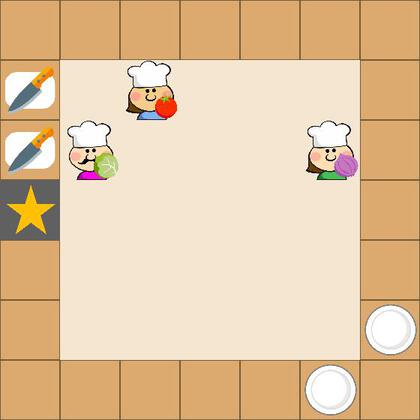}}
    ~    
    \subcaptionbox{(e) After placing the lettuce on the cutting board, the pink robot executes \textbf{\emph{Chop}}.  }
        [0.23\linewidth]{\includegraphics[scale = 0.20]{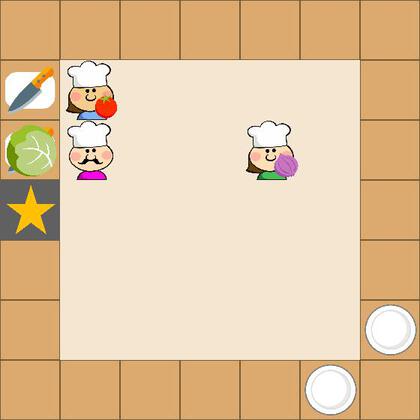}}
    ~    
    \subcaptionbox{(f) After placing the tomato on the cutting board, the blue robot executes \textbf{\emph{Chop}}.  }
        [0.23\linewidth]{\includegraphics[scale = 0.20]{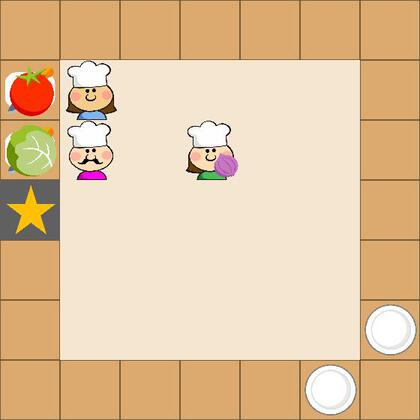}}
    ~    
    \subcaptionbox{(g) After finishing chopping the lettuce, the pink robot executes \textbf{\emph{Get-Lettuce}} to pick it up.}
        [0.23\linewidth]{\includegraphics[scale = 0.20]{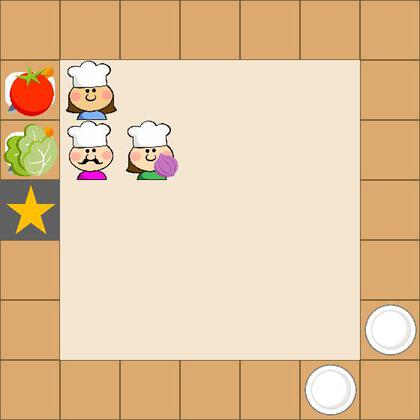}}
    ~    
    \subcaptionbox{(h) With the lettuce in hand, the pink robot executes \textbf{\emph{Get-Plate-1}}.}
        [0.23\linewidth]{\includegraphics[scale = 0.20]{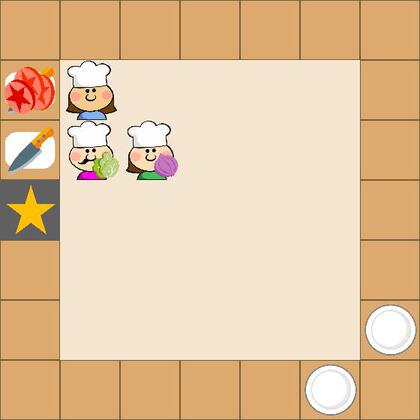}}
    \end{figure*}

\clearpage
\textbf{\emph{Map B:}} In this map, the decentralized policies trained by our method learn the collaboration such that the pink robot focuses on transporting items from right to left, while the other two robots cooperatively prepare the salad.

\begin{figure*}[h!]
    \centering
    \captionsetup[subfigure]{labelformat=empty}
    \centering
    \subcaptionbox{(a) The blue robot executes \textbf{\emph{Go-Cut-board-1}}. The green robot executes \textbf{\emph{Go-Cut-board-2}}. The pink robot executes \textbf{\emph{Get-Lettuce}}.}
        [0.23\linewidth]{\includegraphics[scale = 0.15]{fig/paper3/Overcooked/3_agent_F.png}}
    ~    
    \subcaptionbox{(b) After getting the lettuce, the pink robot executes \textbf{\emph{Go-Counter}}.}
        [0.23\linewidth]{\includegraphics[scale = 0.20]{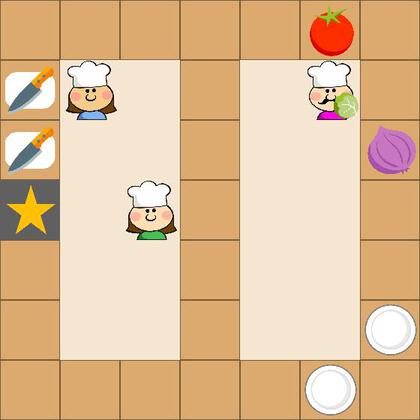}}
    ~    
    \subcaptionbox{(c) After putting the lettuce on the counter, the pink robot executes \textbf{\emph{Get-Onion}}.}
        [0.23\linewidth]{\includegraphics[scale = 0.20]{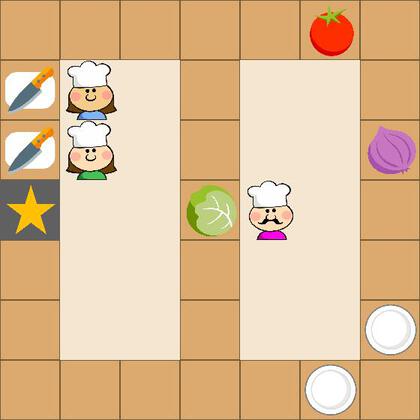}}
    ~    
    \subcaptionbox{(d) The green robot executes \textbf{\emph{Get-Lettuce}}.}
        [0.23\linewidth]{\includegraphics[scale = 0.20]{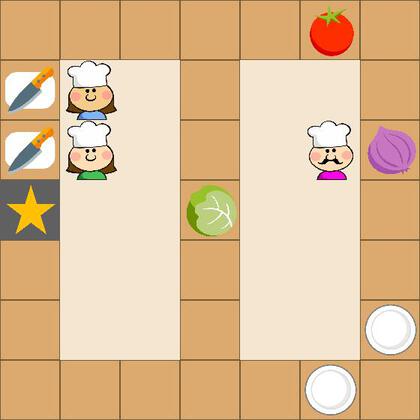}}
    ~    
    \subcaptionbox{(e) After getting the onion, the pink robot executes \textbf{\emph{Go-Counter}}.}
        [0.23\linewidth]{\includegraphics[scale = 0.20]{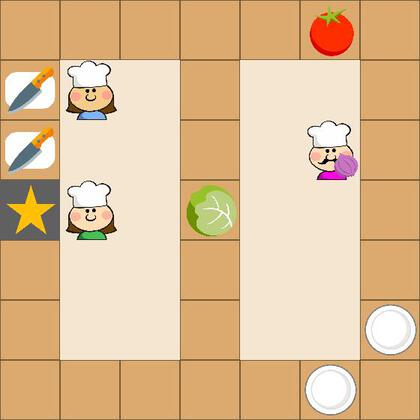}}
    ~    
    \subcaptionbox{(f) After getting the lettuce, the green robot executes \textbf{\emph{Go-Cut-Board-2}}. Meanwhile, the pink robot puts the onion on the counter, and then executes \textbf{\emph{Get-Tomato}}. The blue robot executes \textbf{\emph{Get-Onion}}. }
        [0.23\linewidth]{\includegraphics[scale = 0.20]{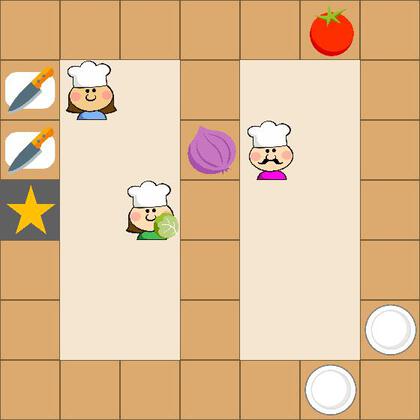}}
    ~    
    \subcaptionbox{(g) After putting the lettuce on the cutting board, the green executes \textbf{\emph{Chop}}. Blue robot executes \textbf{\emph{Go-Cut-Board-1}} with onion in hand.}
        [0.23\linewidth]{\includegraphics[scale = 0.20]{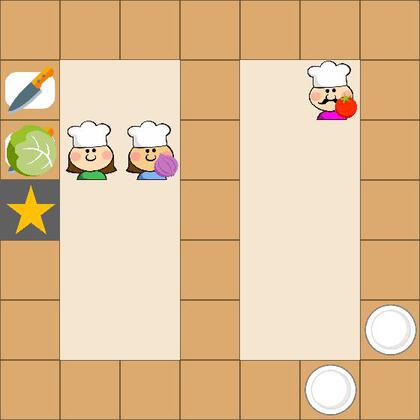}}
    ~    
    \subcaptionbox{(h) The blue robot executes \textbf{\emph{Chop}} to cut the onion to pieces.}
        [0.23\linewidth]{\includegraphics[scale = 0.20]{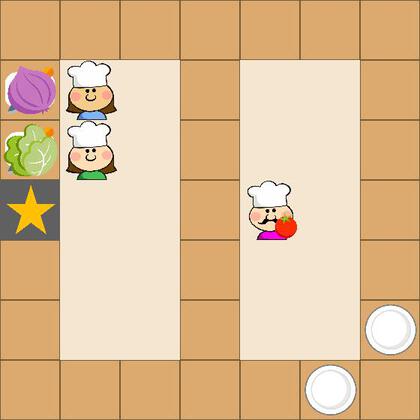}}
\end{figure*}

\clearpage

\begin{figure*}[h!]
    \centering
    \captionsetup[subfigure]{labelformat=empty}
    \centering
    ~    
    \subcaptionbox{(i) After putting the tomato on the counter, the pink robot executes \textbf{\emph{Get-Plate-2}}. The green robot executes \textbf{\emph{Get-Tomato}}. }
        [0.23\linewidth]{\includegraphics[scale = 0.20]{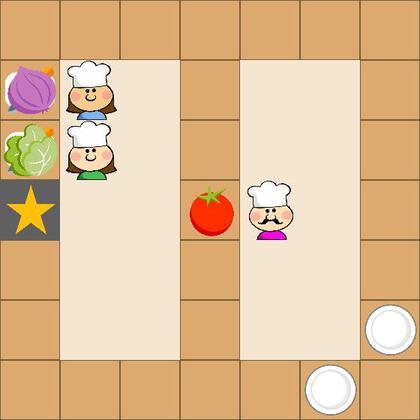}}
    ~    
    \subcaptionbox{(j) The blue robot finishes chopping the onion, and then picks it up by executing \textbf{\emph{Get-Onion}} again.}
        [0.23\linewidth]{\includegraphics[scale = 0.20]{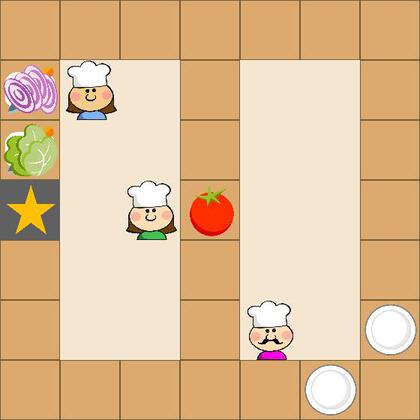}}
    ~    
    \subcaptionbox{(k) The blue robot moves down to make room for the green robot to chop the tomato later on. The green robot moves towards the upper cutting board.}
        [0.23\linewidth]{\includegraphics[scale = 0.20]{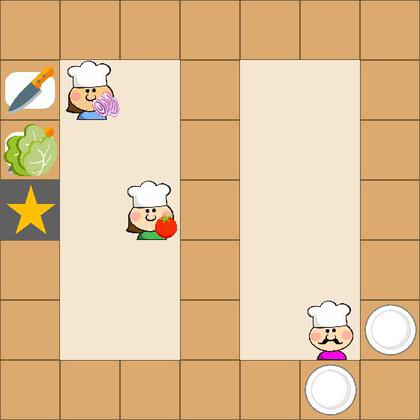}}
    ~    
    \subcaptionbox{(l) After getting the plate, the pink robot executes \textbf{\emph{Go-Counter}}.}
        [0.23\linewidth]{\includegraphics[scale = 0.20]{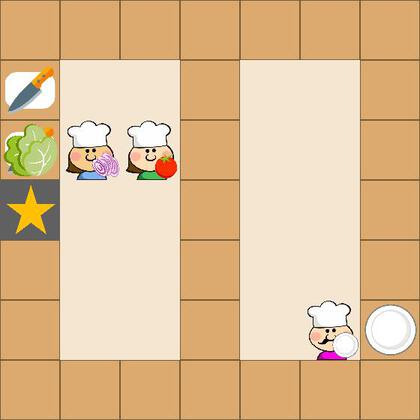}}
    ~    

    \subcaptionbox{(m) After putting the tomato on the cutting board, the green robot executes \textbf{\emph{Chop}}. }
        [0.23\linewidth]{\includegraphics[scale = 0.20]{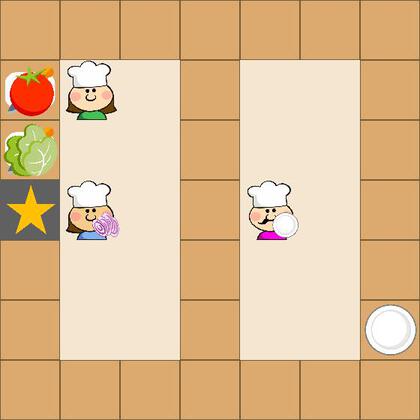}}
    ~    
    \subcaptionbox{(n) The pink robot puts the plate on the counter. The blue robot executes \textbf{\emph{Go-Counter}} to get the plate.}
        [0.23\linewidth]{\includegraphics[scale = 0.20]{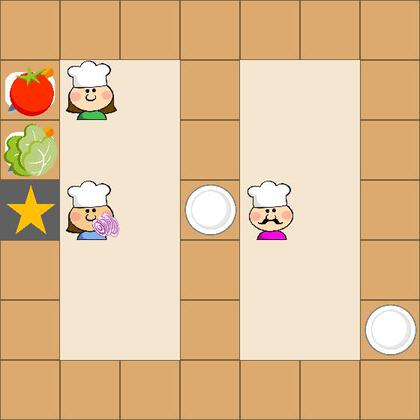}}
    ~    
    \subcaptionbox{(o) The green robot finishes chopping the tomato, while the blue robot puts chopped onion on the plate.}
        [0.23\linewidth]{\includegraphics[scale = 0.20]{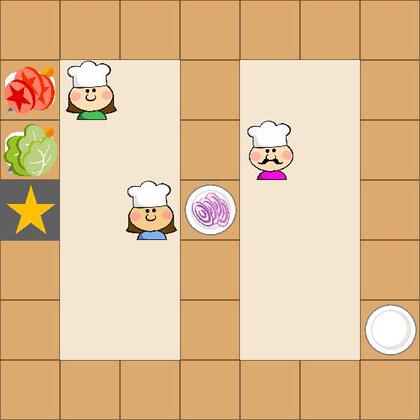}}
    ~    
    \subcaptionbox{(p) The green robot executes \emph{Go-Cut-Board-2} to make room for the blue robot to merge the tomato into the plate.}
        [0.23\linewidth]{\includegraphics[scale = 0.20]{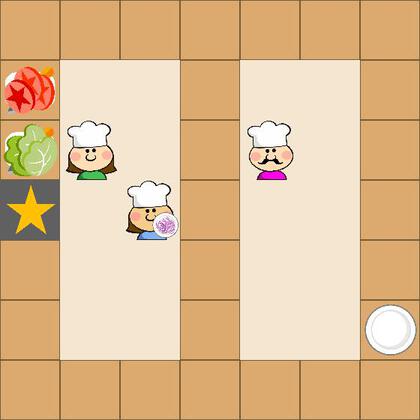}}

    ~    
    \subcaptionbox{(q) The blue robot executes \textbf{\emph{Get-Lettuce}}. The green robot executes \textbf{\emph{Go-Cut-Board-1}}.}
        [0.23\linewidth]{\includegraphics[scale = 0.20]{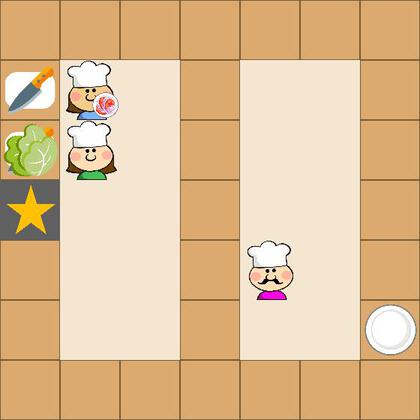}}
    ~    
    \subcaptionbox{(r) After putting the lettuce on the plate, the blue robot executes \textbf{\emph{Deliver}}.}
        [0.23\linewidth]{\includegraphics[scale = 0.20]{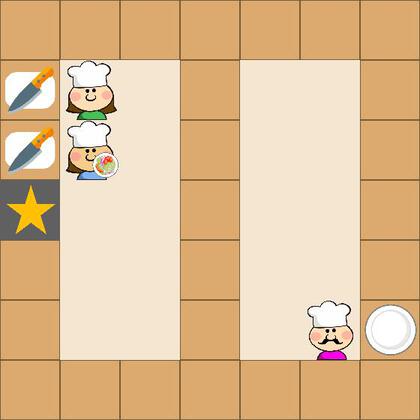}}
    ~
    \subcaptionbox{(s) The blue robot successfully delivers the tomato-lettuce-onion salad.}
        [0.23\linewidth]{\includegraphics[scale = 0.20]{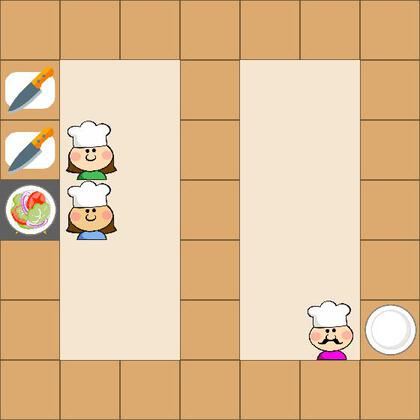}}
    ~
    \caption{Visualization of running decentralized policies learned by Mac-IAICC in Overcooked-B.}
    \label{fig:Overcooked_visual_mapC}
\end{figure*}

\clearpage
\subsection*{Warehouse Tool Delivery}

\textbf{\emph{Warehouse A:}} 

\begin{figure*}[h!]
    \centering
    \captionsetup[subfigure]{labelformat=empty}
    \centering
    \subcaptionbox{(a) Initial State.\vspace{2mm}}
        [0.30\linewidth]{\includegraphics[scale=0.26]{fig/paper3/WTD/wtd_a_small.png}}
    \quad
    \centering
    \subcaptionbox{(b) Mobile robots moves towards the table by running \emph{\textbf{Get-Tool}}, and arm robot runs \emph{\textbf{Search-Tool(0)}} to find Tool-0.\vspace{2mm}}
        [0.30\linewidth]{\includegraphics[scale=0.18]{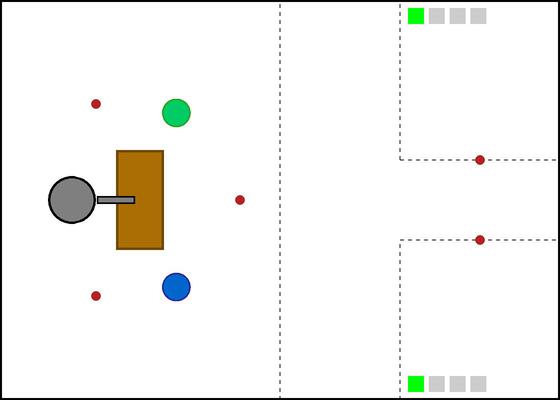}}
    \quad
    \centering
    \subcaptionbox{(c) Mobile robots wait there and the robot arm keeps looking for Tool-0.\vspace{2mm}}
        [0.30\linewidth]{\includegraphics[scale=0.18]{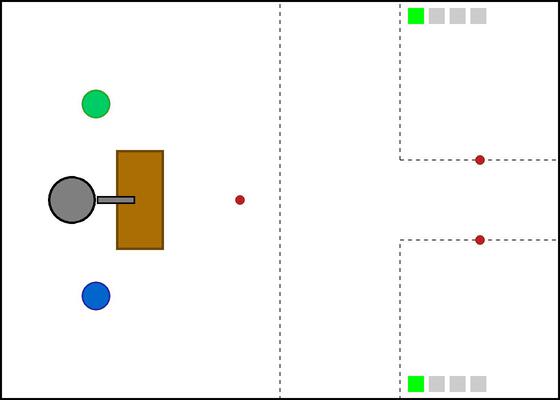}}
    \quad
    \centering
    \subcaptionbox{(d) Arm robot executes \emph{\textbf{Pass-to-M(1)}} to pass Tool-0 to the blue robot.\vspace{2mm}}
        [0.30\linewidth]{\includegraphics[scale=0.18]{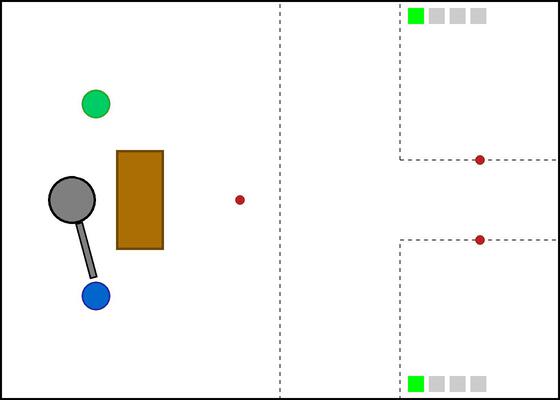}}
    \quad
    \centering
    \subcaptionbox{(e) Arm robot executes \emph{\textbf{Search-Tool(0)}} to find Tool-0, and blue robot moves to workshop-1 by executing \emph{\textbf{Go-W(1)}}.\vspace{2mm}}
        [0.30\linewidth]{\includegraphics[scale=0.18]{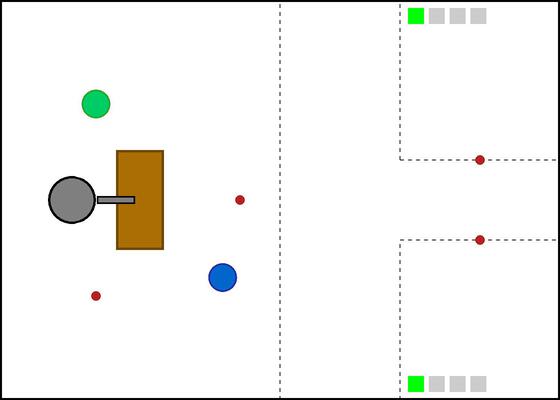}}
    \quad
    \centering
    \subcaptionbox{(f) Blue robot successfully delivers Tool-0 to workshop-1.\vspace{2mm}}
        [0.30\linewidth]{\includegraphics[scale=0.18]{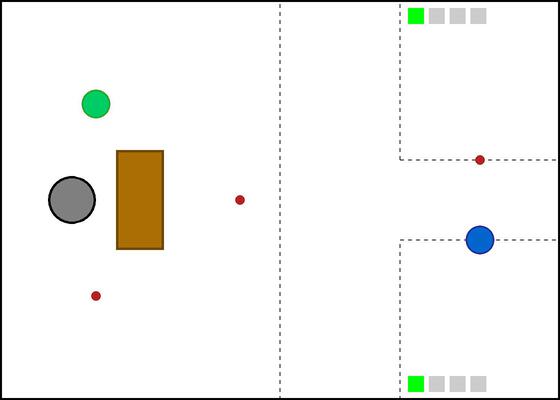}}
    \quad
    \centering
    \subcaptionbox{(g) Blue robot runs \emph{\textbf{Get-Tool}} to go back table, and arm robot executes \emph{\textbf{Pass-to-M(0)}} to pass Tool-0 to green robot.\vspace{2mm}}
        [0.30\linewidth]{\includegraphics[scale=0.18]{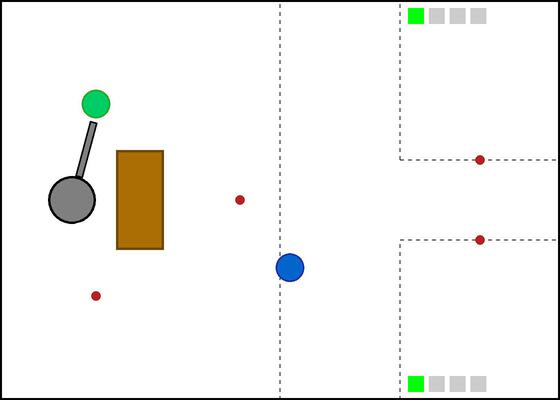}}
    \quad
    \centering
    \subcaptionbox{(h) Green robot executes \emph{\textbf{Go-W(0)}} and arm robot runs \emph{\textbf{Search-Tool(1)}}. \vspace{2mm}}
        [0.30\linewidth]{\includegraphics[scale=0.18]{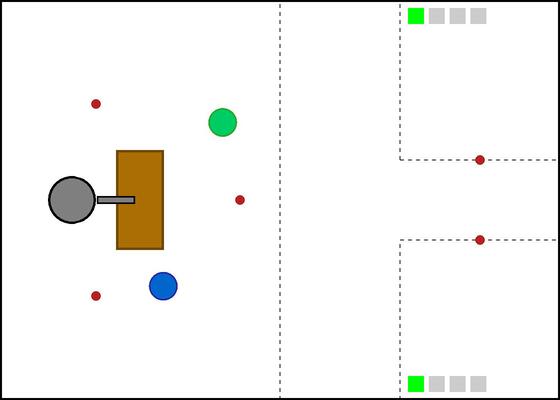}}
    \quad
    \centering
    \subcaptionbox{(i) Green robot successfully delivers Tool-0 to workshop-0. Human-0 and human-1 finish subtask-0 and start to do subtask-1 with delivered Tool-0.\vspace{2mm}}
        [0.30\linewidth]{\includegraphics[scale=0.18]{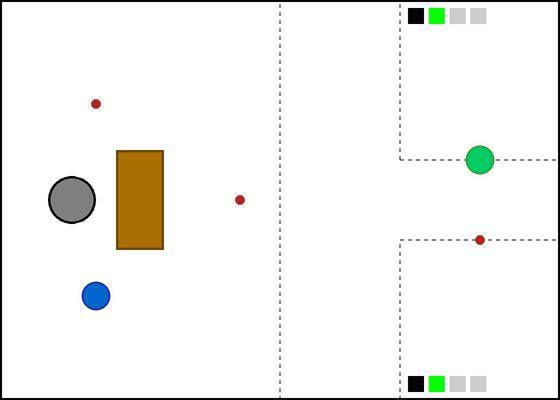}}
\end{figure*}
\begin{figure*}[h!]
    \centering
    \captionsetup[subfigure]{labelformat=empty}
    \centering
    \subcaptionbox{(j) Green robot runs \emph{\textbf{Get-Tool}} to go back table, and arm robot executes \emph{\textbf{Pass-to-M(1)}} to pass a Tool-1 to blue robot.\vspace{2mm}}
        [0.30\linewidth]{\includegraphics[scale=0.18]{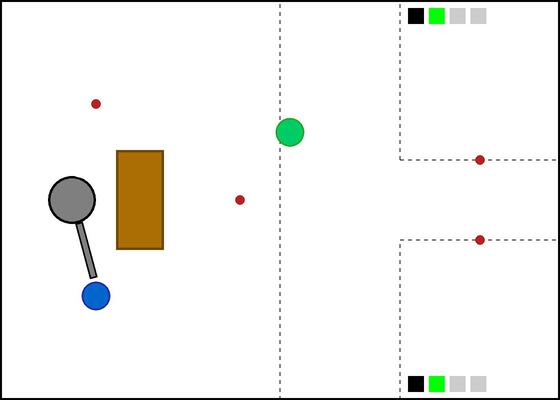}}
    \quad
    \centering
    \subcaptionbox{(k) Blue robot executes \emph{\textbf{Go-W(1)}} and arm robot runs \emph{\textbf{Search-Tool(1)}}. \vspace{2mm}}
        [0.30\linewidth]{\includegraphics[scale=0.18]{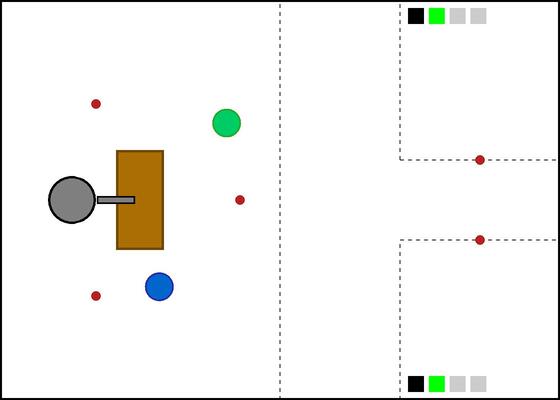}}
    \quad
    \centering
    \subcaptionbox{(l)  Blue robot successfully delivers a Tool-1 to workshop-1.\vspace{2mm}}
        [0.30\linewidth]{\includegraphics[scale=0.18]{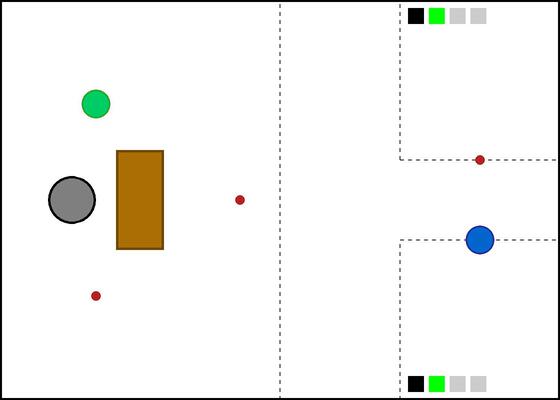}}
    \quad
    \centering
    \subcaptionbox{(m) Arm robot executes \emph{\textbf{Pass-to-M(0)}} to pass Tool-1 to green robot. Blue robot runs \emph{\textbf{Get-Tool}} to go back table. \vspace{2mm}}
        [0.30\linewidth]{\includegraphics[scale=0.18]{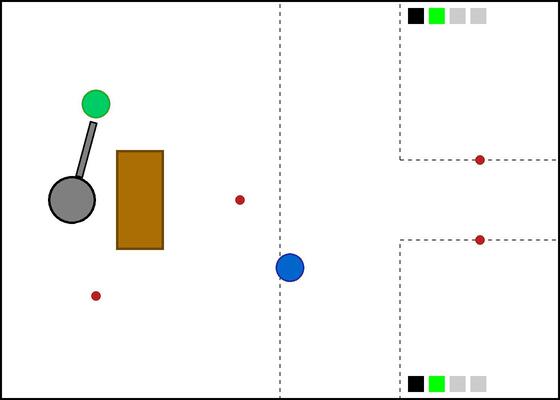}}
    \quad
    \centering
    \subcaptionbox{(n) Green robot delivers Tool-1 to workshop-0. Human-0 and human-1 finish subtask-1 and start to do subtask-2 with delivered Tool-1.\vspace{2mm}}
        [0.30\linewidth]{\includegraphics[scale=0.18]{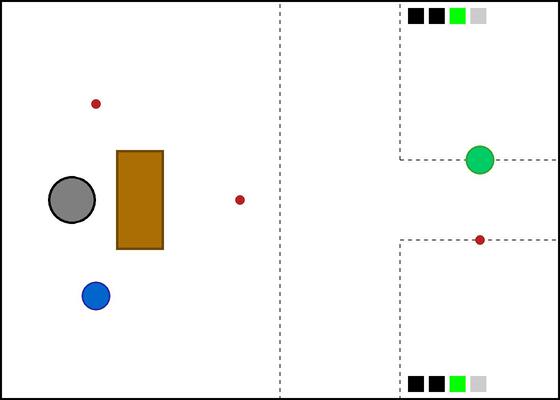}}
    \quad
    \centering
    \subcaptionbox{(o) Arm robot executes \emph{\textbf{Pass-to-M(1)}} to pass Tool-2 to blue robot. Green robot runs \emph{\textbf{Get-Tool}} to go back table.\vspace{2mm}}
        [0.30\linewidth]{\includegraphics[scale=0.18]{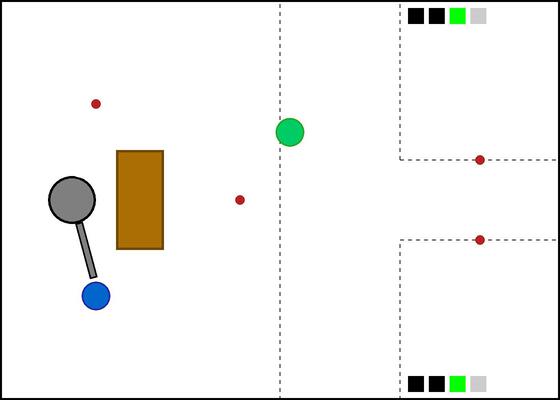}}
    \quad
    \centering
    \subcaptionbox{(p) Blue robot executes \emph{\textbf{Go-W(1)}}. Arm robot runs \emph{\textbf{Search-Tool(2)}}. \vspace{2mm}}
        [0.30\linewidth]{\includegraphics[scale=0.18]{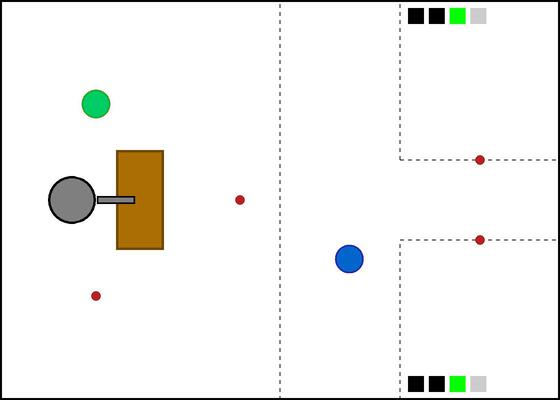}}
    \quad
    \centering
    \subcaptionbox{(q) Blue robot successfully delivers Tool-2 to human-0. \vspace{2mm}}
        [0.30\linewidth]{\includegraphics[scale=0.18]{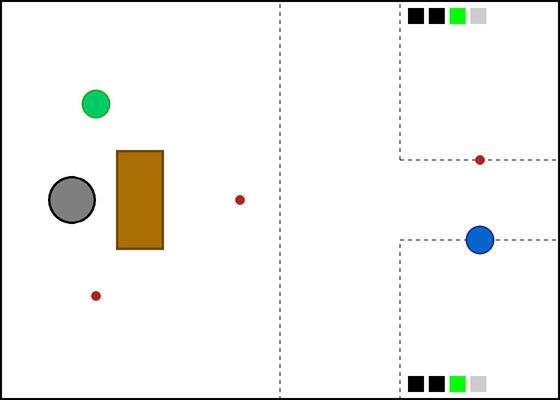}}
    \quad
    \centering
    \subcaptionbox{(r) Arm robot executes \emph{\textbf{Pass-to-M(0)}} to pass Tool-2 to green robot. Blue robot runs \emph{\textbf{Get-Tool}} to go back table.\vspace{2mm}}
        [0.30\linewidth]{\includegraphics[scale=0.18]{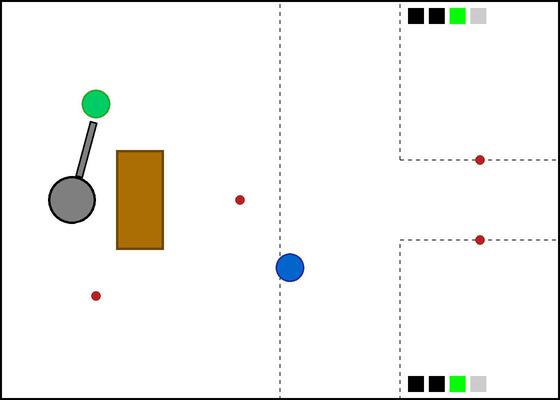}}
    \quad
    \centering
    \subcaptionbox{(s) Green robot directly goes to workshop-0 by running \emph{\textbf{Go-W(0)}} and finishes the last tool delivery for human-0. The entire task is done.}
        [0.9\linewidth]{\includegraphics[scale=0.18]{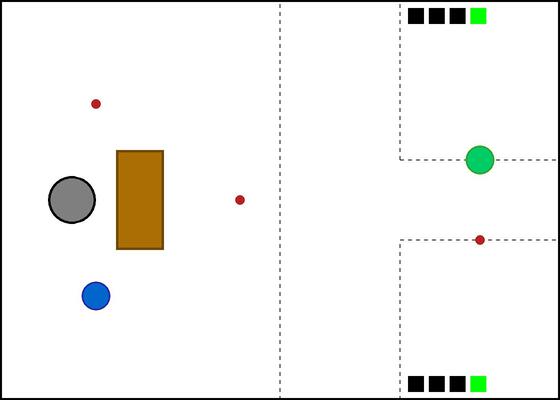}}
    \label{wtd_a_behavior}
\end{figure*}

\clearpage
\textbf{\emph{Warehouse-B:}}  

\begin{figure*}[h!]
    \centering
    \captionsetup[subfigure]{labelformat=empty}
    \centering
    \subcaptionbox{(a) Initial State.\vspace{2mm}}
        [0.30\linewidth]{\includegraphics[scale=0.26]{fig/paper3/WTD/wtd_b_small.png}}
    ~
    \centering
    \subcaptionbox{(b) Mobile robots moves towards the table by running \emph{\textbf{Get-Tool}}, and arm robot runs \emph{\textbf{Search-Tool(0)}} to find Tool-0.\vspace{2mm}}
        [0.30\linewidth]{\includegraphics[scale=0.18]{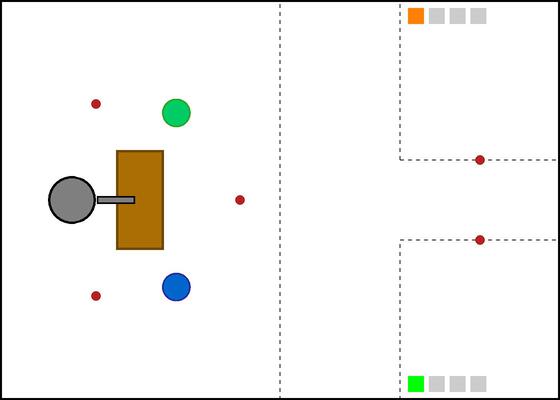}}
    ~
    \centering
    \subcaptionbox{(c) Mobile robots wait there and the robot arm keeps looking for Tool-0.\vspace{2mm}}
        [0.30\linewidth]{\includegraphics[scale=0.18]{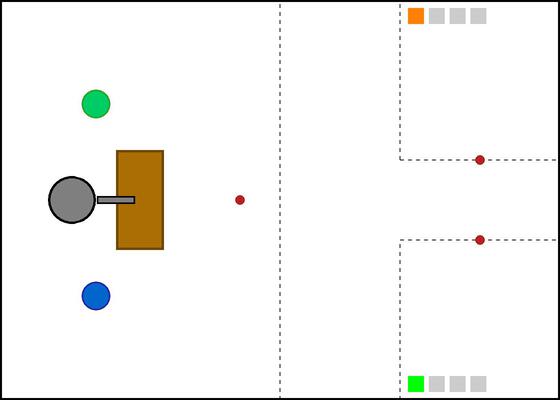}}
    ~
    \centering
    \subcaptionbox{(d) Arm robot executes \emph{\textbf{Pass-to-M(1)}} to pass Tool-0 to the blue robot.\vspace{2mm}}
        [0.30\linewidth]{\includegraphics[scale=0.18]{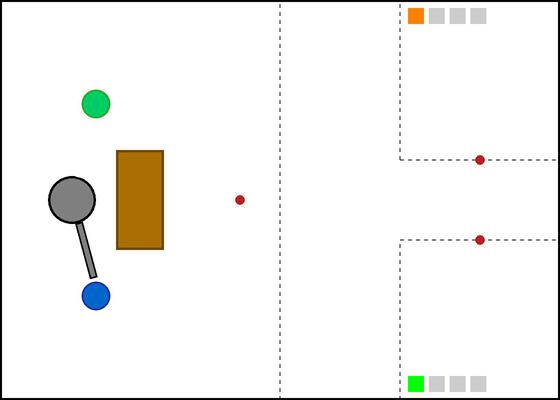}}
    ~
    \centering
    \subcaptionbox{(e) Arm robot runs \emph{\textbf{Search-Tool(1)}} to find Tool-1. Blue robot executes \emph{\textbf{Go-W(0)}} to go to workshop-0. \vspace{2mm}}
        [0.30\linewidth]{\includegraphics[scale=0.18]{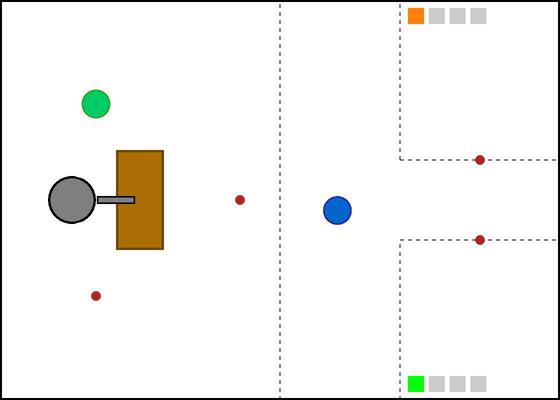}}  
    ~
    \centering
    \subcaptionbox{(f) Blue robot successfully delivers Tool-0 to workshop-0. \vspace{2mm}}
        [0.30\linewidth]{\includegraphics[scale=0.18]{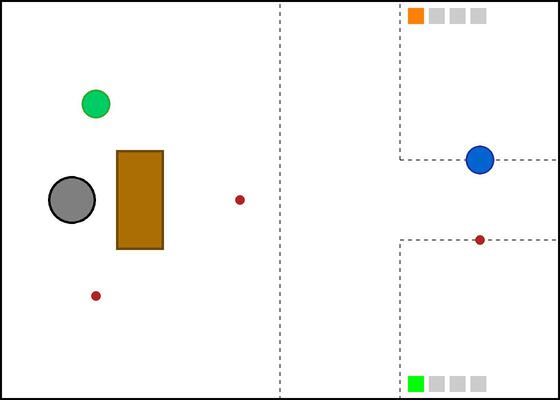}}
    ~
    \centering
    \subcaptionbox{(g)  Blue robot runs \emph{\textbf{Get-Tool}} to go back table. Human-0 finishes subtask-0 and starts to do subtask-1.\vspace{2mm}}
        [0.30\linewidth]{\includegraphics[scale=0.18]{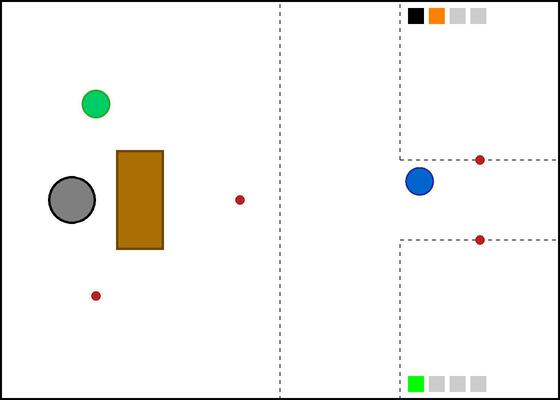}}
    ~
    \centering
    \subcaptionbox{(h) Arm robot executes \emph{\textbf{Pass-to-M(0)}} to pass Tool-1 to green robot.  \vspace{2mm}}
        [0.30\linewidth]{\includegraphics[scale=0.18]{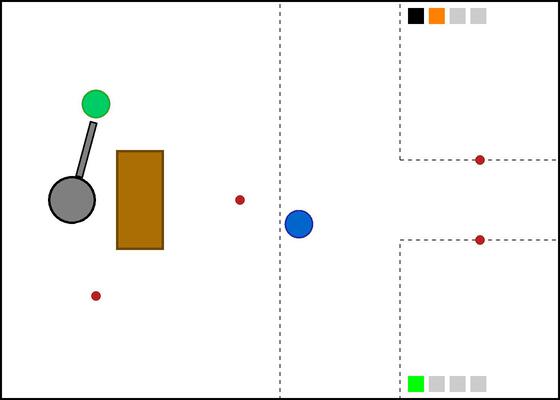}}
    ~
    \centering
    \subcaptionbox{(i) Arm robot runs \emph{\textbf{Search-Tool(0)}} to find Tool-0. Green robot moves to workshop-0 by executing \emph{\textbf{Go-W(0)}}.
    \vspace{2mm}}
        [0.30\linewidth]{\includegraphics[scale=0.18]{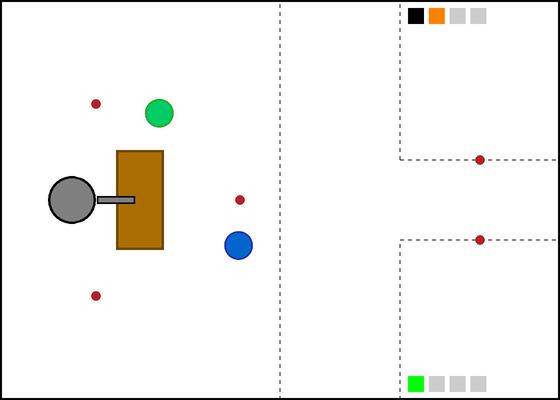}}
    
\end{figure*}
\begin{figure*}[h!]
    \centering
    \captionsetup[subfigure]{labelformat=empty}          
    ~
    \centering
    \subcaptionbox{(j) Green robot successfully delivers Tool-1 to workshop-0. \vspace{2mm}}
        [0.30\linewidth]{\includegraphics[scale=0.18]{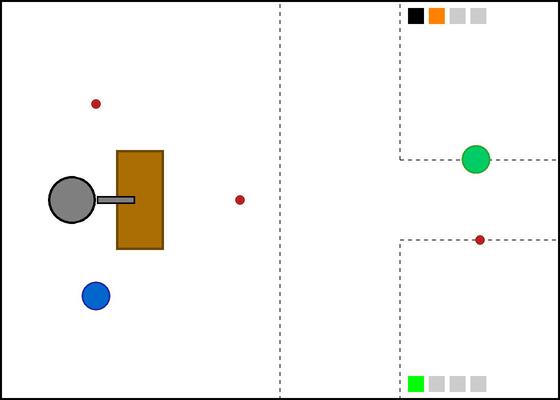}}
    ~
    \centering
    \subcaptionbox{(k) Arm robot executes \emph{\textbf{Pass-to-M(1)}} to pass Tool-0 to blue robot. Green robot runs \emph{\textbf{Get-Tool}} to go back table. \vspace{2mm}}
        [0.30\linewidth]{\includegraphics[scale=0.18]{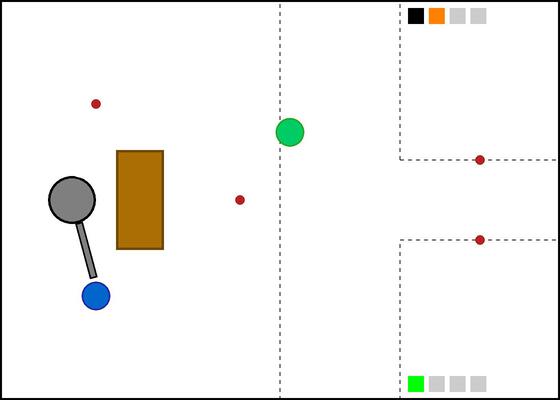}}  
    ~
    \centering
    \subcaptionbox{(l) Arm robot runs \emph{\textbf{Search-Tool(2)}} to find Tool-2. Blue robot moves to workshop-1 by executing \emph{\textbf{Go-W(1)}}. Human-0 finishes subtask-1 and starts to do subtask-2.\vspace{2mm}}
        [0.30\linewidth]{\includegraphics[scale=0.18]{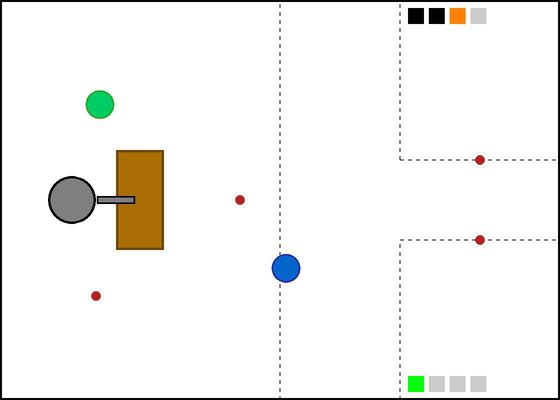}}
    \centering        
    ~
    \centering
    \subcaptionbox{(m) Blue robot successfully delivers Tool-0 to workshop-1. \vspace{2mm}}
        [0.30\linewidth]{\includegraphics[scale=0.18]{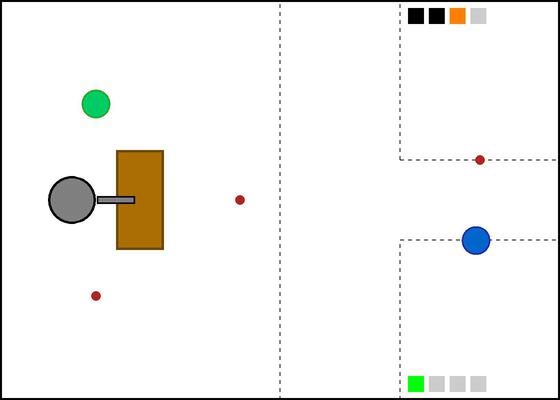}}
    ~
    \centering
    \subcaptionbox{(n) Arm robot executes \emph{\textbf{Pass-to-M(0)}} to pass Tool-2 to green robot. Blue robot runs \emph{\textbf{Get-Tool}} to go back table.\vspace{2mm}}
        [0.30\linewidth]{\includegraphics[scale=0.18]{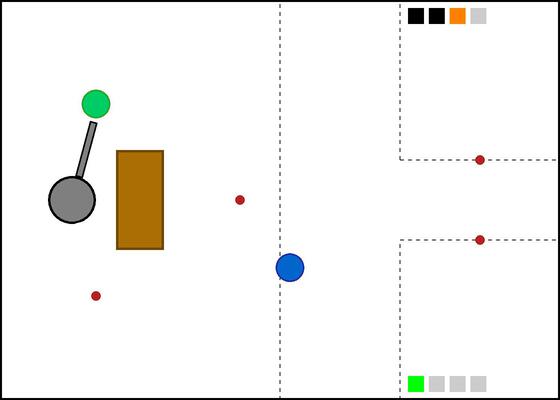}}
    ~
    \centering
    \subcaptionbox{(o) Arm robot runs \emph{\textbf{Search-Tool(1)}} to find Tool-1. Green robot moves to workshop-0 by executing \emph{\textbf{Go-W(0)}}.\vspace{2mm}}
        [0.30\linewidth]{\includegraphics[scale=0.18]{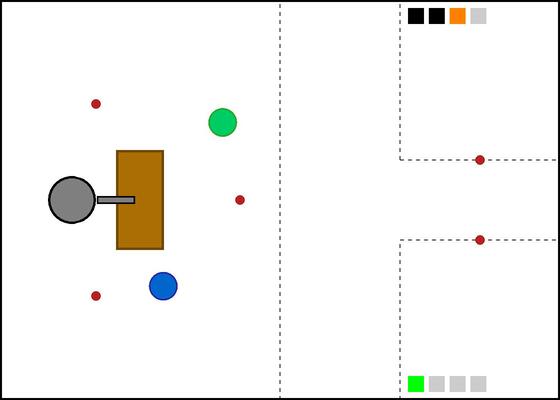}}
    ~
    \subcaptionbox{(p) Green robot successfully delivers Tool-2 to workshop-0.  \vspace{2mm}}
        [0.30\linewidth]{\includegraphics[scale=0.18]{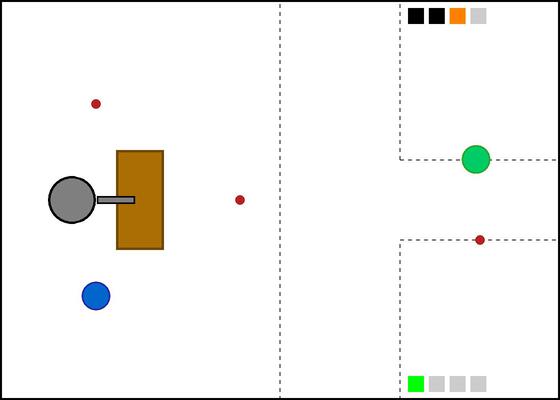}}
    ~
    \centering
    \subcaptionbox{(q) Green robot moves to workshop-1 by executing \emph{\textbf{Go-W(1)}} to observe human-1's status. Human-0 finishes subtask-2 and starts to do subtask-3.\vspace{2mm}}
        [0.30\linewidth]{\includegraphics[scale=0.18]{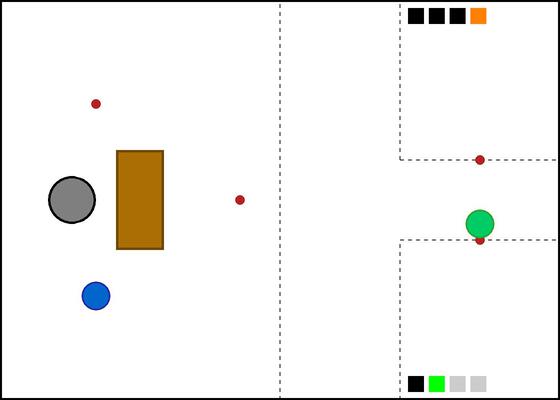}}
    ~
    \centering
    \subcaptionbox{(r) Green robot reaches workshop-1.\vspace{2mm}}
        [0.30\linewidth]{\includegraphics[scale=0.18]{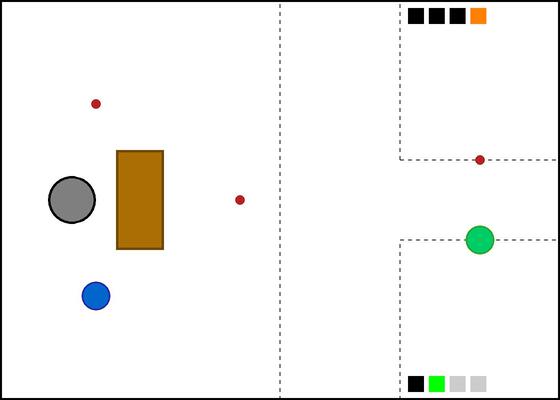}}
\end{figure*}
\begin{figure*}[h!]
    \centering
    \captionsetup[subfigure]{labelformat=empty}
    \centering
    \subcaptionbox{(s) Arm robot executes \emph{\textbf{Pass-to-M(1)}} to pass Tool-1 to blue robot. Green robot runs \emph{\textbf{Get-Tool}} to go back table.\vspace{2mm}}
        [0.30\linewidth]{\includegraphics[scale=0.18]{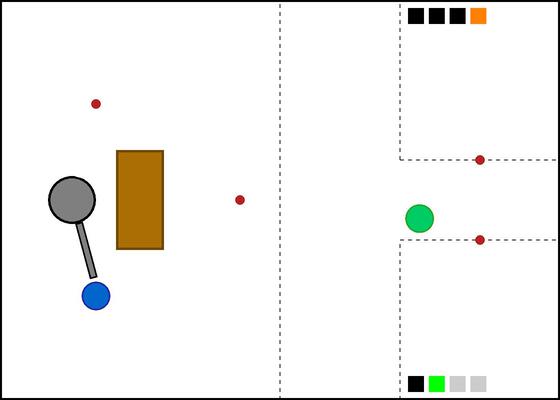}}
    ~
    \centering
    \subcaptionbox{(t) Arm robot runs \emph{\textbf{Search-Tool(2)}} to find Tool-2. Blue robot moves to workshop-1 by executing \emph{\textbf{Go-W(1)}}. \vspace{2mm}}
        [0.30\linewidth]{\includegraphics[scale=0.18]{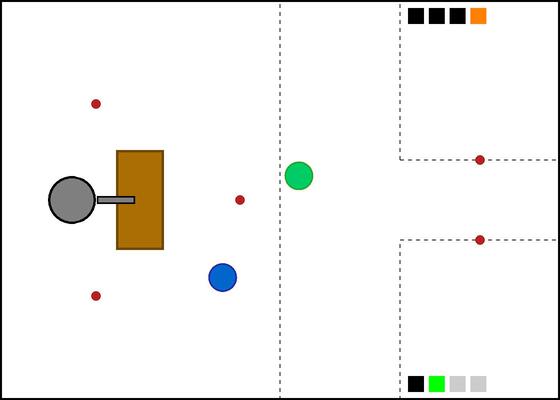}}
    ~
    \centering
    \subcaptionbox{(u) Blue robot successfully delivers Tool-1 to workshop-1. \vspace{2mm}}
        [0.30\linewidth]{\includegraphics[scale=0.18]{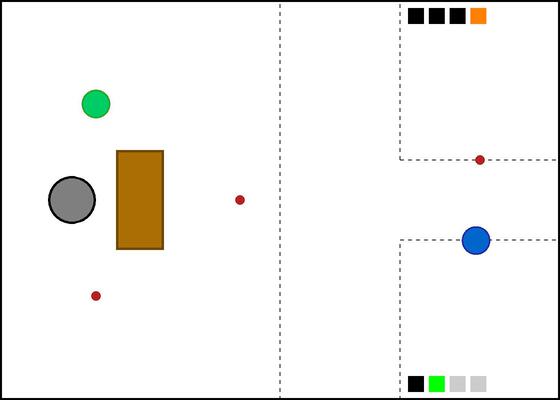}}
    ~
    \centering
    \subcaptionbox{(v) Blue robot runs \emph{\textbf{Get-Tool}} to go back table. Arm robot executes \emph{\textbf{Pass-to-M(0)}} to pass Tool-2 to green robot. Blue robot runs \emph{\textbf{Get-Tool}} to go back table.\vspace{2mm}}
        [0.30\linewidth]{\includegraphics[scale=0.18]{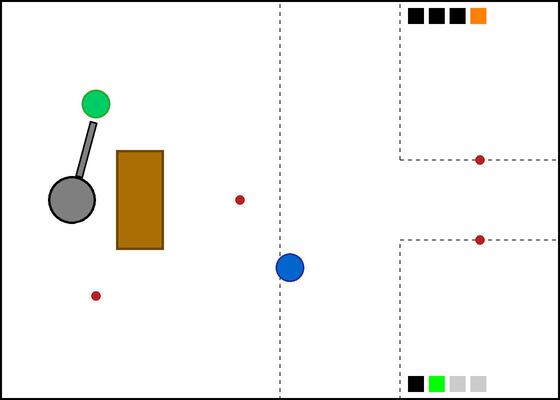}}
    ~
    \centering
    \subcaptionbox{(w) Green robot moves to workshop-1 by executing \emph{\textbf{Go-W(1)}}.\vspace{2mm}}
        [0.30\linewidth]{\includegraphics[scale=0.18]{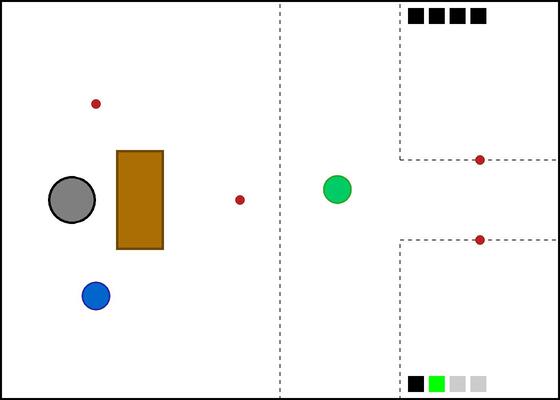}}
    ~
    \centering
    \subcaptionbox{(x) Human-1 finishes subtask-1 and start subtask-2. \vspace{2mm}}
        [0.30\linewidth]{\includegraphics[scale=0.18]{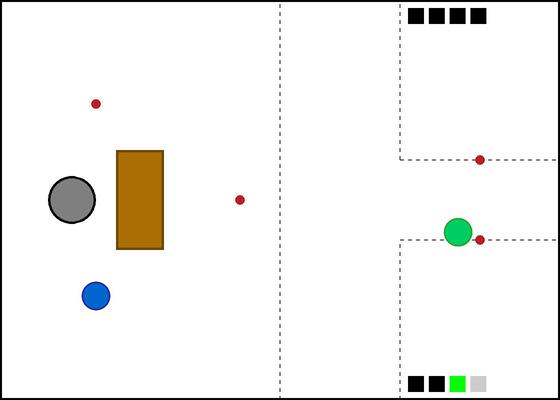}}
    ~
    \centering
    \subcaptionbox{(y) Green robot successfully delivers Tool-2 to workshop-1. Humans have received all tools, and for robots, the task is done.\vspace{2mm}}
        [0.90\linewidth]{\includegraphics[scale=0.18]{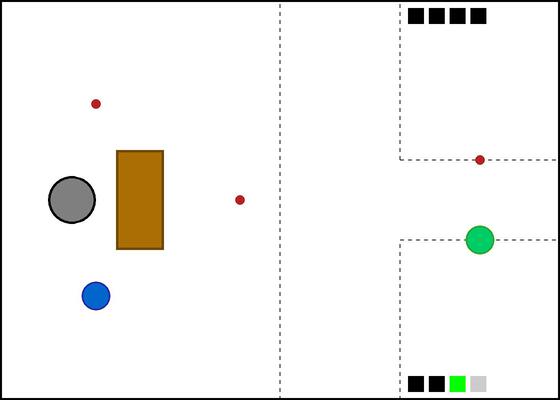}}
    \label{wtd_b_behavior}
\end{figure*}

\clearpage
\textbf{\emph{Warehouse-C:}}.  

\begin{figure*}[h!]
    \centering
    \captionsetup[subfigure]{labelformat=empty}
    \centering
    \subcaptionbox{(a) Initial State.\vspace{2mm}}
        [0.30\linewidth]{\includegraphics[scale=0.24]{fig/paper3/WTD/wtd_c_small.png}}
    ~
    \centering
    \subcaptionbox{(b) Green robot moves towards the table by running \emph{\textbf{Get-Tool}}. Blue robot moves to workshop-0 by executing \emph{\textbf{Go-W(0)}}. Arm robot runs \emph{\textbf{Search-Tool(0)}} to find Tool-0.\vspace{2mm}}
        [0.30\linewidth]{\includegraphics[scale=0.16]{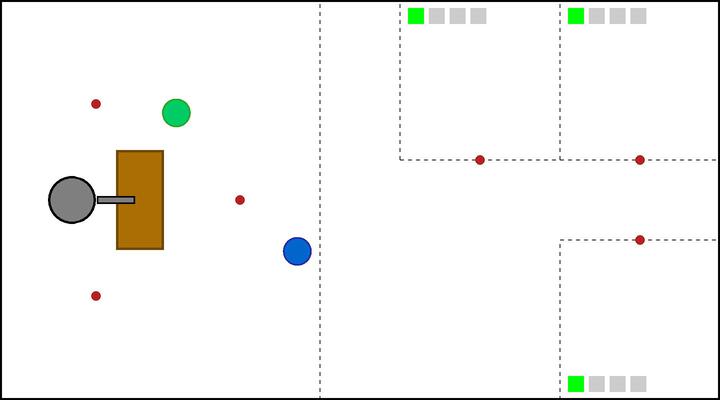}}
    ~
    \centering
    \subcaptionbox{(c) Green robot waits there and arm robot keeps looking for Tool-0.\vspace{2mm}}
        [0.30\linewidth]{\includegraphics[scale=0.16]{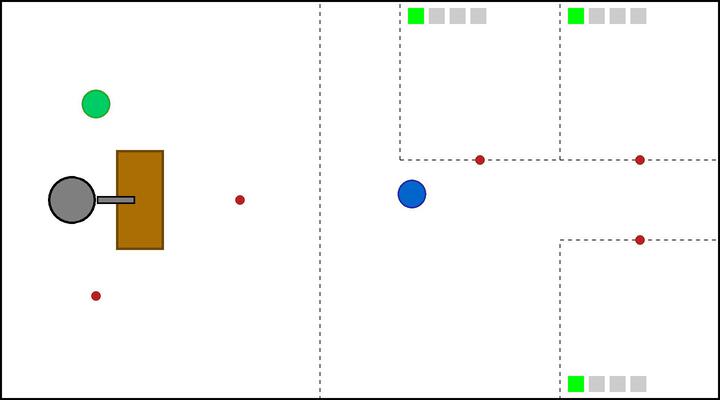}}
    ~
    \centering
    \subcaptionbox{(d) Blue robot reaches workshop-0.\vspace{2mm}}
        [0.30\linewidth]{\includegraphics[scale=0.16]{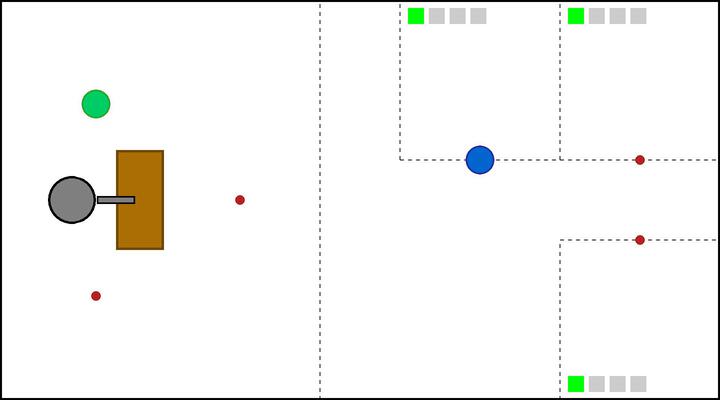}}
    ~
    \centering
    \subcaptionbox{(e) Blue robot runs \emph{\textbf{Get-Tool}} to go back table. \vspace{2mm}}
        [0.30\linewidth]{\includegraphics[scale=0.16]{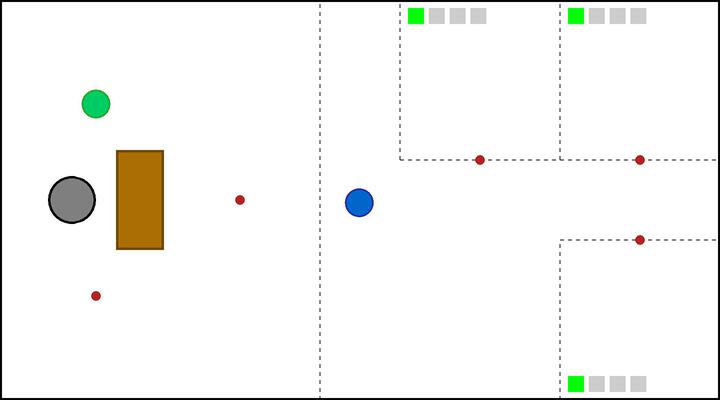}}  
    ~
    \centering
    \subcaptionbox{(f) Arm robot executes \emph{\textbf{Pass-to-M(0)}} to pass Tool-0 to green robot. \vspace{2mm}}
        [0.30\linewidth]{\includegraphics[scale=0.16]{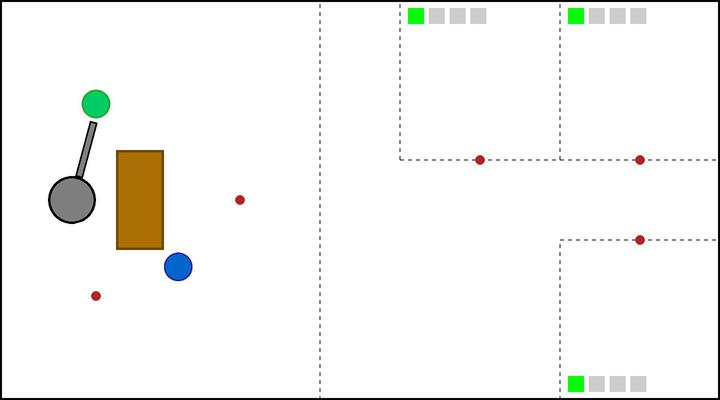}}
    ~
    \centering
    \subcaptionbox{(g)  Arm robot runs \emph{\textbf{Search-Tool(0)}} to find the 2nd Tool-0.\vspace{2mm}}
        [0.30\linewidth]{\includegraphics[scale=0.16]{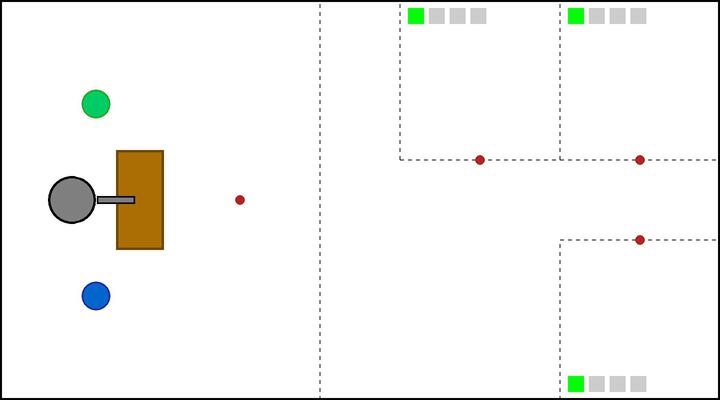}}
    ~
    \centering
    \subcaptionbox{(h) Arm robot executes \emph{\textbf{Pass-to-M(0)}} to pass the 2nd Tool-0 to green robot.  \vspace{2mm}}
        [0.30\linewidth]{\includegraphics[scale=0.16]{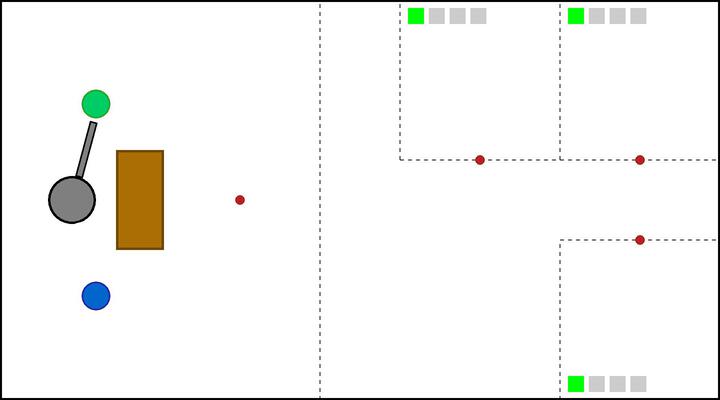}}
    ~
    \centering
    \subcaptionbox{(i) Arm robot runs \emph{\textbf{Search-Tool(0)}} to find the the 3rd Tool-0. Green robot moves to workshop-1 by executing \emph{\textbf{Go-W(1)}}.
    \vspace{2mm}}
        [0.30\linewidth]{\includegraphics[scale=0.16]{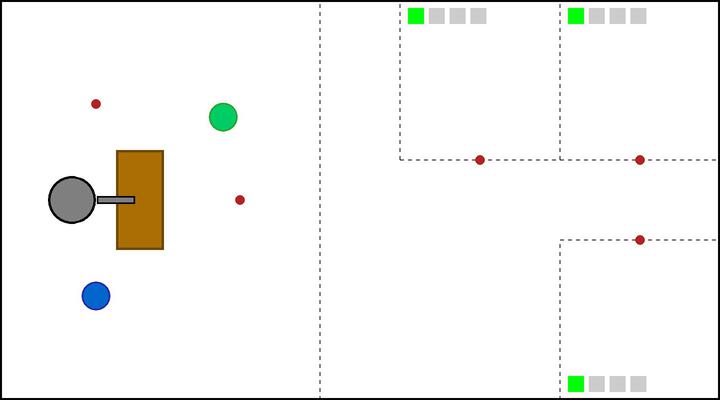}}
\end{figure*}
\begin{figure*}[h!]
    \centering
    \captionsetup[subfigure]{labelformat=empty}          
    ~
    \centering
    \subcaptionbox{(j) Green robot successfully delivers Tool-0 to workshop-1. \vspace{2mm}}
        [0.30\linewidth]{\includegraphics[scale=0.16]{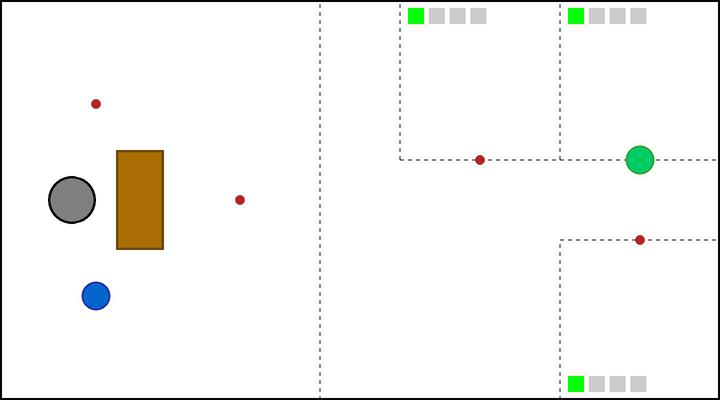}}
    ~
    \centering
    \subcaptionbox{(k) Arm robot executes \emph{\textbf{Pass-to-M(1)}} to pass the 3rd Tool-0 to blue robot. Green robot moves to workshop-0 by executing \emph{\textbf{Go-W(0)}}. \vspace{2mm}}
        [0.30\linewidth]{\includegraphics[scale=0.16]{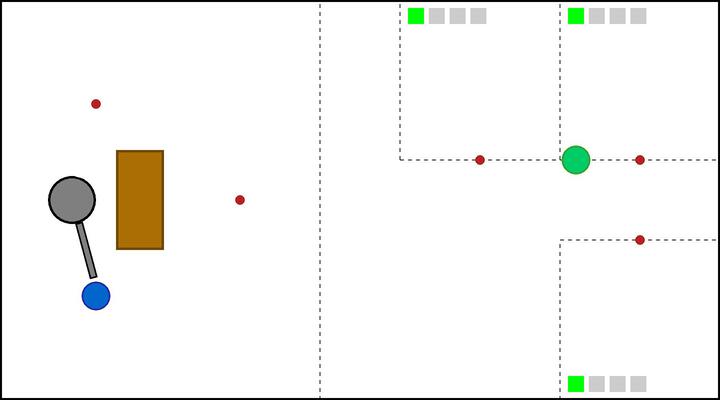}}  
    ~
    \centering
    \subcaptionbox{(l) Arm robot runs \emph{\textbf{Search-Tool(1)}} to find Tool-1. Blue robot moves to workshop-2 by executing \emph{\textbf{Go-W(2)}}. \vspace{2mm}}
        [0.30\linewidth]{\includegraphics[scale=0.16]{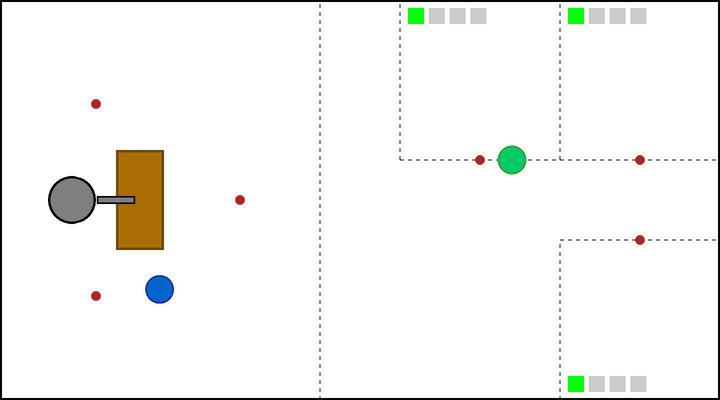}}
    ~
    \centering
    \subcaptionbox{(m) Green robot successfully delivers a Tool-0 to workshop-0. \vspace{2mm}}
        [0.30\linewidth]{\includegraphics[scale=0.16]{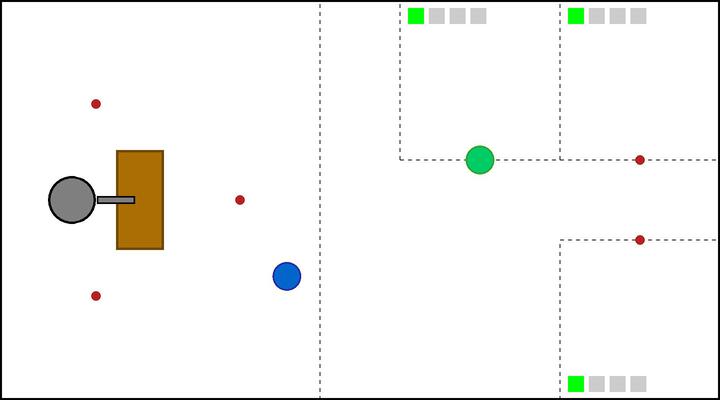}}
    ~
    \centering
    \subcaptionbox{(n) Blue robot successfully delivers a Tool-0 to workshop-2. Arm robot runs \emph{\textbf{Search-Tool(1)}} to find another Tool-1.\vspace{2mm}}
        [0.30\linewidth]{\includegraphics[scale=0.16]{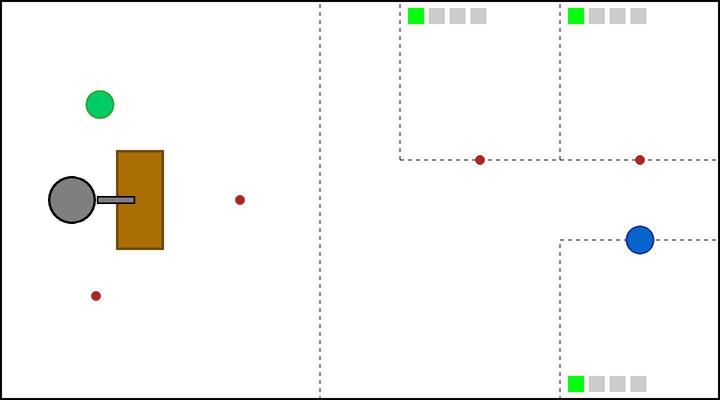}}
    ~
    \centering
    \subcaptionbox{(o) Blue robot runs \emph{\textbf{Get-Tool}} to go back table. All humans finish subtask-0 and start to do subtask-1.\vspace{2mm}}
        [0.30\linewidth]{\includegraphics[scale=0.16]{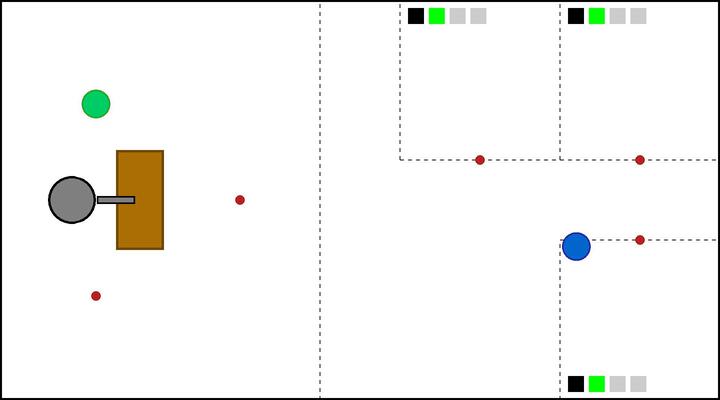}}
    ~
    \subcaptionbox{(p)Arm robot executes \emph{\textbf{Pass-to-M(0)}} to pass a Tool-1 to green robot.  \vspace{2mm}}
        [0.30\linewidth]{\includegraphics[scale=0.16]{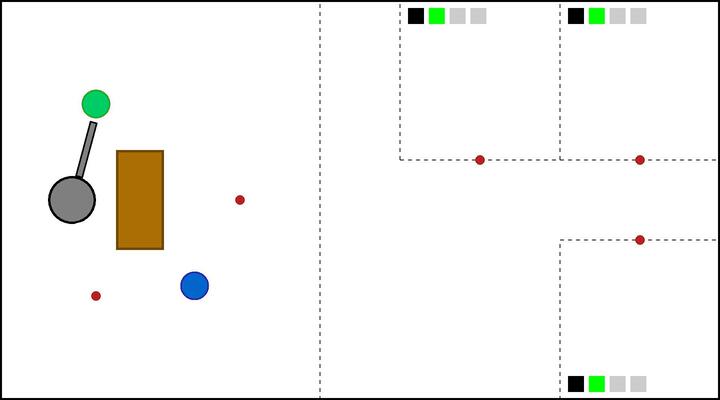}}
    ~
    \centering
    \subcaptionbox{(q) Arm robot executes \emph{\textbf{Pass-to-M(0)}} to pass another Tool-1 to green robot.\vspace{2mm}}
        [0.30\linewidth]{\includegraphics[scale=0.16]{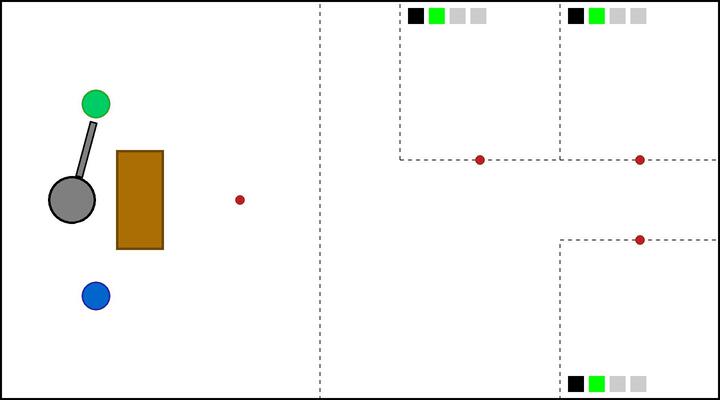}}
    ~
    \centering
    \subcaptionbox{(r) Green robot moves to workshop-1 by executing \emph{\textbf{Go-W(1)}}. Arm robot runs \emph{\textbf{Search-Tool(1)}} to find the 3rd Tool-1.\vspace{2mm}}
        [0.30\linewidth]{\includegraphics[scale=0.16]{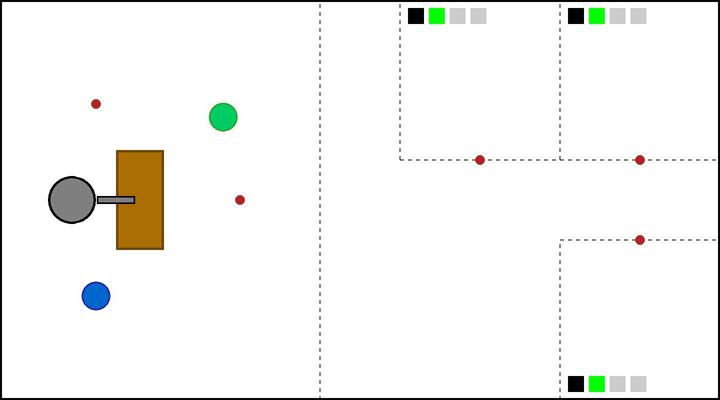}}
\end{figure*}
\begin{figure*}[h!]
    \centering
    \captionsetup[subfigure]{labelformat=empty}          
    ~
    \centering
    \subcaptionbox{(s) Green robot successfully delivers a Tool-0 to workshop-0. \vspace{2mm}}
        [0.30\linewidth]{\includegraphics[scale=0.16]{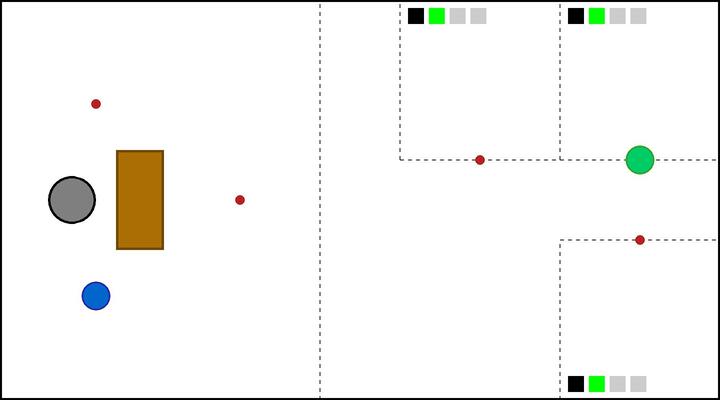}}
    ~
    \centering
    \subcaptionbox{(t) Green robot moves to workshop-0 by executing \emph{\textbf{Go-W(0)}}. Arm robot executes \emph{\textbf{Pass-to-M(1)}} to pass the 3rd Tool-1 to blue robot. \vspace{2mm}}
        [0.30\linewidth]{\includegraphics[scale=0.16]{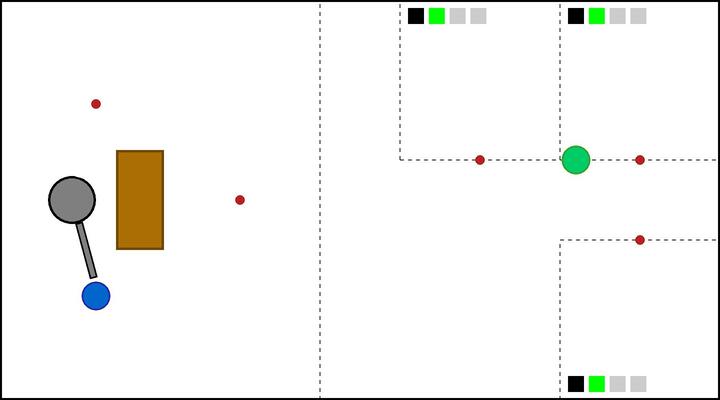}}
    ~
    \centering
    \subcaptionbox{(u) Arm robot runs \emph{\textbf{Search-Tool(2)}} to find Tool-2. Green robot successfully delivers a Tool-1 to workshop-0. Blue robot moves to workshop-2 by executing \emph{\textbf{Go-W(2)}}.\vspace{2mm}}
        [0.30\linewidth]{\includegraphics[scale=0.16]{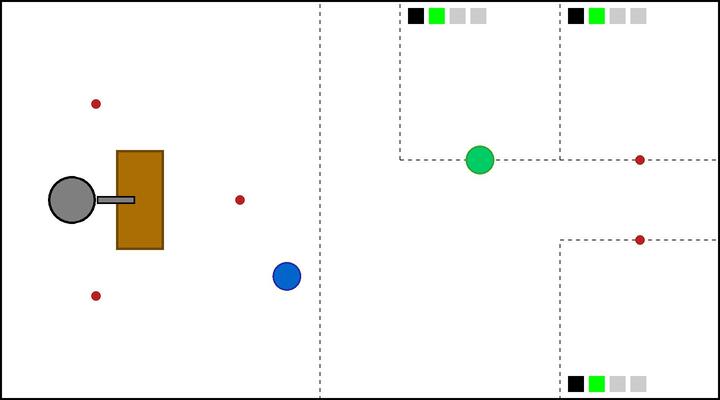}}
    ~
    \centering
    \subcaptionbox{(v) Green robot runs \emph{\textbf{Get-Tool}} to go back table.\vspace{2mm}}
        [0.30\linewidth]{\includegraphics[scale=0.16]{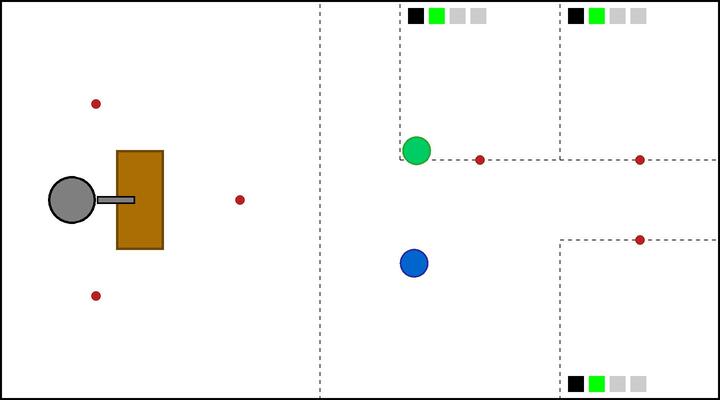}}
    ~
    \centering
    \subcaptionbox{(w) Arm robot runs \emph{\textbf{Search-Tool(2)}} to find another Tool-2. Blue robot successfully delivers a Tool-1 to workshop-2. \vspace{2mm}}
        [0.30\linewidth]{\includegraphics[scale=0.16]{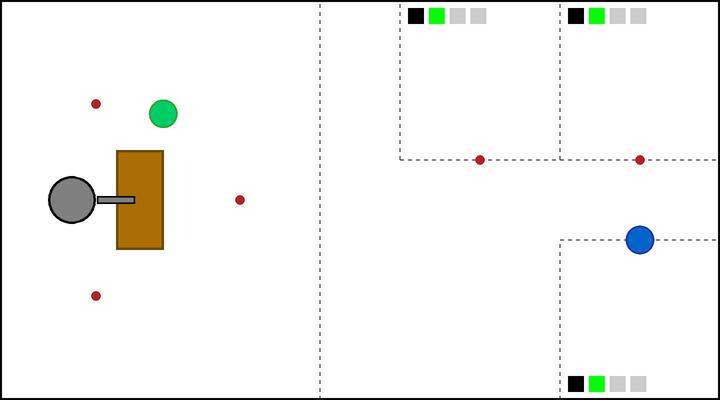}}
    ~
    \centering
    \subcaptionbox{(x) Blue robot runs \emph{\textbf{Get-Tool}} to go back table. \vspace{2mm}}
        [0.30\linewidth]{\includegraphics[scale=0.16]{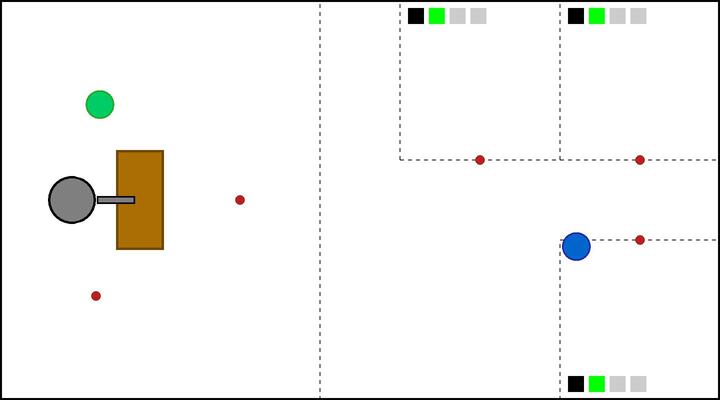}}
    ~
    \centering
    \subcaptionbox{(y) Arm robot executes \emph{\textbf{Pass-to-M(0)}} to pass a Tool-2 to green robot.\vspace{2mm}}
        [0.30\linewidth]{\includegraphics[scale=0.16]{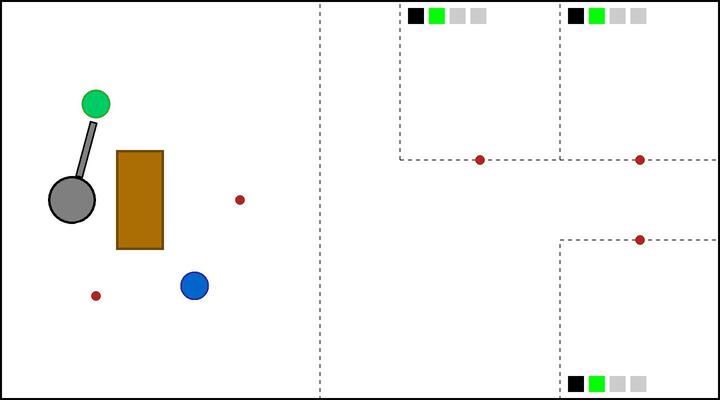}}  
    ~
    \centering
    \subcaptionbox{(z) Arm robot executes \emph{\textbf{Pass-to-M(0)}} to pass another Tool-2 to green robot. All humans finish subtask-1 and start to do subtask-2.\vspace{2mm}}
        [0.30\linewidth]{\includegraphics[scale=0.16]{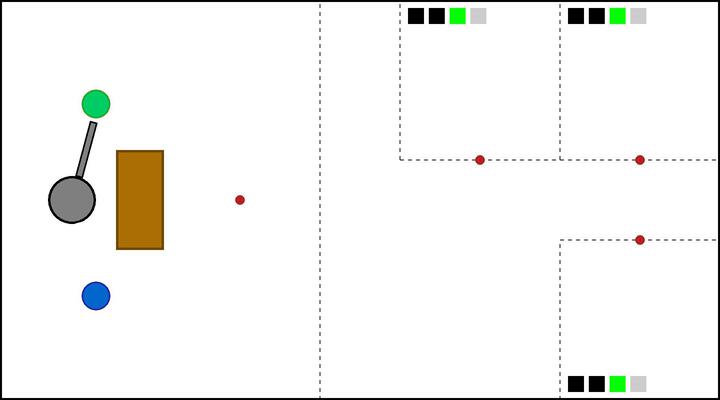}}
    ~
    \centering
    \subcaptionbox{(A) Arm robot runs \emph{\textbf{Search-Tool(2)}} to find the 3rd Tool-2. Green robot moves to workshop-1 by executing \emph{\textbf{Go-W(1)}}.\vspace{2mm}}
        [0.30\linewidth]{\includegraphics[scale=0.16]{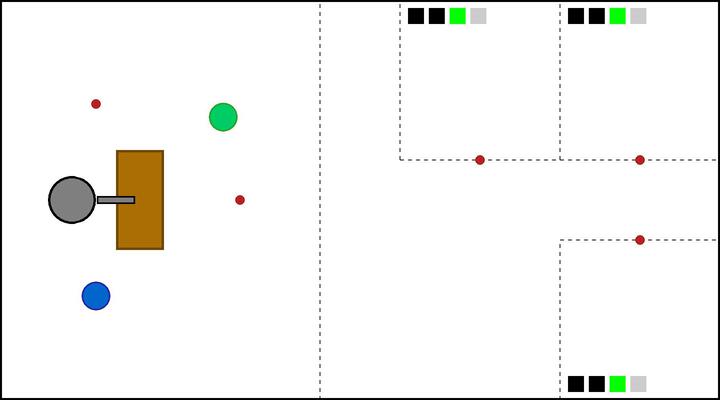}}
\end{figure*}
\begin{figure*}[h!]
    \centering
    \captionsetup[subfigure]{labelformat=empty}          
    ~
    \centering
    \subcaptionbox{(B) Green robot successfully delivers a Tool-2 to workshop-1. \vspace{2mm}}
        [0.30\linewidth]{\includegraphics[scale=0.16]{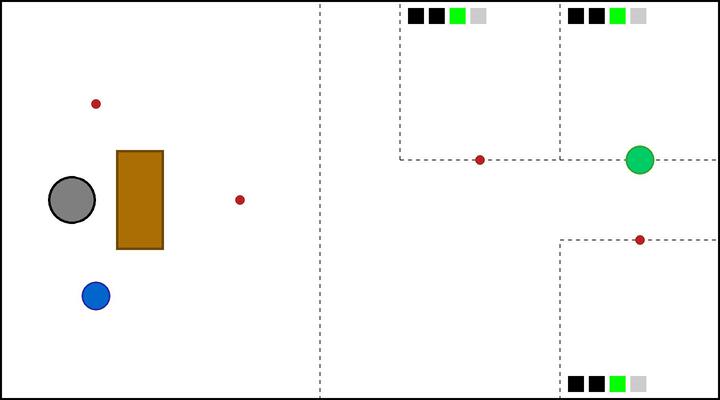}}
    ~
    \centering
    \subcaptionbox{(C) Arm robot executes \emph{\textbf{Pass-to-M(1)}} to pass the 3rd Tool-2 to blue robot.\vspace{2mm}}
        [0.30\linewidth]{\includegraphics[scale=0.16]{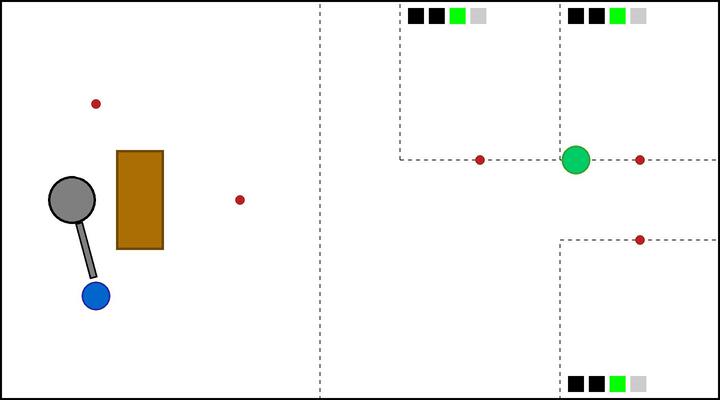}}  
    ~
    \centering
    \subcaptionbox{(D) Green robot successfully delivers a Tool-2 to workshop-0. Blue robot moves to workshop-2 by executing \emph{\textbf{Go-W(2)}}.\vspace{2mm}}
        [0.30\linewidth]{\includegraphics[scale=0.16]{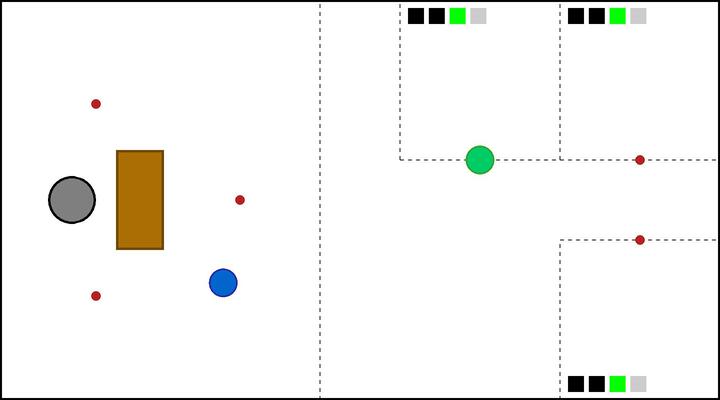}}
    ~
    \centering
    \subcaptionbox{(E) Blue robot successfully delivers a Tool-2 to workshop-2. Humans have received all tools, and for robots, the task is done.\vspace{2mm}}
        [0.9\linewidth]{\includegraphics[scale=0.16]{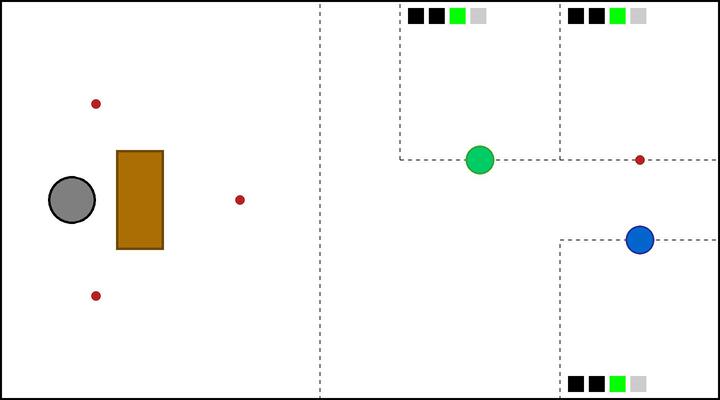}}
    \label{wtd_c_behavior}
\end{figure*}

\clearpage
\textbf{\emph{Warehouse-D:}}  

\begin{figure*}[h!]
    \centering
    \captionsetup[subfigure]{labelformat=empty}
    \centering
    \subcaptionbox{(a) Initial State.\vspace{2mm}}
        [0.30\linewidth]{\includegraphics[scale=0.24]{fig/paper3/WTD/wtd_e_small.png}}
    ~
    \centering
    \subcaptionbox{(b) Mobile robots move towards the table by running \emph{\textbf{Get-Tool}}. Arm robot runs \emph{\textbf{Search-Tool(0)}} to find the 1st Tool-0.\vspace{2mm}}
        [0.30\linewidth]{\includegraphics[scale=0.16]{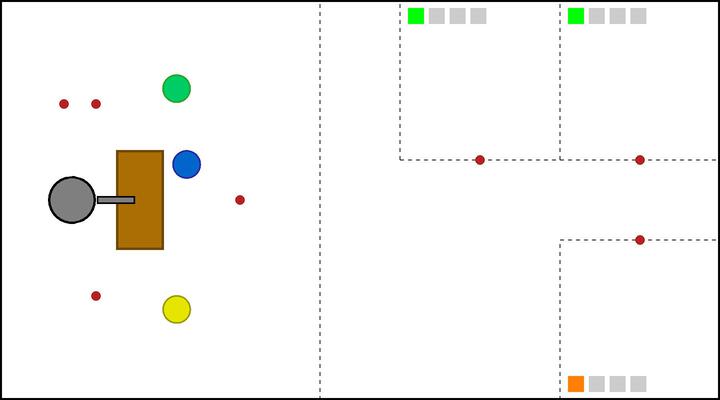}}
    ~
    \centering
    \subcaptionbox{(c) Mobile robots wait there and the robot arm keeps looking for the 1st Tool-0.\vspace{2mm}}
        [0.30\linewidth]{\includegraphics[scale=0.16]{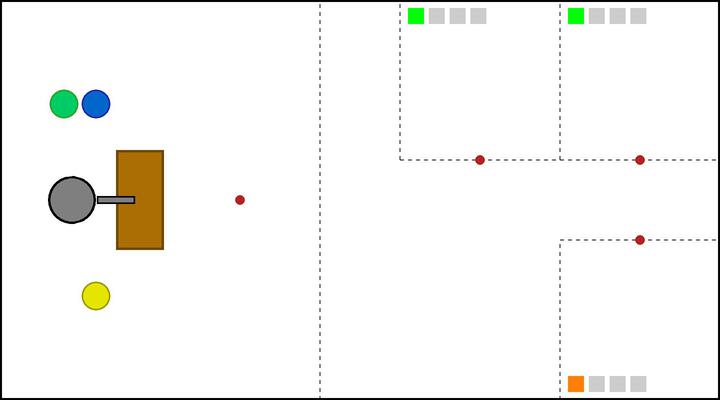}}
    ~
    \centering
    \subcaptionbox{(d) Arm robot executes \emph{\textbf{Pass-to-M(0)}} to pass a Tool-0 to green robot.\vspace{2mm}}
        [0.30\linewidth]{\includegraphics[scale=0.16]{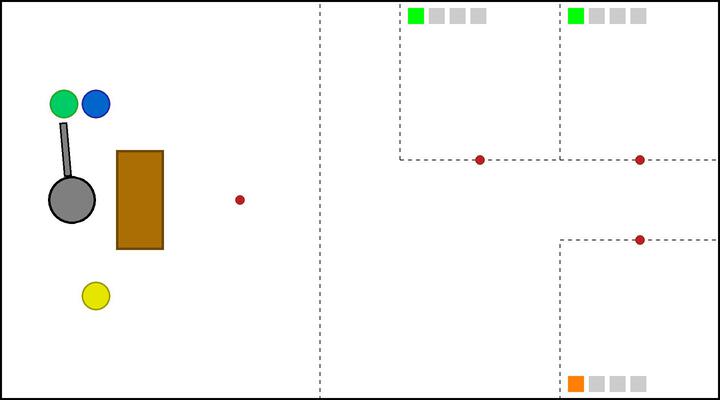}}
    ~
    \centering
    \subcaptionbox{(e) Arm robot runs \emph{\textbf{Search-Tool(0)}} to find the 2nd Tool-0. Green robot moves to workshop-0 by executing \emph{\textbf{Go-W(0)}}.\vspace{2mm}}
        [0.30\linewidth]{\includegraphics[scale=0.16]{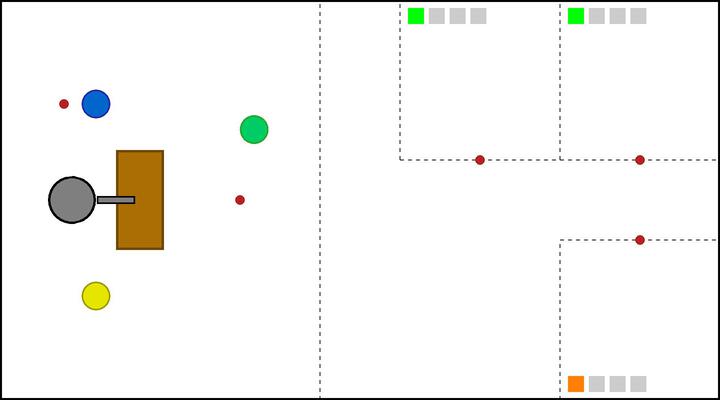}}  
    ~
    \centering
    \subcaptionbox{(f) Green robot successfully delivers Tool-0 to workshop-0.  \vspace{2mm}}
        [0.30\linewidth]{\includegraphics[scale=0.16]{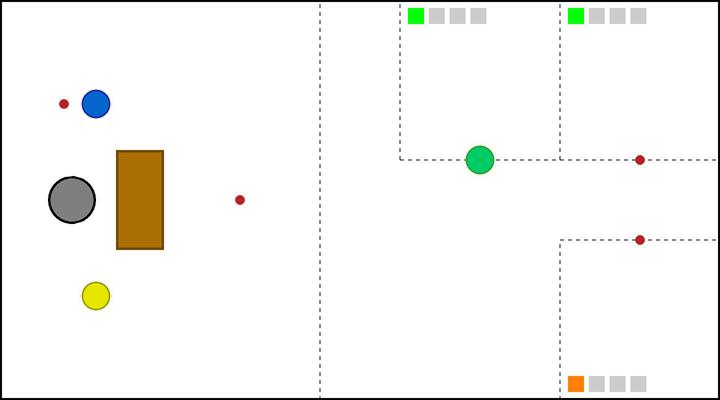}}
    ~
    \centering
    \subcaptionbox{(g) Green robot runs \emph{\textbf{Get-Tool}} to go back table. Arm robot executes \emph{\textbf{Pass-to-M(1)}} to pass a Tool-0 to blue robot.
\vspace{2mm}}
        [0.30\linewidth]{\includegraphics[scale=0.16]{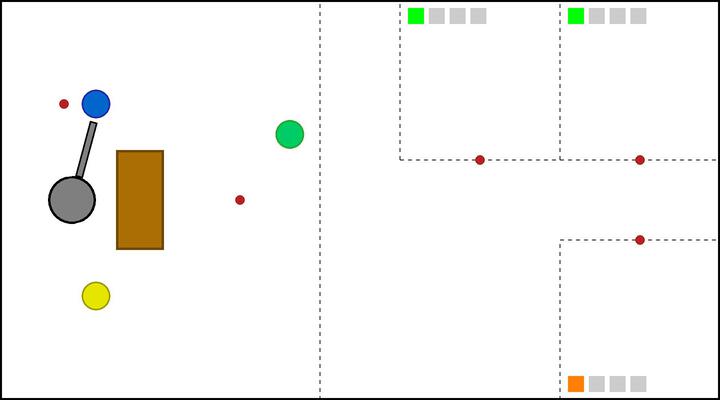}}
    ~
    \centering
    \subcaptionbox{(h) Arm robot runs \emph{\textbf{Search-Tool(0)}} to find the 3rd Tool-0. Blue robot moves to workshop-1 by executing \emph{\textbf{Go-W(1)}}. Yellow robot moves to workshop-0 by executing \emph{\textbf{Go-W(0)}}. \vspace{2mm}}
        [0.30\linewidth]{\includegraphics[scale=0.16]{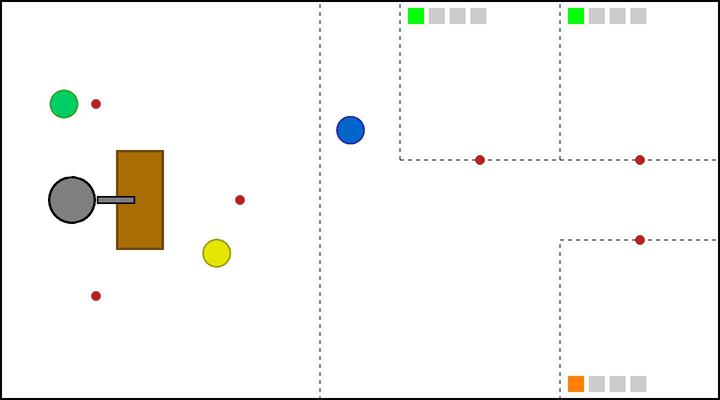}}
    ~
    \centering
    \subcaptionbox{(i) Blue robot successfully delivers a Tool-0 to workshop-1. Yellow robot reaches workshop-0 and observes that human-0 has got Tool-0. Human-2 finishes subtask-0 and waits for Tool-0.
    \vspace{2mm}}
        [0.30\linewidth]{\includegraphics[scale=0.16]{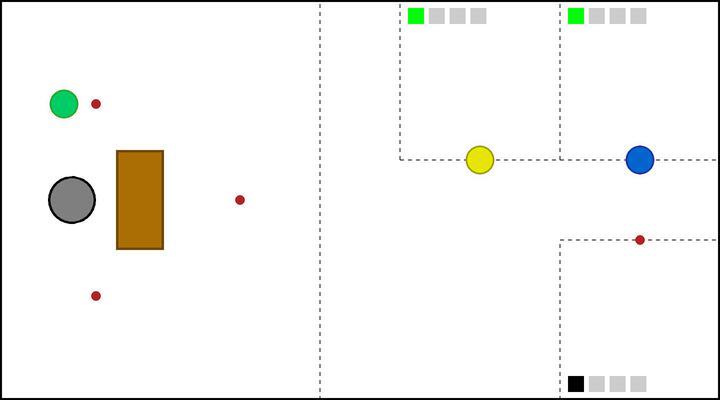}}
\end{figure*}
\begin{figure*}[h!]
    \centering
    \captionsetup[subfigure]{labelformat=empty}          
    ~
    \centering
    \subcaptionbox{(j) Arm robot executes \emph{\textbf{Pass-to-M(0)}} to pass a Tool-0 to green robot. Yellow and blue robots run \emph{\textbf{Get-Tool}} to go back table.\vspace{2mm}}
        [0.30\linewidth]{\includegraphics[scale=0.16]{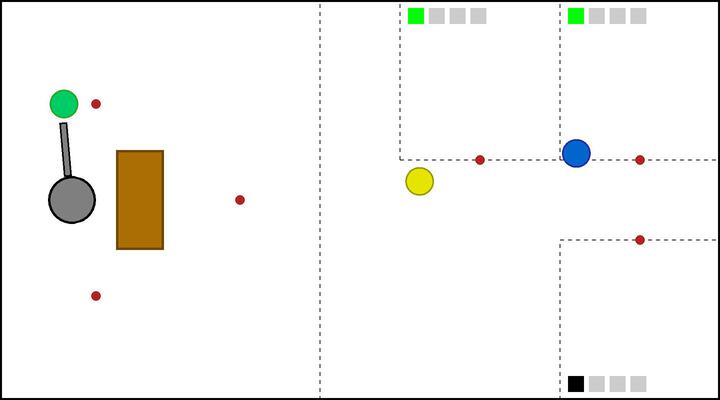}}
    ~
    \centering
    \subcaptionbox{(k) Arm robot runs \emph{\textbf{Search-Tool(1)}} to find the 1st Tool-1. Green robot moves to workshop-0 by executing \emph{\textbf{Go-W(0)}}. \vspace{2mm}}
        [0.30\linewidth]{\includegraphics[scale=0.16]{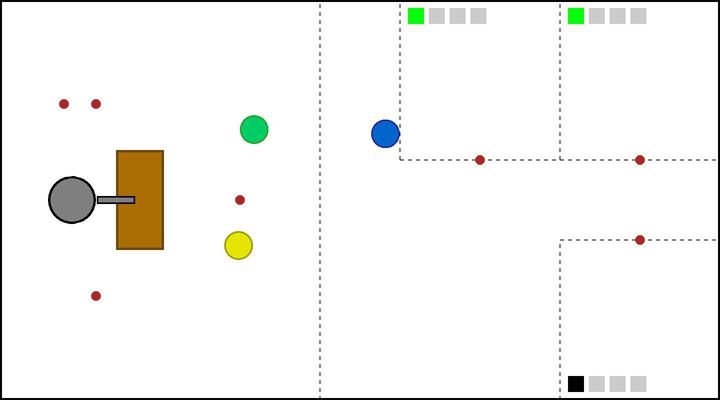}}  
    ~
    \centering
    \subcaptionbox{(l) Green robot reaches workshop-0 and observes that human-0 does not need Tool-0.  \vspace{2mm}}
        [0.30\linewidth]{\includegraphics[scale=0.16]{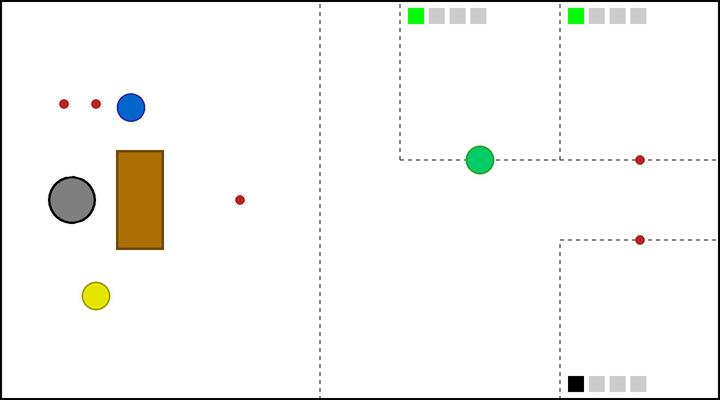}}
    \centering
    \subcaptionbox{(m) Green robot runs \emph{\textbf{Get-Tool}} to go back table. \vspace{2mm}}
        [0.30\linewidth]{\includegraphics[scale=0.16]{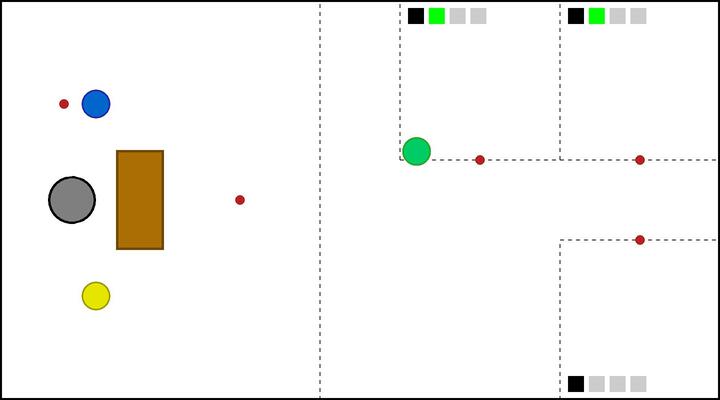}}
    ~
    \centering
    \subcaptionbox{(n) Arm robot executes \emph{\textbf{Pass-to-M(2)}} to pass a Tool-1 to yellow robot.\vspace{2mm}}
        [0.30\linewidth]{\includegraphics[scale=0.16]{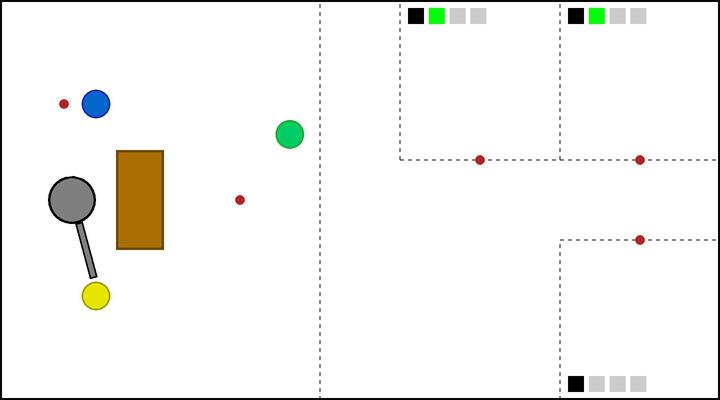}}
    ~
    \centering
    \subcaptionbox{(o) Arm robot runs \emph{\textbf{Search-Tool(1)}} to find the 2nd Tool-1. Yellow robot moves to workshop-0. \vspace{2mm}}
        [0.30\linewidth]{\includegraphics[scale=0.16]{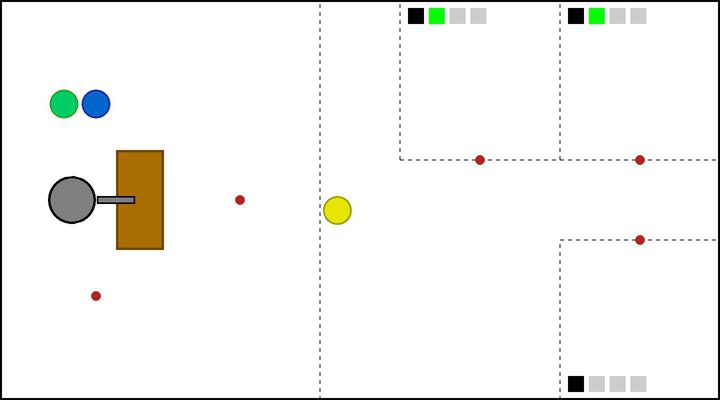}}
    ~
    \subcaptionbox{(p) Yellow robot successfully delivers a Tool-1 to workshop-0.   \vspace{2mm}}
        [0.30\linewidth]{\includegraphics[scale=0.16]{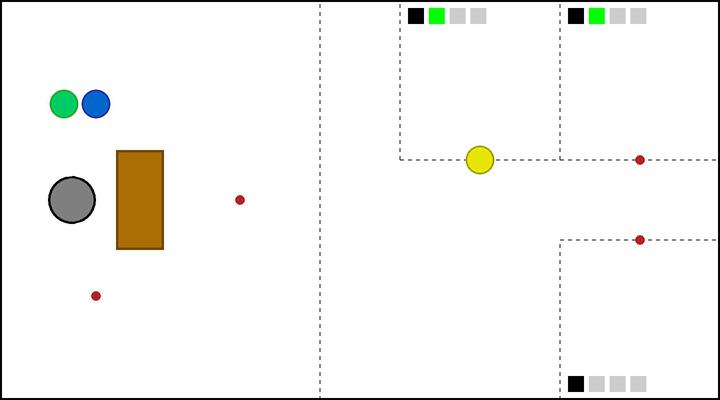}}
    ~
    \centering
    \subcaptionbox{(q) Yellow robot moves to workshop-1 by executing \emph{\textbf{Go-W(1)}}. \vspace{2mm}}
        [0.30\linewidth]{\includegraphics[scale=0.16]{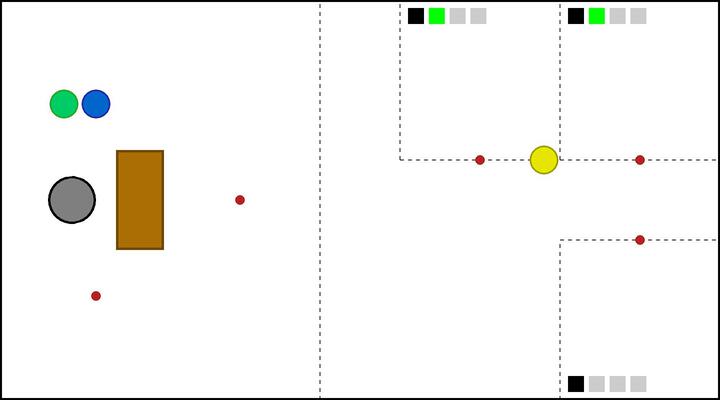}}
    ~
    \centering
    \subcaptionbox{(r) Arm robot executes \emph{\textbf{Pass-to-M(0)}} to pass a Tool-1 to green robot. \vspace{2mm}}
        [0.30\linewidth]{\includegraphics[scale=0.16]{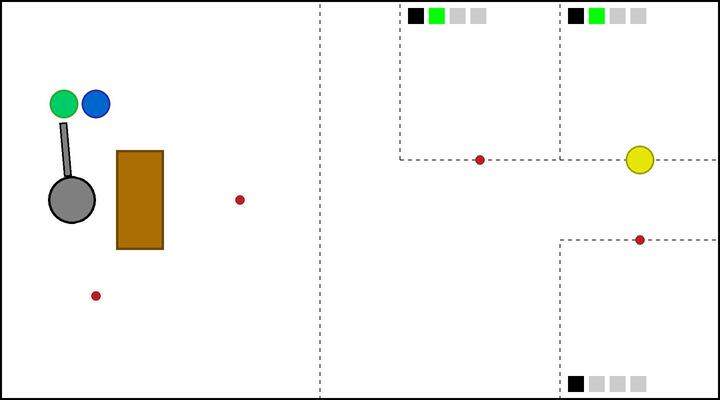}}
    ~
    \centering
    \subcaptionbox{(s) Arm robot runs \emph{\textbf{Search-Tool(1)}} to find 3st Tool-1. Yellow robot runs \emph{\textbf{Get-Tool}} to go back table. Green robot moves to workshop-0 by executing \emph{\textbf{Go-W(0)}}. \vspace{2mm}}
        [0.30\linewidth]{\includegraphics[scale=0.16]{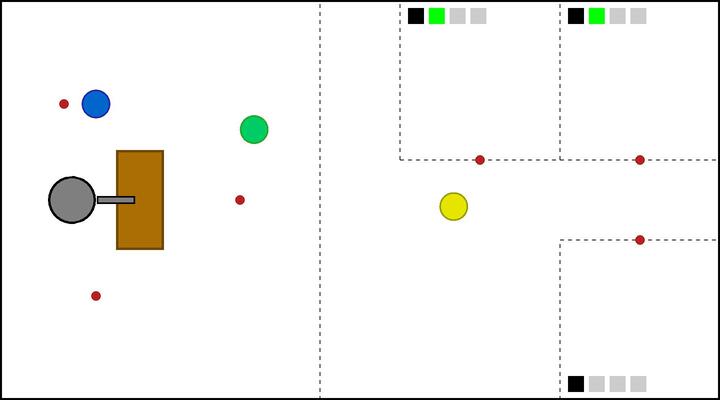}}
    ~
    \centering
    \subcaptionbox{(t) Green robot successfully delivers a Tool-1 to workshop-0.  \vspace{2mm}}
        [0.30\linewidth]{\includegraphics[scale=0.16]{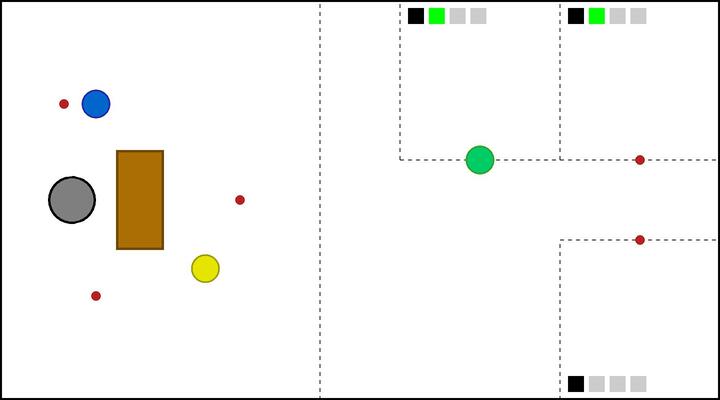}}
    ~
    \centering
    \subcaptionbox{(u) Green robot moves to workshop-2 by executing \emph{\textbf{Go-W(2)}}.\vspace{2mm}}
        [0.30\linewidth]{\includegraphics[scale=0.16]{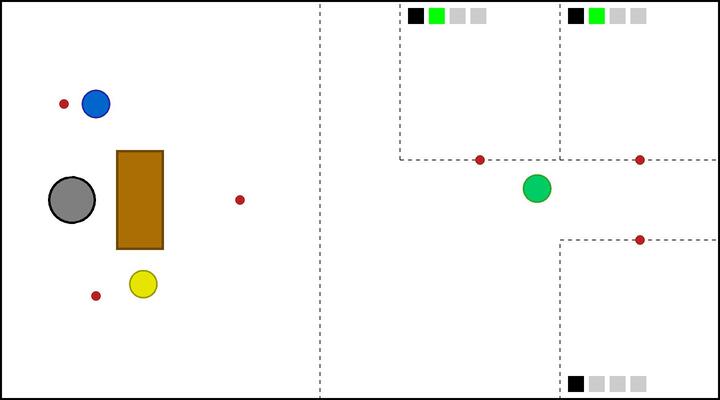}}
\end{figure*}
\begin{figure*}[h!]
    \centering
    \captionsetup[subfigure]{labelformat=empty}          
    ~
    \centering
    \subcaptionbox{(v) Green robot successfully delivers a Tool-0 to workshop-2. Human-2 finishes subtask-0 and starts to do subtask-1. Arm robot executes \emph{\textbf{Pass-to-M(1)}} to pass a Tool-1 to blue robot.\vspace{2mm}}
        [0.30\linewidth]{\includegraphics[scale=0.16]{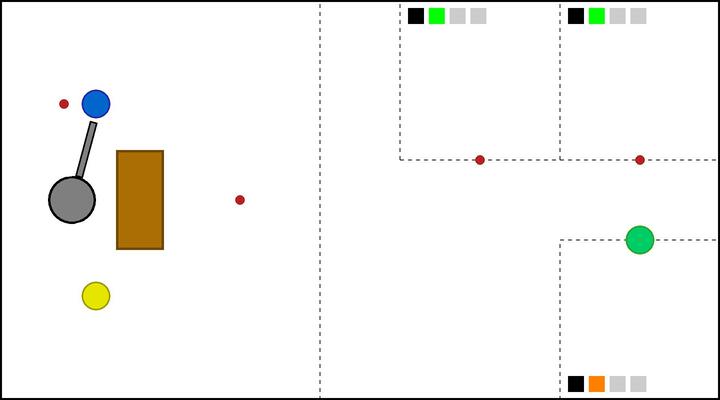}}
    ~
    \centering
    \subcaptionbox{(w) Green robot runs \emph{\textbf{Get-Tool}} to go back table.  Arm robot runs \emph{\textbf{Search-Tool(2)}} to find the 1st Tool-2. Blue robot moves to workshop-1 by executing \emph{\textbf{Go-W(1)}}. \vspace{2mm}}
        [0.30\linewidth]{\includegraphics[scale=0.16]{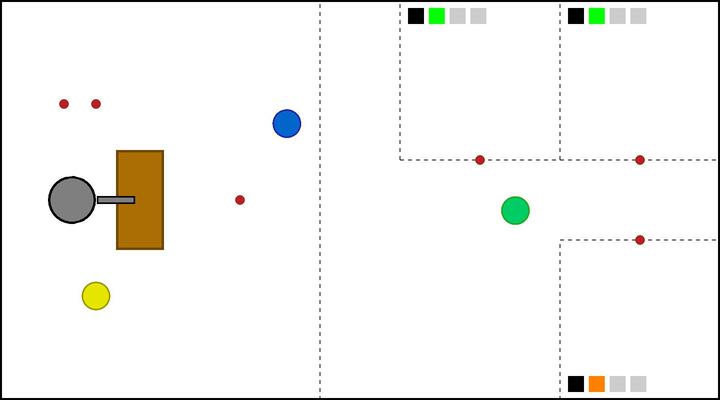}}
    ~
    \centering
    \subcaptionbox{(x) Blue robot successfully delivers a Tool-1 to workshop-1.  \vspace{2mm}}
        [0.30\linewidth]{\includegraphics[scale=0.16]{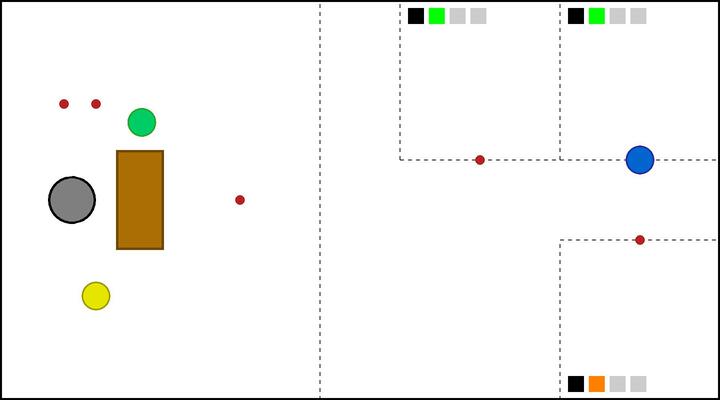}}
    ~
    \centering
    \subcaptionbox{(y) Blue robot runs \emph{\textbf{Get-Tool}} to go back table. Arm robot executes \emph{\textbf{Pass-to-M(2)}} to pass a Tool-2 to yellow robot.\vspace{2mm}}
        [0.30\linewidth]{\includegraphics[scale=0.16]{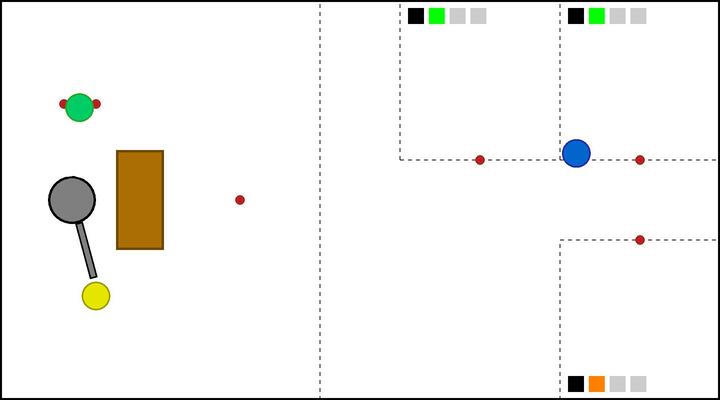}}  
    ~
    \centering
    \subcaptionbox{(z) Arm robot runs \emph{\textbf{Search-Tool(2)}} to find the 2nd Tool-2. Yellow robot moves to workshop-1 by executing \emph{\textbf{Go-W(1)}}.\vspace{2mm}}
        [0.30\linewidth]{\includegraphics[scale=0.16]{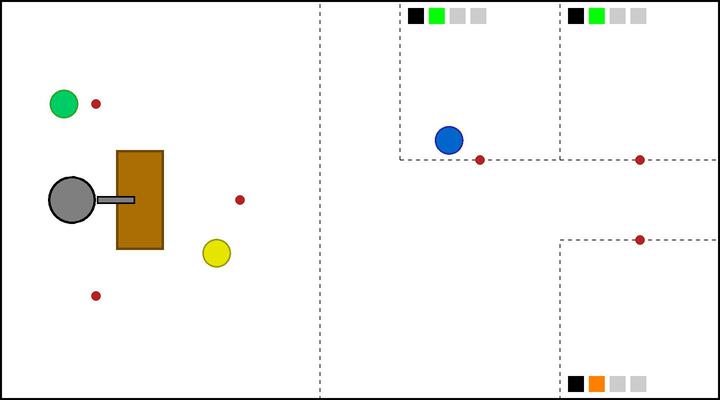}}
    ~
    \centering
    \subcaptionbox{(A) Yellow robot successfully delivers a Tool-2 to workshop-0. Human-0 and human-1 finish subtask-1 and start to do subtask-2. \vspace{2mm}}
        [0.30\linewidth]{\includegraphics[scale=0.16]{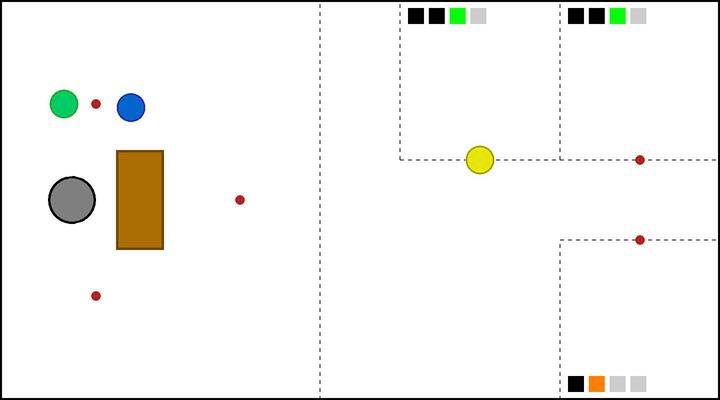}}
    \centering
    \subcaptionbox{(B) Yellow robot moves to workshop-1 by executing \emph{\textbf{Go-W(1)}}. \vspace{2mm}}
        [0.30\linewidth]{\includegraphics[scale=0.16]{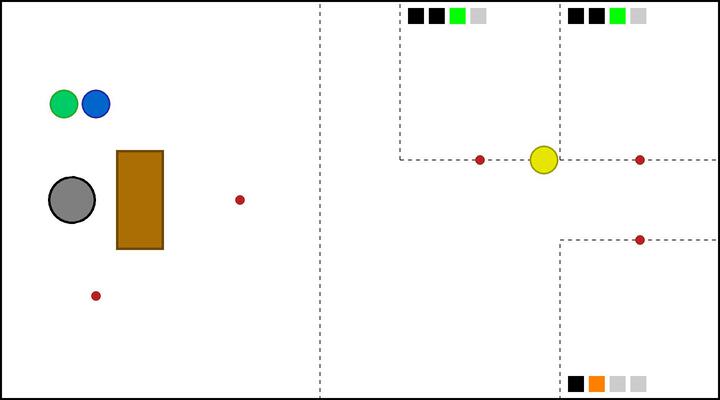}}
    ~
    \centering
    \subcaptionbox{(C) Arm robot executes \emph{\textbf{Pass-to-M(0)}} to pass a Tool-2 to green robot. Yellow robot reaches workshop-1 but it does not have any tools.\vspace{2mm}}
        [0.30\linewidth]{\includegraphics[scale=0.16]{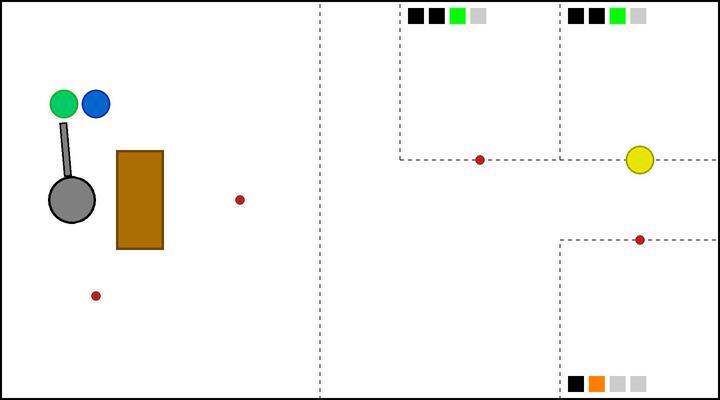}}  
    ~
    \centering
    \subcaptionbox{(D) Arm robot runs \emph{\textbf{Search-Tool(2)}} to find the 3rd Tool-2. Green robot moves to workshop-0 by executing \emph{\textbf{Go-W(0)}}. Yellow robot runs \emph{\textbf{Get-Tool}} to go back table.\vspace{2mm}}
        [0.30\linewidth]{\includegraphics[scale=0.16]{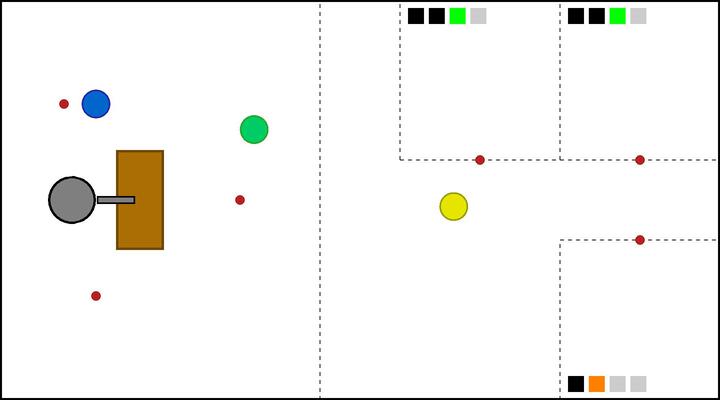}}
\end{figure*}
\begin{figure*}[h!]
    \centering
    \captionsetup[subfigure]{labelformat=empty}
    \centering
    ~
    \centering
    \subcaptionbox{(E) Green robot reaches workshop-0 and observes that human-0 does not need Tool-2.  Human-2 finishes subtask-1 and starts to do subtask-2.\vspace{2mm}}
        [0.30\linewidth]{\includegraphics[scale=0.16]{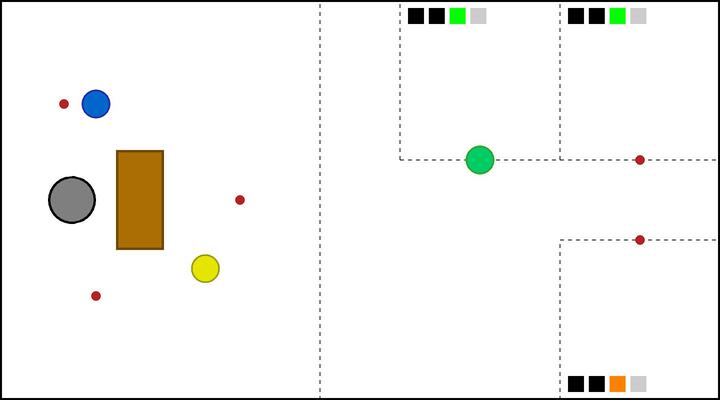}}
    ~
    \centering
    \subcaptionbox{(F) Green robot moves to workshop-2 by executing \emph{\textbf{Go-W(2)}}.\vspace{2mm}}
        [0.30\linewidth]{\includegraphics[scale=0.16]{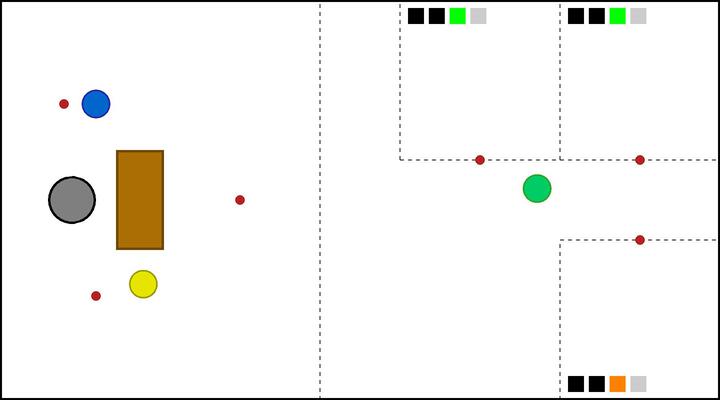}}  
    ~
    \centering
    \subcaptionbox{(G) Green robot successfully delivers a Tool-2 to workshop-2. Arm robot executes \emph{\textbf{Pass-to-M(1)}} to pass a Tool-2 to blue robot. \vspace{2mm}}
        [0.30\linewidth]{\includegraphics[scale=0.16]{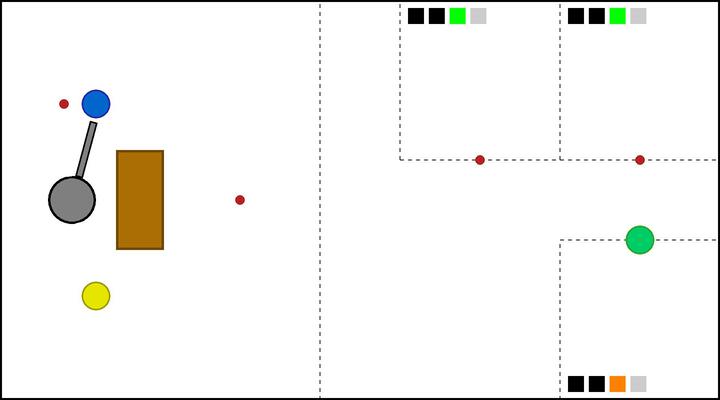}}
    ~
    \centering
    \subcaptionbox{(H) Blue robot moves to workshop-1 by executing \emph{\textbf{Go-W(1)}}.\vspace{2mm}}
        [0.35\linewidth]{\includegraphics[scale=0.16]{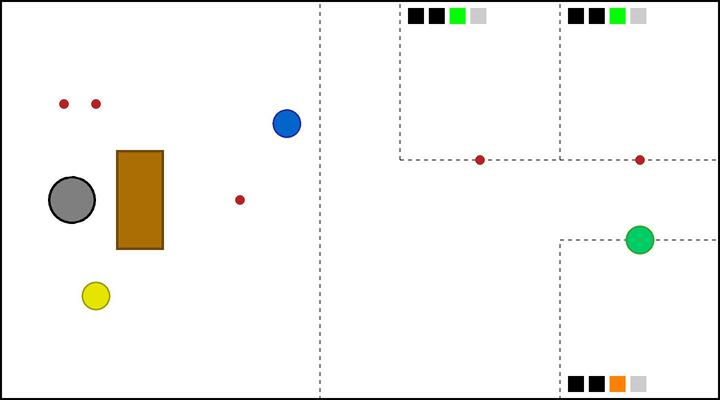}}
    ~
    \centering
    \subcaptionbox{(I) Blue robot successfully delivers a Tool-2 to workshop-1. Humans have received all tools, and for robots, the task is done.\vspace{2mm}}
        [0.35\linewidth]{\includegraphics[scale=0.16]{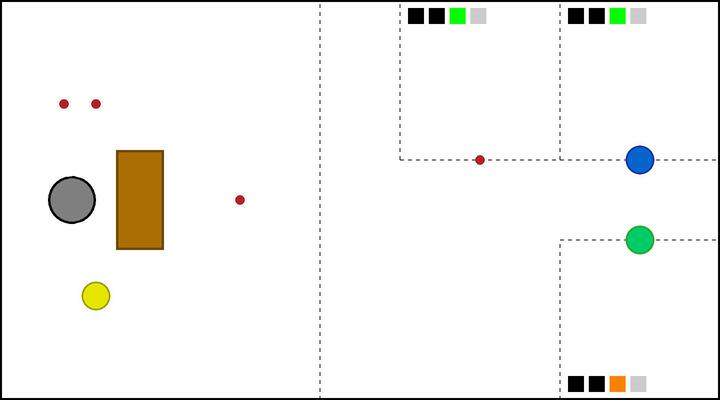}}
    \label{wtd_e_behavior}
\end{figure*}

\clearpage
\textbf{\emph{Warehouse-E:}}  

\begin{figure*}[h!]
    \centering
    \captionsetup[subfigure]{labelformat=empty}
    \centering
    \subcaptionbox{(a) Initial State.\vspace{2mm}}
        [0.30\linewidth]{\includegraphics[scale=0.24]{fig/paper3/WTD/wtd_d_small.png}}
    ~
    \centering
    \subcaptionbox{(b) Green and yellow robots move towards the table by running \emph{\textbf{Get-Tool}}. Blue robot moves to workshop-3 by executing \emph{\textbf{Go-W(3)}}. Arm robot runs \emph{\textbf{Search-Tool(0)}} to find the 1st Tool-0.\vspace{2mm}}
        [0.30\linewidth]{\includegraphics[scale=0.16]{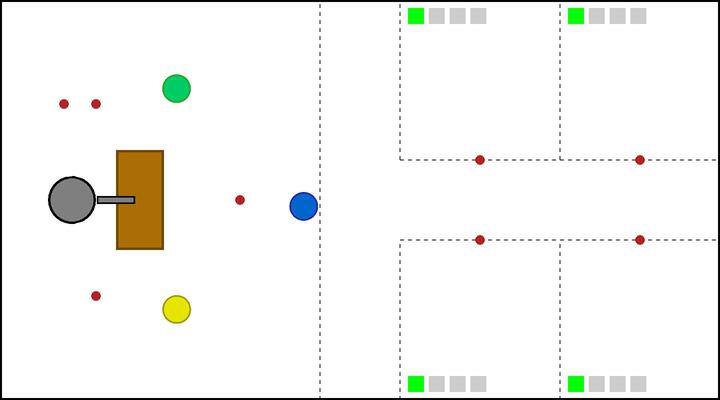}}
    ~
    \centering
    \subcaptionbox{(c) Blue robot reaches workshop-3.\vspace{2mm}}
        [0.30\linewidth]{\includegraphics[scale=0.16]{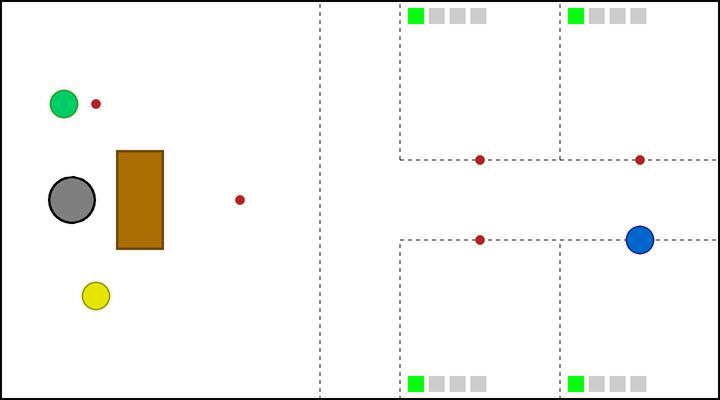}}
    ~
    \centering
    \subcaptionbox{(d) Blue robot moves to workshop-1 by executing \emph{\textbf{Go-W(1)}}.\vspace{2mm}}
        [0.30\linewidth]{\includegraphics[scale=0.16]{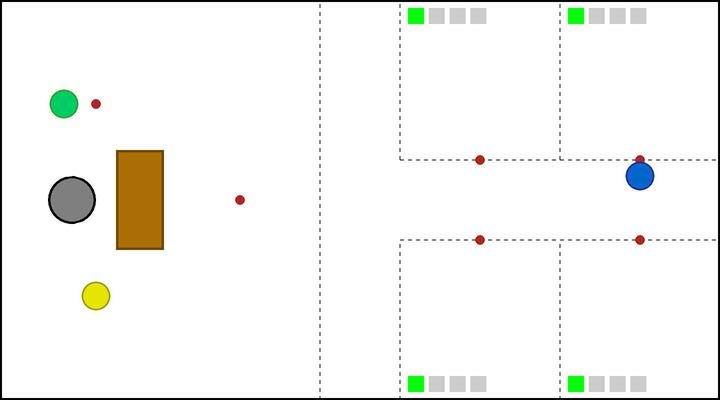}}
    ~
    \centering
    \subcaptionbox{(e) Blue robot reaches workshop-1. Arm robot executes \emph{\textbf{Pass-to-M(2)}} to pass a Tool-0 to yellow robot. \vspace{2mm}}
        [0.30\linewidth]{\includegraphics[scale=0.16]{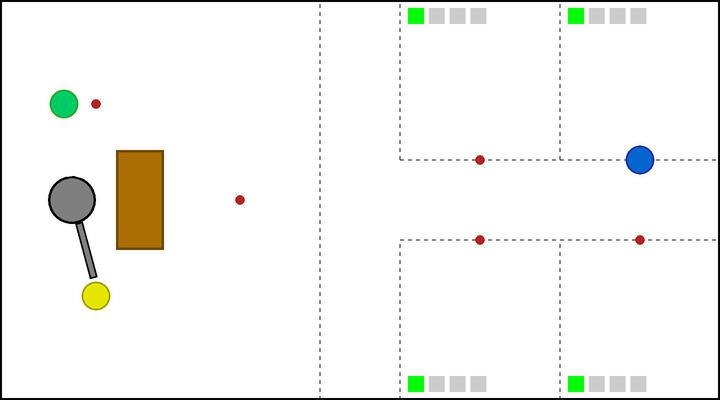}}  
    ~
    \centering
    \subcaptionbox{(f) Arm robot runs \emph{\textbf{Search-Tool(0)}} to find the 2nd Tool-0. Blue robot runs \emph{\textbf{Get-Tool}} to go back table. Yellow robot moves to workshop-1 by executing \emph{\textbf{Go-W(1)}}. \vspace{2mm}}
        [0.30\linewidth]{\includegraphics[scale=0.16]{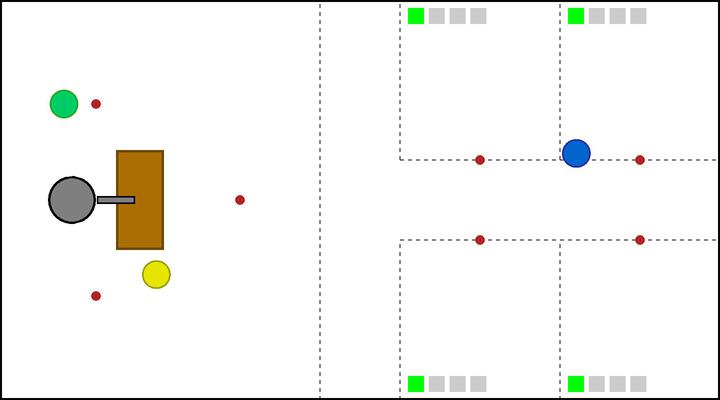}}
    ~
    \centering
    \subcaptionbox{(g)  Yellow robot successfully delivers a Tool-0 to workshop-0. \vspace{2mm}}
        [0.30\linewidth]{\includegraphics[scale=0.16]{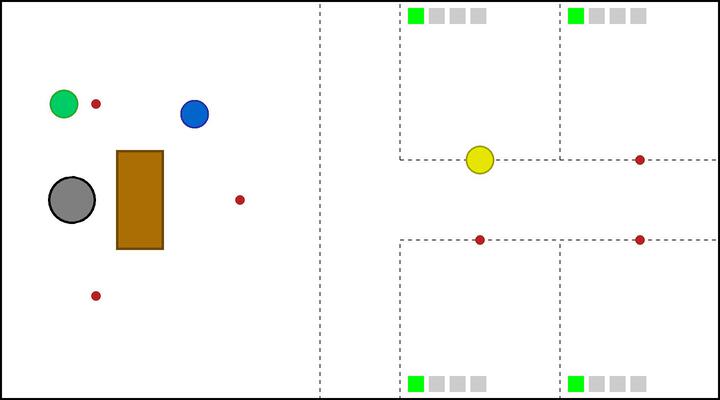}}
    ~
    \centering
    \subcaptionbox{(h) Arm robot executes \emph{\textbf{Pass-to-M(0)}} to pass a Tool-0 to green robot. Yellow robot runs \emph{\textbf{Get-Tool}} to go back table.  \vspace{2mm}}
        [0.30\linewidth]{\includegraphics[scale=0.16]{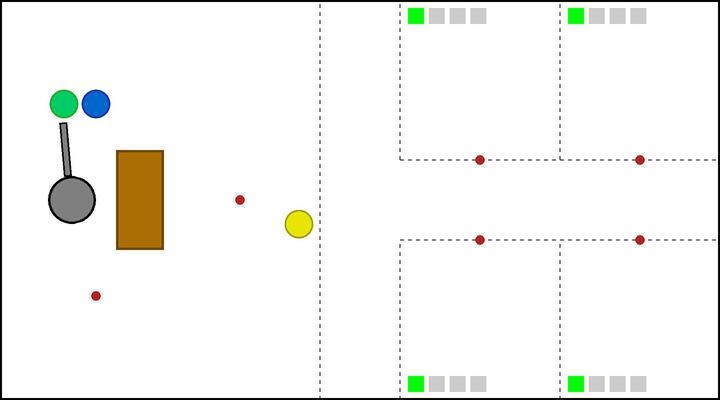}}
    ~
    \centering
    \subcaptionbox{(i) Arm robot runs \emph{\textbf{Search-Tool(0)}} to find the 3rd Tool-0. Green robot moves to workshop-1 by executing \emph{\textbf{Go-W(1)}}.
    \vspace{2mm}}
        [0.30\linewidth]{\includegraphics[scale=0.16]{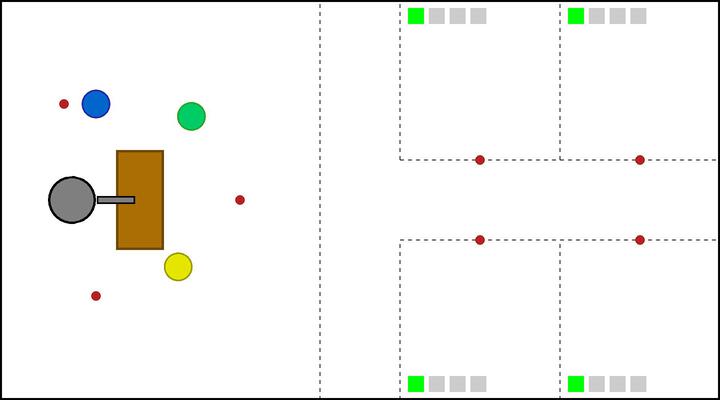}}
\end{figure*}
\begin{figure*}[h!]
    \centering
    \captionsetup[subfigure]{labelformat=empty}          
    ~
    \centering
    \subcaptionbox{(j) Green robot successfully delivers a Tool-0 to workshop-1. Arm robot executes \emph{\textbf{Pass-to-M(2)}} to pass a Tool-0 to yellow robot.\vspace{2mm}}
        [0.30\linewidth]{\includegraphics[scale=0.16]{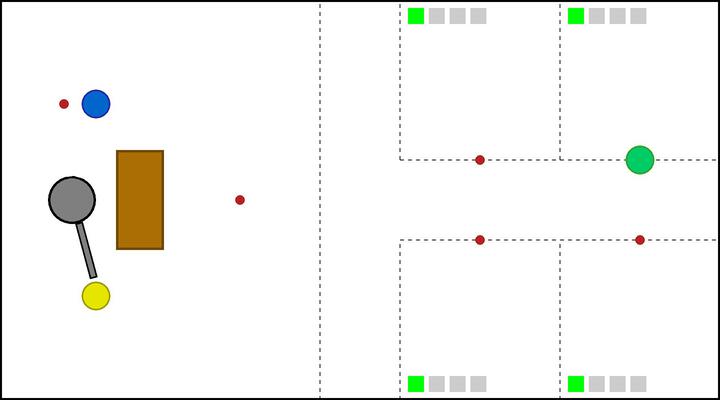}}
    ~
    \centering
    \subcaptionbox{(k) Arm robot runs \emph{\textbf{Search-Tool(1)}} to find the 1st Tool-1. Yellow robot moves to workshop-2 by executing \emph{\textbf{Go-W(2)}}. Green robot runs \emph{\textbf{Get-Tool}} to go back table. \vspace{2mm}}
        [0.30\linewidth]{\includegraphics[scale=0.16]{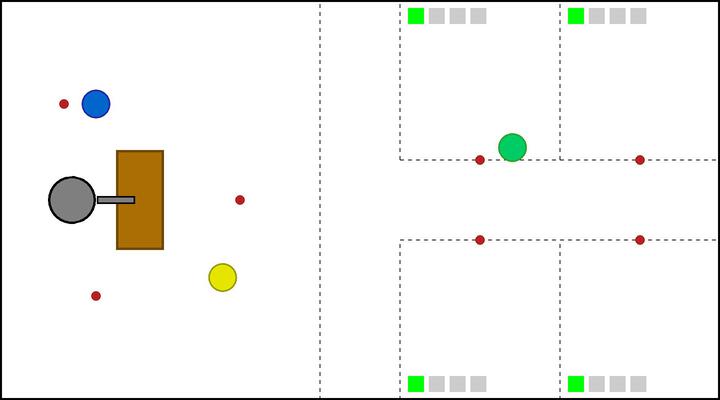}}  
    ~
    \centering
    \subcaptionbox{(l) Yellow robot successfully delivers a Tool-0 to workshop-2.  \vspace{2mm}}
        [0.30\linewidth]{\includegraphics[scale=0.16]{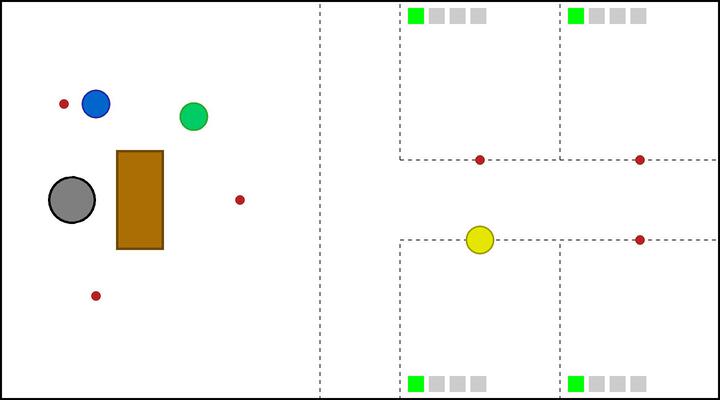}}
    ~
    \centering
    \subcaptionbox{(m) Yellow robot runs \emph{\textbf{Get-Tool}} to go back table. Arm robot executes \emph{\textbf{Pass-to-M(1)}} to pass the a Tool-1 to blue robot. Human-0, human-1, and human-2 finish subtask-0 and start to do subtask-1.\vspace{2mm}}
        [0.30\linewidth]{\includegraphics[scale=0.16]{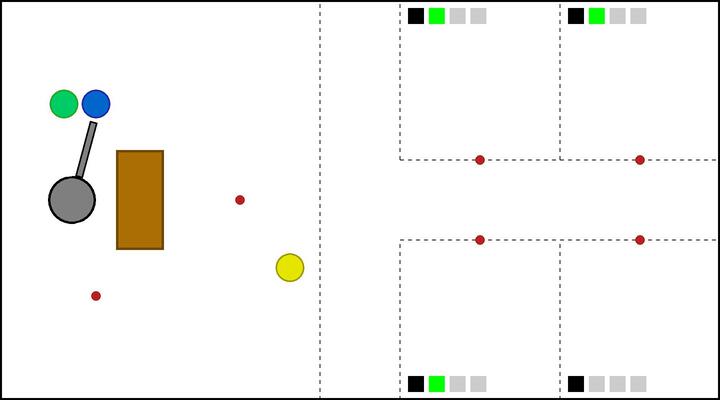}}
    ~
    \centering
    \subcaptionbox{(n) Arm robot runs \emph{\textbf{Search-Tool(1)}} to find the 2nd Tool-1. Blue robot moves to workshop-1 by executing \emph{\textbf{Go-W(1)}}.\vspace{2mm}}
        [0.30\linewidth]{\includegraphics[scale=0.16]{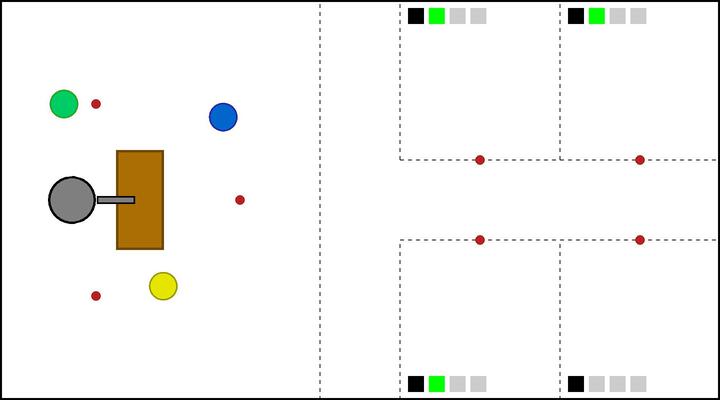}}
    ~
    \centering
    \subcaptionbox{(o) Blue robot successfully delivers a Tool-1 to workshop-1.  \vspace{2mm}}
        [0.30\linewidth]{\includegraphics[scale=0.16]{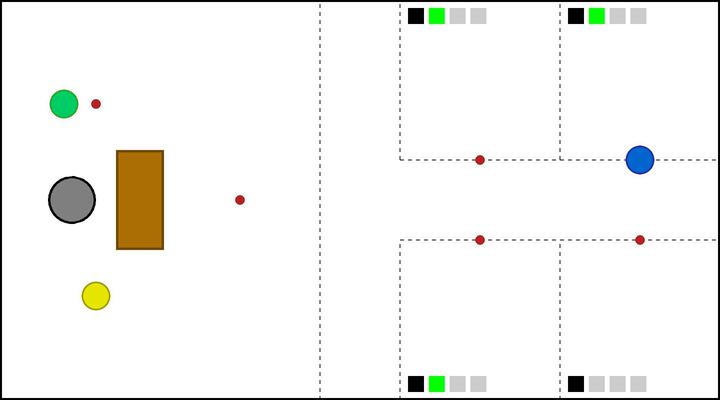}}
    ~
    \subcaptionbox{(p)Arm robot executes \emph{\textbf{Pass-to-M(2)}} to pass a Tool-1 to yellow robot. Blue robot runs \emph{\textbf{Get-Tool}} to go back table. \vspace{2mm}}
        [0.30\linewidth]{\includegraphics[scale=0.16]{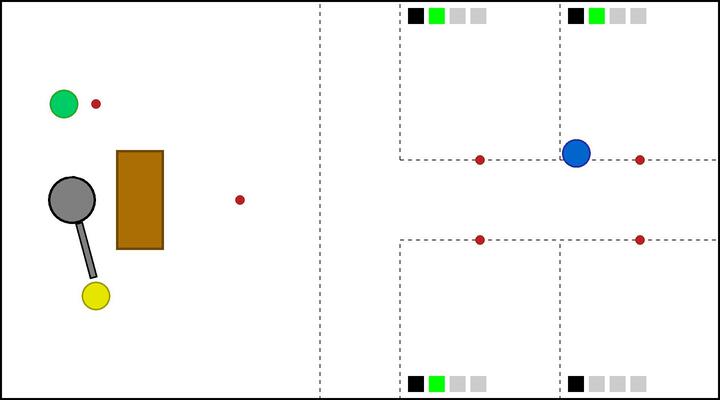}}
    ~
    \centering
    \subcaptionbox{(q) Arm robot runs \emph{\textbf{Search-Tool(0)}} to find 4th Tool-0. Yellow robot moves to workshop-0 by executing \emph{\textbf{Go-W(0)}}.\vspace{2mm}}
        [0.30\linewidth]{\includegraphics[scale=0.16]{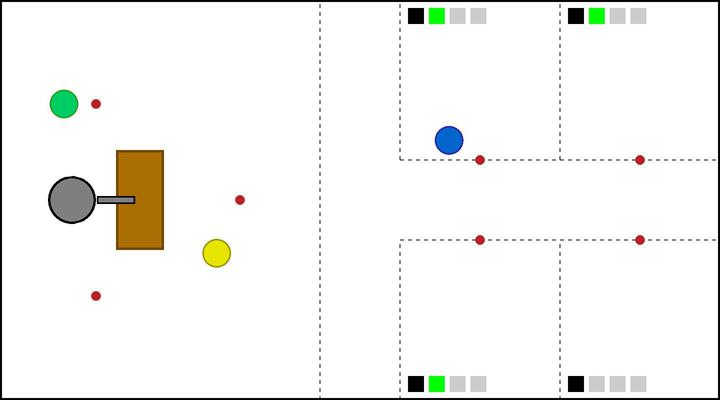}}
    ~
    \centering
    \subcaptionbox{(r) Yellow robot successfully delivers a Tool-1 to workshop-0. \vspace{2mm}}
        [0.30\linewidth]{\includegraphics[scale=0.16]{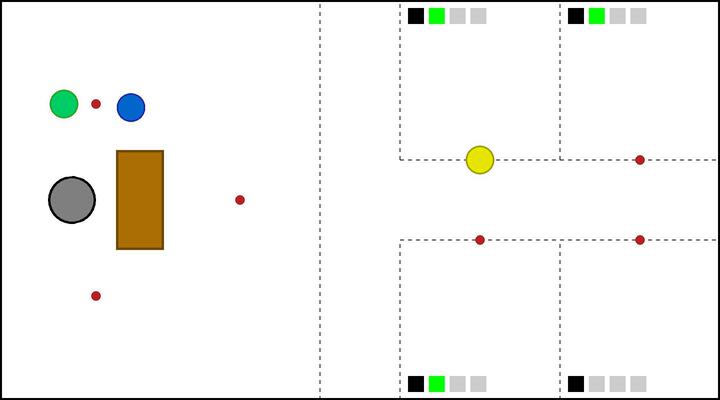}}
\end{figure*}
\begin{figure*}[h!]
    \centering
    \captionsetup[subfigure]{labelformat=empty}          
    ~
    \centering
    \subcaptionbox{(s) Yellow robot runs \emph{\textbf{Get-Tool}} to go back table. Arm robot executes \emph{\textbf{Pass-to-M(0)}} to pass a Tool-0 to green robot. \vspace{2mm}}
        [0.30\linewidth]{\includegraphics[scale=0.16]{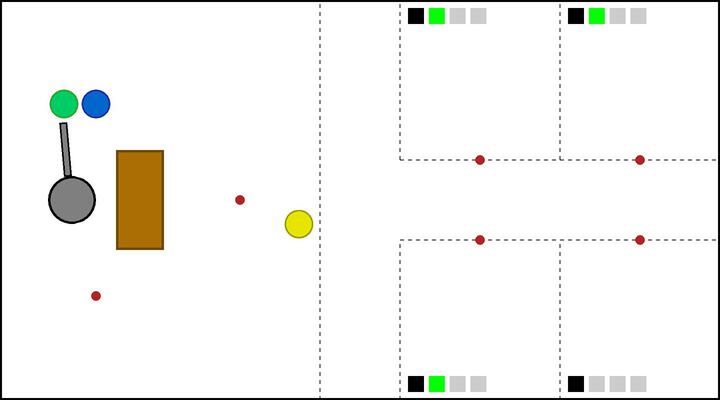}}
    ~
    \centering
    \subcaptionbox{(t) Arm robot runs \emph{\textbf{Search-Tool(1)}} to find the 3rd Tool-1. Green robot moves to workshop-1 by executing \emph{\textbf{Go-W(1)}}. \vspace{2mm}}
        [0.30\linewidth]{\includegraphics[scale=0.16]{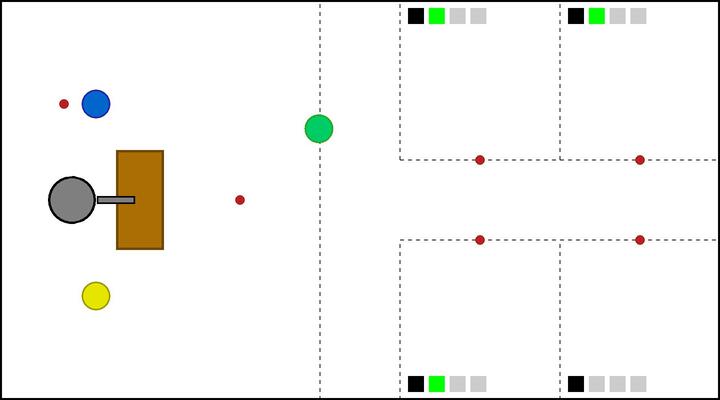}}
    ~
    \centering
    \subcaptionbox{(u) Green robot reaches workshop-1 and observes that human-1 already had Tool-0 and it moves to workshop-3 by executing \emph{\textbf{Go-W(3)}}. Arm robot runs \emph{\textbf{Pass-to-M(2)}} to pass a Tool-1 to yellow robot. \vspace{2mm}}
        [0.30\linewidth]{\includegraphics[scale=0.16]{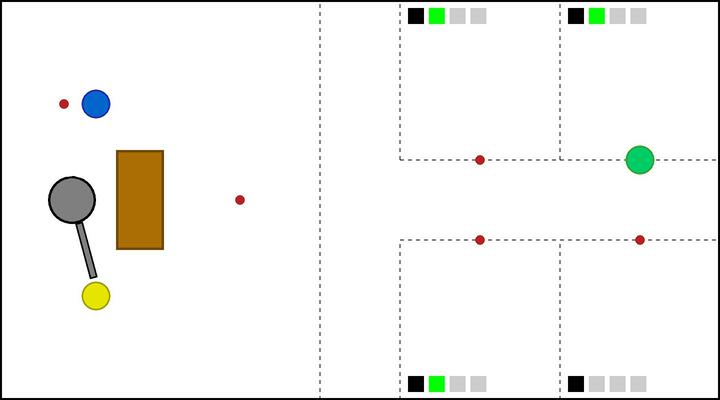}}
    ~
    \centering
    \subcaptionbox{(v) Arm robot runs \emph{\textbf{Search-Tool(1)}} to find the 4th Tool-1. Green robot delivers a Tool-0 to workshop-3. Human-3 finishes subtask-0 and starts to do subtask-1.  Yellow robot moves to workshop-2 by executing \emph{\textbf{Go-W(2)}}.\vspace{2mm}}
        [0.30\linewidth]{\includegraphics[scale=0.16]{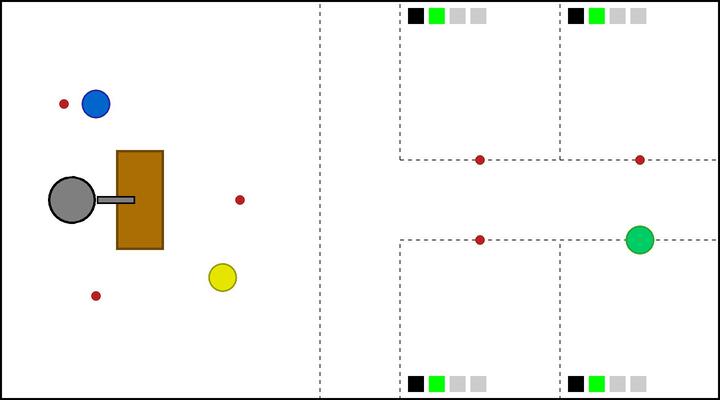}}
    ~
    \centering
    \subcaptionbox{(w) Yellow robot successfully delivers a Tool-1 to workshop-2. Green robot runs \emph{\textbf{Get-Tool}} to go back table. \vspace{2mm}}
        [0.30\linewidth]{\includegraphics[scale=0.16]{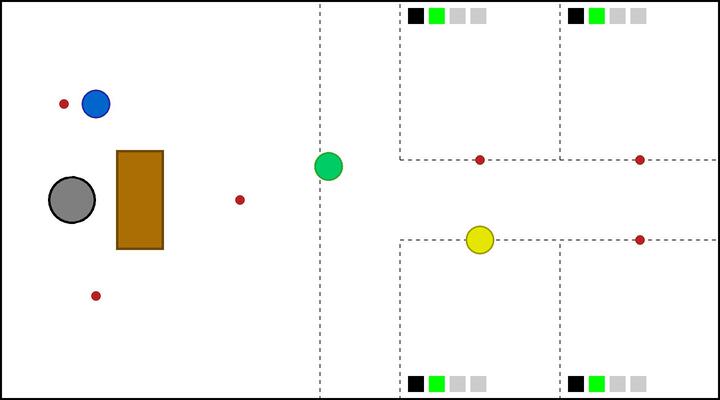}}
    ~
    \centering
    \subcaptionbox{(x) Arm robot executes \emph{\textbf{Pass-to-M(1)}} to pass a Tool-1 to blue robot. Yellow robot runs \emph{\textbf{Get-Tool}} to go back table.\vspace{2mm}}
        [0.30\linewidth]{\includegraphics[scale=0.16]{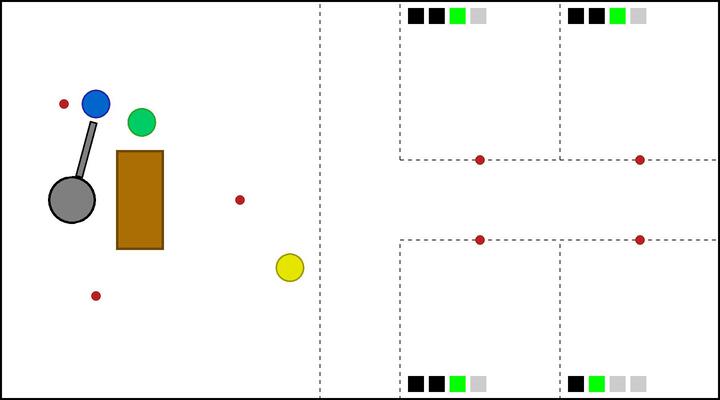}}
    ~
    \centering
    \subcaptionbox{(y) Arm robot runs \emph{\textbf{Search-Tool(2)}} to find the 1st Tool-2. Blue robot moves to workshop-1 by executing \emph{\textbf{Go-W(1)}}. \vspace{2mm}}
        [0.30\linewidth]{\includegraphics[scale=0.16]{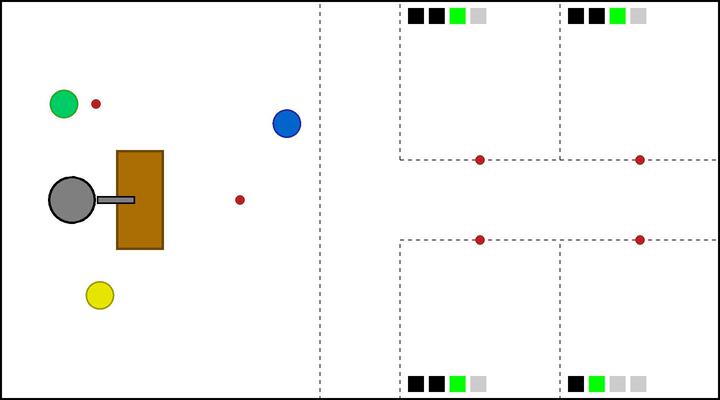}}  
    ~
    \centering
    \subcaptionbox{(z) Blue robot reaches workshop-1 and observes that human-1 does not need Tool-1.\vspace{2mm}}
        [0.30\linewidth]{\includegraphics[scale=0.16]{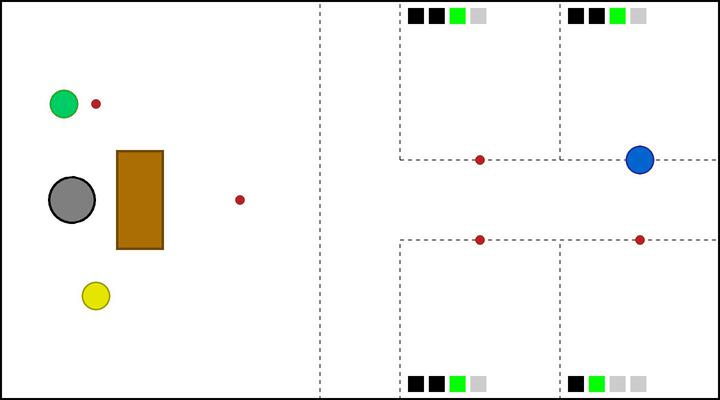}}
    ~
    \centering
    \subcaptionbox{(A) Arm robot executes \emph{\textbf{Pass-to-M(2)}} to pass a Tool-2 to yellow robot. Blue robot moves to workshop-3 by executing \emph{\textbf{Go-W(3)}}.\vspace{2mm}}
        [0.30\linewidth]{\includegraphics[scale=0.16]{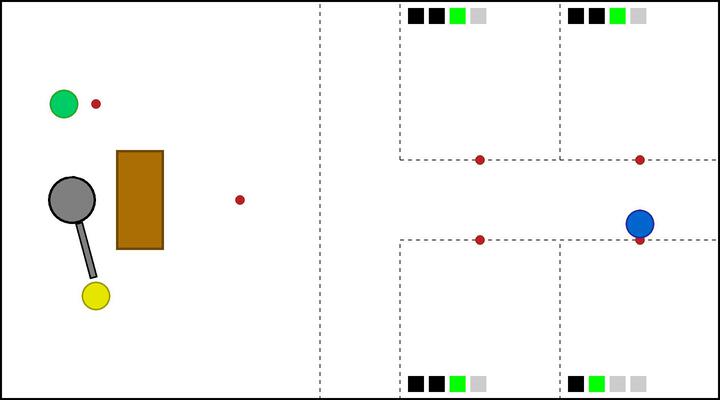}}  
\end{figure*}
\begin{figure*}[h!]
    \centering
    \captionsetup[subfigure]{labelformat=empty}          
    ~
    \centering
    \subcaptionbox{(B) Arm robot runs \emph{\textbf{Search-Tool(2)}} to find the 2nd Tool-2. Blue robot successfully delivers a Tool-1 to workshop-3. \vspace{2mm}}
        [0.30\linewidth]{\includegraphics[scale=0.16]{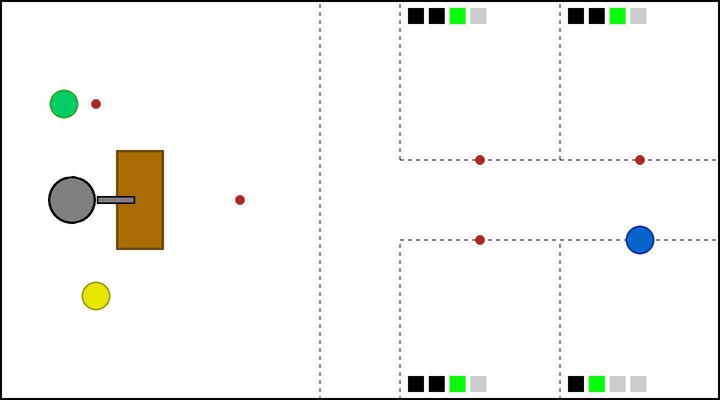}}
    ~
    \centering
    \subcaptionbox{(C) Blue robot moves to workshop-1 by executing \emph{\textbf{Go-W(1)}}.\vspace{2mm}}
        [0.30\linewidth]{\includegraphics[scale=0.16]{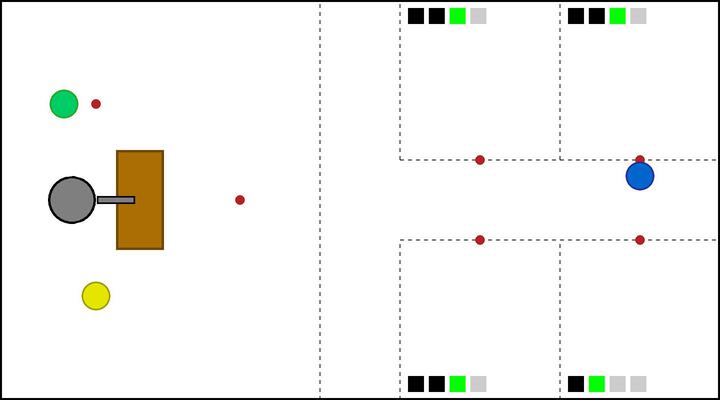}}  
    ~
    \centering
    \subcaptionbox{(D) Blue robot reaches workshop-1. \vspace{2mm}}
        [0.30\linewidth]{\includegraphics[scale=0.16]{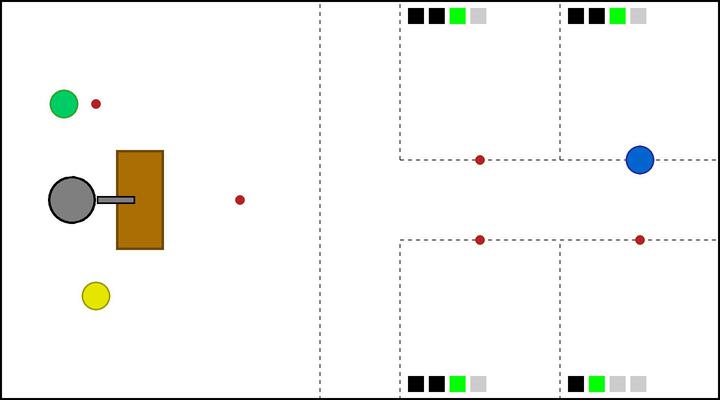}}
    ~
    \centering
    \subcaptionbox{(E) Blue robot runs \emph{\textbf{Get-Tool}} to go back table.\vspace{2mm}}
        [0.30\linewidth]{\includegraphics[scale=0.16]{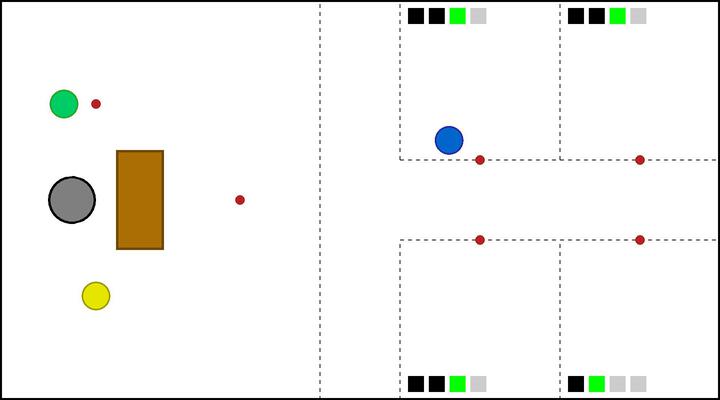}}
    ~
    \centering
    \subcaptionbox{(F) Arm robot executes \emph{\textbf{Pass-to-M(2)}} to pass the a Tool-2 to yellow robot.\vspace{2mm}}
        [0.30\linewidth]{\includegraphics[scale=0.16]{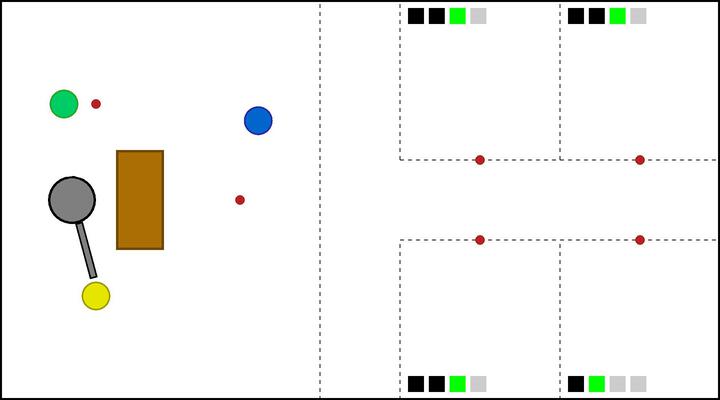}}  
    ~
    \centering
    \subcaptionbox{(G) Arm robot runs \emph{\textbf{Search-Tool(2)}} to find the 3rd Tool-2. Yellow robot moves to workshop-1 by executing \emph{\textbf{Go-W(1)}}. \vspace{2mm}}
        [0.30\linewidth]{\includegraphics[scale=0.16]{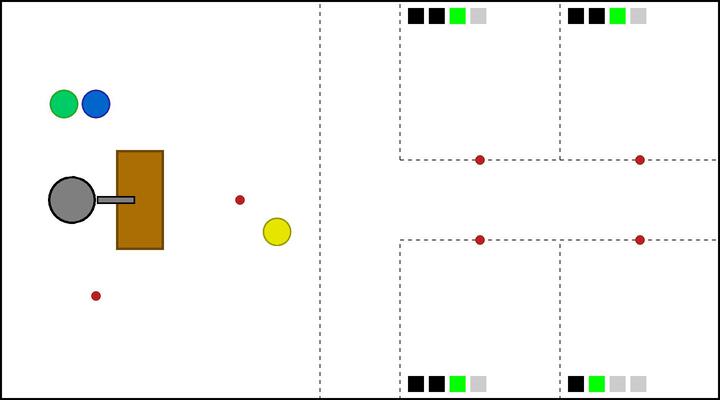}}
    ~
    \centering
    \subcaptionbox{(H) Yellow robot reaches workshop-0 and observes that human-0 has got a Tool-2.\vspace{2mm}}
        [0.30\linewidth]{\includegraphics[scale=0.16]{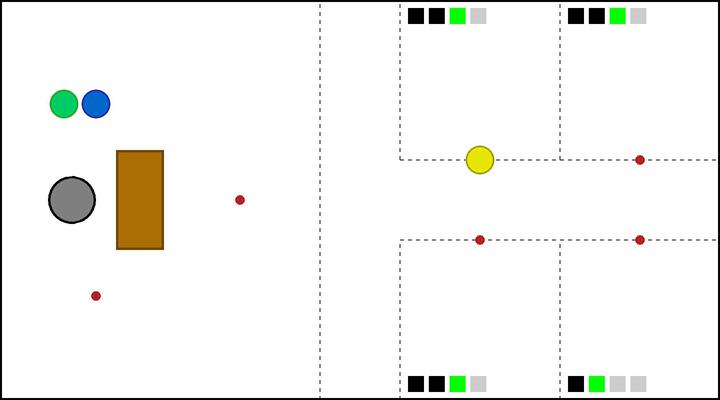}}
    ~
    \centering
    \subcaptionbox{(I) Yellow robot moves to workshop-2 by executing \emph{\textbf{Go-W(2)}}.\vspace{2mm}}
        [0.30\linewidth]{\includegraphics[scale=0.16]{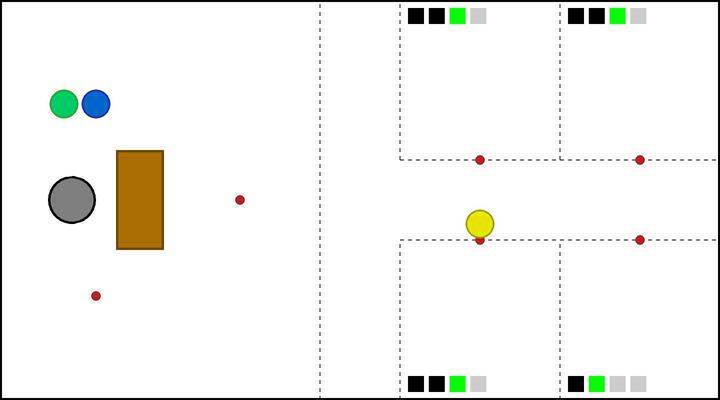}}
    ~
    \centering
    \subcaptionbox{(J) Yellow robot successfully delivers a Tool-2 to workshop-2. \vspace{2mm}}
        [0.30\linewidth]{\includegraphics[scale=0.16]{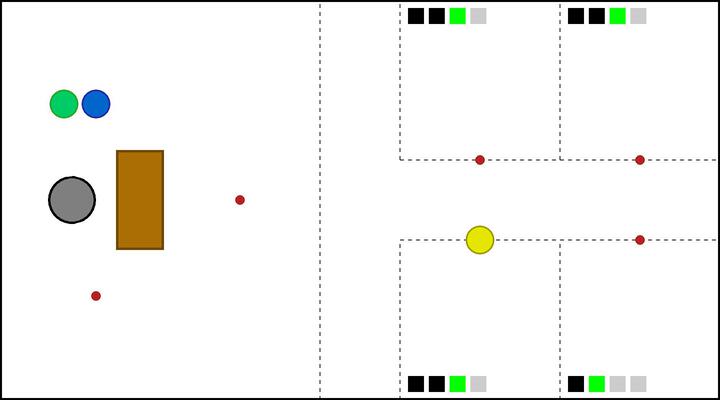}}
    ~
    \centering
    \subcaptionbox{(K) Arm robot executes \emph{\textbf{Pass-to-M(0)}} to pass a Tool-2 to green robot. Yellow robot runs \emph{\textbf{Get-Tool}} to go back table.\vspace{2mm}}
        [0.30\linewidth]{\includegraphics[scale=0.16]{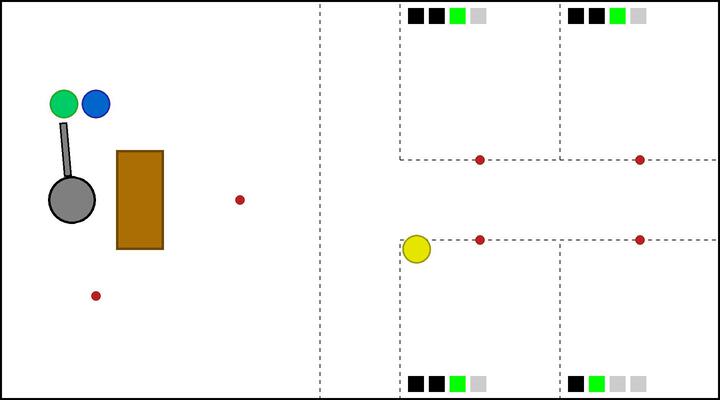}}
    ~
    \centering
    \subcaptionbox{(L) Arm robot runs \emph{\textbf{Search-Tool(2)}} to find the 4th Tool-2. Green robot moves to workshop-1 by executing \emph{\textbf{Go-W(1)}}.\vspace{2mm}}
        [0.30\linewidth]{\includegraphics[scale=0.16]{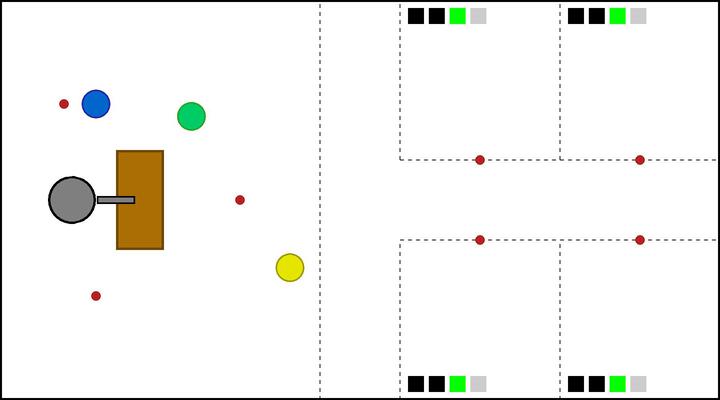}}
    ~
    \centering
    \subcaptionbox{(M) Arm robot runs \emph{\textbf{Pass-to-M(2)}} to pass a Tool-2 to yellow robot. Green robot delivers a Tool-2 to workshop-1. Human-0, human-1 and human-2 finish subtask-2 and start subtask-3.\vspace{2mm}}
        [0.30\linewidth]{\includegraphics[scale=0.16]{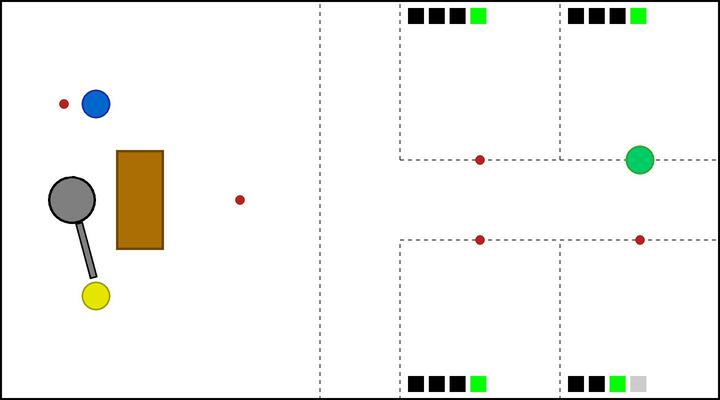}}
\end{figure*}
\begin{figure*}[h!]
    \centering
    \captionsetup[subfigure]{labelformat=empty}          
    ~
    \centering
    \subcaptionbox{(N) Yellow robot moves to workshop-0 by executing \emph{\textbf{Go-W(0)}}. Green robot moves to workshop-3 by executing \emph{\textbf{Go-W(3)}}. \vspace{2mm}}
        [0.30\linewidth]{\includegraphics[scale=0.16]{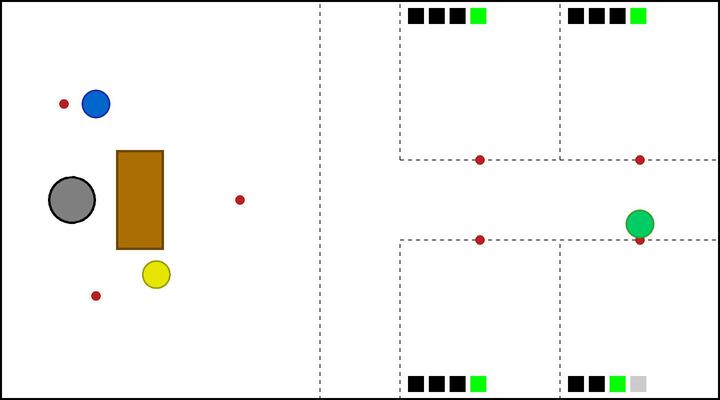}}
    ~
    \centering
    \subcaptionbox{(O) Yellow robot reaches workshop-0 and observes that human-0 does not need Tool-2.\vspace{2mm}}
        [0.30\linewidth]{\includegraphics[scale=0.16]{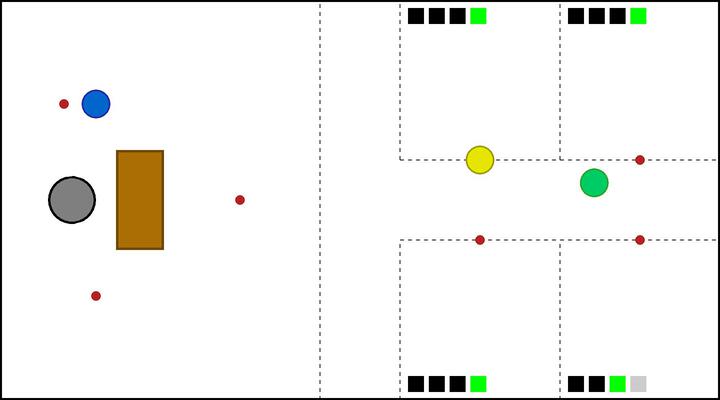}}
    ~
    \centering
    \subcaptionbox{(P) Yellow and green robot move to workshop-3 by executing \emph{\textbf{Go-W(3)}}. \vspace{2mm}}
        [0.30\linewidth]{\includegraphics[scale=0.16]{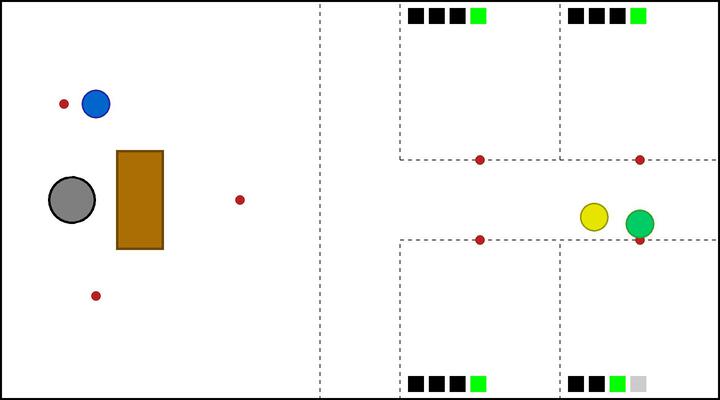}}
    ~
    \centering
    \subcaptionbox{(Q) Yellow robot successfully delivers a Tool-2 to workshop-3.  Humans have received all tools, and for robots, the task is done.\vspace{2mm}}
        [0.90\linewidth]{\includegraphics[scale=0.16]{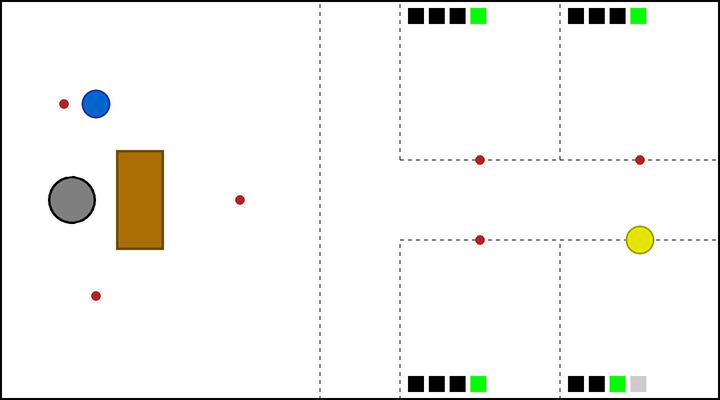}}
    \label{wtd_e_behavior}
\end{figure*}

\clearpage
\section*{Hyper-Parameters}
\addcontentsline{toc}{section}{Hyper-Parameters}  

In this section, we list the hyper-parameters used for generating the results in this thesis. We choose the best performance of each method depending on its final converged value as the first priority and the sample efficiency as the second.\\

\begin{table}[h!]
    \caption {Hyper-parameters used for value-based methods in Capture Target $4\times4$ - $30\times30$.}
    \centering
    \begin{tabular}{lcc}
    \toprule
        Parameter & Dec-Q & Mac-Dec-Q\\
    \cmidrule(r){2-3}
        Training Episodes  & 20K & 20K\\
        Learning rate & 0.001 & 0.001\\
        Batch size & 32 & 32\\
        Replay-buffer size (step) & 50K & 50K \\
        Train freq (step) & 5 & 5\\
        Target-net update freq (step) & 5K & 5K\\
        $\epsilon_{\text{start}}$ & 1 & 1 \\
        $\epsilon_{\text{end}}$ & 0.1 & 0.1 \\
        $\epsilon_{\text{decay}}$ (episode) & 4K & 4K \\
    \bottomrule
    \end{tabular} 
\end{table}

\begin{table}[h!]
    \caption {Hyper-parameters used for value-based methods in Box Pushing $4\times4$ - $8\times8$.}
    \centering
    \begin{tabular}{lccccc}
    \toprule
        Parameter & Dec-Q & MacDec-Q & MacCen-Q & MacCen-Q & MacDec-DDRQN\\
        & &  &  & (unCondi)  & \\
    \cmidrule(r){2-6}
        Training Episodes  & 15K & 15K & 15K & 15K & 15K\\
        Learning rate & 0.001 & 0.001 & 0.001 & 0.001 & 0.001\\
        Batch size & 16 & 128 & 128 & 128 & 128\\
        Replay-buffer size & 1K(epi) & 100K & 100K & 100K & 80K \\
        $\,\,\,\,\,\,\,\,\,\,\,$ (step) &  & \\
        Train freq (step) & 10 & 10 & 10 & 10 & 10\\
        Trace length (step) & N/A & 10 & 10 & 10 & 10\\
        Target-net update & 5K & 5K & 5K & 5K & 5K\\
        $\,\,\,\,\,\,\,$ freq (step) &  & \\
        $\epsilon_{\text{start}}$ & 1 & 1 & 1 & 1 & 1 \\
        $\epsilon_{\text{end}}$ & 0.1 & 0.1 & 0.1 & 0.1 & 0.1 \\
        $\epsilon_{\text{decay}}$ (episode) & 4K & 4K & 4K & 4K & 4K \\
    \bottomrule
    \end{tabular} 
\end{table}

\begin{table}[h!]
    \caption {Hyper-parameters used for value-based methods in Box Pushing $10\times10$.}
    \centering
    \begin{tabular}{lccccc}
    \toprule
        Parameter & Dec-Q & MacDec-Q & MacCen-Q & MacCen-Q & MacDec-DDRQN\\
        & &  &  & (unCondi)  & \\
    \cmidrule(r){2-6}
        Training Episodes  & 15K & 15K & 15K & 15K & 15K\\
        Learning rate & 0.001 & 0.001 & 0.001 & 0.001 & 0.001\\
        Batch size & 16 & 128 & 128 & 128 & 128\\
        Replay-buffer size & 1K(epi) & 100K & 100K & 100K & 80K \\
        $\,\,\,\,\,\,\,\,\,\,\,$ (step) &  & \\
        Train freq (step) & 14 & 14 & 14 & 14 & 14\\
        Trace length (step) & N/A & 14 & 14 & 14 & 14\\
        Target-net update & 5K & 5K & 5K & 5K & 5K\\
        $\,\,\,\,\,\,\,$ freq (step) &  & \\
        $\epsilon_{\text{start}}$ & 1 & 1 & 1 & 1 & 1 \\
        $\epsilon_{\text{end}}$ & 0.1 & 0.1 & 0.1 & 0.1 & 0.1 \\
        $\epsilon_{\text{decay}}$ (episode) & 4K & 4K & 4K & 4K & 4K \\
    \bottomrule
    \end{tabular} 
\end{table}

\begin{table}[h!]
    \caption {Hyper-parameters used for value-based methods in Box Pushing $20\times20$.}
    \centering
    \begin{tabular}{lccccc}
    \toprule
        Parameter & Dec-Q & MacDec-Q & MacCen-Q & MacCen-Q & MacDec-DDRQN\\
        & &  &  & (unCondi)  & \\
    \cmidrule(r){2-6}
        Training Episodes  & 15K & 15K & 15K & 15K & 15K\\
        Learning rate & 0.001 & 0.001 & 0.001 & 0.001 & 0.001\\
        Batch size & 16 & 128 & 128 & 128 & 128\\
        Replay-buffer size & 1K(epi) & 100K & 100K & 100K & 80K \\
        $\,\,\,\,\,\,\,\,\,\,\,$ (step) &  & \\
        Train freq (step) & 35 & 35 & 35 & 35 & 35\\
        Trace length (step) & N/A & 35 & 35 & 35 & 35\\
        Target-net update & 5K & 5K & 5K & 5K & 5K\\
        $\,\,\,\,\,\,\,$ freq (step) &  & \\
        $\epsilon_{\text{start}}$ & 1 & 1 & 1 & 1 & 1 \\
        $\epsilon_{\text{end}}$ & 0.1 & 0.1 & 0.1 & 0.1 & 0.1 \\
        $\epsilon_{\text{decay}}$ (episode) & 6K & 6K & 6K & 6K & 6K \\
    \bottomrule
    \end{tabular} 
\end{table}

\begin{table}[h!]
    \caption {Hyper-parameters used for value-based methods in Box Pushing $30\times30$.}
    \centering
    \begin{tabular}{lccccc}
    \toprule
        Parameter & Dec-Q & MacDec-Q & MacCen-Q & MacCen-Q & MacDec-DDRQN\\
        & &  &  & (unCondi)  & \\
    \cmidrule(r){2-6}
        Training Episodes  & 15K & 15K & 15K & 15K & 15K\\
        Learning rate & 0.001 & 0.001 & 0.001 & 0.001 & 0.001\\
        Batch size & 16 & 128 & 128 & 128 & 128\\
        Replay-buffer size & 1K(epi) & 100K & 100K & 100K & 80K \\
        $\,\,\,\,\,\,\,\,\,\,\,$ (step) &  & \\
        Train freq (step) & 45 & 45 & 45 & 45 & 45\\
        Trace length (step) & N/A & 45 & 45 & 45 & 45\\
        Target-net update & 5K & 5K & 5K & 5K & 5K\\
        $\,\,\,\,\,\,\,$ freq (step) &  & \\
        $\epsilon_{\text{start}}$ & 1 & 1 & 1 & 1 & 1 \\
        $\epsilon_{\text{end}}$ & 0.1 & 0.1 & 0.1 & 0.1 & 0.1 \\
        $\epsilon_{\text{decay}}$ (episode) & 6K & 6K & 6K & 6K & 6K \\
    \bottomrule
    \end{tabular} 
\end{table}

\begin{table}[h!]
    \caption {Hyper-parameters used for value-based methods in Warehouse with one human.}
    \centering
    \begin{tabular}{lccc}
    \toprule
        Parameter & MacDec-Q & MacCen-Q & Parallel-MacDec-DDRQN\\
    \cmidrule(r){2-4}
        Training Episodes  & 40K & 40K & 40K \\
        Learning rate & 0.0006 & 0.0006 & 0.0006 \\
        Batch size & 16 & 16 & 16 \\
        Replay-buffer size (episode) & 1K & 1K & 1K \\
        Train freq (step) & 30 & 30 & 30 \\
        Target-net update & 5K & 5K & 5K \\
        $\,\,\,\,\,\,\,$ freq (step) &  &\\
        $\epsilon_{\text{start}}$ & 1 & 1 & 1\\
        $\epsilon_{\text{end}}$ & 0.1 & 0.1 & 0.1 \\
        $\epsilon_{\text{decay}}$ (episode) & 6K & 6K & 6K \\
    \bottomrule
    \end{tabular} 
\end{table}

\begin{table}[h!]
    \caption {Hyper-parameters used for actor-critic methods in Box Pushing $8\times8$.}
    \centering
    \begin{tabular}{lcccccc}
    \toprule
        Parameter & IAC & CAC & Mac-IAC & Mac-CAC & Mac-NIACC & Mac-IAICC \\
    \cmidrule(r){2-7}
        Training Episodes & 40K & 40K & 40K & 40K & 40K & 40K\\
        Actor Learning rate &0.001 &0.001 &0.001 &0.0005 &0.0005 &0.0003\\
        Critic Learning rate &0.003 &0.003 &0.003 &0.003 &0.001 &0.003\\
        Episodes per train &8 &8 &16 &48 &48 &48\\
        Target-net update &32 &32 &32 &48 &144 &144\\
        $\,\,\,\,$  freq (episode) & \\
        N-step TD &3 &0 &5 &3 &0 &0\\
        $\epsilon_{\text{start}}$ & 1 & 1 & 1 & 1 & 1 & 1\\
        $\epsilon_{\text{end}}$ & 0.01 & 0.01 & 0.01 & 0.01 & 0.01 & 0.01\\
        $\epsilon_{\text{decay}}$ (episode) & 4K & 4K & 4K & 4K & 4K & 4K \\
    \bottomrule
    \end{tabular} 
\end{table}

\begin{table}[h!]
    \caption {Hyper-parameters used for actor-critic methods in Box Pushing $10\times10$.}
    \centering
    \begin{tabular}{lcccccc}
    \toprule
        Parameter & IAC & CAC & Mac-IAC & Mac-CAC & Mac-NIACC & Mac-IAICC \\
    \cmidrule(r){2-7}
        Training Episodes & 40K & 40K & 40K & 40K & 40K & 40K\\
        Actor Learning rate &0.001 &0.001 &0.001 &0.001 &0.0005 &0.0003\\
        Critic Learning rate &0.003 &0.003 &0.001 &0.003 &0.001 &0.003\\ 
        Episodes per train &8 &8 &32 &48 &48 &32\\
        Target-net update &64 &32 &32 &96 &144 &64\\
        $\,\,\,\,$  freq (episode) & \\
        N-step TD &0 &0 &5 &3 &0 &0\\
        $\epsilon_{\text{start}}$ & 1 & 1 & 1 & 1 & 1 & 1\\
        $\epsilon_{\text{end}}$ & 0.01 & 0.01 & 0.01 & 0.01 & 0.01 & 0.01\\
        $\epsilon_{\text{decay}}$ (episode) & 6K & 6K & 6K & 6K & 6K & 6K \\
    \bottomrule
    \end{tabular} 
\end{table}

\begin{table}[h!]
    \caption {Hyper-parameters used for actor-critic methods in Box Pushing $12\times12$.}
    \centering
    \begin{tabular}{lcccccc}
    \toprule
        Parameter & IAC & CAC & Mac-IAC & Mac-CAC & Mac-NIACC & Mac-IAICC \\
    \cmidrule(r){2-7}
        Training Episodes & 40K & 40K & 40K & 40K & 40K & 40K\\
        Actor Learning rate &0.001 &0.001 &0.001 &0.0005 &0.0005 &0.0003\\
        Critic Learning rate &0.003 &0.003 &0.003 &0.0005 &0.001 &0.003\\
        Episodes per train &8 &8 &8 &32 &48 &32\\
        Target-net update &128 &128 &64 &64 &96 &128\\
        $\,\,\,\,$  freq (episode) & \\
        N-step TD &0 &0 &5 &3 &0 &0\\
        $\epsilon_{\text{start}}$ & 1 & 1 & 1 & 1 & 1 & 1\\
        $\epsilon_{\text{end}}$ & 0.01 & 0.01 & 0.01 & 0.01 & 0.01 & 0.01\\
        $\epsilon_{\text{decay}}$ (episode) & 6K & 6K & 6K & 6K & 6K & 6K \\
    \bottomrule
    \end{tabular} 
\end{table}

\begin{table}[h!]
    \caption {Hyper-parameters used for actor-critic methods in Overcooked-A.}
    \centering
    \begin{tabular}{lcccccc}
    \toprule
        Parameter & IAC & CAC & Mac-IAC & Mac-CAC & Mac-NIACC & Mac-IAICC \\
    \cmidrule(r){2-7}
        Training Episodes & 100K & 100K & 100K & 100K & 100K & 100K\\
        Actor Learning rate &0.0003 &0.0003 &0.0003 &0.0001 &0.0003 &0.0003\\
        Critic Learning rate &0.003 &0.003 &0.003 &0.003 &0.003 &0.003\\     
        Episodes per train &4 &8 &4 &8 &4 &8\\
        Target-net update &8 &16 &8 &32 &16 &32\\
        $\,\,\,\,$  freq (episode) & \\
        N-step TD &5 &5 &5 &5 &5 &5\\
        $\epsilon_{\text{start}}$ & 1 & 1 & 1 & 1 & 1 & 1\\
        $\epsilon_{\text{end}}$ & 0.05 & 0.05 & 0.05 & 0.05 & 0.05 & 0.05 \\
        $\epsilon_{\text{decay}}$ (episode) & 20K & 20K & 20K & 20K & 20K & 20K  \\
    \bottomrule
    \end{tabular} 
\end{table}

\begin{table}[h!]
    \caption {Hyper-parameters used for actor-critic methods in Overcooked-B.}
    \centering
    \begin{tabular}{lcccccc}
    \toprule
        Parameter & IAC & CAC & Mac-IAC & Mac-CAC & Mac-NIACC & Mac-IAICC \\
    \cmidrule(r){2-7}
        Training Episodes & 120K & 120K & 120K & 120K & 120K & 120K\\
        Actor Learning rate &0.0003 &0.0003 &0.0003 &0.0001 &0.0003 &0.0003\\
        Critic Learning rate &0.003 &0.003 &0.003 &0.003 &0.003 &0.003\\     
        Episodes per train &4 &4 &4 &4 &8 &4\\
        Target-net update &8 &16 &8 &16 &16 &32\\
        $\,\,\,\,$  freq (episode) & \\
        N-step TD &5 &5 &5 &3 &5 &5\\
        $\epsilon_{\text{start}}$ & 1 & 1 & 1 & 1 & 1 & 1\\
        $\epsilon_{\text{end}}$ & 0.05 & 0.05 & 0.05 & 0.05 & 0.05 & 0.05 \\
        $\epsilon_{\text{decay}}$ (episode) & 20K & 20K & 20K & 20K & 20K & 20K  \\
    \bottomrule
    \end{tabular} 
\end{table}

\begin{table}[h!]
    \caption {Hyper-parameters used for actor-critic methods in Warehouse-A.}
    \centering
    \begin{tabular}{lcccc}
    \toprule
        Parameter & Mac-IAC & Mac-CAC & Mac-NIACC & Mac-IAICC \\
    \cmidrule(r){2-5}
        Training Episodes & 40K & 40K & 40K & 40K\\
        Actor Learning rate &0.0003 &0.0003 &0.0003 &0.0005\\
        Critic Learning rate &0.003 &0.003 &0.003 &0.0005\\  
        Episodes per train &4 &4 &4 &4\\
        Target-net update freq &32 &32 &32 &32\\
        $\,\,\,\,\,\,\,\,\,\,\,\,$ (episode) & \\
        N-step TD &5 &5 &3 &5\\
        $\epsilon_{\text{start}}$ & 1 & 1 & 1 & 1\\
        $\epsilon_{\text{end}}$ & 0.01 & 0.01 & 0.01 & 0.01\\
        $\epsilon_{\text{decay}}$ (episode) & 10K & 10K & 10K & 10K \\
    \bottomrule
    \end{tabular} 
\end{table}

\begin{table}[h!]
    \caption {Hyper-parameters used for actor-critic methods in Warehouse-A for ablation.}
    \centering
    \begin{tabular}{lcccc}
    \toprule
        Parameter & Mac-IAC & Mac-CAC & Mac-NIACC & Mac-IAICC \\
    \cmidrule(r){2-5}
        Training Episodes & 40K & 40K & 40K & 40K\\
        Actor Learning rate &0.0005 &0.0005 &0.0003 &0.0005\\
        Critic Learning rate &0.0005 &0.001 &0.003 &0.0005\\ 
        Episodes per train &16 &8 &8 &4\\
        Target-net update freq &16 &64 &64 &64\\
        $\,\,\,\,\,\,\,\,\,\,\,\,$ (episode) & \\
        N-step TD &5 &5 &5 &5\\
        $\epsilon_{\text{start}}$ & 1 & 1 & 1 & 1\\
        $\epsilon_{\text{end}}$ & 0.05 & 0.05 & 0.05 & 0.05\\
        $\epsilon_{\text{decay}}$ (episode) & 10K & 10K & 10K & 10K \\
    \bottomrule
    \end{tabular} 
\end{table}

\begin{table}[h!]
    \caption {Hyper-parameters used for actor-critic methods in Warehouse-B.}
    \centering
    \begin{tabular}{lcccc}
    \toprule
        Parameter & Mac-IAC & Mac-CAC & Mac-NIACC & Mac-IAICC \\
    \cmidrule(r){2-5}
        Training Episodes  & 100K & 100K & 100K & 100K\\
        Actor Learning rate &0.0005 &0.0003 &0.0003 &0.0005\\
        Critic Learning rate &0.001 &0.003 &0.003 &0.0005\\
        Episodes per train &4 &4 &4 &4\\
        Target-net update freq &32 &16 &32 &32\\
        $\,\,\,\,\,\,\,\,\,\,\,\,$ (episode) & \\
        N-step TD &5 &5 &5 &5\\
        $\epsilon_{\text{start}}$ & 1 & 1 & 1 & 1\\
        $\epsilon_{\text{end}}$ & 0.05 & 0.05 & 0.05 & 0.05\\
        $\epsilon_{\text{decay}}$ (episode) & 10K & 10K & 10K & 10K \\
    \bottomrule
    \end{tabular} 
\end{table}

\begin{table}[h!]
    \caption {Hyper-parameters used for actor-critic methods in Warehouse-C.}
    \centering
    \begin{tabular}{lcccc}
    \toprule
        Parameter & Mac-IAC & Mac-CAC & Mac-NIACC & Mac-IAICC \\
    \cmidrule(r){2-5}
        Training Episodes & 40K & 40K & 40K & 40K\\
        Actor Learning rate &0.0005 &0.0005 &0.0003 &0.0003\\
        Critic Learning rate &0.0005 &0.001 &0.003 &0.003\\
        Episodes per train &8 &4 &16 &4\\
        Target-net update freq &64 &64 &64 &32\\
        $\,\,\,\,\,\,\,\,\,\,\,\,$ (episode) & \\
        N-step TD &5 &5 &5 &5\\
        $\epsilon_{\text{start}}$ & 1 & 1 & 1 & 1\\
        $\epsilon_{\text{end}}$ & 0.01 & 0.01 & 0.01 & 0.01\\
        $\epsilon_{\text{decay}}$ (episode) & 10K & 10K & 10K & 10K \\
    \bottomrule
    \end{tabular} 
\end{table}

\begin{table}[h!]
    \caption {Hyper-parameters used for actor-critic methods in Warehouse-D.}
    \centering
    \begin{tabular}{lcccc}
    \toprule
        Parameter & Mac-IAC & Mac-CAC & Mac-NIACC & Mac-IAICC \\
    \cmidrule(r){2-5}
        Training Episodes  & 80K & 80K & 80K & 80K\\
        Actor Learning rate &0.0005 &0.0003 &0.0003 &0.0003\\
        Critic Learning rate &0.001 &0.003 &0.003 &0.003\\
        Episodes per train &8 &8 &8 &8\\
        Target-net update freq &64 &64 &64 &64\\
        $\,\,\,\,\,\,\,\,\,\,\,\,$ (episode) & \\
        N-step TD &5 &5 &5 &5\\
        $\epsilon_{\text{start}}$ & 1 & 1 & 1 & 1\\
        $\epsilon_{\text{end}}$ & 0.01 & 0.01 & 0.01 & 0.01\\
        $\epsilon_{\text{decay}}$ (episode) & 10K & 10K & 10K & 10K \\
    \bottomrule
    \end{tabular} 
\end{table}

\begin{table}[h!]
    \caption {Hyper-parameters used for actor-critic methods in Warehouse-E.}
    \centering
    \begin{tabular}{lcccc}
    \toprule
        Parameter & Mac-IAC & Mac-CAC & Mac-NIACC & Mac-IAICC \\
    \cmidrule(r){2-5}
        Training Episodes  & 80K & 80K & 80K & 80K\\
        Actor Learning rate &0.0003 &0.0003 &0.0005 &0.0003\\
        Critic Learning rate &0.003 &0.003 &0.005 &0.003\\
        Episodes per train &4 &8 &4 &8\\
        Target-net update freq &16 &64 &32 &64\\
        $\,\,\,\,\,\,\,\,\,\,\,\,$ (episode) & \\
        N-step TD &5 &5 &5 &5\\
        $\epsilon_{\text{start}}$ & 1 & 1 & 1 & 1\\
        $\epsilon_{\text{end}}$ & 0.01 & 0.01 & 0.01 & 0.01\\
        $\epsilon_{\text{decay}}$ (episode) & 10K & 10K & 10K & 10K \\
    \bottomrule
    \end{tabular} 
\end{table}

\begin{table}[h!]
    \caption {Hyper-parameters used for value-based methods in Box Pushing $8\times8$.}
    \centering
    \begin{tabular}{lccc}
    \toprule
        Parameter & Mac-Dec-Q & Mac-Cen-Q & MacDec-DDRQN\\
    \cmidrule(r){2-4}
        Training Episodes  & 40K & 40K & 40K \\
        Learning rate & 0.001 & 0.001 & 0.001 \\
        Batch size & 64 & 64 & 32 \\
        Replay-buffer size (step) & 100K & 100K & 100K \\
        Train freq (step) & 10 & 10 & 10 \\
        Trace length (step) & 10 & 10 & 10 \\
        Target-net update freq (step) & 5K & 5K & 5K \\
        $\epsilon_{\text{start}}$ & 1 & 1 & 1 \\
        $\epsilon_{\text{end}}$ & 0.05 & 0.05 & 0.05 \\
        $\epsilon_{\text{decay}}$ (episode) & 4K & 4K & 4K \\
    \bottomrule
    \end{tabular} 
\end{table}

\begin{table}[h!]
    \caption {Hyper-parameters used for value-based methods in Box Pushing $10\times10$.}
    \centering
    \begin{tabular}{lccc}
    \toprule
        Parameter & Mac-Dec-Q & Mac-Cen-Q & MacDec-DDRQN \\
    \cmidrule(r){2-4}
        Training Episodes  & 40K & 40K & 40K \\
        Learning rate & 0.001 & 0.001 & 0.001\\
        Batch size & 32 & 128 & 64\\
        Replay-buffer size (step) & 100K & 100K & 100K\\
        Train freq (step) & 14 & 14 & 14\\
        Trace length (step) & 14 & 14 & 14\\
        Target-net update freq (step) & 5K & 5K & 5K\\
        $\epsilon_{\text{start}}$ & 1 & 1 & 1\\
        $\epsilon_{\text{end}}$ & 0.05 & 0.05 & 0.05 \\
        $\epsilon_{\text{decay}}$ (episode) & 6K & 6K & 6K \\
    \bottomrule
    \end{tabular} 
\end{table}

\begin{table}[h!]
    \caption {Hyper-parameters used for value-based methods in Box Pushing $12\times12$.}
    \centering
    \begin{tabular}{lccc}
    \toprule
        Parameter & Mac-Dec-Q & Mac-Cen-Q & MacDec-DDRQN\\
    \cmidrule(r){2-4}
        Training Episodes  & 40K & 40K & 40K\\
        Learning rate & 0.001 & 0.001 & 0.001\\
        Batch size & 32 & 128 & 64\\
        Replay-buffer size (step) & 100K & 100K & 100K\\
        Train freq (step) & 20 & 20 & 20 \\
        Trace length (step) & 20 & 20 & 20 \\
        Target-net update freq (step) & 5K & 5K & 5K\\
        $\epsilon_{\text{start}}$ & 1 & 1 & 1 \\
        $\epsilon_{\text{end}}$ & 0.05 & 0.05 & 0.05\\
        $\epsilon_{\text{decay}}$ (episode) & 6K & 6K & 6K \\
    \bottomrule
    \end{tabular} 
\end{table}

\begin{table}[h!]
    \caption {Hyper-parameters used for value-based methods in Overcooked-A.}
    \centering
    \begin{tabular}{lccc}
    \toprule
        Parameter & Mac-Dec-Q & Mac-Cen-Q & MacDec-DDRQN \\
    \cmidrule(r){2-4}
        Training Episodes  & 100K & 100K & 100K \\
        Learning rate & 0.0005 & 0.00003 & 0.0001\\
        Batch size & 64 & 64 & 32 \\
        Replay-buffer size (episode) & 1K & 1K & 1K\\
        Train freq (step) & 64 & 64 & 128\\
        Target-net update freq (step) & 5K & 5K & 5K\\
        $\epsilon_{\text{start}}$ & 1 & 1 & 1 \\
        $\epsilon_{\text{end}}$ & 0.05 & 0.05 & 0.05\\
        $\epsilon_{\text{decay}}$ (episode) & 20K & 20K & 20K\\
    \bottomrule
    \end{tabular} 
\end{table}

\begin{table}[h!]
    \caption {Hyper-parameters used for value-based methods in Overcooked-B.}
    \centering
    \begin{tabular}{lccc}
    \toprule
        Parameter & Mac-Dec-Q & Mac-Cen-Q & MacDec-DDRQN \\
    \cmidrule(r){2-4}
        Training Episodes  & 100K & 100K & 100K \\
        Learning rate & 0.0005 & 0.0001 & 0.0001\\
        Batch size & 32 & 32 & 32\\
        Replay-buffer size (episode) & 3K & 500 & 1K\\
        Train freq (step) & 64 & 64 & 128\\
        Target-net update freq (step) & 5K & 5K & 5K\\
        $\epsilon_{\text{start}}$ & 1 & 1 & 1 \\
        $\epsilon_{\text{end}}$ & 0.05 & 0.05 & 0.05\\
        $\epsilon_{\text{decay}}$ (episode) & 20K & 20K & 20K\\
    \bottomrule
    \end{tabular} 
\end{table}

\begin{table}[h!]
    \caption {Hyper-parameters used for value-based methods in Warehouse-A.}
    \centering
    \begin{tabular}{lccc}
    \toprule
        Parameter & Mac-Dec-Q & Mac-Cen-Q & Parallel-MacDec-DDRQN \\
    \cmidrule(r){2-4}
        Training Episodes  & 40K & 40K & 40K\\
        Learning rate & 0.0001 & 0.0001 & 0.0001\\
        Batch size & 64 & 64 & 64 \\
        Replay-buffer size (episode) & 2K & 2K & 4K\\
        Train freq (step) & 128 & 128 & 128\\
        Target-net update freq (step) & 5K & 5K & 20K\\
        $\epsilon_{\text{start}}$ & 1 & 1 & 1 \\
        $\epsilon_{\text{end}}$ & 0.05 & 0.05 & 0.05\\
        $\epsilon_{\text{decay}}$ (episode) & 10K & 10K & 10K\\
    \bottomrule
    \end{tabular} 
\end{table}

\begin{table}[h!]
    \caption {Hyper-parameters used for value-based methods in Warehouse-B.}
    \centering
    \begin{tabular}{lccc}
    \toprule
        Parameter & Mac-Dec-Q & Mac-Cen-Q & Parallel-MacDec-DDRQN \\
    \cmidrule(r){2-4}
        Training Episodes  & 40K & 40K & 40K \\
        Learning rate & 0.0001 & 0.0001 & 0.0001\\
        Batch size & 64 & 64 & 64\\
        Replay-buffer size (episode) & 2K & 2K & 4K\\
        Train freq (step) & 128 & 128 & 128\\
        Target-net update freq (step) & 5K & 5K & 20K\\
        $\epsilon_{\text{start}}$ & 1 & 1 & 1\\
        $\epsilon_{\text{end}}$ & 0.05 & 0.05 & 0.05\\
        $\epsilon_{\text{decay}}$ (episode) & 10K & 10K & 10K\\
    \bottomrule
    \end{tabular} 
\end{table}

\begin{table}[h!]
    \caption {Hyper-parameters used for value-based methods in Warehouse-C.}
    \centering
    \begin{tabular}{lccc}
    \toprule
        Parameter & Mac-Dec-Q & Mac-Cen-Q & Parallel-MacDec-DDRQN\\
    \cmidrule(r){2-4}
        Training Episodes  & 80K & 80K & 80K\\
        Learning rate & 0.00005 & 0.00005 & 0.0001\\
        Batch size & 64 & 64 & 64\\
        Replay-buffer size (episode) & 2K & 2K & 4K\\
        Train freq (step) & 128 & 128 & 128\\
        Target-net update freq (step) & 5K & 5K & 40K\\
        $\epsilon_{\text{start}}$ & 1 & 1 & 1\\
        $\epsilon_{\text{end}}$ & 0.05 & 0.05 & 0.05\\
        $\epsilon_{\text{decay}}$ (episode) & 10K & 10K & 10K \\
    \bottomrule
    \end{tabular} 
\end{table}

\bibliographystyle{unsrt}

\bibliography{bib/references}

\appendix

\printindex

\end{document}